%% file: main-arxiv.tex
\documentclass[sigconf, nonacm, table]{acmart}

\newcommand\vldbpagestyle{plain}

\usepackage{xspace}
\usepackage{booktabs} 
\usepackage{xcolor} 
\usepackage{xspace}
\usepackage{booktabs} 
\usepackage{graphicx} 
\usepackage{caption}
\usepackage{subcaption}
\usepackage[labelfont=bf]{caption}
\usepackage{tikz}
\usepackage{bbm}
\usepackage{makecell}
\usepackage{balance}
\usepackage{multicol}

\def\checkmark{\tikz\fill[scale=0.4](0,.35) -- (.25,0) -- (1,.7) -- (.25,.15) -- cycle;}

\newcommand{\etal}{\emph{et al.}\xspace}
\newcommand*{\eg}{e.g.,\xspace}
\newcommand*{\ie}{i.e.,\xspace}
\newcommand*{\numexps}{29,736\xspace}

\newcommand*{\mvi}{\texttt{MVI}\xspace}

\newcommand*{\nl}{\texttt{null}\xspace}

\newcommand*{\sys}{\texttt{Shades-of-Null}\xspace}

\newcommand*{\diabetes}{\texttt{diabetes}\xspace}
\newcommand*{\german}{\texttt{german}\xspace}
\newcommand*{\folk}{\texttt{folk-income}\xspace}
\newcommand*{\law}{\texttt{law-school}\xspace}
\newcommand*{\bank}{\texttt{bank}\xspace}
\newcommand*{\heart}{\texttt{heart}\xspace}
\newcommand*{\folkemp}{\texttt{folk-employment}\xspace}

\newcommand*{\deletion}{\texttt{deletion}\xspace}
\newcommand*{\mode}{\texttt{median-mode}\xspace}
\newcommand*{\dummy}{\texttt{median-dummy}\xspace}
\newcommand*{\missforest}{\texttt{miss-forest}\xspace}
\newcommand*{\clustering}{\texttt{clustering}\xspace}
\newcommand*{\datawig}{\texttt{datawig}\xspace}
\newcommand*{\automl}{\texttt{auto-ml}\xspace}
\newcommand*{\gain}{\texttt{gain}\xspace}
\newcommand*{\hivae}{\texttt{hi-vae}\xspace}
\newcommand*{\notmiwae}{\texttt{not-miwae}\xspace}
\newcommand*{\mnarpvae}{\texttt{mnar-pvae}\xspace}
\newcommand*{\boostclean}{\texttt{boostclean}\xspace}
\newcommand*{\nomi}{\texttt{nomi}\xspace}
\newcommand*{\tdm}{\texttt{tdm}\xspace}
\newcommand*{\editgain}{\texttt{edit-gain}\xspace}

\newcommand*{\mcar}{\texttt{MCAR}\xspace}
\newcommand*{\mar}{\texttt{MAR}\xspace}
\newcommand*{\mnar}{\texttt{MNAR}\xspace}

\newcommand{\rev}[1]{#1}
\newcommand{\revtwo}[1]{#1}
\newcommand{\submit}[1]{}
\newcommand{\arxiv}[1]{#1}

\begin{document}
\title{Still More Shades of Null: An Evaluation Suite for Responsible Missing Value Imputation}

\author{Falaah Arif Khan}
\authornote{Co-first authors.}
\email{fa2161@nyu.edu}
\affiliation{%
  \institution{New York University}
  \city{New York}
  \country{USA}
}

\author{Denys Herasymuk}
\authornotemark[1]
\email{herasymuk@ucu.edu.ua}
\affiliation{%
  \institution{Ukrainian Catholic University}
  \city{Lviv}
  \country{Ukraine}
}

\author{Nazar Protsiv}
\email{protsiv.pn@ucu.edu.ua}
\affiliation{%
  \institution{Ukrainian Catholic University}
  \city{Lviv}
  \country{Ukraine}
}

\author{Julia Stoyanovich}
\email{stoyanovich@nyu.edu}
\affiliation{%
  \institution{New York University}
  \city{New York}
  \country{USA}
}

\begin{abstract}
\input{sections/abstract}
\end{abstract}

\maketitle

\pagestyle{\vldbpagestyle}

\input{sections/intro}
\input{sections/related-work}
\input{sections/benchmark}
\input{sections/combined}

\input{sections/shift}

\input{sections/time}
\input{sections/conclusion-new}

\input{sections/limitations}

\bibliographystyle{ACM-Reference-Format}
\bibliography{sample}

\begin{appendix}

\clearpage
\input{appendix/appendix-overview}
\input{appendix/benchmark-additional}
\input{appendix/exp-details-additional}
\include{appendix/single-mechanism-additional}
\include{appendix/shift-additional}
\input{appendix/correlation}

\end{appendix}

\end{document}

%% file: sections/abstract.tex
Data missingness is a practical challenge of sustained interest to the scientific community. 
In this paper, we present \sys, an evaluation suite for responsible missing value imputation. Our work is novel in two ways (i) we model realistic and socially-salient missingness scenarios that go beyond Rubin's classic  Missing Completely at Random (\mcar), Missing At Random (\mar) and Missing Not At Random (\mnar) settings, to include multi-mechanism missingness (when different missingness patterns co-exist in the data) and missingness shift (when the missingness mechanism changes between training and test) (ii) we evaluate imputers holistically, based on imputation quality \rev{and imputation fairness}, as well as on the predictive performance, fairness and stability of the models that are trained and tested on the data post-imputation.

We use \sys to conduct a large-scale empirical study involving \rev{\numexps} experimental pipelines, and find that while there is no single best-performing imputation approach for all missingness types, interesting trade-offs arise between predictive performance, fairness and stability, based on the combination of missingness scenario, imputer choice, and the architecture of the predictive model. We make \sys publicly available, to enable researchers to rigorously evaluate missing value imputation methods on a wide range of metrics in plausible and socially meaningful scenarios.

%% file: sections/intro.tex
\section{Introduction}
\label{sec:intro}

As AI becomes more widely deployed into society, data --- most importantly, openly accessible high quality AI-ready data --- becomes a precious shared commodity. Among the factors affecting data quality is data missingness, a prevailing practical challenge  of sustained interest to the data management, statistics and data science communities, and to the scientific community writ large~\cite{rubin1987statistical, rubin1976inference, pigott_review, Mitra2023_structured, zhou2024_Comprehensive_review, Shadbahr_nature_medicine, graham2009missing, le2021goodimputation, Zhang2021AssessingFI, Howell2007TheTO,DBLP:books/acm/IlyasC19}.

\rev{Debates on handling missing values in data management date back to the field’s inception, with classic discussions such as~\citet{DBLP:journals/sigmod/Date84a}. At the operational level, missing values are typically denoted by \nl, but hidden missing values can exist (\eg `AL' being selected by default in a job application). At the semantic level, \nl can have multiple meanings—unknown, inapplicable, or intentionally withheld.}
\rev{This paper does not engage in the semantic debate or consider hidden missing values~\cite{disguised_missingness}. Instead, we focus on a specific case: a dataset $X$ (a single relation) where some features are missing, marked by \nl, indicating that the feature has a real-world value but is unobserved in $X$. Our goal is to use $X$ in a machine learning (ML) setting, either for model training or inference. Since ML models cannot handle \nl directly, missing values must be imputed as part of data preprocessing.}
 
As our starting point, we will use Rubin's missingness framework~\cite{rubin1976inference} that, nearly  50 years since it was proposed, still remains the most popular approach to modeling missing data. Consider a dataset $X$ of $n$ samples, each with $p$ features, and an indicator $R$ such that $R_{i,j}=1$ when the value of the $j$'s feature of $X_i$ is missing: $X_{i,j} \text{ is null}$, and $R_{i,j}=0$ when that the value is observed: $X_{i,j} \text{is not null}$. Rubin identified three data missingness scenarios:

\paragraph{Missing Completely at Random (\textbf{\mcar})}  
\rev{In a job applicant dataset with salary and years of experience, \mcar holds if salary is missing due to administrative errors, unrelated to the salary itself or work experience. That is: $P(R|X) = P(R)$.}  

\paragraph{Missing at Random (\textbf{\mar})}  
\rev{If job applicants with fewer years of experience are more likely to withhold their salary, and this can be explained by observed covariates (\ie years of experience), then \mar holds. Here, missingness depends only on observed features, not the missing values themselves: $P(R|X) = P(R|X_{\text{obs}})$.}

\paragraph{Missing Not at Random (\textbf{\mnar})}  
\rev{Consider a job applicant whose salary depends on geographic location and skills test results---neither captured in the data---rather than years of experience. Suppose applicants with lower salaries are more likely to withhold this information, hoping for a higher offer. In this case, \mnar holds because missingness is correlated with the missing value itself and \emph{cannot} be explained by observed covariates (\ie years of experience): $P(R|X) \not= P(R|X_{\text{obs}})$.} 

\paragraph{Missing value imputation (\textbf{\mvi})}
\rev{Rubin's framework has shaped a vast body of work on missing value imputation, extensively reviewed in several comprehensive surveys~\cite{abdelaal2023rein,hasan_mvi_review,le2022survey, Mena2022_survey, pintz2022survey, emmanuel2021survey, multifidelity_survey, mvi_review, pigott_review, zhou2024_Comprehensive_review,Alabadla_review,missing_data_review,Joel_Review_Missing_Data_for_ML,DBLP:books/acm/IlyasC19}. \mvi methods fall into three main categories:  
(1) \emph{Statistical methods}, such as median or mode imputation~\cite{schafer2002missing};  
(2) \emph{Learning-based impute-then-classify}, which iteratively impute missing values using k-nearest neighbors~\cite{batista2003analysis}, clustering~\cite{gajawada2012missing_clustering}, decision trees~\cite{twala2009empirical}, or ensembles~\cite{stekhoven2012missforest};  
(3) \emph{Joint data cleaning and model training}, integrating imputation with model learning~\cite{karlavs2020CPClean, krishnan2017boostclean, krishnan2016activeclean}, based on Rubin's multiple imputation framework~\cite{rubin1987multiple}.}

\paragraph{Beyond Rubin's framework: Mixing scenarios and dealing with missingness shift.} 
\rev{Rubin's framework, while analytically clean, does not fully capture real-world missingness. First, \textsc{MCAR} assumptions rarely hold, and real-world data often falls on a continuum between \textsc{MAR} and \textsc{MNAR}, depending on collection methods~\cite{graham2009missing}. Second, missingness mechanisms frequently \emph{co-exist within a dataset} (affecting different features or tuples), leading to \emph{multi-mechanism missingness}~\cite{zhou2024_Comprehensive_review}. For instance,~\citet{Mitra2023_structured} introduce the \emph{data missingness life cycle}, showing how data integration from diverse sources creates \emph{structured missingness} beyond Rubin’s model. Third, in \emph{data-centric AI}, missingness assumptions valid during training may shift post-deployment, a phenomenon termed \emph{missingness shift}, analogous to data distribution shift~\cite{zhou2023_missingnessshift}.}

\paragraph{\rev{Missingness as a form of bias.}} 
\rev{Consider the job applicant screening example with gender and age as features. Female applicants who suspect wage discrimination may withhold salary information more often than men, hoping to narrow the gender pay gap. This leads to more missing salary values for women, where missingness depends on the observed covariate (gender), aligning with \mar. This reflects \emph{pre-existing bias}, where data encodes historical societal discrimination~\cite{friedman_nissenabaum_bias}.  For another example, suppose disability status is included as a feature. Applicants with disabilities may be more likely to omit this information. If disability status is uncorrelated with other features, this scenario aligns with \mnar, with missingness itself acting as a proxy for disadvantage.}

\rev{When handling missing values, data scientists must also address \emph{technical bias}~\cite{friedman_nissenabaum_bias}, where incorrect technical choices create disparities in predictive accuracy, often amplifying pre-existing bias. A key example is imputing missing values under incorrect assumptions, which can worsen disparities in classifier performance~\cite{DBLP:journals/debu/SchelterS20,DBLP:journals/cacm/StoyanovichAHJS22,schelter2019fairprep,guha_icde}.  For instance, if job applicant salaries are missing under \mar or \mnar (\eg older women withhold salaries due to perceived discrimination), imputing them under \mcar could further depress salary estimates, reinforcing the gender wage gap and ageism, and leading to discriminatory outcomes.}  

\paragraph{\rev{Missing value imputation can impact model arbitrariness.}} \rev{Missingness is an indication of uncertainty in the data.  \mvi methods ``resolve'' this uncertainty at the tuple level, but they may induce a change in the data distribution in ways that impacts the stability of predictions of a model trained on this data. In some cases, the resulting models produce vastly different --- and even \emph{arbitrary} --- predictions under small perturbations in the input~\cite{coston2023validity, creel2022algorithmic,DBLP:journals/datamine/RheaMDSSSKS22,DBLP:conf/aies/RheaMDSSSS22}.  For example, if a job applicant's salary is imputed in vastly different ways upon two consecutive applications for the same position, and this, in turn, impacts the hiring decision, then the decision-making process violates the principle of process fairness (\eg~\cite{statefarm1983,tyler2006why}). Importantly, instability and accuracy are orthogonal: models can be accurate in expectation while still being unstable~\cite{long2023arbitrariness}.}

\paragraph{\rev{Research gap}} \rev{Despite numerous \mvi techniques being proposed each year, there has been limited systematic progress in assessing them across key performance aspects, including imputation correctness, predictive accuracy, and fairness—measured as disparities in imputation quality or model performance across groups. Moreover, while missingness signals uncertainty, there has been no comprehensive evaluation of the \emph{stability} of models trained on cleaned data. Crucially, realistic modeling of missingness, identifying bias sources, and selecting appropriate stakeholder groups and fairness metrics must be grounded in the specific context of use~\cite{mitchell2021algorithmic,fuster2022predictably,obermeyer2019dissecting,kochling2020discriminated}. For instance, age-based discrimination is relevant in both hiring and lending, yet older applicants face disadvantages (and legal protections) in hiring, while younger applicants are disadvantaged in lending. Thus, \mvi techniques must be evaluated in societally meaningful scenarios.}  

\paragraph{Summary of contributions}  
We implemented an experimental benchmark called \sys to rigorously and comprehensively evaluate state-of-the-art \mvi techniques on a variety of realistic missingness scenarios (including single- and multiple-mechanism missingness and missingness shift), on a suite of evaluation metrics (including fairness and stability), in the context of data pre-processing in a machine learning pipeline.

Our work is (1) \emph{novel}: to the best of our knowledge, the settings of multi-mechanism missingness and missingness shifts have not been empirically studied before; (2) \emph{comprehensive}: we evaluate a suite of \rev{15 \mvi techniques} on 7 benchmark datasets using 6 model types, running a total of \rev{\textbf{\numexps}} pipelines, and is the first study of such scale in the missing data domain, to the best of our knowledge; (3) \emph{normatively grounded}: we focus on decision-making contexts such as lending, hiring, and healthcare, where missingness is socially salient. Mitigating social harm such as algorithmic discrimination is a leading concern in these domains~\cite{barocas2016big}, and we evaluate the impact of \mvi approaches on downstream model fairness and stability (which have been understudied in the context of missing data), in addition to classically studied imputation quality and model correctness metrics.

While developing the \sys evaluation suite, we found and fixed several bugs in existing \mvi implementations, including data leakage and omitted hyperparameter tuning.   See \submit{full version of the paper for details~\cite{shades_arXiv}.}\arxiv{Appendix~\ref{apdx:enhancements} for details.}
We make \sys publicly available \arxiv{\footnote{\url{https://github.com/FalaahArifKhan/data-cleaning-stability}}} 
and hope to enable researchers to comprehensively evaluate new \mvi methods on a wide range of evaluation metrics, under plausible and socially meaningful missingness scenarios.

%% file: sections/related-work.tex
\section{Related Work}
\label{sec:related-work}

\paragraph{Missing value imputation techniques} Learning-based approaches have become increasingly popular, and include k-nearest neighbors, decision trees, support vector machines, clustering, and ensembles~\cite{batista2003analysis, mvi_review, hasan_mvi_review}. ~\citet{zhou2024_Comprehensive_review} and \citet{Liu2023handling}  review deep learning-based approaches (variational auto-encoders and generative adversarial networks) and representation learning  (graph neural networks and diffusion-based methods). Multiple imputation~\cite{schafer2002missing, zhou2024_Comprehensive_review} and expectation maximization~\cite{pigott_review, mvi_uncertainty} are also influential, but too computationally expensive to be popular in practice~\cite{Alabadla_review}. 

\rev{MNAR-specific techniques, like not-MIWAE~\cite{ipsen2020not} and GINA~\cite{ma2021identifiable}, tackle the challenge of \mnar data by employing identifiable generative models that effectively account for complex missingness mechanisms. Recent methods, including NOMI~\cite{wang2024missing} and TDM~\cite{zhao2023transformed}, introduce advancements like uncertainty-driven networks and transformed distribution matching, which enhance both imputation accuracy and computational efficiency.}

Beyond impute-then-classify, the data management community has proposed holistic methods like CPClean~\cite{karlavs2020CPClean} and ActiveClean~\cite{krishnan2016activeclean}, that jointly perform data cleaning and model training, deriving from the multiple imputation framework~\cite{rubin1987multiple}. These methods detect and repair a variety of errors including outliers, mislabels, duplicates, and missing values, and hence are less directly optimized to model missingness, instead focusing on improving data quality holistically. BoostClean~\cite{krishnan2017boostclean} aims to reduce the human effort in error repair by learning efficiently from a few gold standard annotations (from a human oracle).

\paragraph{Evaluating \mvi techniques} \rev{We are aware of several surveys of \mvi techniques, all conducted with a strong empirical focus~\cite{mvi_review, Alabadla_review, farhangfar2008impact, miao2022experimental,shahbazian2023generative}. \citet{miao2022experimental} compare 19 \mvi methods on 15 datasets, and while our results corroborate their findings (see Section~\ref{sec:shift-f1}), their evaluation is limited to imputation quality and overall accuracy (but not fairness or stability). Other empirical studies have been primarily focused on medical datasets, and only evaluate missingness under \mcar ~\cite{mvi_review, Alabadla_review, farhangfar2008impact, shahbazian2023generative}. Further, most proposed methods only evaluate imputation quality, using metrics such as MAE, MSE, RMSE, and AUC~\cite{hasan_mvi_review, Joel_Review_Missing_Data_for_ML}, although some also evaluate overall predictor accuracy ~\cite{mvi_review}.}
Additionally, the performance of \mvi techniques under multi-mechanism missingness~\cite{zhou2024_Comprehensive_review} and missingness shifts~\cite{zhou2023_missingnessshift} remains unexplored in prior work, despite these conditions being more likely to occur in practice due to distribution shifts in production deployments~\cite{graham2009missing}.

Notably, overwhelming evidence in the literature indicates that there is no single ``best-performing'' \mvi approach on accuracy ~\cite{hasan_mvi_review, mvi_review,Shadbahr_nature_medicine,missing_data_review, farhangfar2008impact}, and that model performance (narrowly measured based on `correctness' thus far) depends on dataset characteristics such as size and correlation between variables~\cite{Alabadla_review} and missingness rates in the train and test sets~\cite{mvi_review, Shadbahr_nature_medicine}.

\paragraph{Fairness and missingness} There has been some recent interest in studying the social harm that can come from poorly chosen \mvi techniques~\cite{caton_imputation_fairness, zhang2022fairness, Zhang2021AssessingFI, zhang2021_imputationfairness, Jeong_fairness_without_imputation, Wang_selection_bias_fairness, martinez2019fairness,feng2024adapting}.  Most empirical studies~\cite{Wang_selection_bias_fairness, zhang2021_imputationfairness, zhang2022fairness, caton_imputation_fairness,jeong2022fairness} have worked with the COMPAS~\cite{larson2016analyzed} and Adult~\cite{Adult} 
datasets, the latter of which has been ``retired'' from community use due to issues with provenance~\cite{folktables_ding2021}. Further, these experimental studies employ randomly-generated missingness: usually by randomly sampling or using a fixed set of columns, and randomly picking rows in which to replace values with \nl. We critique this approach, since detecting and mitigating unfairness requires broader socio-technical thinking, such as having higher rates of missingness for minority groups and in features that are highly correlated with sensitive attributes (called \emph{proxy} variables in the fairness literature)~\cite{caton_imputation_fairness}. 

\rev{A notable exception is~\citet{martinez2019fairness}, who map social mechanisms such as prejudicial access and self-reporting bias to missingness categories like missing-by-design and item non-response. They also analyze feature correlations to study the effects of different missingness types. We adopt a similar methodology to simulate realistic missingness in this work but identify conceptual limitations in their fairness framing.   The authors state: ``The surprising result was to find that, [...] the examples with missing values seem to be fairer than the rest.'' \textbf{However, asserting that some rows of data are more or less fair is misguided}, as fairness is not a property of individual samples (\eg job applicants) but of the model (\eg in hiring decisions), which determines fairness through inclusion or exclusion in positive outcomes. We reinterpret their findings to suggest that excluding samples with missing values can increase model unfairness, reinforcing the case against deletion as a missing data strategy.}
 
\rev{\citet{zhang2021_imputationfairness} evaluate \mvi methods on \emph{imputation fairness}, defined as the difference in imputation accuracy between privileged and disadvantaged groups. They find that imputation unfairness increases with  higher missingness disparity, higher overall missingness rates, and greater data imbalance across groups.  Further, they find that varying missingness mechanisms for the same imputation method impacts prediction fairness. Their analysis is limited to randomly generated \nl values in COMPAS. We extend their work to additional datasets, missingness scenarios, and alternative imputation fairness definitions.}

\rev{In a follow-up work, \citet{zhang2022fairness} introduce \emph{imputation fairness risk} and provide bounds for ``correctly specified'' imputation methods. While this is a commendable theoretical contribution in a largely unexplored area, we question its broader implications: imputation quality metrics do not fully capture downstream model performance~\cite{Zhang2021AssessingFI}. In other words, a classifier can perform well despite poor imputation quality~\cite{Shadbahr_nature_medicine}. This raises a key question: Does minimizing imputation unfairness reduce model unfairness? Our empirical findings suggest it does not, as discussed in Section~\ref{sec:imputation-quality}.}

\rev{Finally,~\citet{jeong2022fairness} propose a decision tree-based method that integrates fairness into model training while handling missing values. Their approach splits only on observed values to mitigate disparities introduced by imputation. Their evaluation is limited to \mcar scenarios (with more missingness for disadvantaged groups).  In contrast, we assess more advanced MVI techniques under diverse missingness scenarios (\mcar, \mar, \mnar, and missingness shift) without applying fairness interventions. Nonetheless, we share their broader motivation of assessing and mitigating unfairness holistically throughout the data lifecycle. Future work could explore different combinations of MVI and fairness interventions.
}

\paragraph{Missingness and stability.} We are not aware of any work investigating the effect of missing value imputation on model stability. 

%% file: sections/benchmark.tex
\section{Benchmark Overview}
\label{sec:benchmark}

\subsection{Methodology for Simulating Missingness}
\label{sec:missingness}

We start with datasets in which there are no \nl values, and then simulate missingness.  We make this choice because we are interested in comparing \mvi performance under single-mechanism versus multi-mechanism missingness, and under missingness shifts, and, to the best of our knowledge, there are no datasets with naturally-occurring documented missingness of this form. 

Our methodology for simulating missingness is based on \textit{evaluation scenarios}, defined by the missingness mechanism during training and testing, shown in Table~\ref{tab:scenarios}: (1) single-mechanism missingness, injected similarly into train and test sets (S1 - S3); (2) single-mechanism missingness, injected differently into train and test sets (missingness shift) (S4 - S9); and (3) missingness is mixed, to include all three missingness mechanisms, and is injected similarly into train and test sets (S10).

\rev{For each dataset in our study (see Section~\ref{sec:datasets}), we designed socially-salient missingness scenarios corresponding to the three missingness mechanisms (\mcar, \mar, \mnar). Following~\cite{Joel_Review_Missing_Data_for_ML, martinez2019fairness}, we identified features for missing value injection, denoted by \( \mathcal{F}^m \), based on their Spearman correlation with the target variable and feature importance scores computed using scikit-learn. These selected features were chosen to reflect plausible missingness patterns. For instance, in the \diabetes dataset, while features like blood pressure or cholesterol levels are expected to be consistently observed, others, such as family history or physical activity, might be omitted or withheld due to privacy concerns or reporting biases. The remaining features, denoted by \( \mathcal{F}^c \), were considered complete, with no missing values.}

\rev{The three missing mechanisms share the same set of selected features (\( \mathcal{F}^m \)), but differ in their injection strategies. For \mcar, the missing values are randomly injected on \( \mathcal{F}^m \). In contrast, the missingness of \mar is based on sensitive attributes within \( \mathcal{F}^c \) to simulate pre-existing bias, as described in Section~\ref{sec:intro}. Specifically, higher rates of missingness were injected to disadvantaged groups wherever possible (in some cases there were too few samples from disadvantaged groups), reflecting realistic disparities caused by unequal access, distrust, or procedural injustice~\cite{akpinar2024impact}. Finally, for \mnar, the missingness is determined by missing values themselves, and the likelihood of missing values depends on the missing features.}

\rev{Table~\ref{tab:simulation-diabetes} presents the selected columns (\( \mathcal{F}^m \)) and injection conditions for the \diabetes dataset, based on the correlation coefficients and feature importance values in Figure~\ref{fig:diabetes-sim}. Additional information on other datasets is available in \arxiv{Appendix~\ref{apdx:extended-methodology}}\submit{full version~\cite{shades_arXiv}}.}

\subsection{Missing Value Imputation (\mvi) Techniques}
\label{sec:mvm-summary}

\rev{As discussed in Section~\ref{sec:related-work}, many competitive \mvi techniques have been proposed. We selected 15 of them, from 8 broad categories based on taxonomies presented in \cite{emmanuel2021survey, jager2021benchmark, miao2022experimental, wang2024missing}, namely: (1) \deletion; (2) statistical: \mode and \dummy; (3) machine learning-based:  \missforest~\cite{missForest} and \clustering\cite{gajawada2012missing_clustering}; (4) discriminative deep learning-based: \datawig~\cite{biessmann2019datawig} and  \automl~\cite{jager2021benchmark}; (5) generative deep learning-based: \gain~\cite{yoon2018gain} and  \hivae~\cite{nazabal2020handling}; (6) MNAR-specific: \notmiwae~\cite{ipsen2020not} and  \mnarpvae~\cite{ma2021identifiable}; (7) multiple imputation: \boostclean~\cite{krishnan2017boostclean}; and (8) other recent: \nomi~\cite{wang2024missing}, \tdm~\cite{zhao2023transformed}, and \editgain~\cite{miao2021efficient}. \arxiv{See Appendix~\ref{apdx:mvm-techniques} for details.}\submit{See full version~\cite{shades_arXiv}} for details.}

\input{tables/evaluation_scenarios}

\subsection{Evaluation Metrics}
\label{sec:benchmark:eval}

Following~\cite{mvi_review, hasan_mvi_review}, we evaluate \mvi techniques in two ways: directly using imputation quality metrics and indirectly based on downstream model performance.

\subsubsection{Imputation Quality} 
Shadbahr~\etal~\cite{Shadbahr_nature_medicine} report that distributional metrics capture downstream model performance better than classically-used discrepancy metrics. To confirm or refute this claim, we use a mix of both. To assess agreement with true values, we compute \emph{Root Mean Square Error (RMSE)} for numerical features and \emph{F1 score} for categorical features. To assess distributional alignment, we compute \emph{KL-divergence} (\ie the Shannon entropy) between the true and the predicted feature distributions, for both numerical and categorical features, measured for the imputed columns only as well as for the full dataset. For categorical features, we obtain the probability distributions using the \texttt{value\_counts} method with normalization from \texttt{pandas}. For numerical features, we use Gaussian kernel density estimation from \texttt{scipy}, with 1000 samples. Finally, to assess imputation fairness~\cite{zhang2021_imputationfairness, zhang2022fairness}, we compute \emph{F1 score difference}, \emph{RMSE difference}, and \emph{KL divergence difference} between privileged (\emph{priv}) and disadvantaged (\emph{dis}) groups.

\subsubsection{Model Performance} 

To assess the impact of \mvi techniques on model correctness, we report the \emph{F1 score} because it is a more reliable metric than accuracy for imbalanced data. 

For evaluating model stability, we report average \emph{Label Stability}~\cite{darling2018toward, khan2023fairness} over the full test set (closely related to the self-consistency metric from ~\citet{cooper2023arbitrariness}), computed per-sample for binary classification as \textit{Label Stability} = $\frac{|B_+ - B_-|}{B}$, where $B_+$ is the number of times the sample is classified into the positive class and $B_-$ is the number of times the sample is classified into the negative class, and $B=B_++B_-$ models are trained by bootstrapping over the train set. We set $B=50$ in all our experiments. 

Lastly, we report model fairness based on group-specific error rates, namely \emph{True Positive Rate Difference (TPRD)}, \emph{True Negative Rate Difference (TNRD)}, \emph{Selection Rate Difference} (SRD), and \emph{Disparate Impact (DI)}.  (Note that DI computes the ratio of selection rates, but we refer to it as DI as is standard in the literature~\cite{DBLP:conf/kdd/FeldmanFMSV15}.) 

\arxiv{$$ TPRD = \frac{TP_{dis}}{TP_{dis}+FN_{dis}} - \frac{TP_{priv}}{TP_{priv}+FN_{priv}}$$} 

\arxiv{$$ TNRD = \frac{TN_{dis}}{TN_{dis}+FP_{dis}} - \frac{TN_{priv}}{TN_{priv}+FP_{priv}}$$}

\arxiv{$$ SRD = \frac{TP_{dis}+FP_{dis}}{N_{dis}} - \frac{TP_{priv}+FP_{priv}}{N_{priv}}$$}

\arxiv{$$ DI=  \frac{TP_{dis}+FP_{dis}}{N_{dis}} \div \frac{TP_{priv}+FP_{priv}}{N_{priv}}$$}

Fairness metrics based on error rates align with formal equality of opportunity, while those based on selection rates (SRD, DI) reflect substantive equality~\cite{DBLP:conf/eaamo/KhanMS22}. The choice of metric depends on the domain and stakeholder~\cite{narayanan2018fairness}. SRD captures absolute disparities (\eg fixed quotas in college admissions), while DI measures relative disparities (\eg the 4/5th rule in U.S. hiring). From the individual’s perspective, TPRD suits opportunity allocation (\eg hiring, loans), ensuring access to positive outcomes, whereas TNRD applies to exclusion decisions (\eg medical diagnoses), emphasizing fairness in avoiding false negatives.

\input{sections/architecture}

\input{sections/workload}

\subsection{Model Types}
We evaluate predictive performance of 6 ML models: (i) decision tree (\texttt{dt$\_$clf}) with a tuned maximum tree depth, minimum samples at a leaf node, number of features used to decide the best split, and criteria to measure the quality of a split; (ii) logistic regression (\texttt{lr$\_$clf}) with tuned regularization penalty, regularization strength, and optimization algorithm; (iii) gradient boosted trees (\texttt{lgbm$\_$clf)} with tuned number of boosted trees, maximum tree depth, maximum tree leaves, and minimum number of samples in a leaf; (iv) random forest (\texttt{rf$\_$clf}) with a tuned number of trees, maximum tree depth, minimum samples required to split a node, and minimum samples at a leaf node (v) neural network, historically called the multi-layer perceptron (\texttt{mlp$\_$clf}) with two hidden layers, each with 100 neurons, and a tuned activation function, optimization algorithm, and learning rate; (vi) a deep table-learning method called GANDALF~\cite{joseph2022gandalf} (\texttt{gandalf$\_$clf)} with a tuned learning rate, number of layers in the feature abstraction layer, dropout rate for the feature abstraction layer, and initial percentage of features to be selected in each Gated Feature Learning Unit (GFLU) stage. \rev{Search grids of hyperparameters for all models are defined in our codebase.}

%% file: tables/evaluation_scenarios.tex
\begin{table}[t!]
\small 
\centering
\caption{Table~\ref{tab:scenarios}: Evaluation Scenarios}
\resizebox{0.9\columnwidth}{!}{ 
\begin{tabular}{c|ccc||ccc}
\hline & \multicolumn{3}{c}{Train} & \multicolumn{2}{c}{\quad\quad\quad\quad Test} \\
\hline
Scenario & \mcar & \mar & \mnar & \mcar & \mar & \mnar \\
\hline S1 & \cellcolor{green!25} $\checkmark$ & & &  \cellcolor{green!25} $\checkmark$ & & \\
S2 & &  \cellcolor{green!25} $\checkmark$ & & &  \cellcolor{green!25} $\checkmark$ & \\
S3 & & &  \cellcolor{green!25} $\checkmark$ & & &  \cellcolor{green!25} $\checkmark$ \\
S4 &  \cellcolor{green!25} $\checkmark$ & & & &  \cellcolor{green!25} $\checkmark$ & \\
S5 &  \cellcolor{green!25} $\checkmark$ & & & & &  \cellcolor{green!25} $\checkmark$ \\
S6 & &  \cellcolor{green!25} $\checkmark$ & &  \cellcolor{green!25} $\checkmark$ & & \\
S7 & &  \cellcolor{green!25} $\checkmark$ & & & &  \cellcolor{green!25} $\checkmark$ \\
S8 & & &  \cellcolor{green!25} $\checkmark$ &  \cellcolor{green!25} $\checkmark$ & & \\
S9 & & &  \cellcolor{green!25} $\checkmark$ & &  \cellcolor{green!25} $\checkmark$ & \\
S10 & \cellcolor{green!25} $\checkmark$ & \cellcolor{green!25} $\checkmark$ &  \cellcolor{green!25} $\checkmark$ & \cellcolor{green!25} $\checkmark$ &  \cellcolor{green!25} $\checkmark$ & \cellcolor{green!25} $\checkmark$ \\
\hline
\end{tabular}
}
\label{tab:scenarios}
\end{table}

%% file: sections/architecture.tex
\subsection{\sys Architecture}
\label{sec:benchmark-archicture}

\begin{figure}[t!]
    \centering
    \includegraphics[width=\columnwidth]{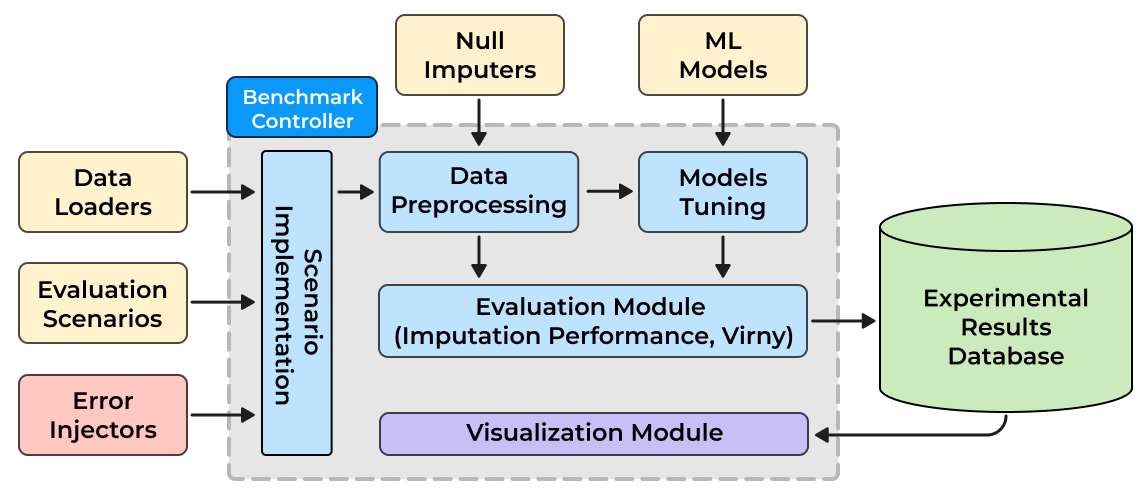}
    \caption{\sys architecture}
    \label{fig:benchmark-architecture}
\end{figure}

The architecture of \sys is shown in Figure~\ref{fig:benchmark-architecture}. Its core component, the \textit{benchmark controller}, executes user-specified missingness scenarios by applying error injectors to input datasets. It then imputes missing values using selected \mvi technique(s) and preprocesses data via standard scaling (numerical) and one-hot encoding (categorical), and then trains ML models with hyperparameter tuning. The evaluation module assesses imputation quality and model performance. For comprehensive profiling, it uses Virny~\cite{herasymuk2024responsible}, a Python library that computes accuracy, stability, and fairness metrics across multiple sensitive attributes and their intersections.

\begin{table}[b!]
\small 
    \centering
    \caption{\rev{Table~\ref{tab:dataset-info}: Dataset Information}}
    \begin{tabular}{llrrc}
        \toprule
        \textbf{name} & \textbf{domain} & \textbf{\# tuples} & \textbf{\# attrs} &\textbf{sensitive attrs} \\
        \midrule
        \diabetes & healthcare & 952 & 17 & sex\\
        \german & finance & 1,000 & 21 & sex, age\\
        \folk & finance & 15,000 & 10 & sex, race\\
        \law & education & 20,798 & 11 & sex, race\\
        \bank & marketing & 40,004 & 13 & age\\
        \heart & healthcare & 70,000 & 11 & sex \\
        \folkemp & hiring & 302,640 & 16 & sex, race\\
        \bottomrule
    \end{tabular}
    \label{tab:dataset-info}
\end{table}

\sys incorporates two optimizations to enhance experimental efficiency. First, it decouples missing value imputation from model training, allowing imputed datasets to be stored and reused in subsequent training and evaluation stages. Second, it supports simultaneous evaluation on multiple test sets (\eg with varying missingness rates or types), significantly reducing running time, and so executing a pipeline with one training set and multiple test sets takes about the same time as with a single test set.

%% file: sections/workload.tex
\subsection{Datasets and Tasks}
\label{sec:datasets}

\rev{As noted in Section~\ref{sec:intro}, we focus on \emph{socially salient} missingness. With this in mind, we selected seven datasets from diverse social decision-making contexts, including lending, hiring, marketing, admissions, and healthcare, summarized in Table~\ref{tab:dataset-info}.  Each dataset involves a binary classification task, where a positive label represents access to a desirable social good (e.g., education, employment, or healthcare). We chose these datasets to ensure broad coverage of (i) social domains, (ii) dataset sizes, and (iii) numerical-to-categorical column ratios.}
\arxiv{Dataset descriptions are deferred to Appendix~\ref{apdx:datasets}}

\begin{table*}[t!]
    \centering
    \small 
    \caption{Table~\ref{tab:simulation-diabetes}: Missingness scenarios for an error rate of 30\% for \diabetes. SoundSleep is a numerical column;  Family$\_$Diabetes, PhysicallyActive and RegularMedicine are categorical columns.}
    \vspace{-0.2cm}
    \begin{tabular}{lllll}
        \toprule
        \textbf{Mechanism} & \textbf{Missing Column ($\mathcal{F}^m)$} & \textbf{Conditional Column ($I$)} & \textbf{Pr($\mathcal{F}^m$ | $I$ is dis)} & \textbf{Pr($\mathcal{F}^m$ | $I$ is priv)} \\
        \midrule
        MCAR &\makecell[tl]{SoundSleep, Family$\_$Diabetes,\\ PhysicallyActive, RegularMedicine} & N/A & 0.3 & 0.3 \\
        \hline
        
        MAR & Family$\_$Diabetes, RegularMedicine & Sex & 0.2 (female) & 0.1 (male)\\
         & PhysicallyActive, SoundSleep & Age & 0.2 ($\geq 40$) & 0.1 ($<40$)\\
        \hline
         
         MNAR & Family$\_$Diabetes & Family$\_$Diabetes & 0.25 (yes) & 0.05 (no) \\
          & RegularMedicine & RegularMedicine & 0.2 (yes) & 0.1 (no)\\
          & PhysicallyActive & PhysicallyActive & \makecell[tl]{ 0.25 (none, $<\frac{1}{2}$ hour)} & \makecell[tl]{0.05 ($>\frac{1}{2}$ hour, $>1$ hour)} \\
          & SoundSleep & SoundSleep & 0.2 ($<5$) & 0.1 ($\geq 5$) \\
        \bottomrule
    \end{tabular}
    \label{tab:simulation-diabetes}
\end{table*}

\begin{figure*}[h!]
\begin{subfigure}[h]{0.275\linewidth}
    \centering
    \includegraphics[width=\linewidth]{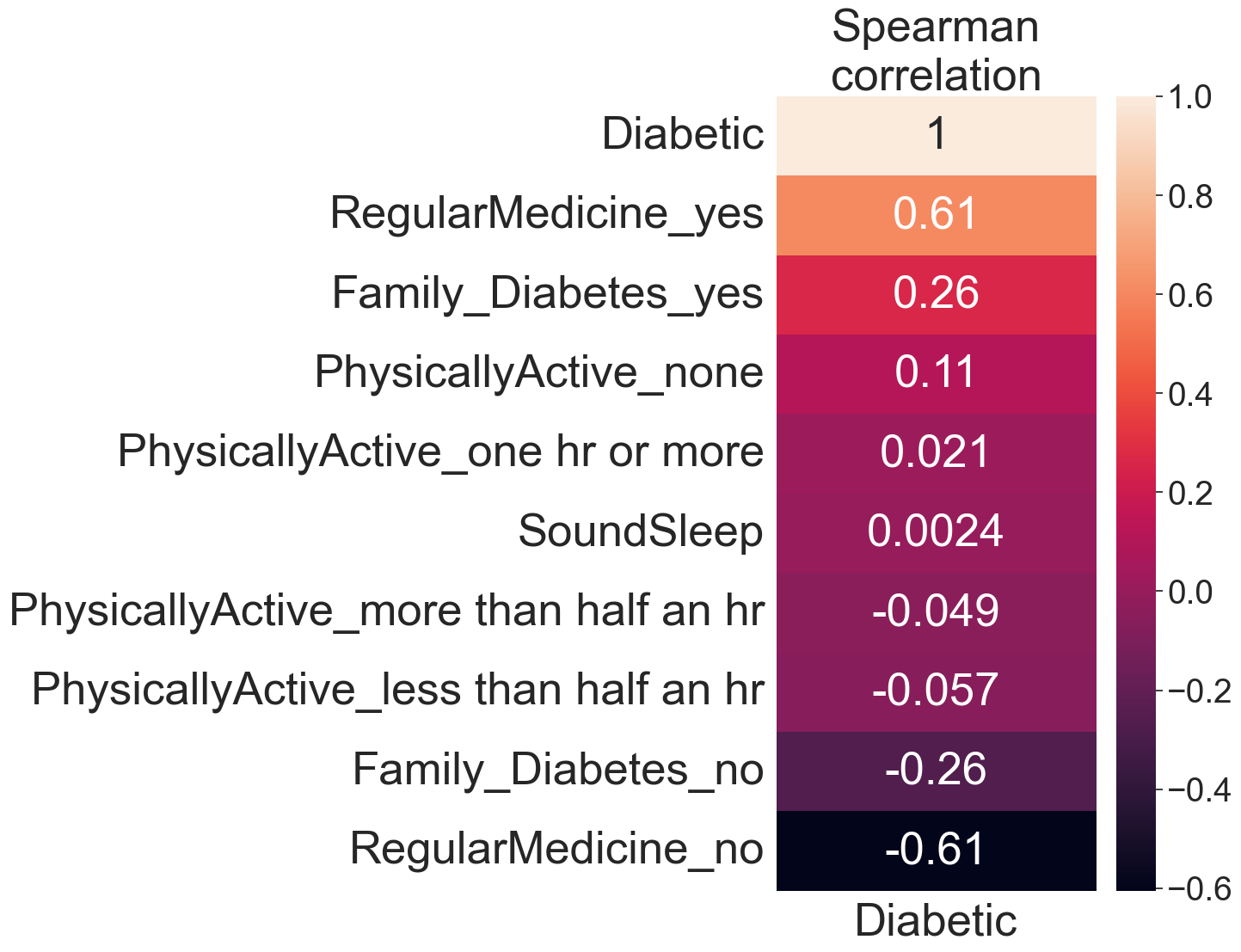}
    \caption{Correlation with label}
\end{subfigure}
\begin{subfigure}[h]{0.65\linewidth}
    \centering
    \includegraphics[width=\linewidth]{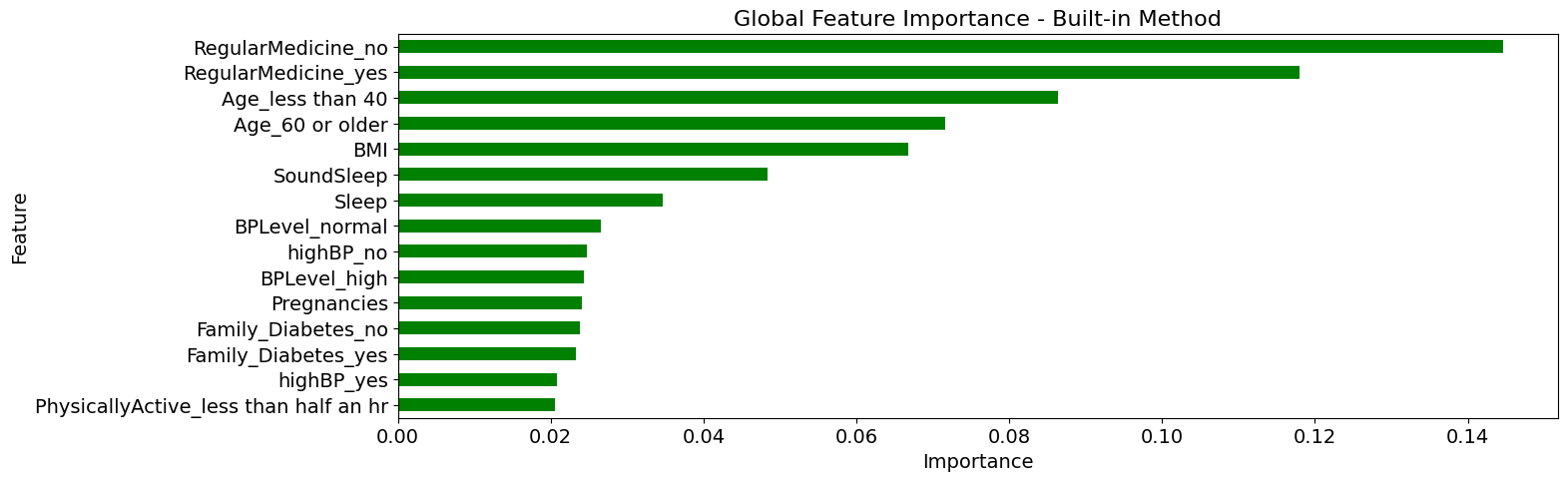}
    \caption{Feature importance}
\end{subfigure}
\caption{EDA for designing missingness scenarios in \diabetes.}
\label{fig:diabetes-sim}
\end{figure*}

%% file: sections/combined.tex
\section{Single and Multi-Mechanism Missingness}
\label{sec:exp1-single}

To simulate single-mechanism missingness (S1-S3 in Table~\ref{tab:scenarios}) we inject 30\% of each training and test sets with nulls, according to the missingness scenarios described in Section~\ref{sec:missingness}. For multi-mechanism or mixed missingness,  when \mcar, \mar and \mnar co-exist (S10 in Table~\ref{tab:scenarios}), we inject 10\% of nulls for each of the three mechanism into both training and test sets, for a total of 30\% nulls. 

To evaluate model correctness, we report results for F1, \arxiv{see  Appendix~\ref{apdx:single-mechanism-fairness-metrics}}\submit{see full version~\cite{shades_arXiv}} for accuracy results. 
For fairness, we use binary group definitions. For datasets with two sensitive attributes, we define the doubly-disadvantaged group as disadvantaged (\emph{dis}) and everyone else as privileged (\emph{priv}). For example: on the \law, \folk and \folkemp datasets, non-White women are the \emph{dis} group, and White women, non-White men and White men are the \emph{priv} group. We report results for TPRD, \arxiv{and defer results for other fairness metrics to Appendix~\ref{apdx:single-mechanism-fairness-metrics}}\submit{see results for other fairness metrics in the full version of the paper~\cite{shades_arXiv}}.
For stability, we used a bootstrap of 50 estimators, each seeing a random 80\% of the training set~\citep{efron1994bootstrap}. Higher values of F1 and label stability are better, and values of TPRD close to zero are better.

\begin{figure}[t!]
\begin{subfigure}[h]{\linewidth}
    \includegraphics[width=\linewidth]{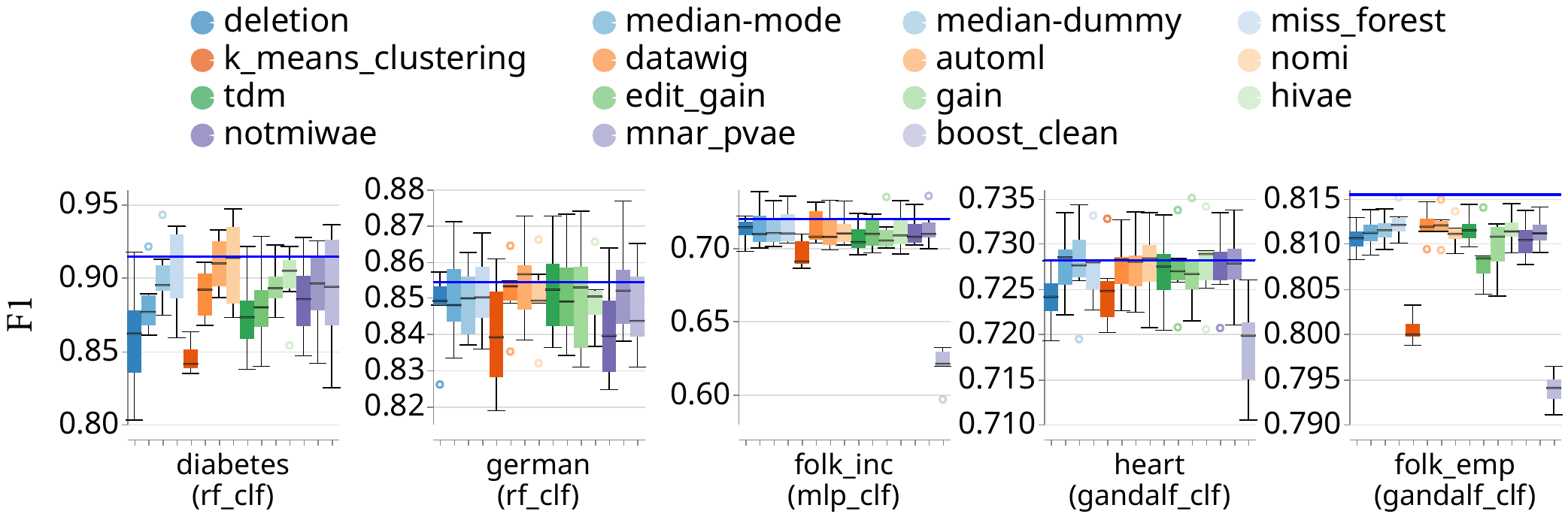}
    \caption{Missing Completely At Random (\mcar)}
\end{subfigure}

\begin{subfigure}[h]{\linewidth}
    \includegraphics[width=\linewidth]{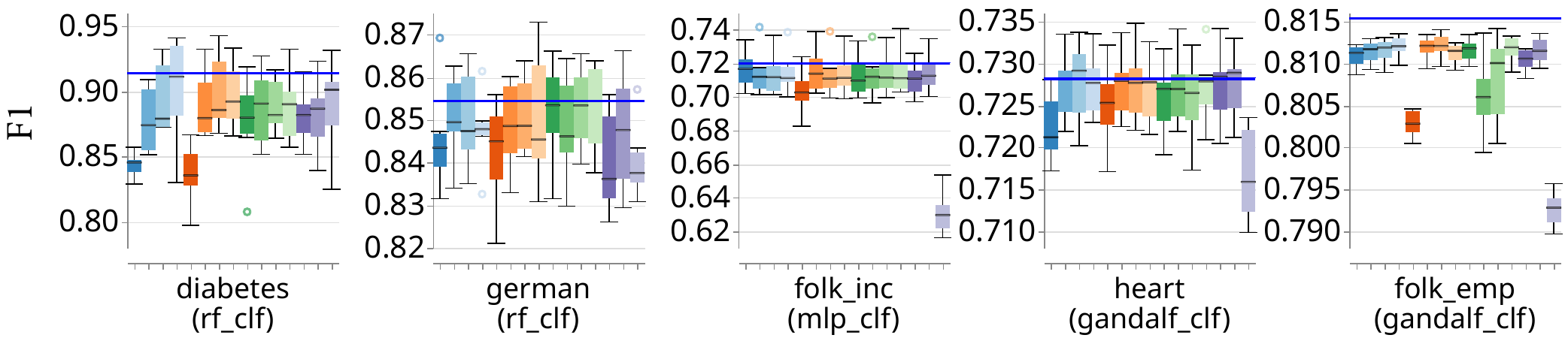}
    \caption{Missing At Random (\mar)}
\end{subfigure}

\begin{subfigure}[h]{\linewidth}
    \includegraphics[width=\linewidth]{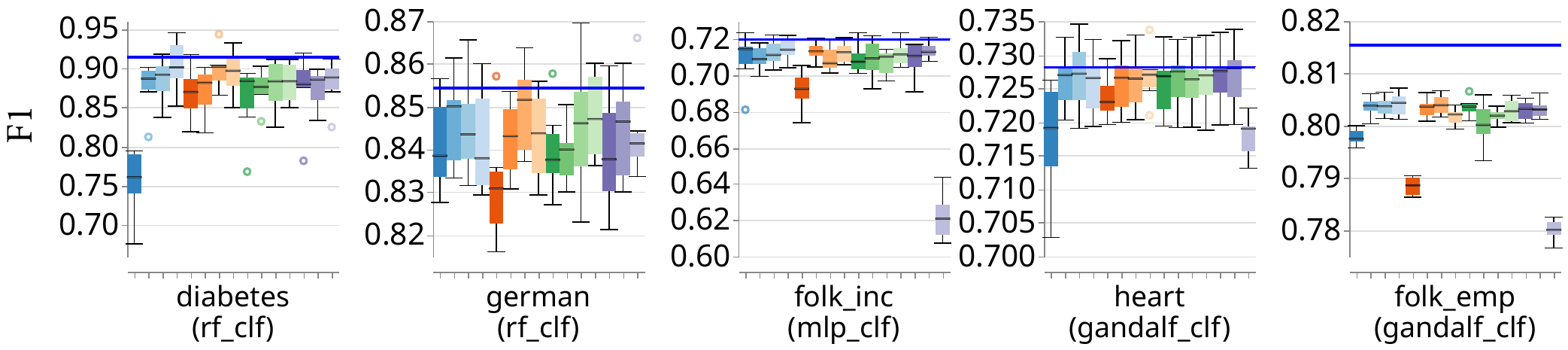}
    \caption{Missing Not At Random (\mnar)}
\end{subfigure}

\begin{subfigure}[h]{\linewidth}
    \centering
    \includegraphics[width=\linewidth]{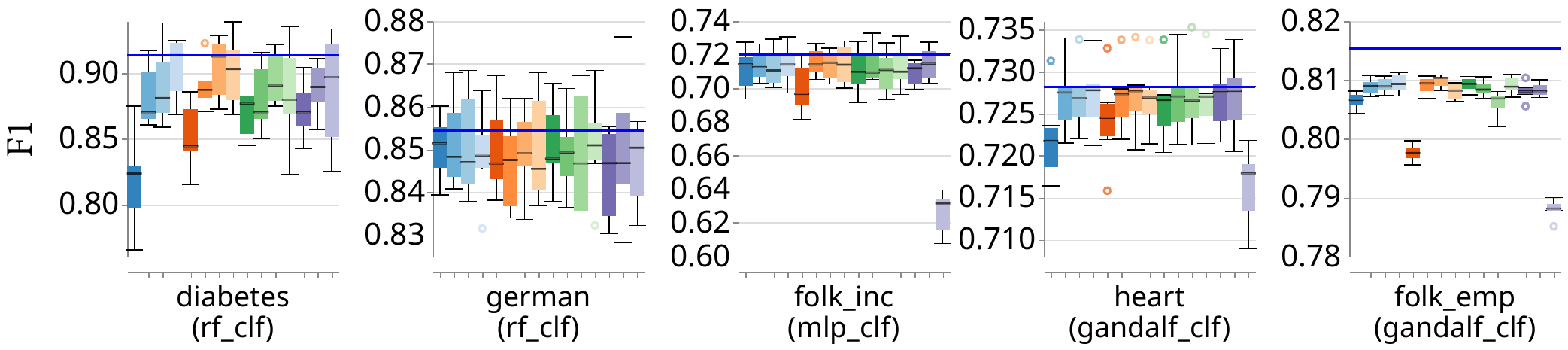}
    \caption{Mixed missingness (\mcar \& \mar \& \mnar)}
\end{subfigure}
\caption{\revtwo{F1 of best performing models (shown in figure) for different imputation strategies (colors in the legend), datasets (x-axis), and missingness mechanisms (subplots). Datasets are in increasing order by size. Blue line shows median performance of the model trained on clean data.}}
\label{fig:exp1-F1}
\end{figure}

\begin{figure}[b!]
\begin{subfigure}[h]{\linewidth}
    \includegraphics[width=\linewidth]{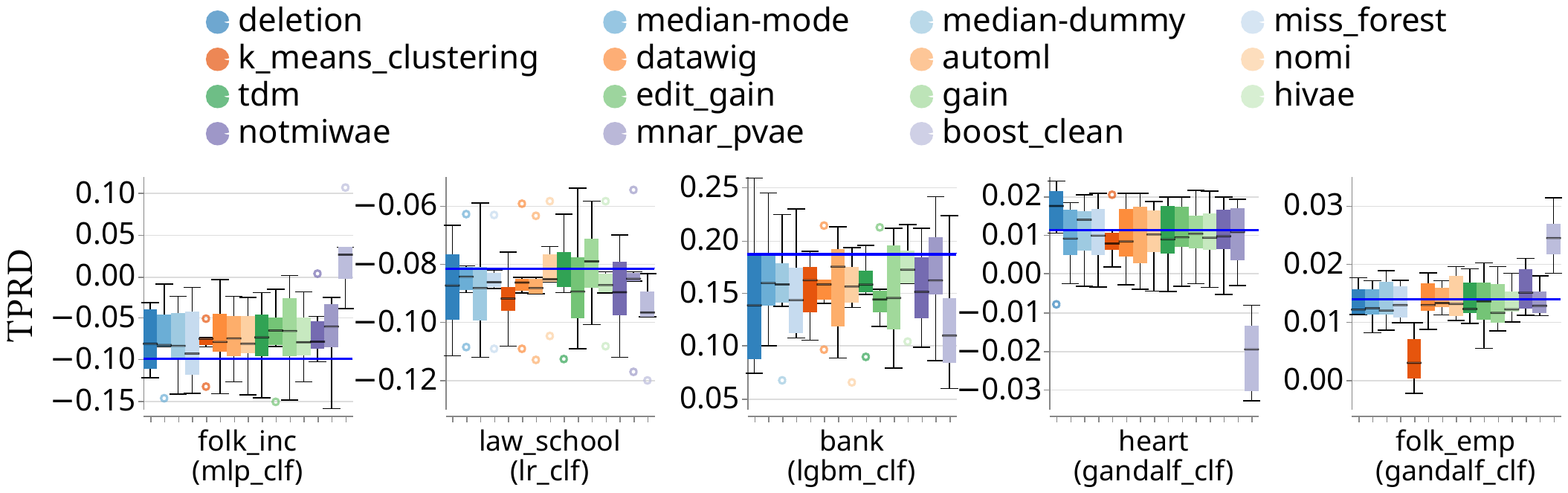}
    \caption{Missing Completely At Random (\mcar)}
\end{subfigure}

\begin{subfigure}[h]{\linewidth}
    \includegraphics[width=\linewidth]{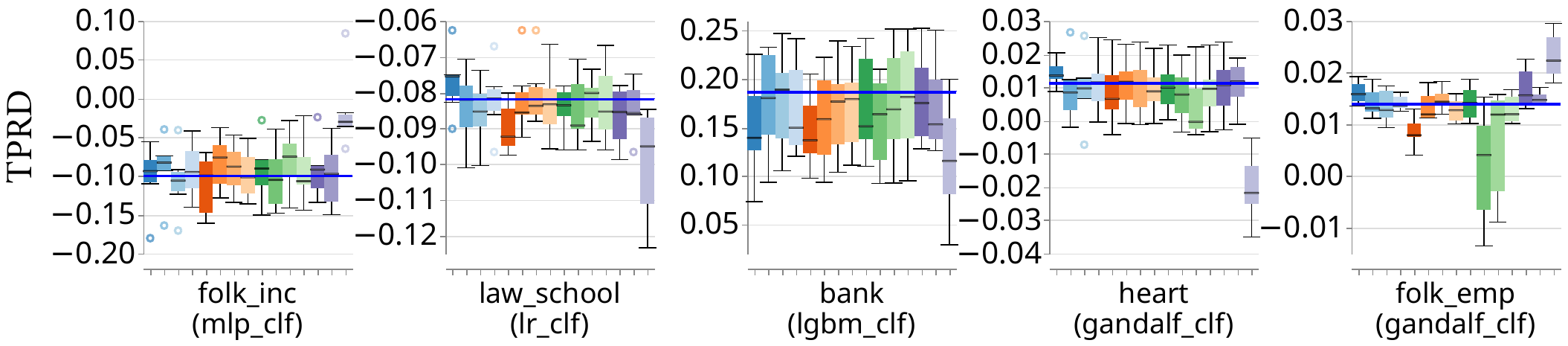}
    \caption{Missing At Random (\mar)}
\end{subfigure}

\begin{subfigure}[h]{\linewidth}
    \includegraphics[width=\linewidth]{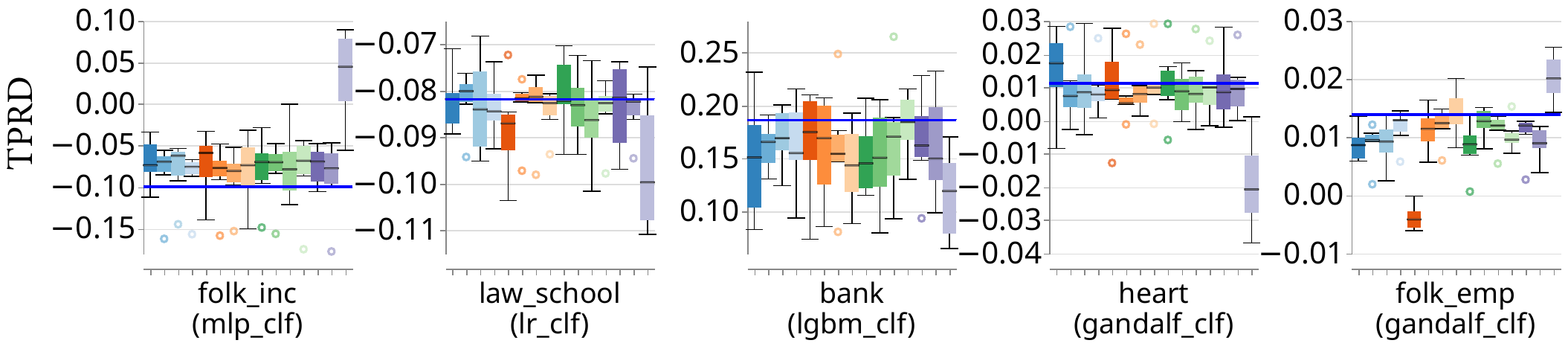}
    \caption{Missing Not At Random (\mnar)}
\end{subfigure}

\begin{subfigure}[h]{\linewidth}
    \includegraphics[width=\linewidth]{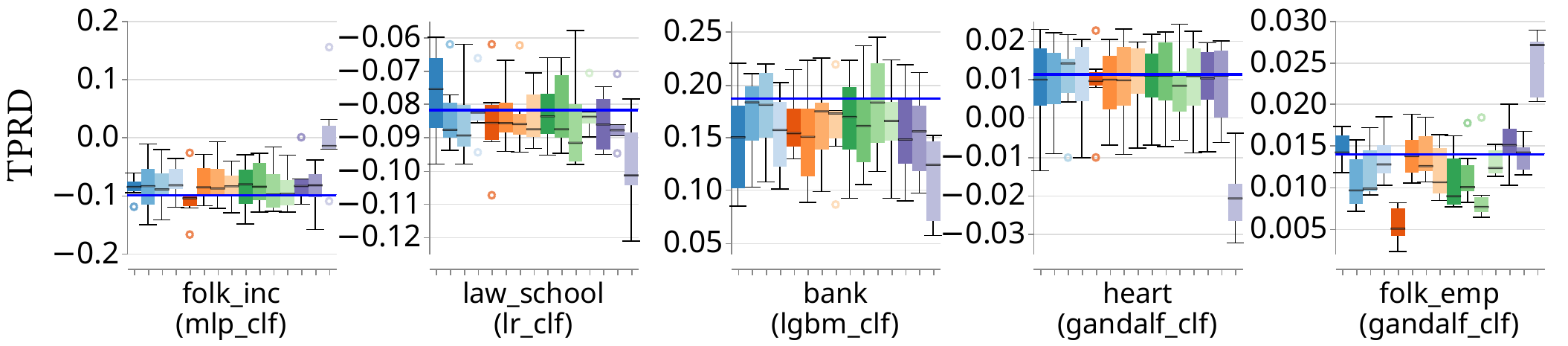}
    \caption{Mixed missingness (\mcar \& \mar \& \mnar)}
\end{subfigure}

\caption{\revtwo{True Positive Rate Difference (unfairness) of best performing models (shown in figure) for different imputation strategies (colors in the legend), datasets ($x$-axis), and missingness mechanisms (subplots). Values close to 0 are desirable. Datasets are in increasing order by size. Blue line shows median TPRD of the model trained on clean data.}}
\label{fig:exp1-TPR}
\end{figure}

\revtwo{Different models are the best-performing on different datasets.  In Figures~\ref{fig:exp1-F1},~\ref{fig:exp1-TPR} and \ref{fig:exp1-label-stability}, we report on the best-performing models (according to F1) for five most representative datasets per experiment, and compare performance against a model trained on clean data.  Complete results are available in \arxiv{Appendix~\ref{apdx:single-mechanism-fairness-metrics}}\submit{the full version~\cite{shades_arXiv}}}. 

\subsection{Correctness of the Predictive Model}
\label{sec:exp1-single-f1}

Figure~\ref{fig:exp1-F1} shows the F1 of models trained with different \mvi techniques. We find interesting trends in \mvi performance based on characteristics of the dataset and missingness type. All techniques are competitive for all missingness mechanisms, including mixed missingness, on \heart and \law. \boostclean, which uses multiple imputation (MI), is otherwise only competitive on small datasets (\diabetes and \german), and only under \mcar and mixed missingness on \german. 
None of the \mvi techniques are able to match the F1 of the model trained on clean data on \folkemp, and this effect is strongest under \mnar (notably, stronger than under mixed missingness). \boostclean shows particularly poor performance on \folk, with a 0.08 decrease in F1 compared to other methods, for all missingness types. We discuss this unexpected performance of MI further in Section~\ref{sec:conclusion}.  

\automl, \datawig and \missforest are generally the best performing \mvi techniques, \rev{with \nomi a close second, offering an optimal balance between imputation accuracy and training time (see \submit{the full paper~\cite{shades_arXiv}}\arxiv{Section~\ref{sec:time} and Appendix~\ref{apdx:mvi-metrics}} for training time analysis).} However, simpler statistical techniques (\eg \mode under \mar) are also competitive. This underscores the need to evaluate novel DL-based and ML-based methods holistically (\eg on a variety of missingness scenarios) to ensure that they justify the additional training overhead and complexity they introduce compared to simple methods.

\rev{Interestingly, \notmiwae and \mnarpvae do not demonstrate superior performance compared to other methods in our socially salient MNAR scenario. Instead, their performance aligns closely with other leading MVI approaches under MNAR conditions. This finding is further discussed in Section~\ref{sec:conclusion}.}

In line with conventional wisdom~\cite{rubin1987statistical, martinez2019fairness, little2015modeling, Joel_Review_Missing_Data_for_ML}, we find that deletion worsens predictive performance.  This effect is strongest for small datasets like \diabetes, with F1 decreasing as much as 0.1 under \mnar, compared to the model trained on clean data. \rev{This is due to \deletion discarding useful information, whereas retaining rows with nulls can still provide valuable signal for model training.}

The F1 score on the \bank dataset is low (0.32), due to severe class imbalance (base rate 0.117, see \submit{full version of the paper~\cite{shades_arXiv}).}\arxiv{Table~\ref{tab:bank-rates} in Appendix~\ref{apdx:exp-details-additional}). Figure~\ref{fig:exp1-accuracy} in Appendix~\ref{apdx:single-mechanism-fairness-metrics} shows that \texttt{lgbm$\_$clf} achieves 0.89 accuracy, slightly outperforming Logistic Regression (0.885) reported in~\cite{le2022survey} in Table 15.} \revtwo{Interestingly, models trained on imputed data can sometimes outperform those trained on clean data, as seen for \german and \heart under \mcar. We hypothesize this occurs when  models trained on cleaned data learns spurious correlations (\eg from noisy or erroneous values), while MVI methods may impute more accurate values, mitigating such artifacts and improving performance.}

\subsection{Fairness of the Predictive Model}
\label{sec:exp1-single-model-fairness}

Figure~\ref{fig:exp1-TPR} shows the effect of \mvi on fairness, according to TPRD. ~\citet{Wang_selection_bias_fairness} posit that models will exhibit more unfairness under \mar and \mnar compared to \mcar, but we only find weak empirical evidence towards this, even for \deletion. A nuance here is that we designed missingness scenarios, described in Section~\ref{sec:missingness}, to be realistic --- including \mar scenarios where people from disadvantaged groups withhold information that could hurt their chances of getting the desired outcome. Hence, dropping these rows can in fact improve fairness under \mar and \mnar, as observed on \bank. 

In contrast to~\citet{Wang_selection_bias_fairness}, we find that the effect of \mvi on fairness is strongly correlated with fairness of the model trained on clean data, corroborating the findings of ~\citet{guha_icde}. Fairness depends on two things: dataset characteristics and model type. All \mvi techniques except for \boostclean have the same model type as the model trained on clean data (because they are impute-then-classify approaches) and generally preserve fairness of that model, under all missingness mechanisms. Notably, this is agnostic to whether the TPRD of the model trained on clean data is low (close to 0.01 on \heart and \folkemp, and 0 on \german) or high (close to -0.1 on~\folk and 0.2 on \bank).

On the other hand, \boostclean, is a joint data cleaning and model training approach and thereby constitutes it own model type, and shows fairness trends that deviate from the model trained on clean data. \boostclean significantly improves fairness on\\ \folk (TPRD close to 0, compared to -0.1 for the clean model) and \bank (TPRD close to 0.1, compared to 0.2 for the clean model), but marginally worsens fairness on \law (TPRD -0.1 compared to -0.08 for the clean model) and \heart (TPRD -0.02 compared to 0.01 for the clean model), for all missingness types.

\begin{figure}[t!]
\begin{subfigure}[h]{\linewidth}
    \centering
    \includegraphics[width=\linewidth]{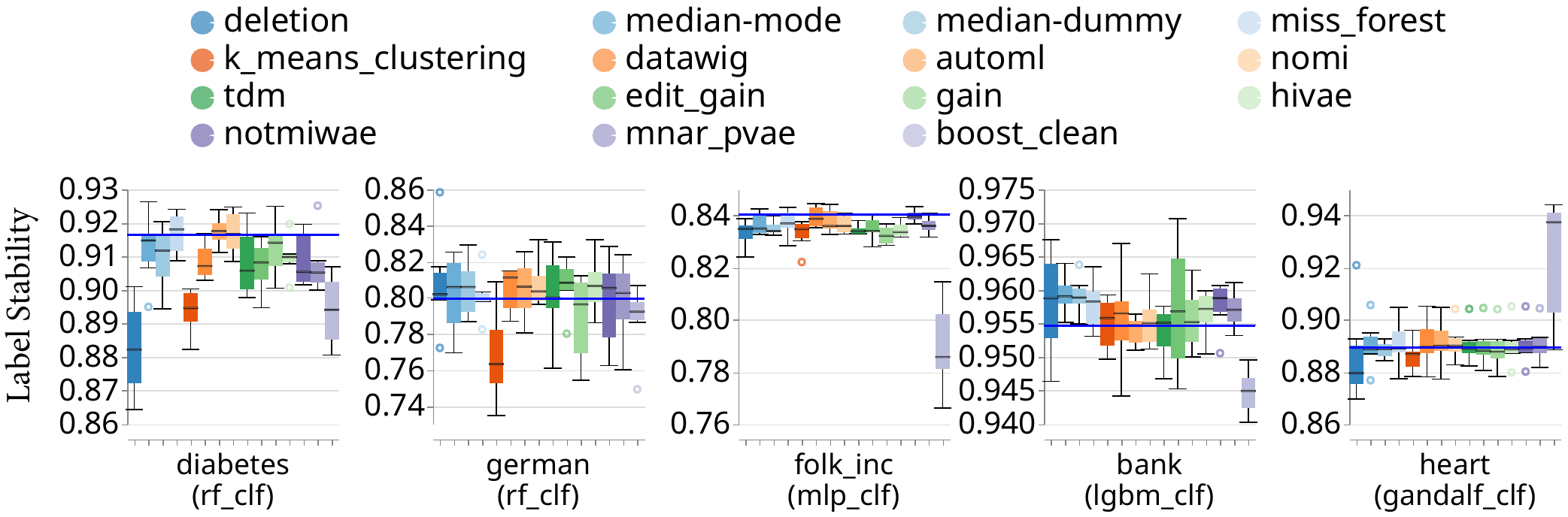}
    \caption{Missing Completely At Random (\mcar)}
\end{subfigure}

\begin{subfigure}[h]{\linewidth}
    \centering
    \includegraphics[width=\linewidth]{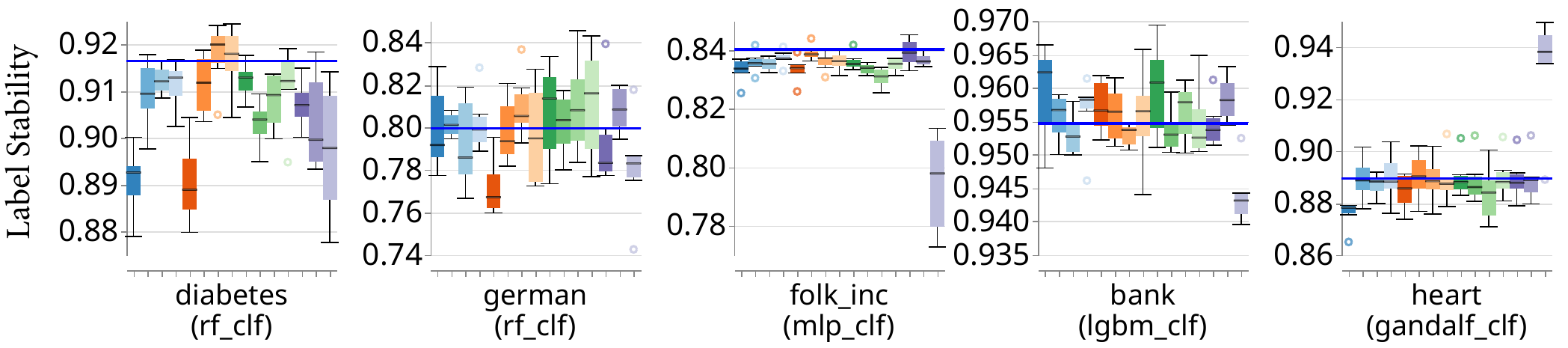}
    \caption{Missing At Random (\mar)}
\end{subfigure}

\begin{subfigure}[h]{\linewidth}
    \centering
    \includegraphics[width=\linewidth]{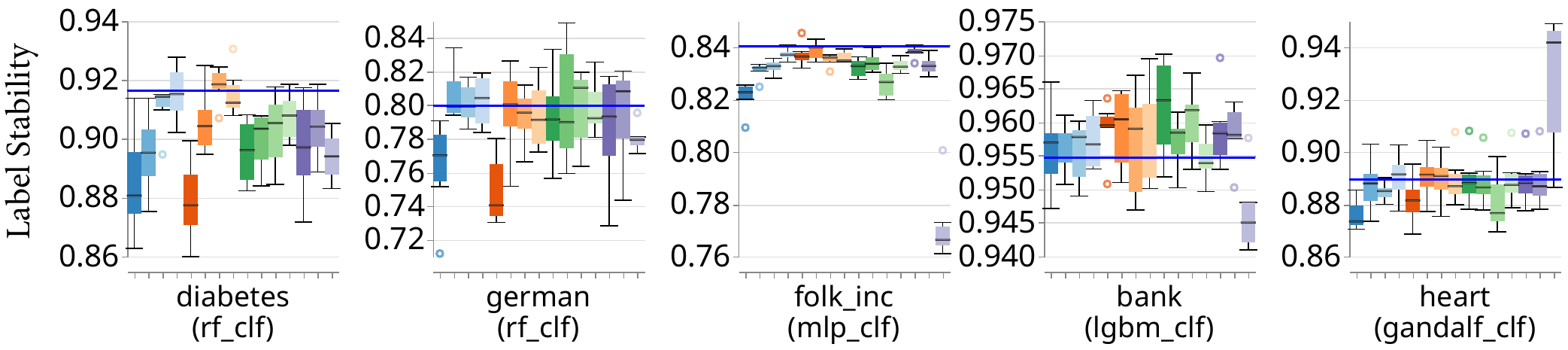}
    \caption{Missing Not At Random (\mnar)}
\end{subfigure}

\begin{subfigure}[h]{\linewidth}
    \centering
    \includegraphics[width=\linewidth]{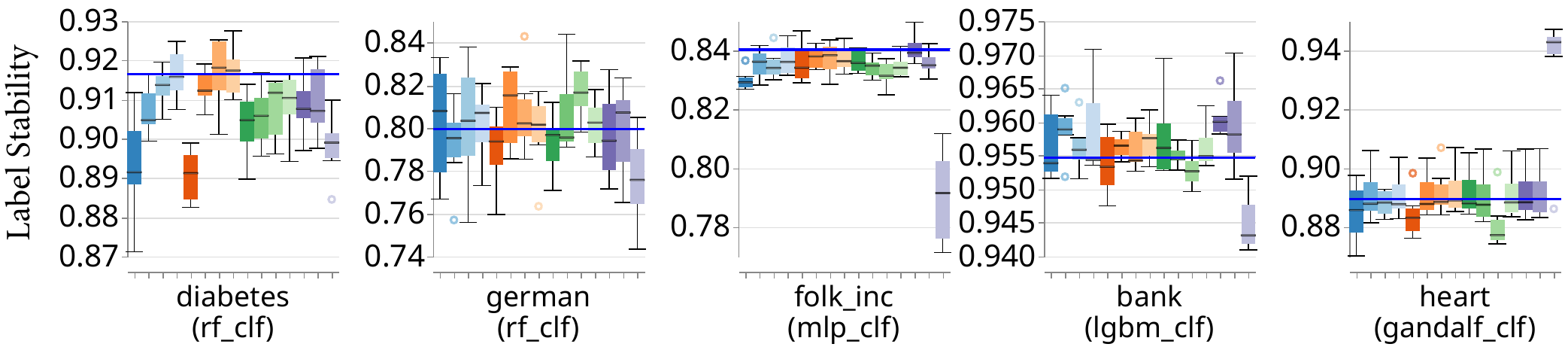}
    \caption{Mixed missingness (\mcar \& \mar \& \mnar)}
\end{subfigure}

\caption{\revtwo{Label Stability of best performing models for different imputation strategies (colors in the legend), datasets (x-axis), and missingness mechanisms (subplots). Values close to 1 are desirable. 
Blue line shows median performance of the model trained on clean data.}}
\label{fig:exp1-label-stability}
\end{figure}

\subsection{Stability of the Predictive Model}
\label{sec:exp1-single-model-stability}


Figure~\ref{fig:exp1-label-stability} shows label stability of models trained with different \mvi techniques. In line with conventional wisdom~\cite{unified_decomposition}, stability depends primarily on dataset characteristics (especially size) and model type. Impute-then-classify methods like \missforest, \automl, and \datawig, which perform best on F1, also match the stability of models trained on clean data across missingness types and dataset sizes. \rev{\nomi, which performs best on accuracy and training time, shows comparable stability to these top methods across datasets.} \clustering, which performed poorly on F1, is likewise unstable on small datasets (\diabetes and \german, with 905 and 1k samples, respectively). A notable exception is \german under mixed missingness, where \clustering is competitive despite underperforming on \mcar, \mar, or \mnar individually. \rev{This may be because imputation accuracy affects data uncertainty, which in turn drives model uncertainty~\cite{gal2016uncertainty}.}

Deletion worsens stability compared to the model trained on clean data for all missingness types on \diabetes, but, notably, only under \mnar on \german. 
\boostclean, which constitutes its own model type, shows a deviation from the stability of the model trained on clean data on all datasets except \bank: worsening stability compared to the clean model on \folk, \law and \folkemp (except under \mcar), but, surprisingly, improving it on \heart, even under mixed missingness.

\input{sections/imputation_quality}

%% file: sections/imputation_quality.tex
\begin{figure}[b!]
\begin{subfigure}[h]{\linewidth}
    \centering
    \includegraphics[width=\linewidth]{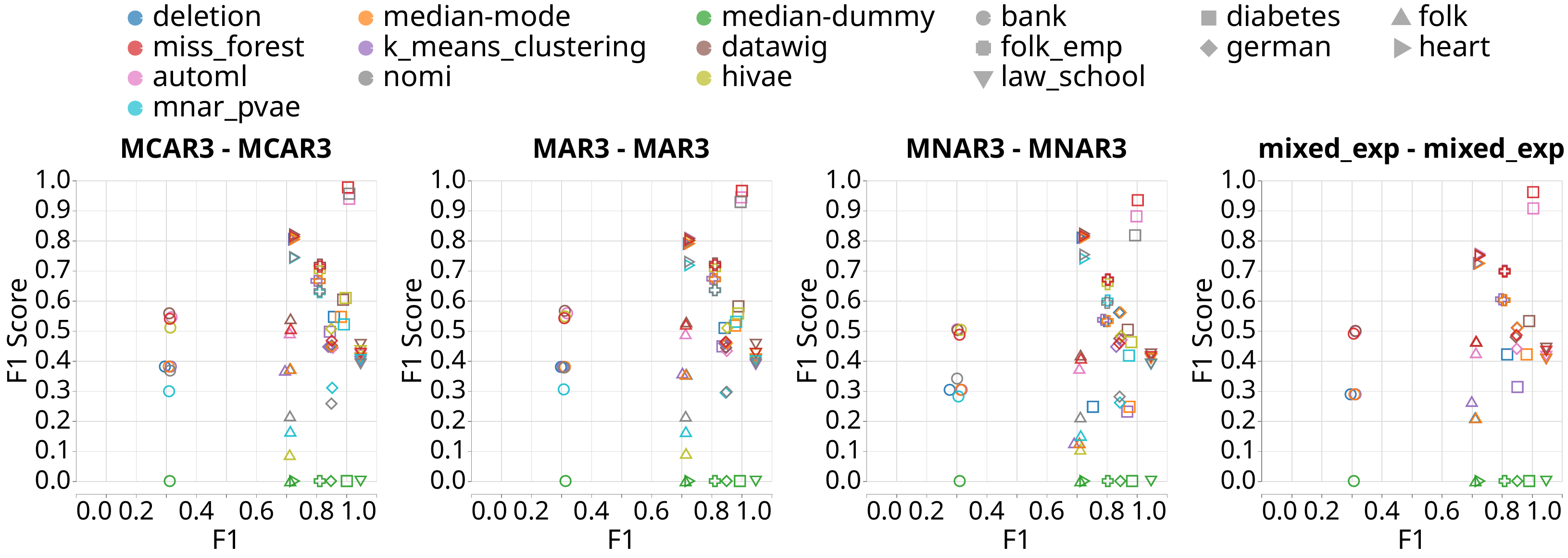}
    \caption{F1 (imputation) vs F1 (model)}
    \label{fig:imputation-f1}
\end{subfigure}

\begin{subfigure}[h]{\linewidth}
    \centering
    \includegraphics[width=\linewidth]{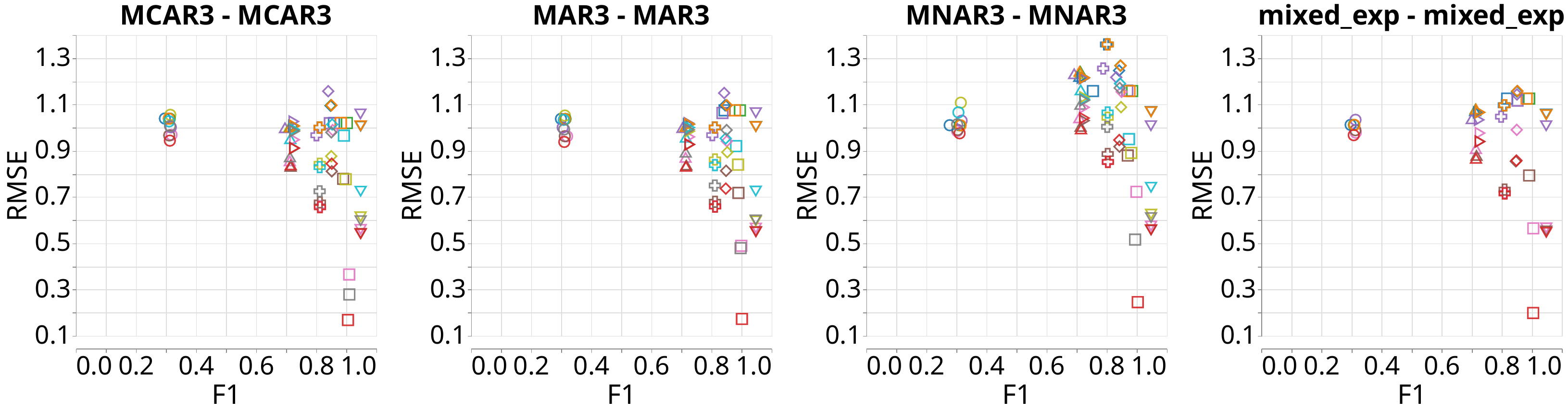}
    \caption{RMSE (imputation) vs F1 (model)}
    \label{fig:imputation-rmse}
\end{subfigure}

\begin{subfigure}[h]{\linewidth}
    \centering
    \includegraphics[width=\linewidth]{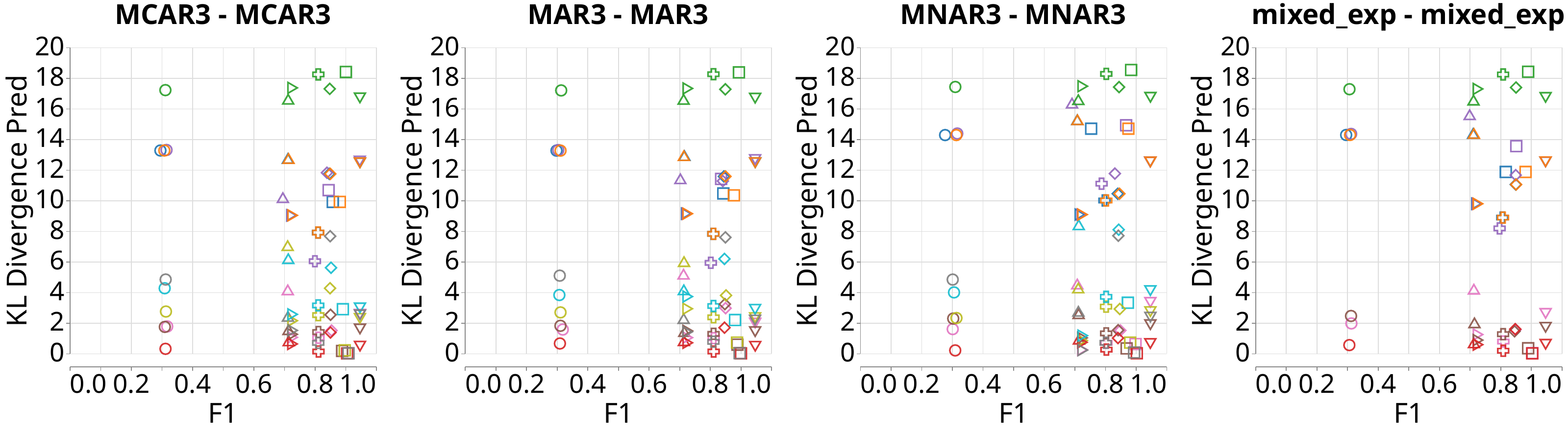}
    \caption{KL divergence (imputation) vs F1 (model)}
    \label{fig:imputation-kld}
\end{subfigure}

\caption{\revtwo{Imputation quality vs. model performance: imputation correctness (F1, RMSE and KL divergence) may not be indicative of model correctness (F1).}}
\label{fig:imputation-quality-no-shift}
\end{figure}

\subsection{Imputation Quality and Fairness}
\label{sec:imputation-quality}

In Figure~\ref{fig:imputation-quality-no-shift}, we report the imputation quality of \rev{10 most accurate \mvi techniques per category} according to F1 score (for categorical columns), RMSE (for numerical columns), and KL divergence (for both numerical and categorical columns, computed over the columns with nulls only) and compare it with the F1 of the downstream model. \rev{\arxiv{See  Appendix~\ref{apdx:single-mechanism-additional}}\submit{See full version~\cite{shades_arXiv}} for an extended comparison of training times and accuracy across all MVI techniques.}

\emph{Imputation Quality.} \mvi techniques with widely varying imputation quality can yield models with similar F1, suggesting that imputation correctness is not a reliable predictor of downstream performance~\cite{Zhang2021AssessingFI}. \rev{For instance, in Figure~\ref{fig:imputation-f1} on \diabetes, \dummy has imputation F1 near 0; \automl, \missforest, \rev{and \nomi} are near 1; others fall between 0.5–0.6, yet all produce models with F1 close to 1.} This pattern holds across datasets, missingness types, and for numerical columns (Figure~\ref{fig:imputation-rmse}). Similar trends are observed for KL divergence (Figure~\ref{fig:imputation-kld}), reinforcing that neither discrepancy-based nor distributional metrics reliably predict downstream model performance, contradicting \citet{Shadbahr_nature_medicine}'s claim.

\begin{figure}[b!]
\begin{subfigure}[h]{\linewidth}
    \includegraphics[width=\linewidth]{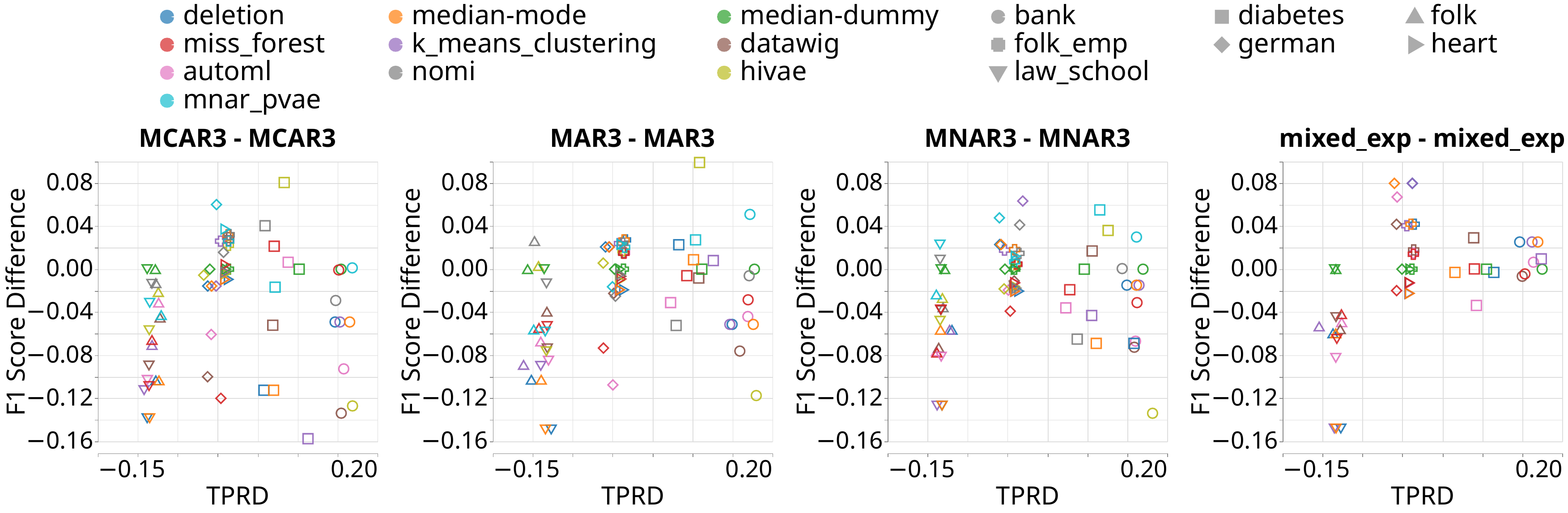}
    \caption{F1 score difference (imputation) vs TPRD (model)}
    \label{fig:imputation-f1-diff}
\end{subfigure}

\begin{subfigure}[h]{\linewidth}
    \includegraphics[width=\linewidth]{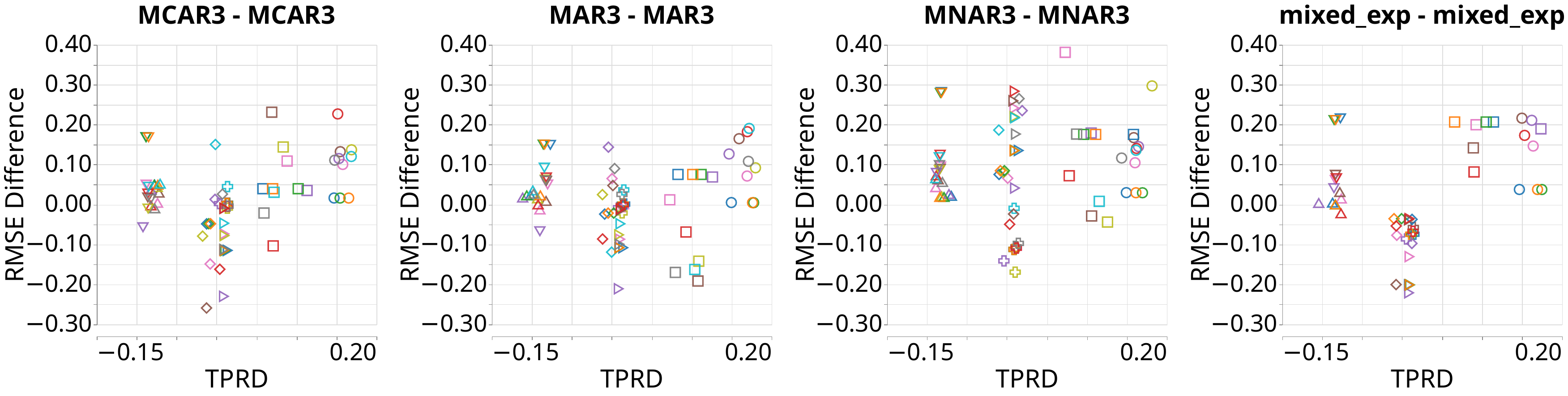}
    \caption{RMSE difference (imputation) vs TPRD (model)}
    \label{fig:imputation-rmse-diff}
\end{subfigure}

\begin{subfigure}[h]{\linewidth}
    \includegraphics[width=\linewidth]{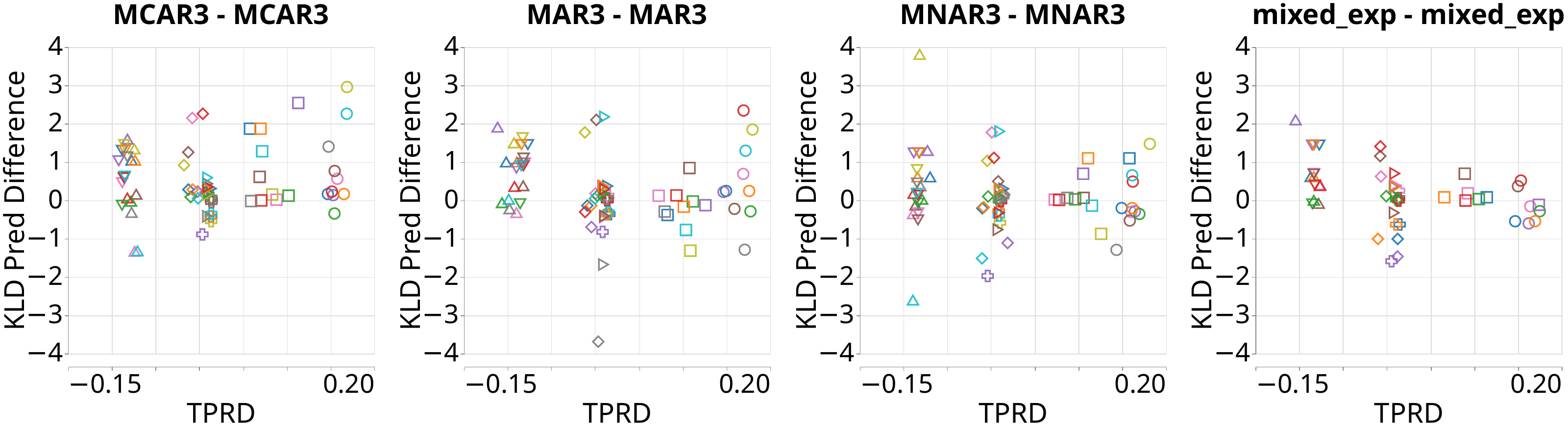}
    \caption{KL divergence difference (imputation) vs TPRD (model)}
    \label{fig:imputation-kld-diff}
\end{subfigure}

\caption{\revtwo{Imputation fairness vs. model fairness: imputation fairness (F1 difference, RMSE difference and KL divergence difference) may not be indicative of model fairness (TPRD).}}
\label{fig:imputation-fairness-no-shift}
\end{figure}

\emph{Imputation Fairness.} In Figure~\ref{fig:imputation-fairness-no-shift}, we report the imputation fairness of different \mvi techniques, according to F1 score difference (for categorical columns), RMSE difference (for numerical columns), and KL divergence difference (for both numerical and categorical columns, computed over the columns with nulls only) and compare it with the fairness of the downstream model, according to TPRD.  We find that, while model fairness is generally agnostic to missingness type (as discussed in Section~\ref{sec:exp1-single-model-fairness}), \textbf{imputation fairness is highly sensitive to missingness mechanism}. For example, in Figure~\ref{fig:imputation-kld-diff}, \missforest has good imputation fairness on \german under \mar (KL difference of -0.4) but significant imputation unfairness under \mcar (KL difference of 2.25), \mnar (KL difference of 1.1) and mixed missingness (KL difference of 1.4). 

Further, \textbf{imputation fairness is insufficiently predictive of model fairness}. For example, on \german under mixed missingness, \dummy has KL difference close to 0, \datawig and \missforest have KL difference between 1 and 1.5, \clustering and \mode have KL difference between -1 and -1.5, but the models trained using all five of these techniques have a TPRD close to -0.02. Conversely, on \diabetes under \mar, \automl, \clustering, \missforest, and \dummy all have near perfect imputation fairness (KL difference close to 0), but different model fairness (TPRD between 0.04 and 0.12). We see similar trends for other imputation fairness metrics such as F1 score difference (Figure~\ref{fig:imputation-f1-diff}) and RMSE difference (Figure~\ref{fig:imputation-rmse-diff}).

%% file: sections/shift.tex
\section{Missingness Shift}
\label{sec:shift}

Next, we evaluate the correctness, fairness, and stability of \rev{10 most effective \mvi techniques from various categories} under missingness shift. We simulate missingness shift in two ways: (i) by varying the missingness mechanism between training and test (S4-9 in Table~\ref{tab:scenarios}); and (ii) by varying the missingness rates between training and test. First, we hold the fraction of nulls in the test set constant (at 30\%) and vary the fraction of nulls in the training set (10\%, 30\% and 50\%). Then, we hold the fraction of nulls in the training set constant (at 30\%) and vary the fraction of nulls in the test set ($10\%$, $20\%$, $30\%$, $40\%$ and $50\%$). Note that we have fewer settings for training missingness rates because varying the test set is less computationally demanding (as discussed in Section~\ref{sec:benchmark-archicture}).
We discuss results on \diabetes, and defer results on other datasets, with fixed and variable training and test missingness rates, to \arxiv{Appendix~\ref{apdx:shift-additional}}\submit{the full version of the paper~\cite{shades_arXiv}}.

\subsection{Correctness of the Predictive Model}
\label{sec:shift-f1}

\begin{figure}[t!]
    \centering
    \includegraphics[width=\linewidth]{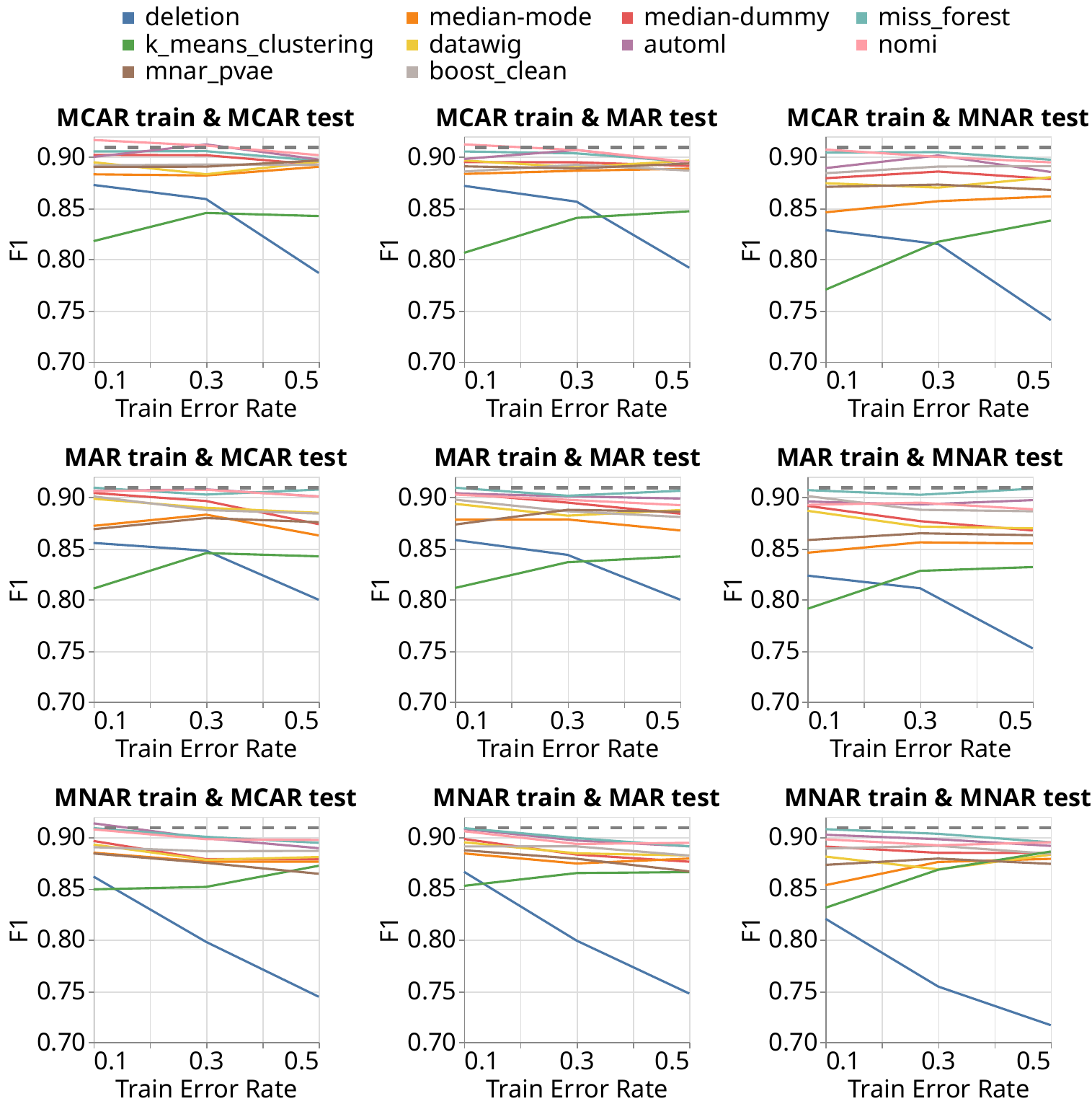}
    \caption{\rev{F1 of the Random Forest model on \diabetes, as a function of training set missingness rate. Dashed line shows performance of the model trained on clean data.}}
    \label{fig:diabetes-train-error-f1}
\end{figure}

\emph{Training set missingness.}  Figure~\ref{fig:diabetes-train-error-f1} shows the F1 of the Random Forest model on \diabetes as a function of training missingness rate. Of all \mvi techniques, \deletion is most strongly affected by missingness: F1 degrades with increasing missingness rate, and this effect is strongest under \mnar. This includes when \mnar is encountered both during training (the bottom row in Figure~\ref{fig:diabetes-train-error-f1} shows the steepest decline in F1 compared to other rows---training missingness) \emph{and} during testing (the right-most column in Figure~\ref{fig:diabetes-train-error-f1} has the lowest F1 compared to other columns ---test missingness).

All other \mvi techniques, including \boostclean, are generally robust to higher training missingness rates, and only show a 5\% decrease in F1 (compared to 10\% decrease with \deletion), even at rates as high as 50\%. \rev{This is because even imperfect imputation provides valuable insights for the model, making deletion a less favorable choice.} A notable exception is \clustering, which, surprisingly and somewhat counter-intuitively, has higher F1 at higher training missingness rates, and is actually better under \mnar than under \mar and \mcar, for all missingness rates and scenarios. 

\emph{Test set missingness.} \arxiv{Figure~\ref{fig:diabetes-test-error-f1} in Appendix~\ref{apdx:f1-shift-additional} shows the F1 of the Random Forest model on \diabetes as a function of test missingness rate.} We find that F1 generally decreases with an increase in test missingness, corroborating the findings of ~\citet{Shadbahr_nature_medicine} \rev{and \citet{miao2022experimental}}. A notable exception is \missforest, which is robust to both forms of missingness shift such as changing missingness rates \rev{(as observed in~\cite{miao2022experimental})} and missingness mechanisms. As for training set missingness, F1 decreases with an increase in test missingness most steeply under \mnar (both during training and test), \rev{further supporting the findings of~\citet{miao2022experimental}.} And, once again, \clustering is an exception to this trend, instead showing invariance to test missingness rates under \mnar train (irrespective of test missingness) but not under \mcar and \mar train.  \submit{See full version of the paper~\cite{shades_arXiv} for complete results.}

\subsection{Fairness of the Predictive Model}
\label{sec:shift-fairness}

\begin{figure}[t!]
    \centering
    \includegraphics[width=\linewidth]{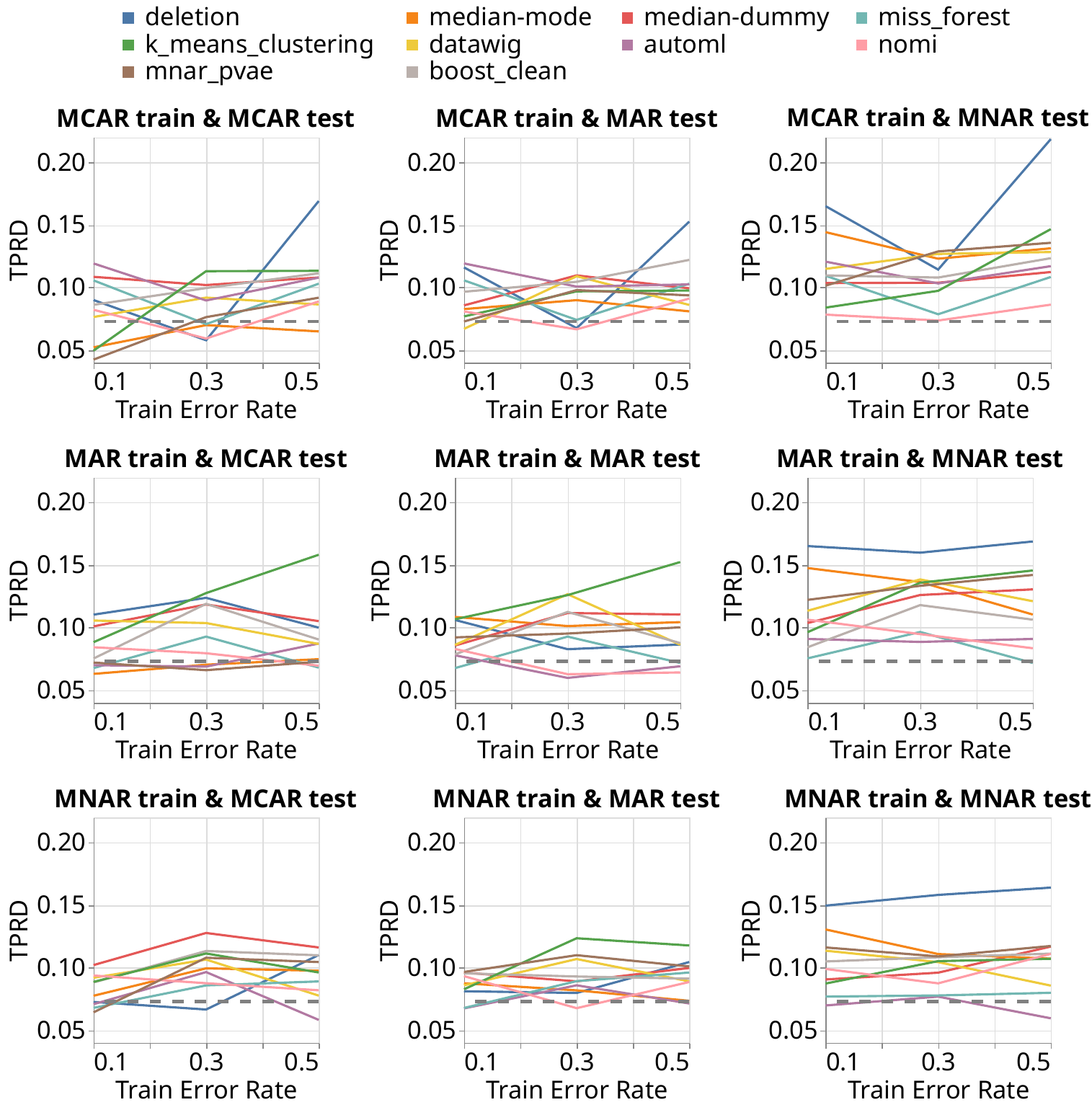}
    \caption{\rev{True Positive Rate Difference of Random Forest on \diabetes, as a function of training missingness rate. Dashed line shows performance of the model trained on clean data.}}
    \label{fig:diabetes-train-error-tprd}
\end{figure}

\emph{Training set missingness.} Figure~\ref{fig:diabetes-train-error-tprd} shows TPRD of Random Forest on \diabetes as a function of training set missingness rate. While we previously found that fairness is generally agnostic to missingness type when it is the same between training and test sets (see Section~\ref{sec:exp1-single-model-fairness}), we find that \textbf{model fairness is highly sensitive to missingness shift} --- in terms of both different missingness rates and different missingness mechanisms between training and test.

Worryingly, there is no consistent trend across \mvi technique, missingness type and training test missingness rate. For example, consider \missforest, which was the most robust to missingness shift according to F1. Under \mcar training, \missforest preserves fairness of the model trained on clean data (shown with a dashed grey line) when training and test missingness rates match (at 30\% training error rate, fixed at 30\% for this experiment), but worsens fairness (higher TPRD) when they are different (at 10\% and 50\% training error rates). Under \mar training, however, we see the opposite behavior, with \missforest preserving clean model fairness at 10\% and 50\% train missingness rates, but worsening fairness when training and test missingness rates are equal (at 30\%). Under \mnar training, TPRD increases with an increase in training missingness rate under \mcar and \mar test, and remains constant when there is no shift in missingness mechanism (under \mnar test).

\emph{Test set missingness.} We measured the impact of test missingness rate on fairness and found that fairness is highly sensitive to such shifts.
\arxiv{Figure~\ref{fig:diabetes-test-error-tprd} in Appendix~\ref{apdx:tprd-shift-additional} shows the TPRD of the Random Forest model on \diabetes as test missingness increases. As before, fairness is highly sensitive to shifts in test missingness.}
For \boostclean, \datawig, \rev{and \mnarpvae}, TPRD generally increases (fairness worsens) as test error rate rises, though not always monotonically. \rev{\missforest, \automl, and \nomi are  robust to increases in test missingness rate under all scenarios}. Simpler methods such as \deletion, \clustering, \mode and \dummy show no consistent trend, even when missingness types remain the same and only missingness rates change between train and test. For example, with \deletion and \clustering, TPRD increases with test missingness in scenarios S1 (\mcar train, \mcar test) and S3 (\mnar train, \mnar test), but decreases in S2 (\mar train, \mar test). \submit{See full version for details~\cite{shades_arXiv}.}

\emph{In summary,} our results corroborate the findings of \citet{guha_icde}, and are a cause for concern as they indicate that the \mvi techniques that perform best during development may not preserve fairness post-deployment, where shifts are likely to occur.

\subsection{Stability of the Predictive Model}
\label{sec:shift:stability}

\begin{figure}[t!]
    \centering
    \includegraphics[width=\linewidth]{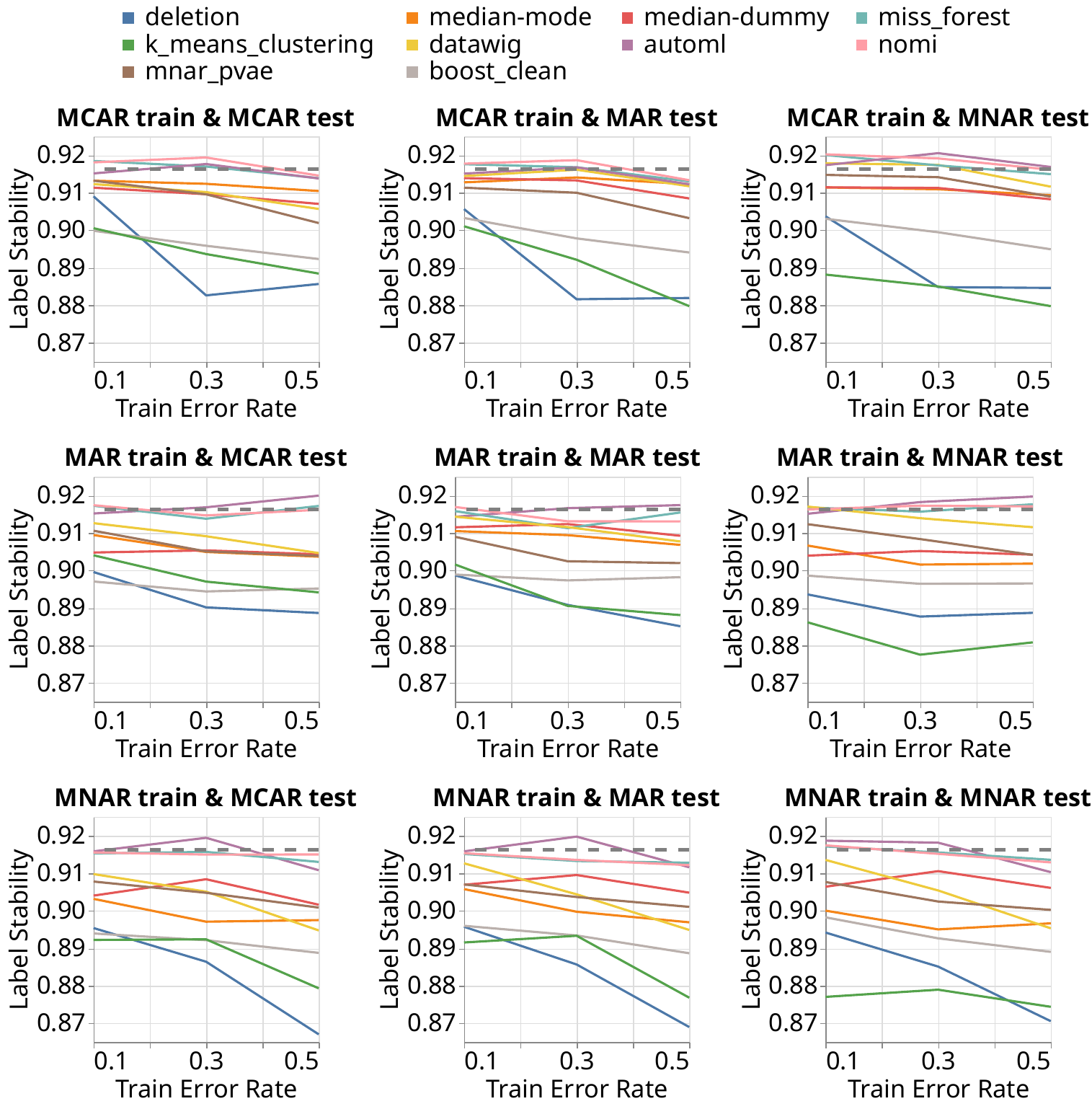}
    \caption{\rev{Label Stability of Random Forest on \diabetes, as a function of training set missingness rate. Dashed line shows performance of the model trained on clean data.}}
    \label{fig:diabetes-train-error-stability}
\end{figure}

\emph{Training set missingness.} Figure~\ref{fig:diabetes-train-error-stability} shows Label Stability of the Random Forest model on \diabetes as a function of training set missingness rate. Under \mcar and \mar missingness, most \mvi techniques with the exception of \boostclean, \deletion, and \clustering show good stability (comparable to the model trained on clean data), and are generally insensitive to missingness rates. \deletion and \clustering are the least stable methods and show a monotonic decrease in stability with increase in missingness rate, with strongest effect under \mnar. \missforest, \automl, \rev{and \nomi} are the most robust \mvi techniques and preserve stability of the clean model under all missingness settings and error rates. \rev{This further highlights how imputation quality directly influences data uncertainty, ultimately impacting the overall uncertainty of the final model, as discussed in Section~\ref{sec:exp1-single-model-stability}.}

\emph{Test set missingness.} \arxiv{Figure~\ref{fig:diabetes-test-error-stability} in Appendix~\ref{apdx:stability-shift-additional} shows the Label Stability of the Random Forest model on \diabetes as a function of test missingness rate.} We find that test missingness rate has little effect on model stability under all settings except \clustering, which shows lower label stability at higher missingness rates, most pronounced in scenario S7 (\mar train, \mnar test).  \submit{See full version of the paper~\cite{shades_arXiv} for complete results.}

%% file: sections/time.tex
\section{\rev{Running Time}}
\label{sec:time}

\begin{table*}[h!]
\rev{
    \caption{\rev{Table 4: Training time (in seconds) of MVI techniques averaged across single- and multi-mechanism scenarios (S1-3, S10). Imputers are sorted by running time on the \texttt{folk\_emp} dataset, and datasets are ordered by the number of rows. Dataset shapes reflect training sets with 30\% rows with nulls. Values represent mean running times across seeds, with standard deviations.}}
    \small 
    \label{tab:4}
    \input{tables/imputation_runtime_avg}
    }
\end{table*}

\rev{Table 4 presents the training time of \mvi techniques for each dataset, averaged across all unique training scenarios (single- and multi-mechanism S1-S3, S10).  Our time efficiency analysis approach aligns with prior work~\cite{wang2024missing, miao2021efficient}, who also focused on training time as inference times are comparably fast across all techniques.}

\rev{Our results reveal that statistical imputers \deletion, \mode, and \dummy are the fastest, while still delivering competitive accuracy for larger datasets like \heart and \texttt{folk\_emp}\arxiv{, as shown in Figures~\ref{fig:exp1-imp-f1} and \ref{fig:exp1-imp-rmse}}. In contrast, \missforest, \datawig, and \automl exhibit the longest training times, with at least one of these methods achieving the highest imputation accuracy in most cases. Interestingly, \automl requires three times more training time than \datawig, the second most computationally intensive technique. This difference is due to the auto-ML nature of \automl, which involves extensive hyperparameter and network architecture tuning.}

\rev{Among non-statistical techniques, \mnarpvae, \editgain, and \nomi are the most efficient. Notably, \nomi delivers accuracy on par with \missforest, \datawig, and \automl, successfully balancing imputation accuracy and training time. A key comparison is between \gain and \editgain. As explained \arxiv{in Appendix~\ref{apdx:mvm-techniques}}\submit{in the full version of the paper~\cite{shades_arXiv}}, \editgain achieves a 28x speedup on \texttt{folk\_emp}, with even greater improvements for smaller datasets, as shown in Table~\ref{tab:4}, while maintaining comparable accuracy to \gain\arxiv{, as shown in Figures~\ref{fig:exp1-imp-f1} and \ref{fig:exp1-imp-rmse}}.}


%% file: tables/imputation_runtime_avg.tex
\begin{tabular}{llllllll}
\toprule
 \textbf{Imputer}         & \makecell[tl]{\textbf{diabetes}\\\textbf{(633, 17)}}   & \makecell[tl]{\textbf{german}\\\textbf{(700, 21)}}   & \makecell[tl]{\textbf{folk$\_$inc}\\\textbf{(12000, 10)}}   & \makecell[tl]{\textbf{law$\_$school}\\\textbf{(16638, 11)}}   & \makecell[tl]{\textbf{bank}\\\textbf{(32003, 13)}}   & \makecell[tl]{\textbf{heart}\\\textbf{(56000, 11)}}   & \makecell[tl]{\textbf{folk$\_$emp}\\\textbf{(242112, 16)}}   \\
\midrule
 median-dummy             & $0.013 \pm 0.000$                                      & $0.014 \pm 0.001$                                    & $0.021 \pm 0.001$                                           & $0.024 \pm 0.001$                                             & $0.027 \pm 0.000$                                    & $0.053 \pm 0.001$                                     & $0.773 \pm 0.024$                                            \\
 median-mode              & $0.012 \pm 0.000$                                      & $0.014 \pm 0.001$                                    & $0.023 \pm 0.001$                                           & $0.025 \pm 0.001$                                             & $0.031 \pm 0.001$                                    & $0.067 \pm 0.003$                                     & $0.933 \pm 0.019$                                            \\
 deletion                 & $0.013 \pm 0.000$                                      & $0.013 \pm 0.001$                                    & $0.024 \pm 0.001$                                           & $0.025 \pm 0.000$                                             & $0.046 \pm 0.001$                                    & $0.087 \pm 0.004$                                     & $1 \pm 0.049$                                                \\
 mnar$\_$pvae             & $8 \pm 0.705$                                          & $14 \pm 11$                                          & $14 \pm 1$                                                  & $22 \pm 11$                                                   & $55 \pm 30$                                          & $42 \pm 6$                                            & $206 \pm 7$                                                  \\
 edit$\_$gain             & $2 \pm 0.119$                                          & $2 \pm 0.141$                                        & $13 \pm 0.221$                                              & $21 \pm 2$                                                    & $30 \pm 1$                                           & $62 \pm 7$                                            & $215 \pm 4$                                                  \\
 nomi                     & $11 \pm 3$                                             & $14 \pm 7$                                           & $22 \pm 2$                                                  & $22 \pm 2$                                                    & $29 \pm 1$                                           & $38 \pm 2$                                            & $356 \pm 20$                                                 \\
 tdm                      & $932 \pm 74$                                           & $1023 \pm 11$                                        & $1297 \pm 22$                                               & $1172 \pm 92$                                                 & $1412 \pm 34$                                        & $1310 \pm 42$                                         & $1449 \pm 31$                                                \\
 notmiwae                 & $161 \pm 101$                                          & $217 \pm 82$                                         & $555 \pm 2$                                                 & $804 \pm 8$                                                   & $665 \pm 284$                                        & $1393 \pm 535$                                        & $2944 \pm 725$                                               \\
 gain                     & $261 \pm 12$                                           & $298 \pm 5$                                          & $1115 \pm 38$                                               & $1484 \pm 30$                                                 & $1964 \pm 36$                                        & $2892 \pm 48$                                         & $6148 \pm 178$                                               \\
 hivae                    & $68 \pm 0.784$                                         & $95 \pm 1$                                           & $745 \pm 15$                                                & $1073 \pm 11$                                                 & $2450 \pm 130$                                       & $3794 \pm 129$                                        & $7163 \pm 317$                                               \\
 k$\_$means$\_$clustering & $10 \pm 0.402$                                         & $13 \pm 0.442$                                       & $27 \pm 0.640$                                              & $323 \pm 7$                                                   & $998 \pm 49$                                         & $1037 \pm 64$                                         & $7427 \pm 778$                                               \\
 miss$\_$forest           & $111 \pm 17$                                           & $244 \pm 86$                                         & $1758 \pm 526$                                              & $2530 \pm 851$                                                & $4307 \pm 1310$                                      & $6337 \pm 1601$                                       & $20358 \pm 4934$                                             \\
 datawig                  & $596 \pm 185$                                          & $277 \pm 59$                                         & $604 \pm 46$                                                & $2361 \pm 492$                                                & $5089 \pm 651$                                       & $7592 \pm 854$                                        & $31060 \pm 3743$                                             \\
 automl                   & $1953 \pm 195$                                         & $1805 \pm 212$                                       & $5559 \pm 581$                                              & $6803 \pm 565$                                                & $13893 \pm 1687$                                     & $19055 \pm 2710$                                      & $104476 \pm 14743$                                           \\
\bottomrule
\end{tabular}

%% file: sections/conclusion-new.tex
\section{Summary of Experimental Findings}
\label{sec:conclusion}

\emph{Do not drop your nulls!} \rev{Building on prior evidence~\cite{rubin1987statistical, martinez2019fairness, little2015modeling, Joel_Review_Missing_Data_for_ML}, we confirm that deletion is the least effective strategy for model accuracy, fairness, and stability—especially when each row holds valuable information. While deletion leads to data loss by design, its suitability depends more on data quality than quantity. If rows are duplicates or contain errors, deletion may be warranted.}
 
\emph{Multiple imputation shows mixed results.}  \rev{There is conflicting evidence on the performance of multiple imputation (MI)~\cite{rebutting_MI}, and our empirical findings are similarly mixed and somewhat unexpected. \citet{feng2023adapting} and \citet{le2021goodimputation} argue that MI outperforms impute-then-classify approaches in predictive performance, while \citet{graham2009missing} and \citet{mcneish_missingness_small_samples} find MI effective even with limited data and small error rates. In contrast, we find that MI (specifically, \boostclean) is \emph{only} competitive on small datasets and is less stable than simpler \mvi techniques. This likely stems from the complexity-stability trade-off~\cite{unified_decomposition, breiman_two_cultures}: MI employs a more complex model class that minimizes empirical loss but exhibits greater prediction variance under small training set perturbations.}

\emph{Fairness is highly missingness-specific.} We find that no \mvi technique is consistently fairness-preserving, corroborating the findings of~ \citet{zhang2021_imputationfairness} and \citet{guha_icde}. Further, we find that fairness is highly sensitive to changes in missingness rates and missingness mechanisms between training and test sets, which are likely to occur in practice, and therefore a cause for ethical concern.

\emph{Imputation quality and fairness metrics often fail to predict the correctness or fairness of downstream models.} \rev{\citet{Shadbahr_nature_medicine} argue that distributional imputation quality metrics better predict model performance than discrepancy metrics. However, we find that neither reliably predicts downstream performance, as strong learners can compensate for poor imputations. Moreover, imputation fairness does not predict model fairness: fair imputers can still yield unfair models, while models trained with fairness-poor \mvi techniques can achieve good downstream fairness.}

\begin{figure*}[t!]
\begin{subfigure}[h]{0.245\linewidth}
    \centering
    \includegraphics[width=\linewidth]{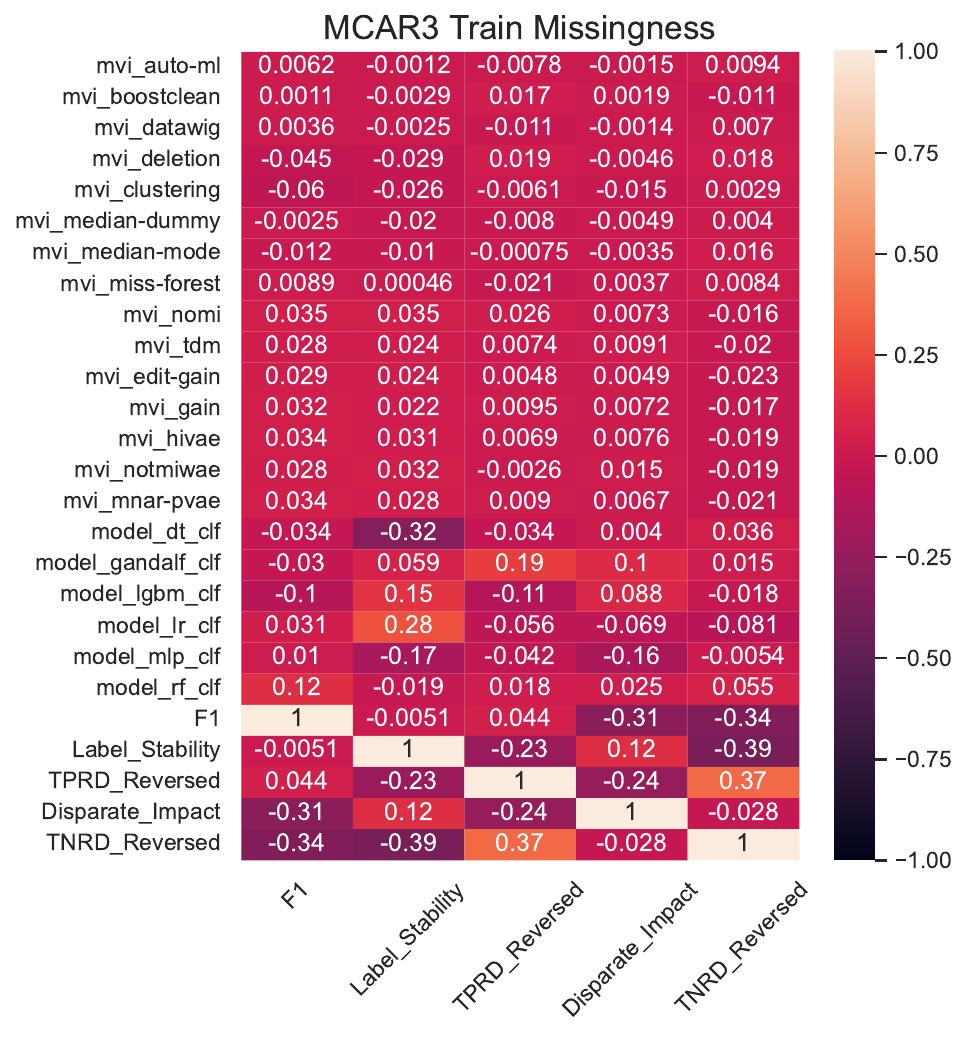}
    \caption{\mcar missingness}
\end{subfigure}
\hfill
\begin{subfigure}[h]{0.245\linewidth}
    \centering
    \includegraphics[width=\linewidth]{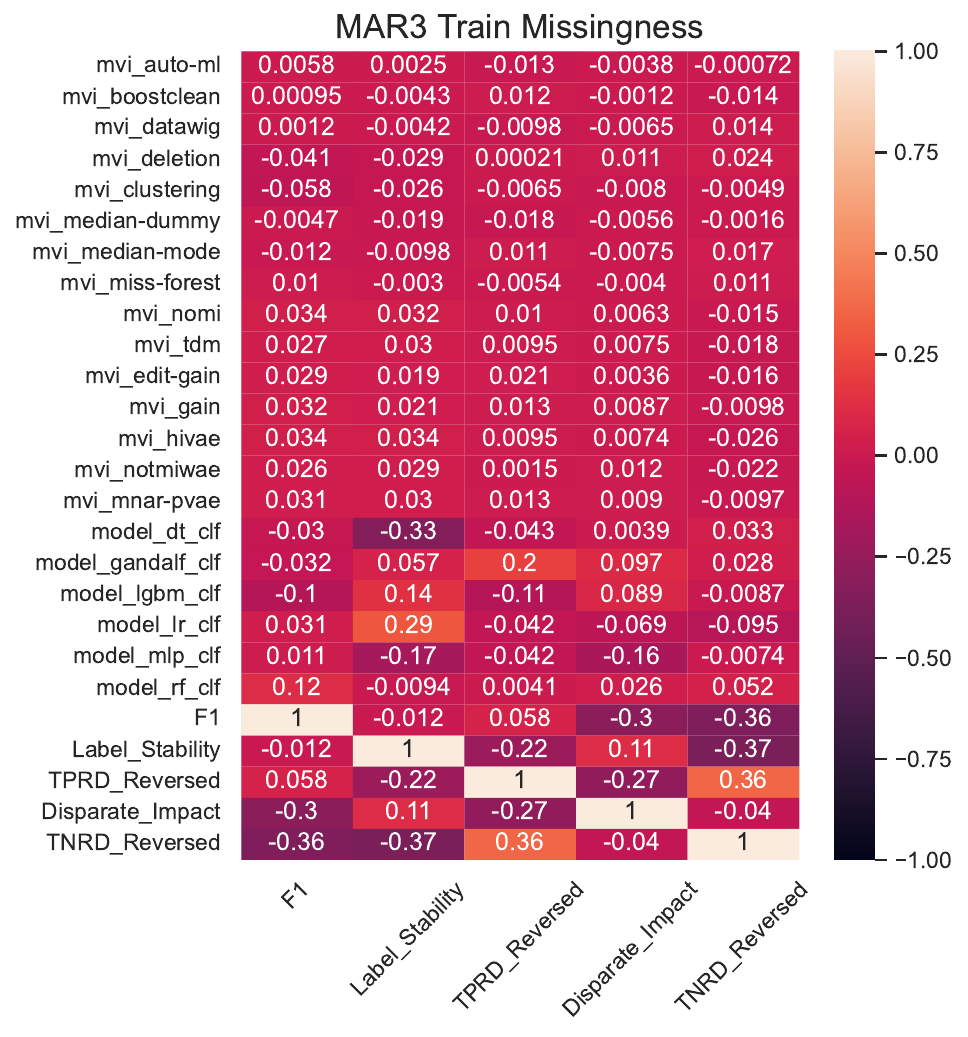}
    \caption{\mar missingness}
\end{subfigure}
\hfill
\begin{subfigure}[h]{0.245\linewidth}
    \centering
    \includegraphics[width=\linewidth]{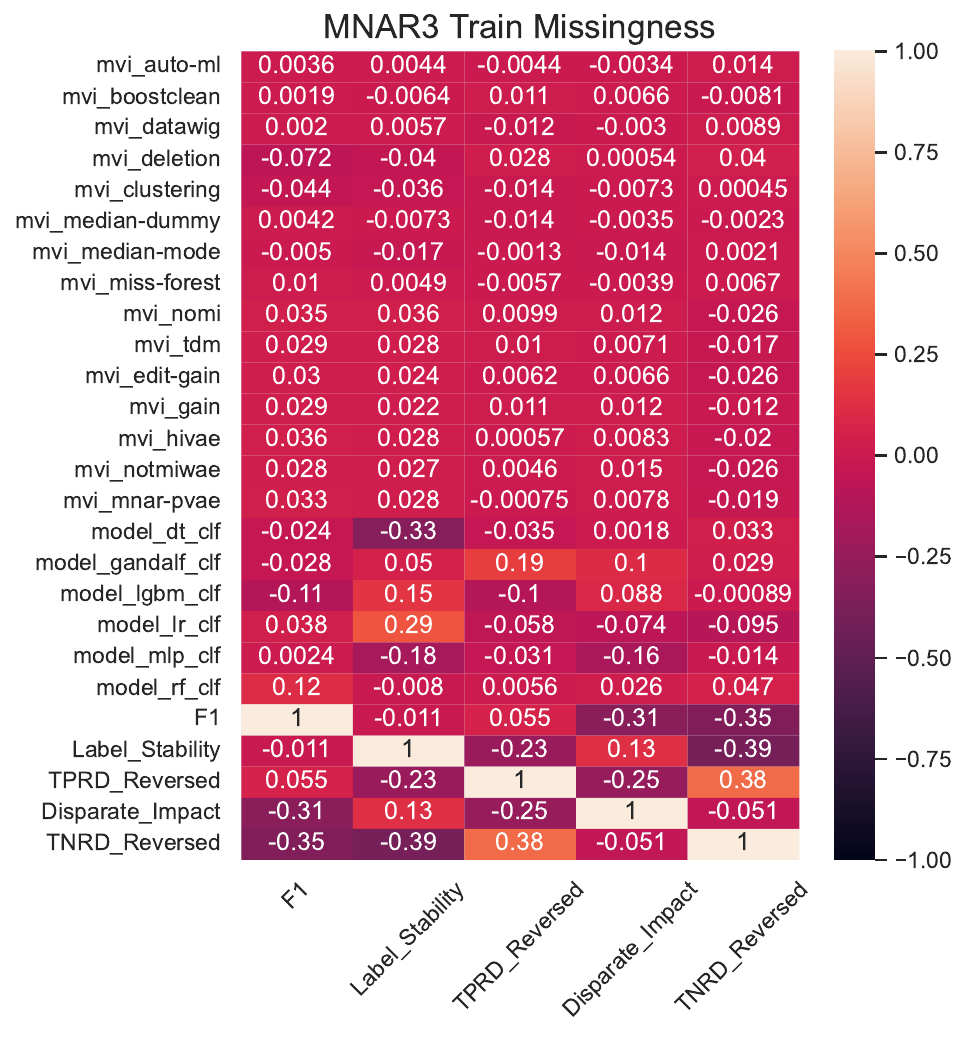}
    \caption{\mnar missingness}
\end{subfigure}
\hfill
\begin{subfigure}[h]{0.245\linewidth}
    \centering
    \includegraphics[width=\linewidth]{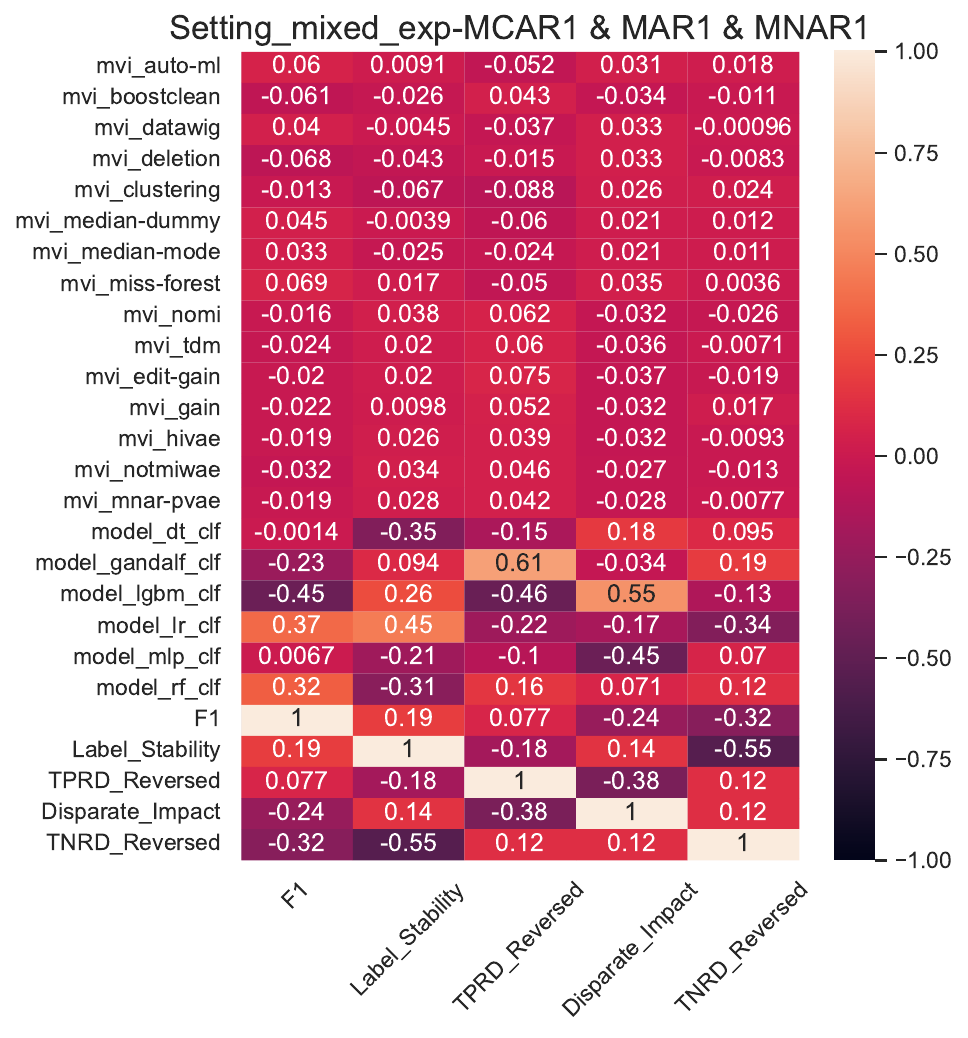}
    \caption{Mixed missingness}
\end{subfigure}

\caption{\rev{Spearman correlation ($\rho$) between \mvi technique, model type, and performance metrics (F1, fairness and stability) for different train missingness mechanisms (subplots). TPRD and TNRD values close to 0 are ideal (fair), so we compute correlations using $TPRD\_Reversed = 1 - |TPRD|$ and $TNRD\_Reversed = 1 - |TNRD|$. Supplementary plots are in \arxiv{Appendix~\ref{sec:apdx-correlations}}\submit{the full version~\cite{shades_arXiv}}.}}
\label{fig:correlations_train}
\end{figure*}

\emph{Model stability depends more on the dataset size and \mvi technique than on the missingness scenario.} We find that for large datasets even simple statistical imputers can preserve stability.  In contrast, for small datasets, only a few \mvi techniques do so, while deletion, statistical imputation and complex ML-based \mvi all worsen stability. 

{\emph{Sensitivity to train and test missingness rates.}  \rev{Model performance (F1) is more affected by test missingness than training missingness. Fairness, however, is highly sensitive to both. Model stability is largely unaffected by test missingness, but most \mvi techniques become more unstable with higher training missingness.}

\rev{\emph{Existing MNAR-specific methods are insufficient.} \mnar is theoretically the hardest setting to model. \mvi techniques, including \mnar-specific \notmiwae and \mnarpvae, perform poorly under \mnar. For example, F1 and stability are more sensitive to missingness rates under \mnar than under \mcar or \mar. This performance gap stems from unrealistic assumptions and limited evaluations, typically on large datasets and a single scenario, lacking the diversity of our benchmark. Advancing \mnar-specific \mvi methods and building comprehensive benchmarks are key future directions.}

\emph{Deep learning outperforms tree-based models on large datasets.}  
\rev{Interest in deep learning for tabular data has surged~\cite{hwang2023recent_tabular, gorishniy2021revisiting_tabular, borisov2022deep_tabular, ye2024closer_tabular, fayaz2022deep_tabular}. Recall that Figures~\ref{fig:exp1-F1}, \ref{fig:exp1-TPR}, and~\ref{fig:exp1-label-stability} show only the best-performing models (by F1) for clarity. Tree-based methods like random forests and gradient-boosted trees outperform deep table-learning models like GANDALF~\cite{joseph2022gandalf} on small datasets (\diabetes, \german), aligning with prior findings~\cite{grinsztajn2022tabular, shwartz2022tabular}. However, GANDALF excels on larger datasets (\heart, \folkemp), highlighting the increasing effectiveness of deep learning across diverse data modalities.}

\emph{Best-performing approaches under single- and multi-mechanism missingness can differ.} \rev{A key contribution of our study is the evaluation under mixed (multi-mechanism) missingness --- a complex pattern likely to arise in practice. We find that performance trends differ between single- and multi-mechanism settings. Figure~\ref{fig:correlations_train} shows that model types like \texttt{gandalf\_clf} and \texttt{lgbm\_clf} are uncorrelated with performance metrics under single-mechanism missingness, but become correlated under mixed missingness: F1 correlations are $\rho=-0.24$ for \texttt{gandalf\_clf} and $\rho=-0.37$ for \texttt{lgbm\_clf}, while fairness (TPRD) correlations are $\rho=0.55$ and $\rho=-0.4$, respectively.}

\emph{It's complicated!}  
\rev{Prior studies~\cite{hasan_mvi_review,mvi_review,Joel_Review_Missing_Data_for_ML, Shadbahr_nature_medicine, missing_data_review} have found no universally best missing value imputation technique for predictive performance. Our findings reinforce this, revealing trade-offs between F1, fairness, and stability that depend on the predictive model’s architecture.  For instance, under multi-mechanism missingness (Figure~\ref{fig:correlations_train}), logistic regression (\texttt{lr$\_$clf}) correlates positively with F1 ($\rho$=0.23) and stability ($\rho$=0.38) but negatively with fairness (TPRD, $\rho$=-0.23). Random forest (\texttt{rf$\_$clf}) shows weak positive correlations with fairness (TPRD, $\rho$=0.16) and F1 ($\rho$=0.25) but a negative correlation with stability ($\rho$=-0.24). Deep table-learning (\texttt{gandalf$\_$clf}) is strongly correlated with fairness (TPRD, $\rho$=0.55), weakly with stability ($\rho$=0.12), and negatively with F1 ($\rho$=-0.24). This is not bad news, but a recognition of the complexity of learning from incomplete data, and of the need for rigorous, holistic evaluation protocols, like those used here, to identify the best imputation method and model architecture for a given task.}

%% file: sections/limitations.tex
\section{Conclusions, Limitations and  Future Work}
\label{sec:limitations}

\emph{Conclusions.} \rev{We introduced \sys, an evaluation suite for responsible missing value imputation. A key contribution is evaluating fairness and stability alongside predictive performance. Additionally, we model realistic missingness scenarios beyond Rubin’s \mcar, \mar, and \mnar, incorporating multi-mechanism missingness and missingness shift.  Through \rev{\numexps} experimental pipelines, we assessed various \mvi methods under realistic missingness conditions, revealing key trends and trade-offs across evaluation metrics.}

\paragraph{Limitations.} We began with clean datasets, designed meaningful missingness scenarios, and simulated them using error injectors. A key limitation is that we evaluate performance on synthetically generated---rather than naturally occurring---missingness. This is common in the field: clean ground truth is rarely available, so we assume that injecting synthetic errors only worsens the effects of existing (unknown) ones. Under this assumption, any observed performance degradation likely underestimates the true degradation relative to an ideal ground truth.

\paragraph{Future work.} \rev{Creating research datasets with naturally occurring missingness shifts is a promising direction. While predictive performance, fairness, and stability are often treated as orthogonal, we uncover trade-offs driven by missingness, imputation choice, and, in some cases, model architecture.  The absence of a universally best imputation method highlights the need for holistic metrics, evaluation procedures, and techniques grounded in the data lifecycle. Missing value imputation remains a critical aspect of responsible data engineering, to which our work makes a contribution.}

%% file: appendix/appendix-overview.tex
\section*{Appendix Overview}

This supplemental material provides significantly more detail regarding the results presented in this paper. The organization is as follows: We first provide additional details on the implementation of the evaluation suite, which include a broad description and configuration of the \mvi techniques used in our study (Appendix~\ref{apdx:mvm-techniques}) and an explanation of the corrections and enhancements made to existing benchmarks and \mvi techniques to ensure the development of our truly fair evaluation suite (Appendix~\ref{apdx:enhancements}). Next, we extend the description of our experimental settings to facilitate reproducibility (Appendix~\ref{apdx:exp-details-additional}). Specifically, Appendix~\ref{apdx:computing-infrastructure} provides detailed information about our experimental setup on the cluster, while Appendix~\ref{apdx:extended-methodology} extends our methodology for simulating missingness in other datasets, including proportions and base rates of protected groups, realistic missingness scenarios, feature correlation coefficients with the target, and feature importance for each dataset. In Appendix~\ref{apdx:single-mechanism-additional}, we offer additional results to those discussed in Section~\ref{sec:exp1-single}. In particular, Appendix~\ref{apdx:mvi-metrics} presents the imputation quality of \mvi techniques across all datasets from the perspectives of accuracy and training time, while Appendix~\ref{apdx:single-mechanism-fairness-metrics} showcases additional metrics for predictive model performance beyond those mentioned in Section~\ref{sec:exp1-single} and includes scatter plots for all imputers that extend scatter plots from Section~\ref{sec:imputation-quality}. Then, Appendix~\ref{apdx:shift-additional} presents supplementary experimental findings on the missingness shift discussed in Section~\ref{sec:shift}. Specifically, Appendix~\ref{apdx:fixed-shift-additional} examines the accuracy, stability, and fairness dimensions of model performance under missingness shift with fixed train and test error rates, highlighting its effects on various model types and datasets. Furthermore, Appendix~\ref{apdx:variable-shift-additional} explores further experimental results on model performance under missingness shift with variable train and test error rates. Finally, Appendix~\ref{sec:apdx-correlations} presents correlation plots for other missingness settings than in Section~\ref{sec:conclusion}.

%% file: appendix/benchmark-additional.tex
\section{Additional details on the evaluation suite implementation}

\subsection{Missing Value Imputation Techniques}
\label{apdx:mvm-techniques}

In our evaluation suite, we evaluate 15 \mvi techniques, classified into 8 broad categories, as outlined in Section~\ref{sec:mvm-summary}. These categories comprehensively cover the spectrum of MVI methods discussed in recent surveys~\cite{emmanuel2021survey, jager2021benchmark, miao2022experimental, abdelaal2023rein} and peer-reviewed papers~\cite{wang2024missing, wu2023differentiable, miao2021efficient}, which propose new MVI techniques and compare against them. In each category, we selected several SOTA methods for our comparison. Each technique is capable of imputing both categorical and numerical features. For all \mvi techniques, we adopt the default parameter settings from original papers or their source code.

\begin{enumerate}
    \item \textbf{\deletion}: We simply drop the rows with missingness indicators.

    \item \textbf{\mode} (\textit{statistical}): We impute missing values with the median of the complete case for numerical columns and mode (most frequently occurring value) for categorical columns. 
    
    \item \textbf{\dummy} (\textit{statistical}): We impute missing values with the median of the complete case for numerical columns and assign a (new) dummy category for nulls in categorical columns.
    
    \item \textbf{\missforest}~\cite{missForest} (\textit{supervised ML}): This approach iteratively trains an RF model on a set of clean samples (no missingness) and predicts the missing values. We tune \textit{RandomForestClassifier} for categorical columns with nulls and \textit{RandomForestRegressor} for numerical columns with nulls. The maximum number of iterations for \missforest is set to 10.

    \item \textbf{\clustering}~\cite{gajawada2012missing_clustering} (\textit{unsupervised ML}): We assign the missing value to a cluster, based on the distance from the cluster center, and then impute the missing value with the mean or mode of the data points in that cluster.\footnote{We used the k-prototypes implementation from \url{https://github.com/nicodv/kmodes}} We implemented this method ourselves, as no existing solution supports null imputation based on clustering for data with both numerical and categorical features.

    \item \textbf{\datawig}~\cite{biessmann2019datawig} (\textit{discriminative DL}): This method implements deep learning modules combined with neural architecture search and end-to-end optimization of the imputation pipeline. We train \datawig for 100 epochs and a single iteration using a batch size of 64. The \textit{final\_fc\_hidden\_units} parameter is tuned for each dataset by selecting the optimal value from the set \{1, 10, 50, 100\}.
    
    \item \textbf{\automl}~\cite{jager2021benchmark} (\textit{discriminative DL}): This approach uses the AutoML library autokeras~\cite{jin2019auto} to implement the discriminative deep learning imputation method. For categorical columns, autokeras’ StructuredDataClassifier is employed, while StructuredDataRegressor is utilized for numerical columns. These classes manage data encoding and optimize model architecture and hyperparameters autonomously. We train \automl for 100 epochs with a validation split of 0.2, using up to 50 trials to tune hyperparameters.

    \item \textbf{\gain}~\cite{yoon2018gain} (\textit{generative DL}): This method adapts the Generative Adversarial Networks (GAN) framework to impute missing data. The generator observes partially observed data and imputes the missing components conditioned on the observed data, outputting a completed data vector. The discriminator distinguishes between observed and imputed components, using a hint vector to focus on specific imputation tasks. The adversarial training ensures that the generator learns to produce imputations that closely match the true data distribution. We train \gain by tuning its hyperparameters based on the following grid: $\alpha \in \{1, 10\}$, hint rate $\in \{0.7, 0.9\}$, batch size $\in \{64, 128\}$, generator learning rate $\in \{1 \times 10^{-5}, 1 \times 10^{-4}, 0.0005\}$, and discriminator learning rate $\in \{1 \times 10^{-6}, 1 \times 10^{-5}, 0.00005\}$.

    \item \textbf{\hivae}~\cite{nazabal2020handling} (\textit{generative DL}): This approach extends the Variational Autoencoder (VAE) framework to handle incomplete and heterogeneous datasets. HI-VAE models mixed numerical (e.g., real-valued, positive real-valued, and count) and categorical (e.g., ordinal and nominal) data types. It employs a hierarchical architecture with a shared latent space to capture correlations among attributes and a likelihood model tailored to each data type. By training on observed data and utilizing an Evidence Lower Bound (ELBO) computed only for observed entries, HI-VAE achieves robust imputation without overfitting. We train \hivae for 2000 epochs using a batch size of 128, with latent dimensions set as $z = 10$, $y = 5$, and $s = 10$, and a learning rate of $1 \times 10^{-3}$.

    \item \textbf{\notmiwae}~\cite{ipsen2020not} (\textit{MNAR-specific}): This method extends deep generative modeling to handle Missing Not At Random (MNAR) data by explicitly modeling the missing data mechanism alongside the data distribution. It leverages importance-weighted variational inference to jointly optimize the parameters of the data and missing mechanisms. The method uses stochastic gradients derived via reparameterization in both latent and data spaces, allowing efficient training. By incorporating prior knowledge about the missingness process into a flexible missing model, \notmiwae achieves robust imputation under MNAR scenarios. We train \notmiwae for a maximum of 100,000 iterations using a batch size of 16, with 128 hidden units, and a latent dimensionality of $L = 10,000$, employing the self-masking process \textit{selfmasking\_known}.

    \item \textbf{\mnarpvae}~\cite{ma2021identifiable} (\textit{MNAR-specific}): This approach introduces an identifiable deep generative model for handling Missing Not At Random (MNAR) data. By leveraging identifiability principles and extending variational autoencoders (VAEs), GINA ensures that the underlying data-generating process can be uniquely recovered. It jointly models the data distribution and the missingness mechanism, integrating auxiliary variables to achieve identifiability under mild assumptions. Through a combination of importance-weighted variational inference and a flexible neural architecture, GINA enables robust and unbiased imputation of MNAR data. We train \mnarpvae (GINA) for 400 epochs and a single iteration using a batch size of 100 and a learning rate of $1 \times 10^{-3}$. The network is configured with an embedding dimension of 20, a latent dimension of 20, encoder and decoder layers each containing 10 units, and uses \textit{Tanh} as the non-linearity.
    
    \item Multiple imputation (\textit{joint}) using \textbf{\boostclean}~\cite{krishnan2017boostclean}: This approach treats the error correction task as a statistical boosting problem where a set of weak learners are composed into a strong learner. To generate the weak learners, BoostClean iteratively selects a single imputation technique, applies it to a training set (with missingness), and fits a new model on the newly imputed training set. We train \boostclean for 5 iterations and tune its underlying prediction model for each iteration.

    \item \textbf{\nomi}~\cite{wang2024missing} (\textit{recent}): This method introduces an uncertainty-driven network for missing data imputation. NOMI integrates three key components: a retrieval module, a Neural Network Gaussian Process Imputator (NNGPI), and an uncertainty-based calibration module. The retrieval module identifies local neighbors of incomplete samples, while the NNGPI combines the probabilistic modeling of Gaussian Processes with the representation power of neural networks to impute missing values and quantify uncertainty. The uncertainty-based calibration module dynamically refines the imputations by balancing predictions across iterations, leveraging uncertainty to improve reliability. We train \nomi with a maximum of 3 iterations, using 10 neighbors for imputation, the $l_2$ similarity metric, a temperature parameter $\tau = 1.0$, and a weighting coefficient $\beta = 0.8$.
    
    \item \textbf{\tdm}~\cite{zhao2023transformed} (\textit{recent}): This approach proposes Transformed Distribution Matching (TDM), a novel approach to missing data imputation that leverages optimal transport in a transformed latent space. TDM employs deep invertible neural networks to map data into a latent space where the geometry better reflects the underlying data structure. It minimizes the Wasserstein distance between the transformed distributions of two batches of data to achieve imputation, while avoiding overfitting through mutual information constraints. We train \tdm for 10,000 iterations with a batch size of 512 and a learning rate of $1 \times 10^{-2}$.

    \item \textbf{\editgain}~\cite{miao2021efficient} (\textit{recent}): This technique introduces an efficient and effective data imputation framework that leverages influence functions to accelerate the training of parametric imputation models. EDIT consists of two key modules: the Imputation Influence Evaluation (IIE) module, which estimates the influence power of samples on the imputation model's predictions, and the Representative Sample Selection (RSS) module, which constructs a minimal representative sample set to satisfy user-specified accuracy guarantees. Additionally, EDIT employs a weighted loss function that emphasizes high-influence samples, boosting imputation accuracy while reducing training cost. We employ \gain~\cite{yoon2018gain} as the underlying imputer for EDIT, as it is the only method shared with us directly by the authors of the original paper. We train \editgain using a batch size of 128 for 30 epochs, with an $\alpha$ parameter value of 1. The initial sample size is set to 6000 for all our datasets, matching the initial sample size of the smallest dataset used in the original paper, since all our datasets are of the same size or smaller.
\end{enumerate}

\subsection{Corrections and Enhancements in Benchmarks and \mvi Techniques}
\label{apdx:enhancements}

Prior to constructing our evaluation suite, we conducted a thorough analysis of the strengths and weaknesses of existing state-of-the-art benchmarks and \mvi techniques by reviewing their codebases. Unfortunately, we identified several methodological flaws and code bugs in some of them. Consequently, alongside creating our software novelty in the evaluation suite, we dedicated considerable effort to correcting these bugs in the \nl imputers and enhancing existing benchmarking approaches to ensure the development of a truly fair evaluation suite.

Although our evaluation suite pursued different goals, \cite{abdelaal2023rein} served as the primary benchmark against which we compared our work. The authors made significant contributions, establishing a benchmark that set a high standard in the field. However, we identified several critical concerns in their codebase that we aimed to address. Firstly, their comparison did not include hyper-parameter tuning for \nl imputers, and they did not utilize seeds to control randomization in the \nl imputers. In contrast, we meticulously tuned hyper-parameters for each \nl imputer in our study, implemented controlled randomization throughout the entire pipeline using seeds, and rigorously tested the evaluation suite with both unit and integration tests to ensure reproducibility. Secondly, a significant methodological flaw in their benchmark was that \nl imputers were fitted on the entire dataset without initially splitting it into training and test sets. The dataset split was introduced only during the model training stage, resulting in a clear leakage when \nl imputers were fitted on samples included in the test set. This issue likely arose because imputers like \missforest and \datawig lack built-in interfaces for fitting only on a training set and then transforming both training and test sets. Presumably, the authors chose to use these imputers without modification. In order to ensure a fair comparison of \mvi techniques as they would be used in production systems, in our study, we standardized our \mvi techniques to meet the following requirements: 1) fitting a \nl imputers only on a training set and then applying it to both training and test sets; 2) enabling hyper-parameter tuning of a \nl imputer (if applicable); 3) controlling randomization of a \nl imputer using seeds (if applicable). Below is a summary of the enhancements we implemented for \mvi techniques used in our evaluation suite:

\begin{itemize}
    \item \textbf{\clustering}: We implemented this method ourselves to meet our imputation requirements, as existing solutions do not support \nl imputation based on clustering for datasets containing missingness in both numerical and categorical features at the same time. For this, we used the \textit{k-prototypes} as a distance function in the imputer.\footnote{\url{https://github.com/nicodv/kmodes}}

    \item \textbf{\missforest}~\cite{missForest}: The main concern with the original implementation was that the \nl imputer was fitted within a \texttt{transform} method. This meant that, when applying the \texttt{transform} method to a test set, the \nl imputer would be fitted on testing samples as well. To adhere to our imputation requirements, we moved the predictor fitting to a \texttt{fit} method and reused the fitted predictors in the \texttt{transform} method. Additionally, we introduced hyper-parameter tuning for the base classifier and regressor, which was lacking in the original implementation.

    \item \textbf{\datawig}~\cite{biessmann2019datawig}: To align this technique with our imputation criteria, we modified the \texttt{complete} method from the\\ \texttt{SimpleImputer}, which imputes \nl values across all categorical and numerical columns in the dataset. We adjusted the methodological approach in the source code and reused built-in \texttt{fit} and \texttt{transform} methods from the \texttt{SimpleImputer} to achieve this.

    \item \textbf{\automl}: We adopted a conceptual idea from \cite{jager2021benchmark} and used their code as a foundation for our implementation. A significant limitation in their source code was its capability to handle only one column with \nl values at a time. To enhance its functionality, we expanded the method to support imputation across multiple categorical and numerical columns. Additionally, we incorporated ideas inspired by \missforest~\cite{missForest}, specifically leveraging initial statistical estimates for each column containing \nl values to train classifiers and regressors used for imputation.

    \item \textbf{\boostclean}~\cite{krishnan2017boostclean}: We enhanced this method by introducing the capability to tune the internal hyper-parameters of \boostclean used for statistical boosting. Furthermore, we implemented the ability to fine-tune the hyper-parameters of the underlying models trained on each dataset that was imputed using \mvi techniques listed in \boostclean.
\end{itemize}

In summary, our evaluation suite offers users not only a robust evaluation framework but also incorporates fixes to the \mvi implementations previously available.

\subsection{Datasets and Tasks}
\label{apdx:datasets}

\emph{\diabetes}\footnote{\url{https://www.kaggle.com/datasets/tigganeha4/diabetes-dataset-2019}}~\cite{diabetes_dataset} was collected in India through a questionnaire including 18 questions related to health, lifestyle, and family background. A total of 952 participants are characterized by 17 attributes (13 categorical, 4 numerical) and a binary target variable that represents whether a person is diabetic. Here, \textit{sex} is the sensitive attribute, with ``female'' as the disadvantaged group.

\emph{\german}\footnote{\url{https://archive.ics.uci.edu/dataset/144/statlog+german+credit+data}}~\cite{misc_statlog_(german_credit_data)_144} is a popular fairness dataset that contains records of creditworthiness assessments, classifying individuals as high or low credit risks. It contains information on 1,000 individuals characterized by 21 attributes (14 categorical, 7 numerical), including credit history, occupation and housing information. Here, \textit{sex} and \textit{age} are the sensitive attributes, with ``female'' and ``age$\leq25$'' as the disadvantaged groups.

Folktables~\cite{folktables_ding2021} is another popular fairness dataset derived from US Census data from all 50 states between 2014-2018. The dataset has several associated tasks, of which we selected two: (i) ACSIncome (\folk) is a binary classification task to predict whether an individual’s annual income is above \$50,000, from 10 features (8 categorical, 2 numerical) including educational attainment, work hours per week, marital status, and occupation. We use data from Georgia from 2018, subsampled to 15k rows. (ii) ACSEmployment (\folkemp) is a binary classification task to predict whether an individual is employed, from 16 features (15 categorical, 1 numerical) including educational attainment, employment status of parent, military status, and nativity. We use data from California from 2018, with 302,640 rows (we do not subsample). In both tasks, \textit{sex} and \textit{race} are the sensitive attributes, with ``female'' and ``non-White'' as the disadvantaged groups.

\emph{\law}\footnote{\url{https://www.kaggle.com/datasets/danofer/law-school-admissions-bar-passage }}
~\cite{wightman1998lsac} was gathered through a survey conducted by the Law School Admission Council (LSAC) across 163 law schools in the US in 1991, and contains admissions records of 20,798 applicants, characterized by 11 attributes (5 categorical, 6 numerical), including LSAT scores and college GPAs. The  task is to predict whether a candidate would pass the bar exam. Here, \textit{sex} and \textit{race} are the sensitive attributes, with ``female'' and ``non-White'' as the disadvantaged groups.

\emph{\bank}\footnote{\url{https://archive.ics.uci.edu/dataset/222/bank+marketing}}
~\cite{Moro2014ADA} contains data from direct marketing campaigns by a Portuguese bank, between 2008 and 2013. It contains information on 40,004 potential customers, with 13 attributes (7 categorical, 6 numerical), including occupation, marital status, education, and a binary target that indicates whether the individual subscribed for a term deposit. Here, \textit{age} is the sensitive attribute, with $<25$ and $>60$ as the disadvantaged group. 

\emph{\heart}\footnote{\url{https://www.kaggle.com/datasets/sulianova/cardiovascular-disease-dataset}} consists of patient measurements with respect to cardiovascular diseases, with information of 70,000 individuals characterized by 11 attributes (6 categorical, 5 numerical), including age, height, weight, blood pressure, and a binary target indicating whether the patient has heart disease. Here, \textit{sex} is the sensitive attribute, with ``female'' as the disadvantaged group. 

%% file: appendix/exp-details-additional.tex
\section{Additional Experimental Details for Reproducibility}
\label{apdx:exp-details-additional}

\subsection{Computing Infrastructure}
\label{apdx:computing-infrastructure}

Our large-scale experimental study incorporated 15 null imputation techniques, 6 model types, 7 datasets, 10 evaluation scenarios, and 6 seeds for controlling randomization across the entire pipeline. This configuration led to an overwhelming total of \numexps experimental pipelines. The complexity was further compounded by the necessity to tune both null imputers and ML models within each pipeline, employing a bootstrap of 50 estimators to assess model stability. Depending on the dataset size, null imputation technique, and model type, the execution time for a single experimental pipeline ranged from 3 minutes to 16.5 hours. Executing such a large number of pipelines would be impossible without a suitable experimental environment comprising a high-performance computing (HPC) cluster for execution and a 3-node MongoDB cluster for storing experimental results.

Our evaluation suite controller provides a useful separation of experimental pipelines, and the MongoDB cluster gives the ability to store results independently from a local file system. To manage the extensive computational demands, we utilized several optimizations (detailed in Section~\ref{sec:benchmark-archicture}) and the SLURM software system, which allowed us to execute up to 100 simultaneous jobs on the HPC cluster. To run experimental pipelines, each SLURM job was allocated 12-24 physical cores or one GPU card, along with 16-96 GB of RAM, depending on the dataset size, imputation method, and model type. For null imputation using \datawig, we employed an \texttt{RTX8000 NVIDIA} GPU card with 48 GB of GPU memory, while all other tasks were handled by a \texttt{2x Intel Xeon Platinum 8268 24C 205W 2.9GHz} processor. All evaluated imputation algorithms were implemented using Python 3.9, utilizing their original source code. Some of this code was generously shared directly by the respective authors, for which we express our sincere gratitude. All dependencies are specified in our repository, along with detailed instructions in the README for installing and running the evaluation suite.

To effectively store the results of the experiments in the MongoDB cluster, we designed a data model with unique GUIDs for each experimental pipeline. This model comprehensively encompasses all facets of each experiment: dataset names, all used seeds, imputer and model hyper-parameters, and more. Since we have multiple options of input variables for each stage of the experimental pipeline, we needed to consolidate and unify the data to enable effective analysis of the results of the experiments. Hence, we established GUIDs for each layer of the experimental pipeline, encompassing identifiers for each imputation stage, model tuning stage, model profiling stage, and experimental session. This data model helps us access data on different levels, unify the structure of all experiments, increase the safety of aggregations and result interpretation, and simplify the creation of visualizations.

\begin{figure*}[h!]
    \begin{minipage}{\linewidth}
        \begin{multicols}{2}
            \subsection{Extended Methodology for Simulating Missingness}
            \label{apdx:extended-methodology}
            
            Similarly to Section~\ref{sec:missingness}, this subsection outlines realistic missingness scenarios based on intrinsic dataset properties. Tables \ref{tab:diabetes-rates}-\ref{tab:folk-emp-rates} report the demographic composition of all datasets, specifically, the proportions and base rates of each protected group. Figures \ref{fig:folk-emp-sim}-\ref{fig:german-sim} display the Spearman correlation coefficients between covariates and the target variable, as well as feature importance derived using scikit-learn’s built-in functionality for datasets not covered in Section~\ref{sec:missingness}. For datasets with numerous features, plots include only the most correlated features. These proportions, base rates, correlation coefficients, and feature importance guided the conditions of our missingness scenarios, aiming to reflect realistic instances of non-response caused by disparate access, distrust, or procedural injustice, where disadvantaged groups exhibit more missing values than privileged groups. Tables~\ref{tab:simulation-folk-emp}-\ref{tab:simulation-german} detail the missingness scenarios for the \folkemp, \folk, \bank, \heart, \law, and \german datasets, respectively.
        \end{multicols}
    \end{minipage}

    \vspace{2cm}
\end{figure*}

\begin{table*}[h!]
    \caption{Table~\ref{tab:diabetes-rates}: Proportions and Base Rates for \diabetes.}
    \vspace{-0.3cm}
    \centering
    \begin{tabular}{|c|c|c|c|}
    \hline
     & overall & gender\_priv & gender\_dis \\ 
    \hline
    Proportions & 1.0 & 0.621 & 0.379 \\ 
    \hline
    Base Rates & 0.291 & 0.272 & 0.321 \\ 
    \hline
    \end{tabular}
    \label{tab:diabetes-rates}
    \vspace{1cm}
\end{table*}

\begin{table*}[h!]
\caption{Table~\ref{tab:german-rates}: Proportions and Base Rates for \german.}
\vspace{-0.3cm}
\centering
    \begin{tabular}{|c|c|c|c|c|c|c|c|}
    \hline
     & overall & sex\_priv & sex\_dis & age\_priv & age\_dis & sex\&age\_priv & sex\&age\_dis \\ 
    \hline
    Proportions & 1.0 & 0.69 & 0.31 & 0.81 & 0.19 & 0.895 & 0.105 \\ 
    \hline
    Base Rates & 0.7 & 0.723 & 0.648 & 0.728 & 0.579 & 0.717 & 0.552 \\ 
    \hline
    \end{tabular}
\label{tab:german-rates}
\vspace{1cm}
\end{table*}

\begin{table*}[h!]
\caption{Table~\ref{tab:folk-rates}: Proportions and Base Rates for \folk.}
\vspace{-0.3cm}
\centering
    \begin{tabular}{|c|c|c|c|c|c|c|c|}
    \hline
     & overall & sex\_priv & sex\_dis & race\_priv & race\_dis & sex\&rac1p\_priv & sex\&race\_dis \\ 
    \hline
    Proportions & 1.0  & 0.511 & 0.489 & 0.678 & 0.322 & 0.829 & 0.171 \\ 
    \hline
    Base Rates & 0.35 & 0.422 & 0.274 & 0.389 & 0.267 & 0.374 & 0.232 \\ 
    \hline
    \end{tabular}
\label{tab:folk-rates}
\vspace{1cm}
\end{table*}

\begin{table*}[h!]
\caption{Table~\ref{tab:law-rates}: Proportions and Base Rates for \law.}
\vspace{-0.3cm}
\centering
    \begin{tabular}{|c|c|c|c|c|c|c|c|}
    \hline
     & overall & male\_priv & male\_dis & race\_priv & race\_dis & male\&race\_priv & male\&race\_dis \\ 
    \hline
    Proportions & 1.0 & 0.561 & 0.439 & 0.841 & 0.159 & 0.917 & 0.083 \\ 
    \hline
    Base Rates & 0.89 & 0.899 & 0.878 & 0.921 & 0.723 & 0.906 & 0.713 \\ 
    \hline
    \end{tabular}
\label{tab:law-rates}
\vspace{1cm}
\end{table*}

\begin{table*}[h!]
    \caption{Table~\ref{tab:bank-rates}: Proportions and Base Rates for \bank.}
    \vspace{-0.3cm}
    \centering
    \begin{tabular}{|c|c|c|c|}
    \hline
     & overall & age\_priv & age\_dis \\ 
    \hline
    Proportions & 1.0 & 0.955 & 0.045 \\ 
    \hline
    Base Rates & 0.117 & 0.106 & 0.341 \\ 
    \hline
    \end{tabular}
    \label{tab:bank-rates}
    \vspace{1cm}
\end{table*}

\begin{table*}[h!]
    \caption{Table~\ref{tab:heart-rates}: Proportions and Base Rates for \heart.}
    \vspace{-0.3cm}
    \centering
    \begin{tabular}{|c|c|c|c|}
    \hline
     & overall & gender\_priv & gender\_dis \\ 
    \hline
    Proportions & 1.0 & 0.35 & 0.65 \\ 
    \hline
    Base Rates & 0.5 & 0.505 & 0.497 \\ 
    \hline
    \end{tabular}
    \label{tab:heart-rates}
    \vspace{1cm}
\end{table*}

\begin{table*}[h!]
\caption{Table~\ref{tab:folk-emp-rates}: Proportions and Base Rates for \folkemp.}
\vspace{-0.3cm}
\centering
    \begin{tabular}{|c|c|c|c|c|c|c|c|}
    \hline
     & overall & sex\_priv & sex\_dis & race\_priv & race\_dis & sex\&rac1p\_priv & sex\&race\_dis \\ 
    \hline
    Proportions & 1.0  & 0.490 & 0.510 & 0.625 & 0.375 & 0.806 & 0.194 \\ 
    \hline
    Base Rates & 0.57 & 0.617 & 0.525 & 0.563 & 0.582 & 0.577 & 0.541 \\ 
    \hline
    \end{tabular}
\label{tab:folk-emp-rates}
\vspace{2cm}
\end{table*}

\begin{table*}[h]
    \centering
    \caption{Table~\ref{tab:simulation-folk-emp}: Missingness scenarios for an error rate of 30\% for \folkemp. AGEP is a numerical column;  DIS, MIL, SCHL are categorical columns.}
    \vspace{-0.2cm}
    \begin{tabular}{lllll}
        \toprule
        \textbf{Mechanism} & \textbf{Missing Column ($\mathcal{F}^m)$} & \textbf{Conditional Column ($I$)} & \textbf{Pr($\mathcal{F}^m$ | $I$ is dis)} & \textbf{Pr($\mathcal{F}^m$ | $I$ is priv)} \\
        \midrule
        MCAR &\makecell[tl]{DIS, MIL, AGEP, SCHL} & N/A & 0.3 & 0.3 \\
        \hline
        
        MAR & MIL, AGEP & SEX & 0.2 (female) & 0.1 (male)\\
         & DIS, SCHL & RAC1P & 0.2 (non-white) & 0.1 (white)\\
        \hline
         
         MNAR & DIS & DIS & 0.25 (with disability) & 0.05 (without disability) \\
          & MIL & MIL & \makecell[tl]{0.05 ($\in$ \{past duty,\\ training\}} & \makecell[tl]{0.25 ($\notin$ \{past duty,\\ training\}}\\
          & AGEP & AGEP & 0.25 (> 50) & 0.05 ($\leq50$) \\
          & SCHL & SCHL & 0.25 (< 21) & 0.05 ($\geq21$) \\
        \bottomrule
    \end{tabular}
    \label{tab:simulation-folk-emp}
    \vspace{1cm}
\end{table*}

\begin{figure*}[h!]
\begin{subfigure}[h]{0.22\linewidth}
    \centering
    \includegraphics[width=\linewidth]{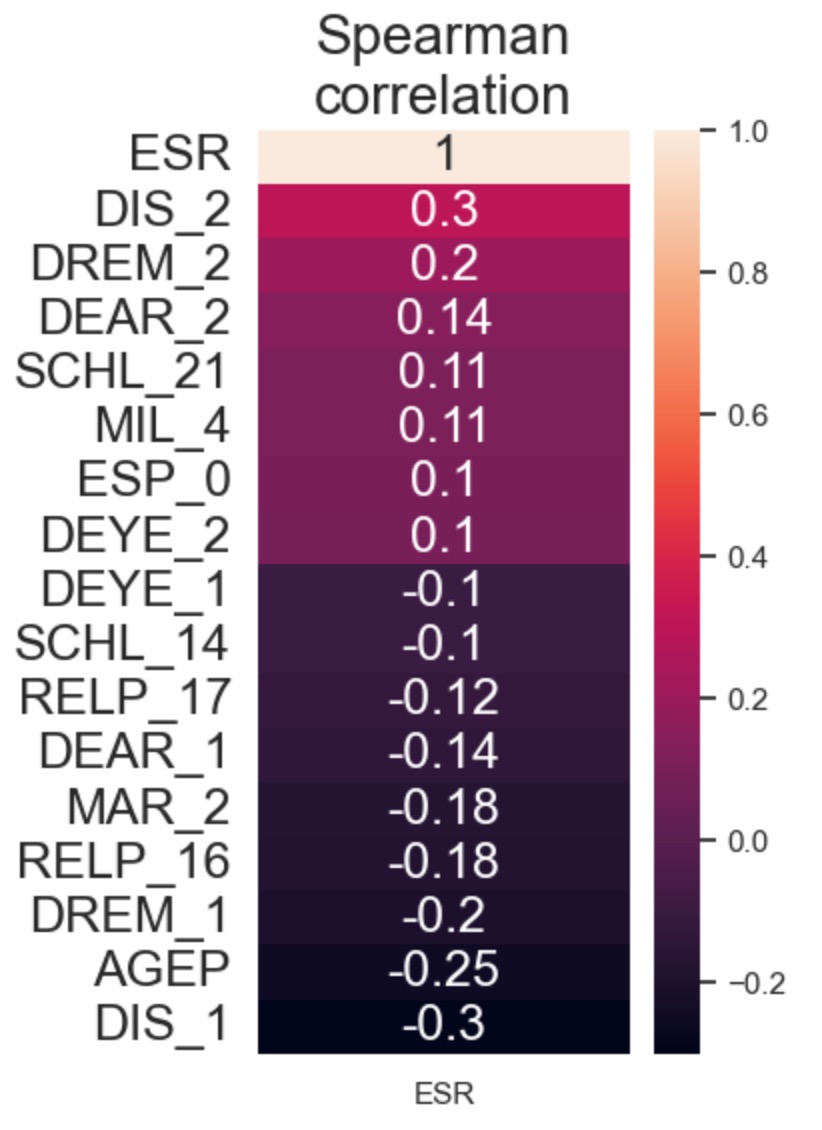}
    \caption{Correlation with label}
\end{subfigure}
\hspace{0.5cm}
\begin{subfigure}[h]{0.52\linewidth}
    \centering
    \includegraphics[width=\linewidth]{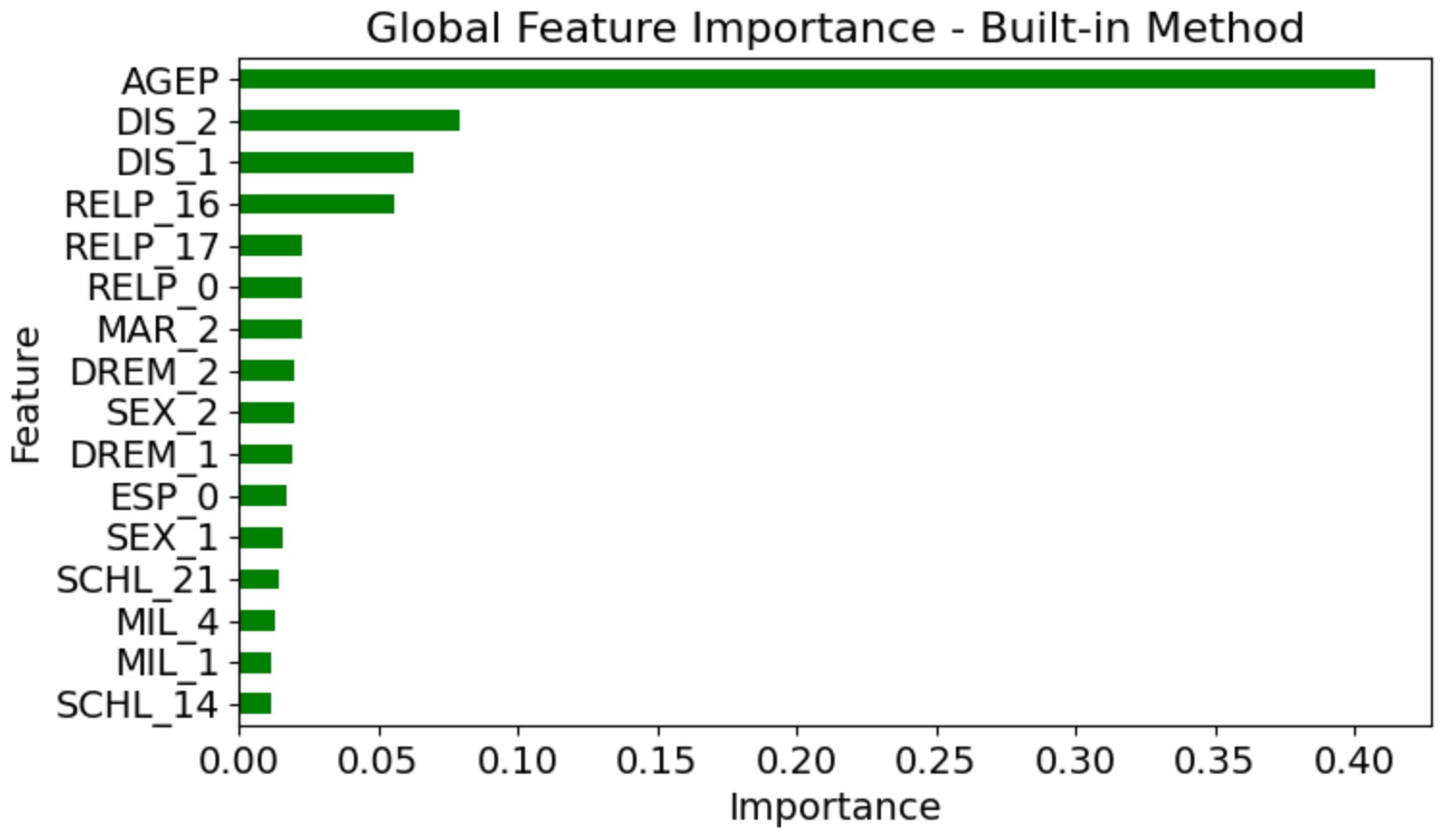}
    \caption{Feature importance}
\end{subfigure}
\vspace{-0.3cm}
\caption{EDA for designing missingness scenarios in \folkemp.}
\label{fig:folk-emp-sim}
\vspace{1cm}
\end{figure*}

\begin{table*}[h]
    \centering
    \caption{Table~\ref{tab:simulation-folk}: Missingness scenarios for an error rate of 30\% for \folk. WKHP, AGEP are numerical columns;  SCHL, MAR are categorical columns.}
    \vspace{-0.2cm}
    \begin{tabular}{lllll}
        \toprule
        \textbf{Mechanism} & \textbf{Missing Column ($\mathcal{F}^m)$} & \textbf{Conditional Column ($I$)} & \textbf{Pr($\mathcal{F}^m$ | $I$ is dis)} & \textbf{Pr($\mathcal{F}^m$ | $I$ is priv)} \\
        \midrule
        MCAR &\makecell[tl]{WKHP, AGEP, SCHL, MAR} & N/A & 0.3 & 0.3 \\
        \hline
        
        MAR & WKHP, SCHL & SEX & 0.2 (female) & 0.1 (male)\\
         & MAR, AGEP & RAC1P & 0.2 (non-white) & 0.1 (white)\\
        \hline
         
         MNAR & MAR & MAR & 0.25 (not married) & 0.05 (married) \\
          & WKHP & WKHP & 0.25 (< 40) & 0.05 ($\geq40$)\\
          & AGEP & AGEP & 0.25 (> 50) & 0.05 ($\leq50$) \\
          & SCHL & SCHL & 0.25 (< 21) & 0.05 ($\geq21$) \\
        \bottomrule
    \end{tabular}
    \label{tab:simulation-folk}
\end{table*}

\begin{figure*}[h!]
\begin{subfigure}[h]{0.2\linewidth}
    \centering
    \includegraphics[width=\linewidth]{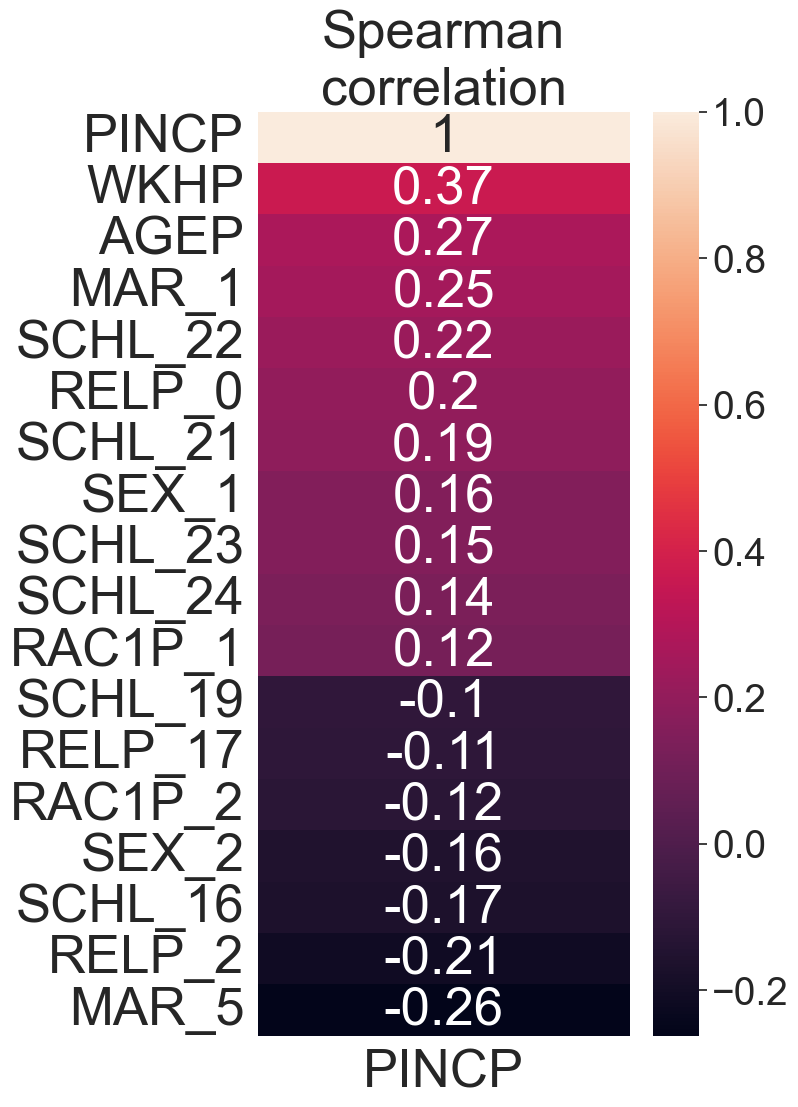}
    \caption{Correlation with label}
\end{subfigure}
\hspace{0.5cm}
\begin{subfigure}[h]{0.52\linewidth}
    \centering
    \includegraphics[width=\linewidth]{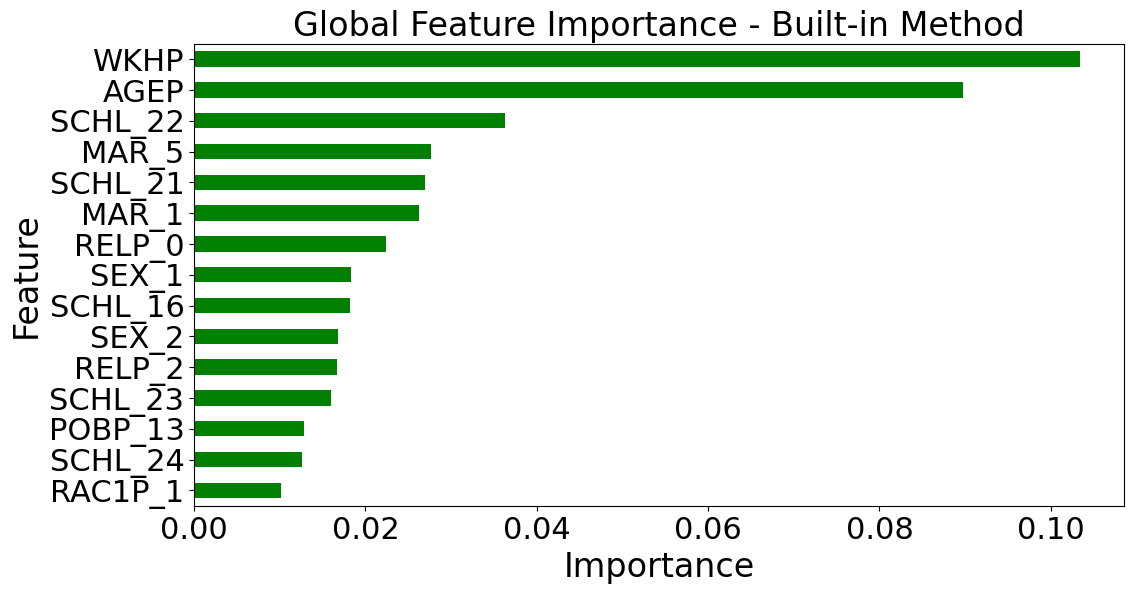}
    \caption{Feature importance}
\end{subfigure}
\vspace{-0.3cm}
\caption{EDA for designing missingness scenarios in \folk.}
\label{fig:folk-sim}
\end{figure*}

\begin{table*}[h]
    \centering
    \caption{Table~\ref{tab:simulation-bank}: Missingness scenarios for an error rate of 30\% for \bank. balance, campaign are numerical columns; education, job are categorical columns.}
    \vspace{-0.2cm}
    \begin{tabular}{lllll}
        \toprule
        \textbf{Mechanism} & \textbf{Missing Column ($\mathcal{F}^m)$} & \textbf{Conditional Column ($I$)} & \textbf{Pr($\mathcal{F}^m$ | $I$ is dis)} & \textbf{Pr($\mathcal{F}^m$ | $I$ is priv)} \\
        \midrule
        MCAR &\makecell[tl]{balance, campaign, education, job} & N/A & 0.3 & 0.3 \\
        \hline
        
        MAR & education, job & age & 0.18 ($\geq30$) & 0.12 (< 30)\\
         & balance, campaign & marital & 0.2 (single) & 0.1 (married)\\
        \hline
         
         MNAR & education & education & 0.2 (tertiary) & 0.1 (secondary) \\
          & job & job & \makecell[tl]{0.2 ($\notin$ \{management,\\ blue-collar\})} & \makecell[tl]{0.1 ($\in$ \{management,\\ blue-collar\})}\\
          & balance & balance & 0.2 ($\leq1000$) & 0.1 (> 1000) \\
          & campaign & campaign & 0.2 ($\leq1$) & 0.1 (> 1) \\
        \bottomrule
    \end{tabular}
    \label{tab:simulation-bank}
\end{table*}

\begin{figure*}[h!]
\begin{subfigure}[h]{0.2\linewidth}
    \centering
    \includegraphics[width=\linewidth]{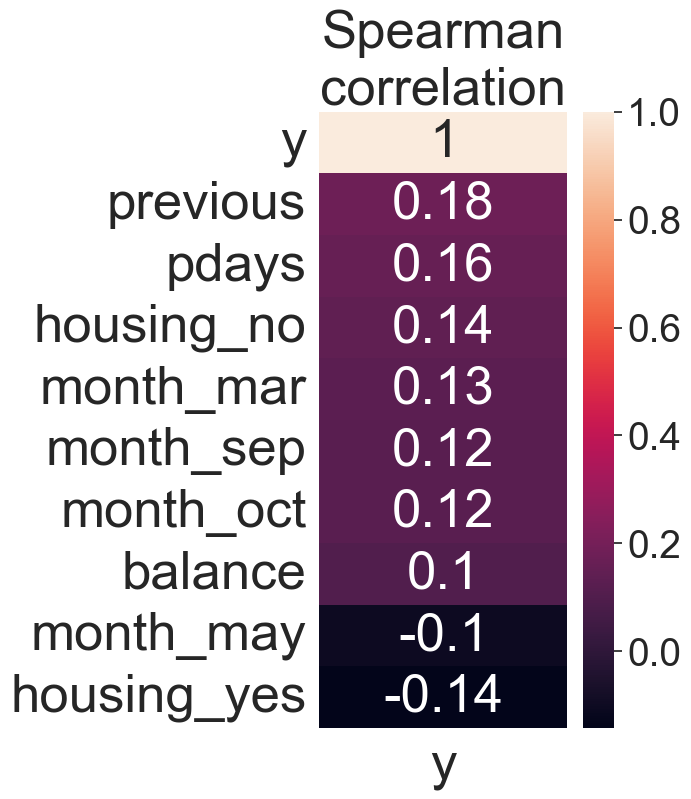}
    \caption{Correlation with label}
\end{subfigure}
\hspace{0.5cm}
\begin{subfigure}[h]{0.47\linewidth}
    \centering
    \includegraphics[width=\linewidth]{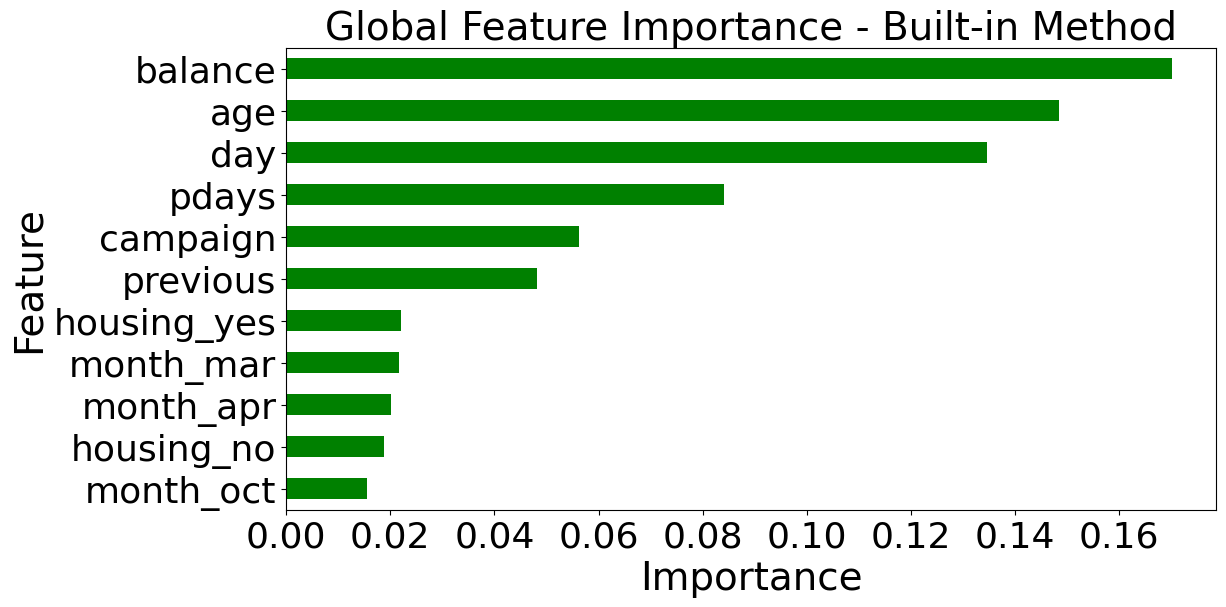}
    \caption{Feature importance}
\end{subfigure}
\vspace{-0.3cm}
\caption{EDA for designing missingness scenarios in \bank.}
\label{fig:bank-sim}
\end{figure*}

\begin{table*}[h]
    \centering
    \caption{Table~\ref{tab:simulation-heart}: Missingness scenarios for an error rate of 30\% for \heart. weight, height are numerical columns; cholesterol, gluc are categorical columns.}
    \vspace{-0.2cm}
    \begin{tabular}{lllll}
        \toprule
        \textbf{Mechanism} & \textbf{Missing Column ($\mathcal{F}^m)$} & \textbf{Conditional Column ($I$)} & \textbf{Pr($\mathcal{F}^m$ | $I$ is dis)} & \textbf{Pr($\mathcal{F}^m$ | $I$ is priv)} \\
        \midrule
        MCAR &\makecell[tl]{weight, height, cholesterol, gluc} & N/A & 0.3 & 0.3 \\
        \hline
        
        MAR & weight, height & gender & 0.2 (female) & 0.1 (male)\\
         & cholesterol, gluc & age & 0.2 ($\geq50$) & 0.1 (< 50)\\
        \hline
         
         MNAR & weight & weight & 0.25 ($\geq75$) & 0.05 (< 75) \\
          & height & height & 0.2 (< 160) & 0.1 ($\geq160$) \\
          & cholesterol & cholesterol & 0.16 (not normal) & 0.14 (normal) \\
          & gluc & gluc & 0.12 (not normal) & 0.18 (normal) \\
        \bottomrule
    \end{tabular}
    \label{tab:simulation-heart}
\end{table*}

\begin{figure*}[h!]
\begin{subfigure}[h]{0.2\linewidth}
    \centering
    \includegraphics[width=\linewidth]{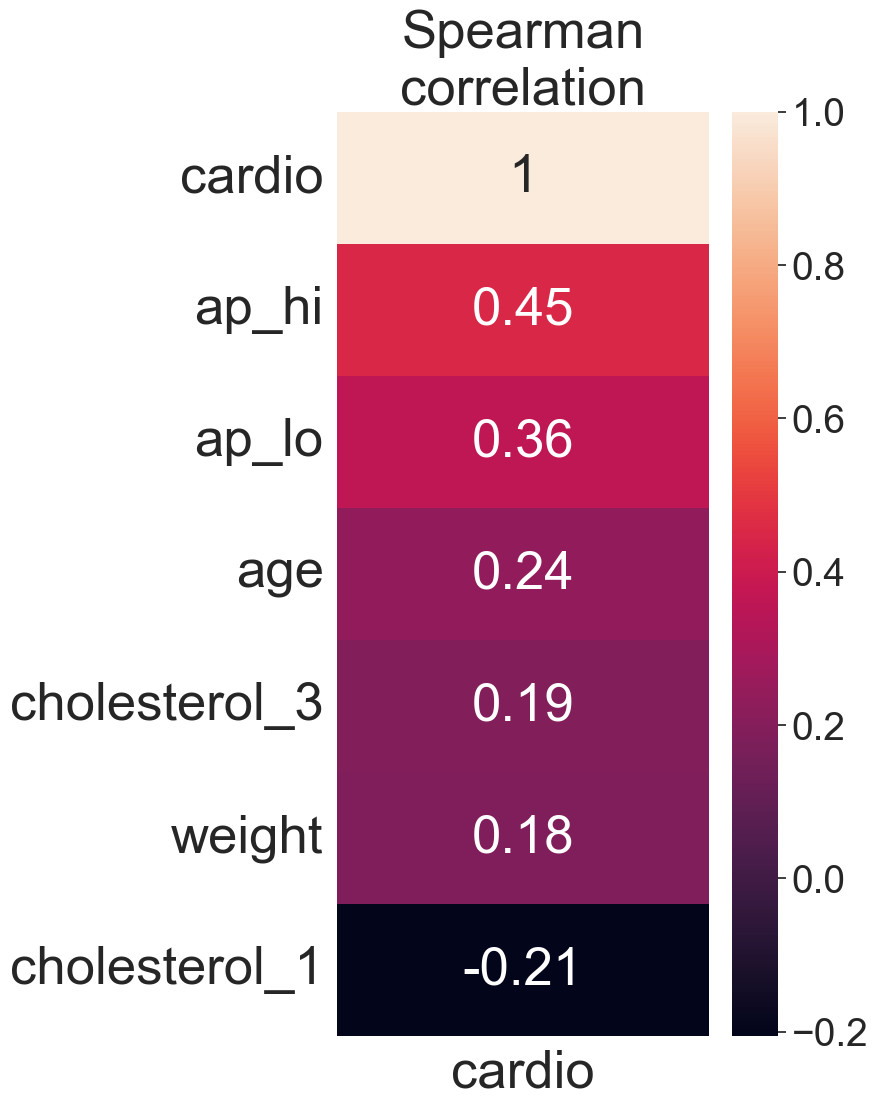}
    \caption{Correlation with label}
\end{subfigure}
\hspace{0.5cm}
\begin{subfigure}[h]{0.51\linewidth}
    \includegraphics[width=\linewidth]{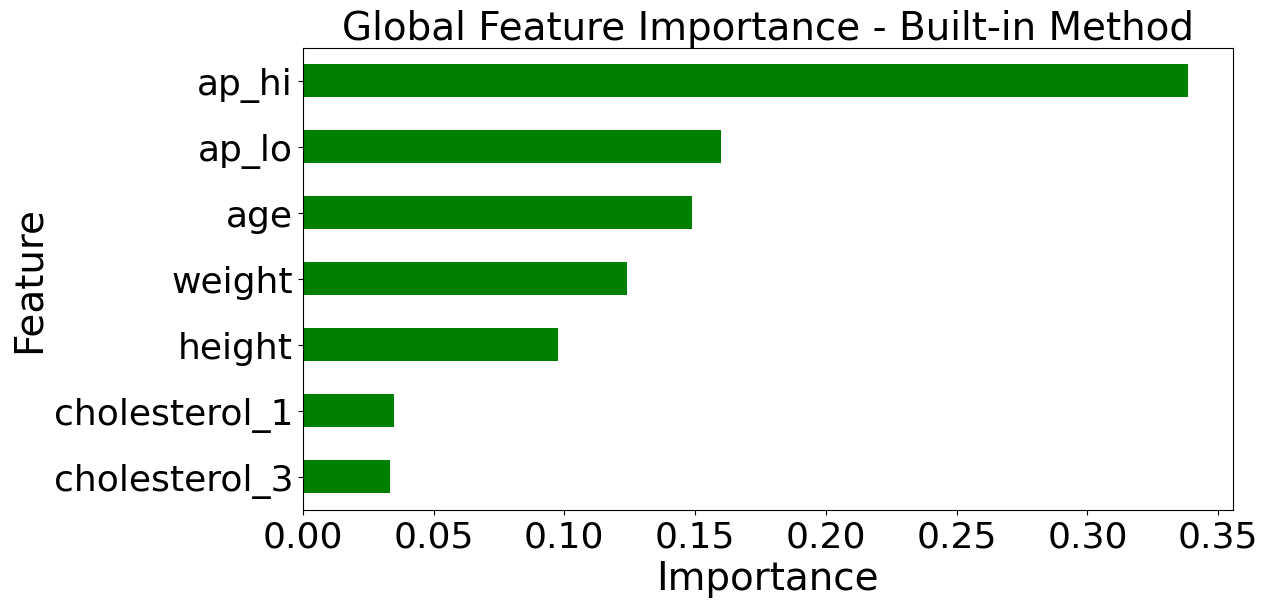}
    \caption{Feature importance}
\end{subfigure}
\vspace{-0.3cm}
\caption{EDA for designing missingness scenarios in \heart.}
\label{fig:bank-sim}
\end{figure*}

\begin{table*}[h]
    \centering
    \caption{Table~\ref{tab:simulation-law}: Missingness scenarios for an error rate of 30\% for \law. zfygpa, ugpa are numerical columns; fam\_inc, tier are categorical columns.}
    \vspace{-0.2cm}
    \begin{tabular}{lllll}
        \toprule
        \textbf{Mechanism} & \textbf{Missing Column ($\mathcal{F}^m)$} & \textbf{Conditional Column ($I$)} & \textbf{Pr($\mathcal{F}^m$ | $I$ is dis)} & \textbf{Pr($\mathcal{F}^m$ | $I$ is priv)} \\
        \midrule
        MCAR & \makecell[tl]{zfygpa, ugpa, fam\_inc, tier} & N/A & 0.3 & 0.3 \\
        \hline
        
        MAR &  \makecell[tl]{ugpa, zfygpa} & male & 0.2 (0) & 0.1 (1)\\
         & am\_inc, tier & race & 0.15 (non-white) & 0.15 (white)\\
        \hline
         
         MNAR 
         & ugpa & ugpa & 0.2 (< 3.0) & 0.1 ($\geq3.0$) \\
          & zfygpa & zfygpa & 0.2 ($\leq0$) & 0.1 (> 0) \\
          & fam\_inc & fam\_inc & 0.2 (< 4) & 0.1 ($\geq4$) \\
          & tier & tier & 0.2 (< 4) & 0.1 ($\geq4$) \\
        \bottomrule
    \end{tabular}
    \label{tab:simulation-law}
\end{table*}

\begin{figure*}[h!]
\begin{subfigure}[h]{0.25\linewidth}
    \centering
    \includegraphics[width=\linewidth]{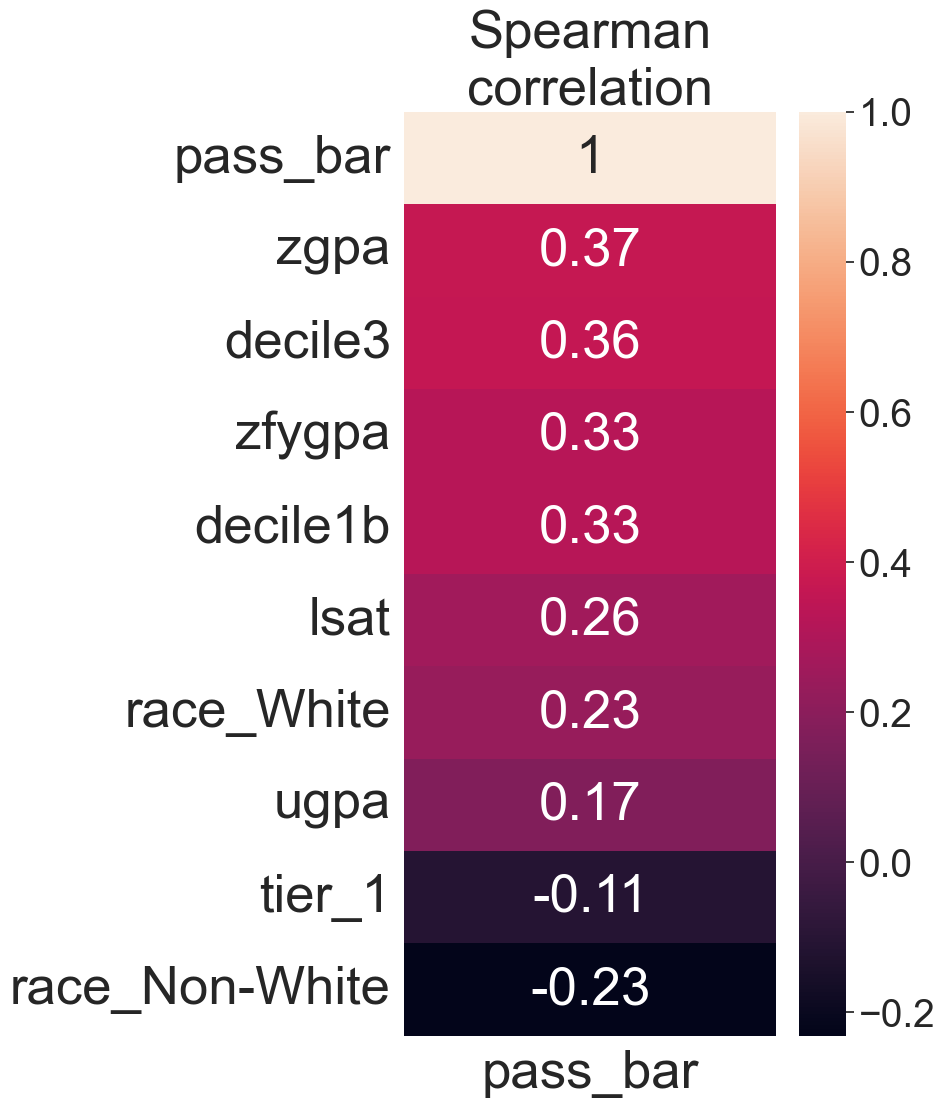}
    \caption{Correlation with label}
\end{subfigure}
\hspace{0.5cm}
\begin{subfigure}[h]{0.5\linewidth}
    \includegraphics[width=\linewidth]{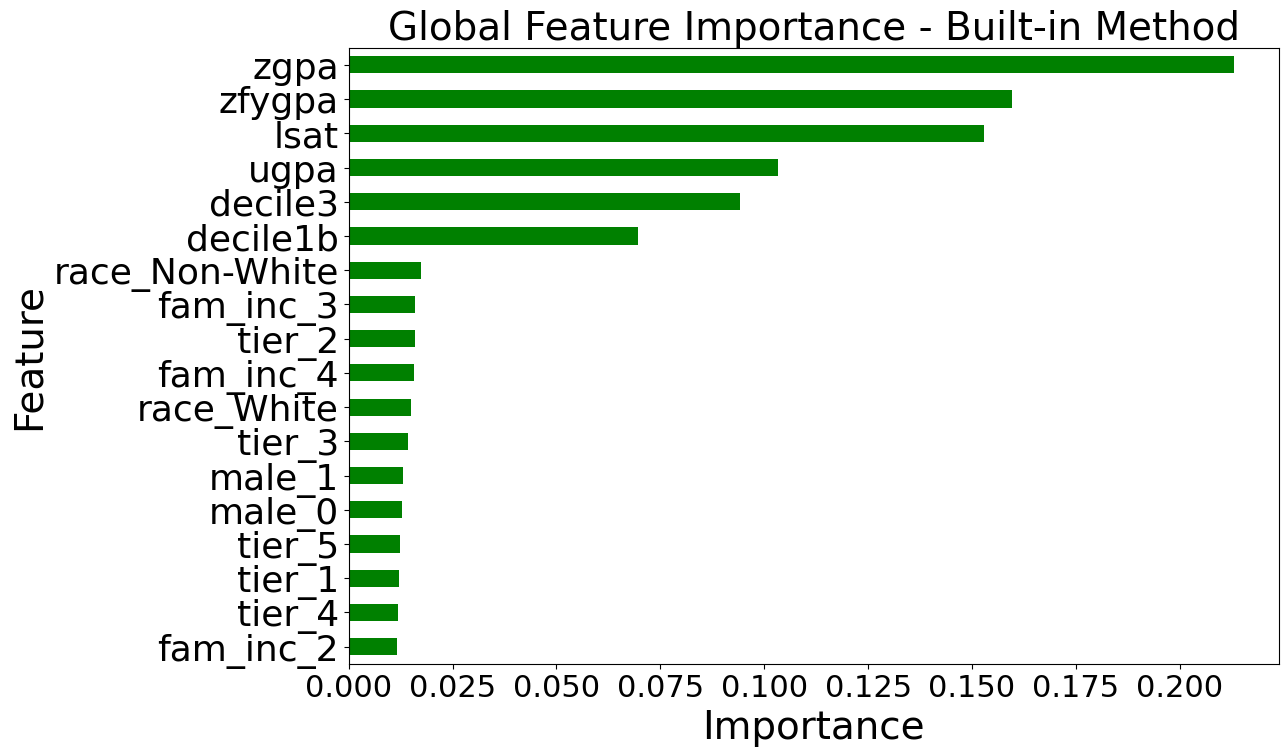}
    \caption{Feature importance}
\end{subfigure}
\vspace{-0.3cm}
\caption{EDA for designing missingness scenarios in \law.}
\label{fig:law-sim}
\end{figure*}

\begin{table*}[h]
    \centering
    \caption{Table~\ref{tab:simulation-german}: Missingness scenarios for an error rate of 30\% for \german. duration, credit-amount are numerical columns;  checking-account, savings-account, employment-since are categorical columns.}
    \vspace{-0.2cm}
    \begin{tabular}{lllll}
        \toprule
        \textbf{Mechanism} & \textbf{Missing Column ($\mathcal{F}^m)$} & \textbf{Conditional Column ($I$)} & \textbf{Pr($\mathcal{F}^m$ | $I$ is dis)} & \textbf{Pr($\mathcal{F}^m$ | $I$ is priv)} \\
        \midrule
        MCAR & \makecell[tl]{duration,\\credit-amount,\\checking-account,\\savings-account,\\employment-since} & N/A & 0.3 & 0.3 \\
        \hline
        
        MAR &  \makecell[tl]{savings-account,\\checking-account,\\ credit-amount} & age & 0.18 ($\leq25$) & 0.12 (> 25)\\
         & \makecell[tl]{employment-since,\\duration} & sex & 0.2 (female) & 0.1 (male)\\
        \hline
         
         MNAR 
         & checking-account & checking-account & 0.25 (no account) & 0.05 (not no account) \\
          & duration & duration & 0.25 ($\leq20$) & 0.05 (> 20) \\
          & savings-account & savings-account & 0.2 (not no savings account) & 0.1 (no savings account) \\
          & employment-since & employment-since & 0.2 ($\in$ \{<1 year, unemployed\}) & 0.1 ($\notin$ \{<1 year, unemployed\}) \\
          & credit-amount & credit-amount & 0.25 (> 5000) & 0.05 ($\leq5000$) \\
        \bottomrule
    \end{tabular}
    \label{tab:simulation-german}
\end{table*}
\begin{figure*}[h!]
\begin{subfigure}[h]{0.36\linewidth}
    \centering
    \includegraphics[width=\linewidth]{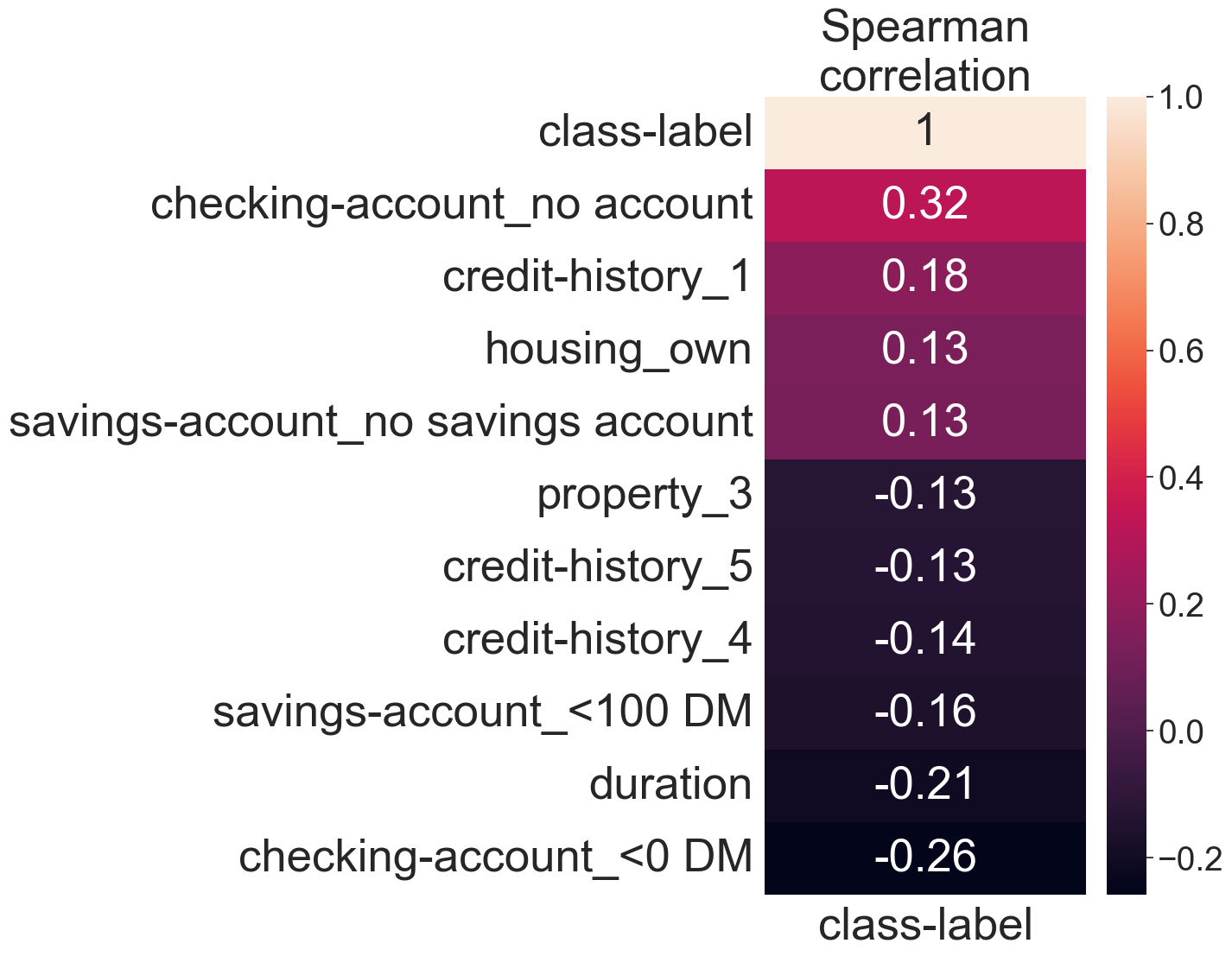}
    \caption{Correlation with label}
\end{subfigure}
\hspace{0.5cm}
\begin{subfigure}[h]{0.6\linewidth}
    \centering
    \includegraphics[width=\linewidth]{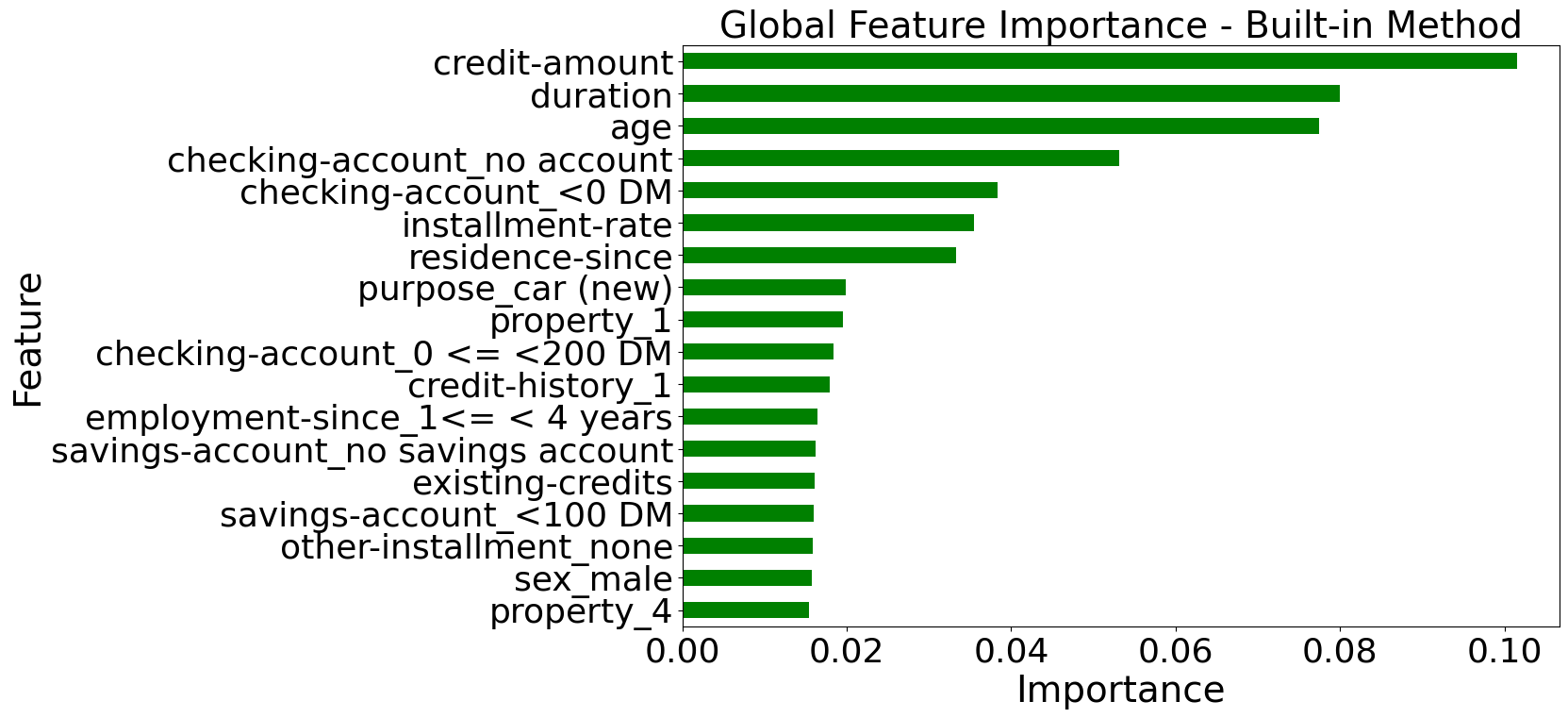}
    \caption{Feature importance}
\end{subfigure}
\vspace{-0.3cm}
\caption{EDA for designing missingness scenarios in \german.}
\label{fig:german-sim}
\end{figure*}

%% file: appendix/single-mechanism-additional.tex
\section{Additional Experimental Results for Single- and Multi-mechanism Missingness}
\label{apdx:single-mechanism-additional}

\balance
\subsection{Accuracy and Efficiency of MVI Techniques}
\label{apdx:mvi-metrics}

\begin{figure}[H]
\begin{subfigure}[h]{\linewidth}
    \includegraphics[width=\linewidth]{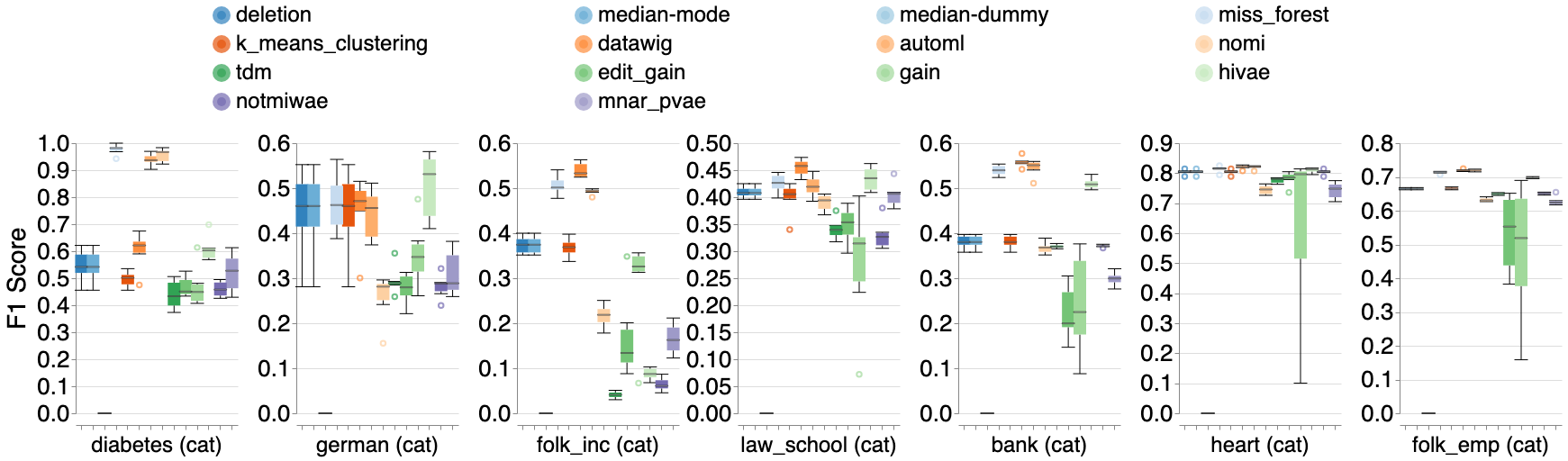}
    \caption{Missing Completely At Random (\mcar)}
\end{subfigure}

\begin{subfigure}[h]{\linewidth}
    \includegraphics[width=\linewidth]{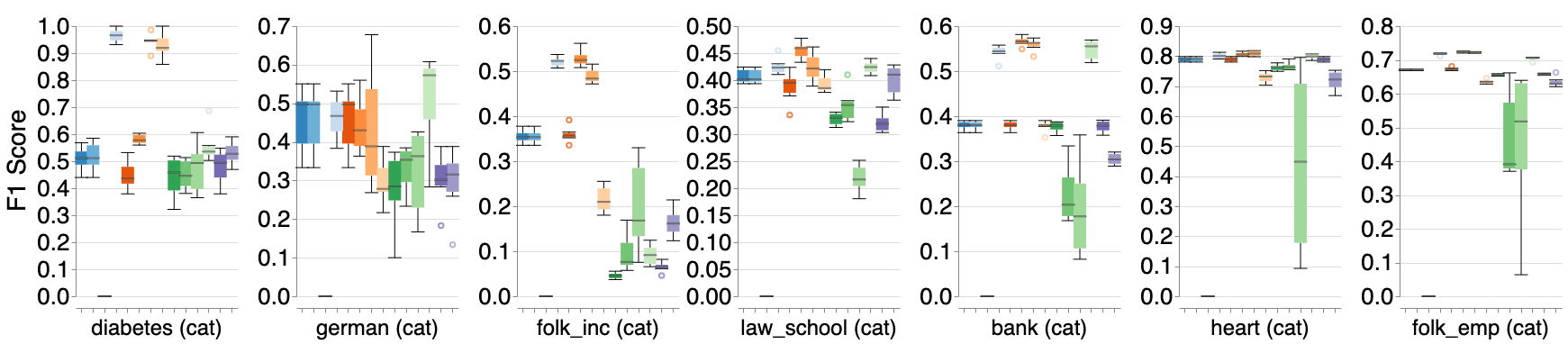}
    \caption{Missing At Random (\mar)}
\end{subfigure}

\begin{subfigure}[h]{\linewidth}
    \includegraphics[width=\linewidth]{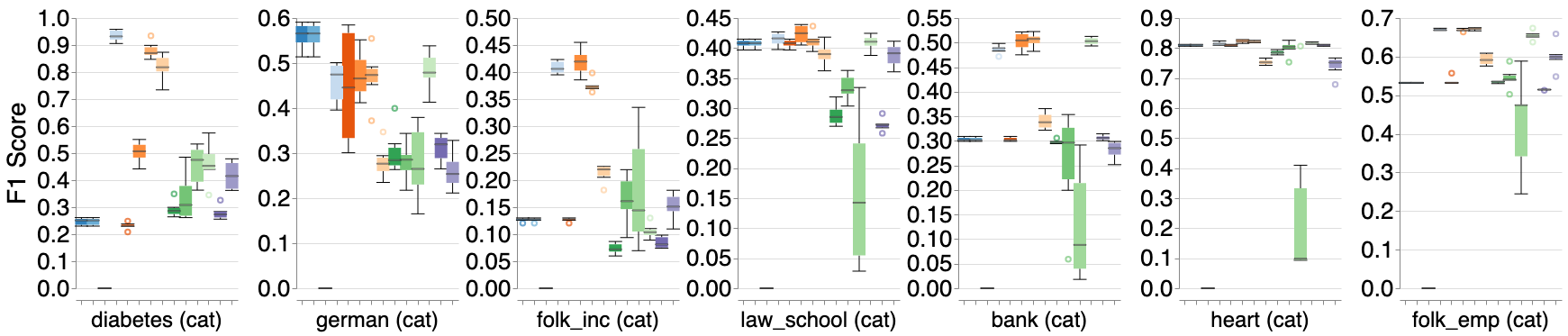}
    \caption{Missing Not At Random (\mnar)}
\end{subfigure}

\begin{subfigure}[h]{\linewidth}
    \centering
    \includegraphics[width=\linewidth]{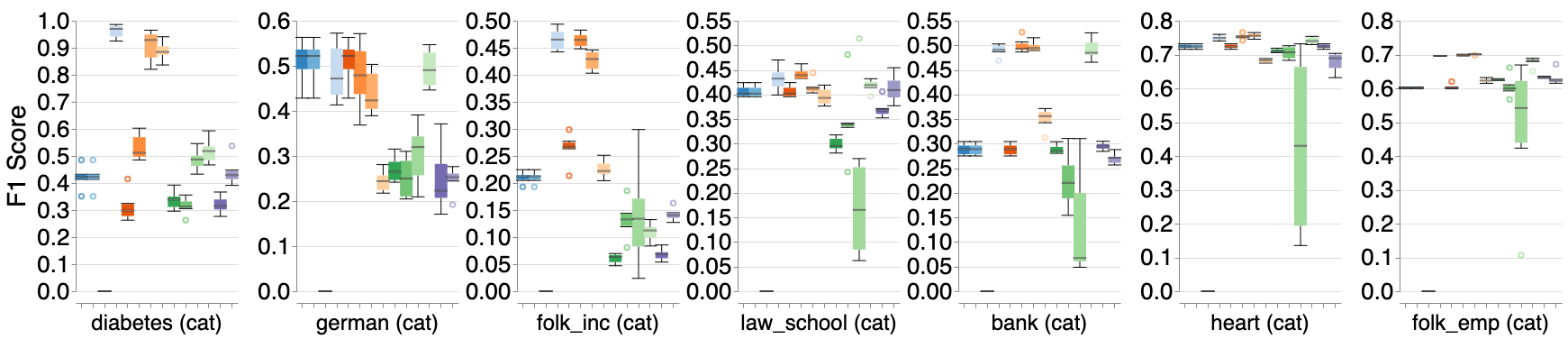}
    \caption{Mixed missingness (\mcar \& \mar \& \mnar)}
\end{subfigure}

\caption{F1 of different imputation strategies (colors in the legend) averaged across categorical columns with nulls per dataset (x-axis) and missingness mechanisms (subplots). Datasets are ordered in increasing order by size.}
\label{fig:exp1-imp-f1}
\end{figure}

\begin{figure}[H]
\begin{subfigure}[h]{\linewidth}
    \includegraphics[width=\linewidth]{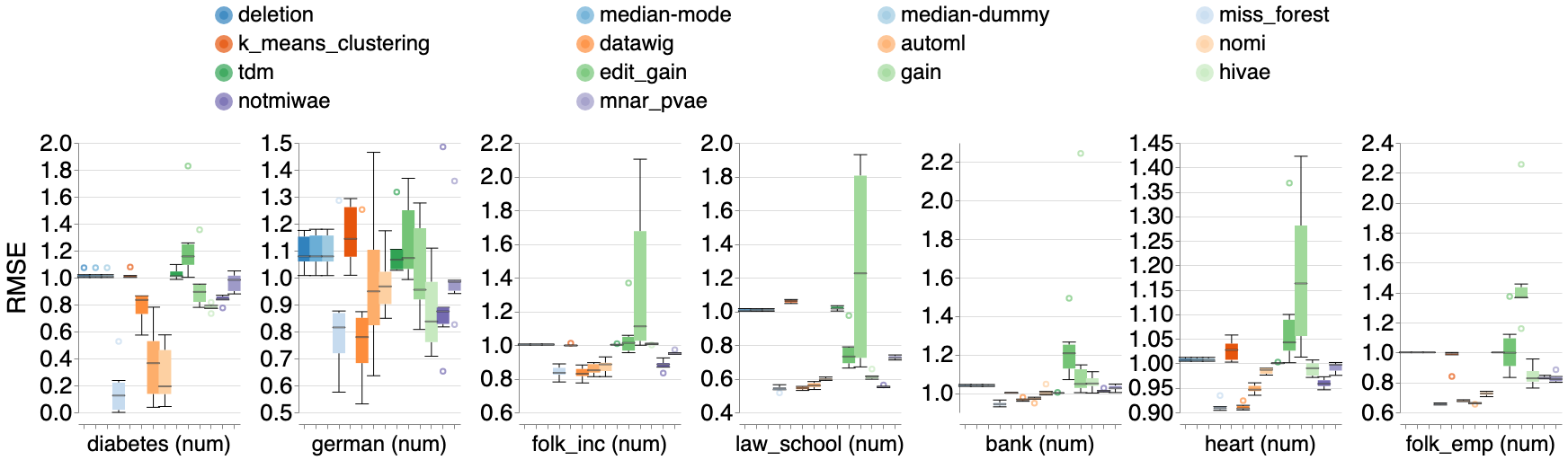}
    \caption{Missing Completely At Random (\mcar)}
\end{subfigure}

\begin{subfigure}[h]{\linewidth}
    \includegraphics[width=\linewidth]{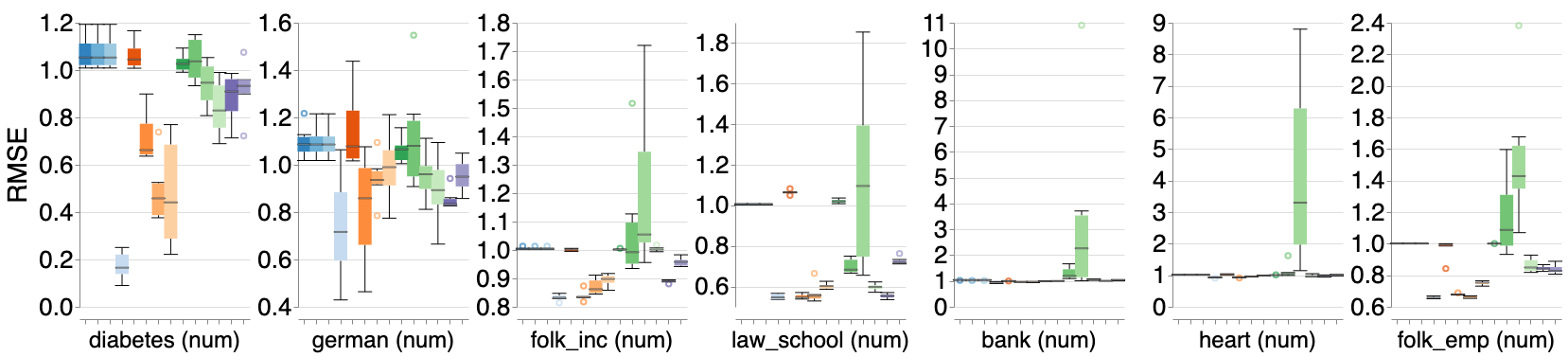}
    \caption{Missing At Random (\mar)}
\end{subfigure}

\begin{subfigure}[h]{\linewidth}
    \includegraphics[width=\linewidth]{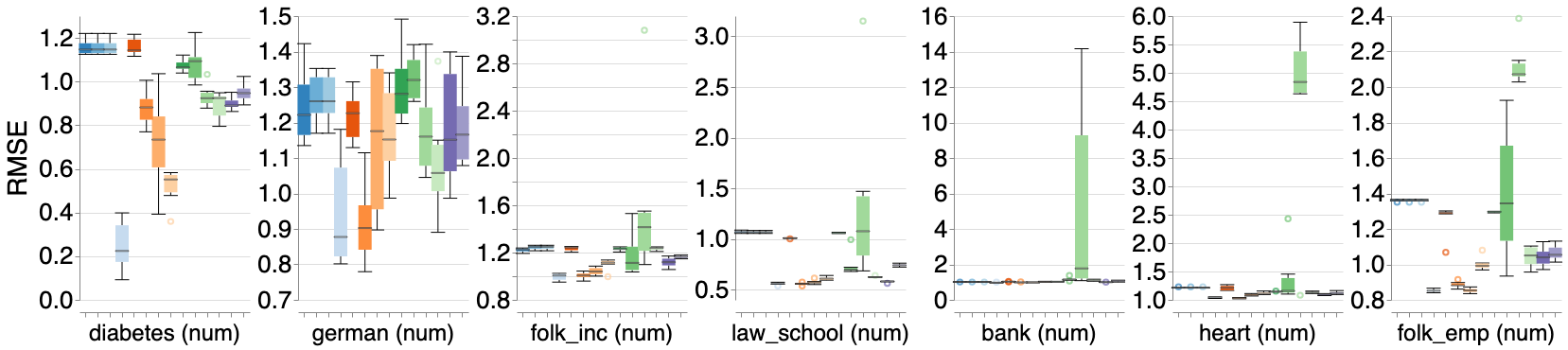}
    \caption{Missing Not At Random (\mnar)}
\end{subfigure}

\begin{subfigure}[h]{\linewidth}
    \centering
    \includegraphics[width=\linewidth]{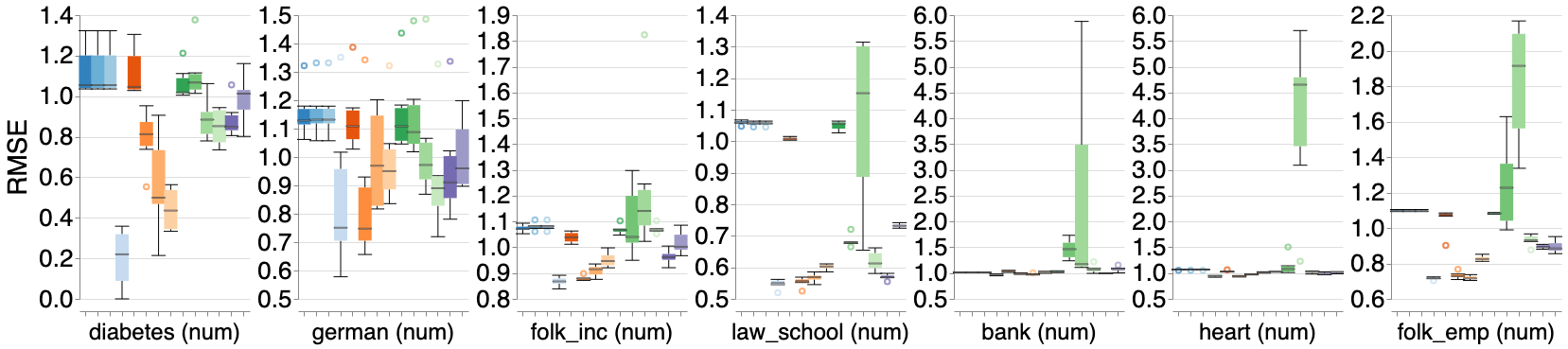}
    \caption{Mixed missingness (\mcar \& \mar \& \mnar)}
\end{subfigure}

\caption{RMSE of different imputation strategies (colors in the legend) averaged across numerical columns with nulls per dataset (x-axis) and missingness mechanisms (subplots). Datasets are ordered in increasing order by size.}
\label{fig:exp1-imp-rmse}
\end{figure}

In this section, we extend Section~\ref{sec:imputation-quality} by presenting additional results on the accuracy and efficiency of MVI techniques across single- and multi-mechanism scenarios (S1-S3, S10). Figures~\ref{fig:exp1-imp-f1} and \ref{fig:exp1-imp-rmse} illustrate imputation accuracy, measured by F1 scores for categorical columns and RMSE for numerical columns, respectively. Tables 18–22 summarize the training time of various MVI techniques (in seconds), averaged across these scenarios. In the tables, imputers are sorted by runtime on the \texttt{folk\_emp} dataset, while datasets are ordered by their number of rows. Dataset shapes indicate training sets consisting of 70\% complete rows and 30\% rows with missing values. In the tables, the reported values represent mean runtimes across different seeds, with standard deviations included. Higher F1 scores and lower RMSE and runtimes are desirable for better performance. Notably, \boostclean is excluded as it is a joint cleaning-and-training technique that does not generate imputed rows.

Table 18 (same as Table 4 from Section~\ref{sec:time}, duplicated for convenience) reveals that statistical imputers such as \deletion,\\ \mode, and \dummy are the fastest, while still delivering competitive accuracy for larger datasets like \heart and \texttt{folk\_emp}, as shown in Figures~\ref{fig:exp1-imp-f1} and \ref{fig:exp1-imp-rmse}. In contrast, \missforest, \datawig, and \automl exhibit the longest training times, with at least one of these methods achieving the highest imputation accuracy in most cases. Interestingly, \automl requires three times more training time than \datawig, the second most computationally intensive technique. This difference is due to the auto-ML nature of \automl, which involves extensive hyperparameter and network architecture tuning.

Among non-statistical techniques, \mnarpvae, \editgain, and \nomi are the most efficient in terms of runtime. Notably, \nomi provides competitive accuracy comparable to \missforest, \datawig, and \automl, making it the most optimal technique in terms of the trade-off between imputation accuracy and training time. A particularly noteworthy comparison is between \gain and \editgain. As explained in Appendix~\ref{apdx:mvm-techniques}, EDIT accelerates the training of parametric imputation models. Applying EDIT to \gain results in a 28x speedup on the \texttt{folk\_emp} dataset, with even greater improvements for smaller datasets, as shown in Table 18. Importantly, with this runtime acceleration \editgain achieves accuracy comparable to \gain on most datasets, as shown in Figures~\ref{fig:exp1-imp-f1} and \ref{fig:exp1-imp-rmse}.

Tables~19–22 provide training times for individual scenarios (S1, S2, S3, and S10). These values are generally consistent with those in Table 18, with minor exceptions: \nomi outperforms \mnarpvae and \editgain in terms of speed for MCAR and MAR scenarios, and \clustering is faster than \hivae in the MAR scenario.

It is worth noting that the training times in Tables 18–22 exhibit substantial standard deviations for many MVI techniques. These deviations arise because the statistics are computed across pipelines executed with different random seeds, resulting in varying training-test splits and model initializations. Consequently, standard deviations can be significant for all MVI techniques. Furthermore, \mnarpvae, \notmiwae, \missforest, \datawig, and \automl show large standard deviations. For \mnarpvae and \notmiwae, their MNAR-specific nature makes them highly sensitive to the data characteristics, leading to variability across splits. In the case of \missforest, \datawig, and \automl, the presence of parameters such as maximum iterations or trials introduces early stopping mechanisms that can be influenced by the data from different splits, further contributing to runtime variability.

\begin{figure*}[h!]
    \label{tab:imp_runtime_all_scen}
    \caption*{Table~18: Training time (in seconds) of MVI techniques averaged across single- and multi-mechanism scenarios (S1-3, S10). Imputers are sorted by runtime on the \texttt{folk\_emp} dataset, and datasets are ordered by the number of rows. Dataset shapes reflect training sets with 30\% rows with nulls. Values represent mean runtime across seeds, with standard deviations included.}
    \input{tables/imputation_runtime_avg}
    \vspace{0.5cm}
\end{figure*}

\newpage

\begin{figure*}[h!]
    \label{tab:imp_runtime_mcar}
    \caption*{Table~19: Training time (in seconds) of MVI techniques for the MCAR scenario (S1). Imputers are sorted by runtime on the \texttt{folk\_emp} dataset, and datasets are ordered by the number of rows. Dataset shapes reflect training sets with 30\% rows with nulls. Values represent mean runtime across seeds, with standard deviations included.}
    \input{tables/imputation_runtime_mcar}
    \vspace{2cm}
\end{figure*}

\begin{figure*}[h!]
    \label{tab:imp_runtime_mar}
    \caption*{Table~20: Training time (in seconds) of MVI techniques for the MAR scenario (S2). Imputers are sorted by runtime on the \texttt{folk\_emp} dataset, and datasets are ordered by the number of rows. Dataset shapes reflect training sets with 30\% rows with nulls. Values represent mean runtime across seeds, with standard deviations included.}
    \input{tables/imputation_runtime_mar}
\end{figure*}

\begin{figure*}[h!]
    \label{tab:imp_runtime_mar}
    \caption*{Table~21: Training time (in seconds) of MVI techniques for the MNAR scenario (S3). Imputers are sorted by runtime on the \texttt{folk\_emp} dataset, and datasets are ordered by the number of rows. Dataset shapes reflect training sets with 30\% rows with nulls. Values represent mean runtime across seeds, with standard deviations included.}
    \input{tables/imputation_runtime_mnar}
    \vspace{2cm}
\end{figure*}

\begin{figure*}[h!]
    \label{tab:imp_runtime_mixed}
    \caption*{Table~22: Training time (in seconds) of MVI techniques for the multi-mechanism scenario (S10). Imputers are sorted by runtime on the \texttt{folk\_emp} dataset, and datasets are ordered by the number of rows. Dataset shapes reflect training sets with 30\% rows with nulls. Values represent mean runtime across seeds, with standard deviations included.}
    \input{tables/imputation_runtime_mixed}
\end{figure*}

\clearpage

\balance
\subsection{Additional Experimental Results for Predictive Model Performance}
\label{apdx:single-mechanism-fairness-metrics}

\begin{figure}[H]
\begin{subfigure}[h]{\linewidth}
    \centering
    \includegraphics[width=\linewidth]{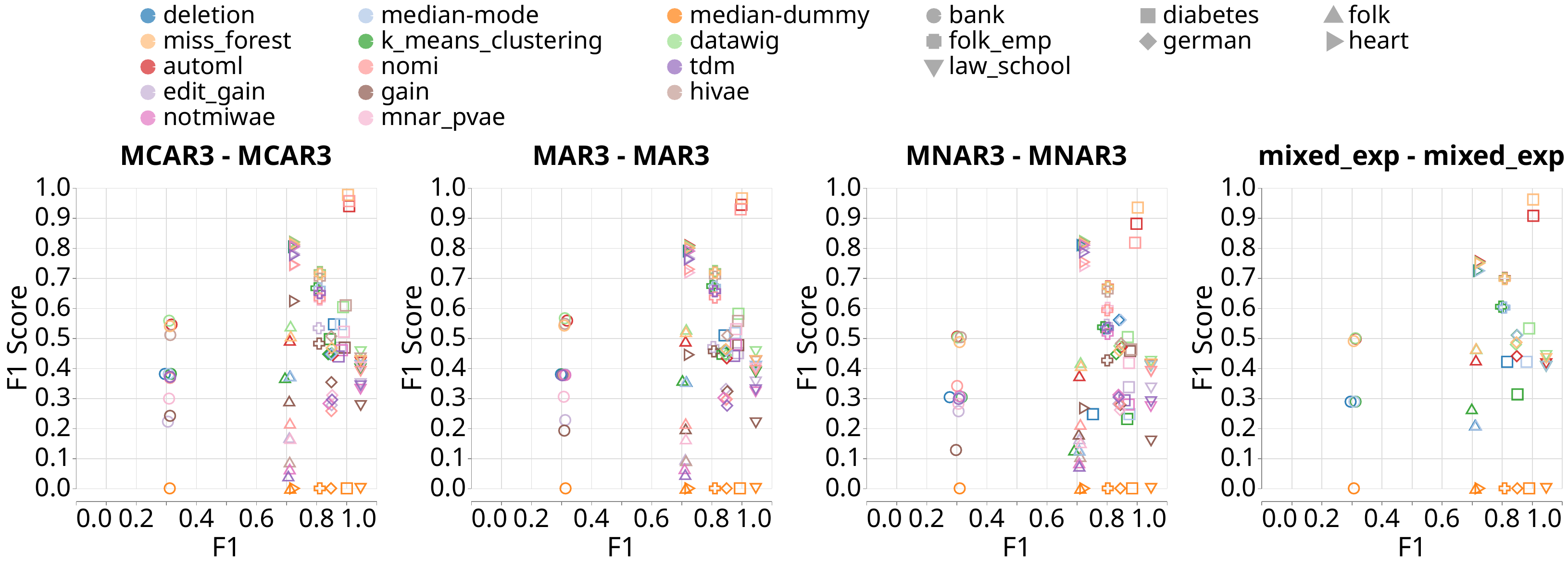}
    \caption{F1 (imputation) vs F1 (model)}
    \label{fig:imputation-f1-extended}
\end{subfigure}

\begin{subfigure}[h]{\linewidth}
    \centering
    \includegraphics[width=\linewidth]{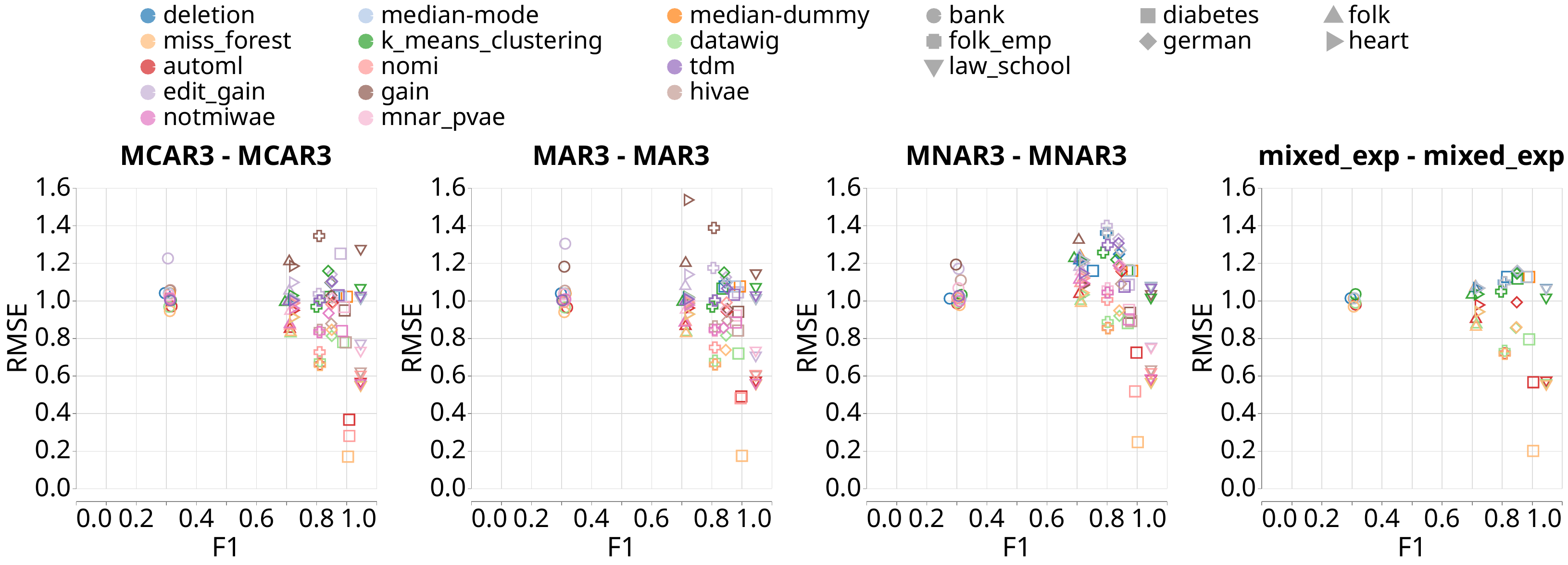}
    \caption{RMSE (imputation) vs F1 (model)}
    \label{fig:imputation-rmse-extended}
\end{subfigure}

\begin{subfigure}[h]{\linewidth}
    \centering
    \includegraphics[width=\linewidth]{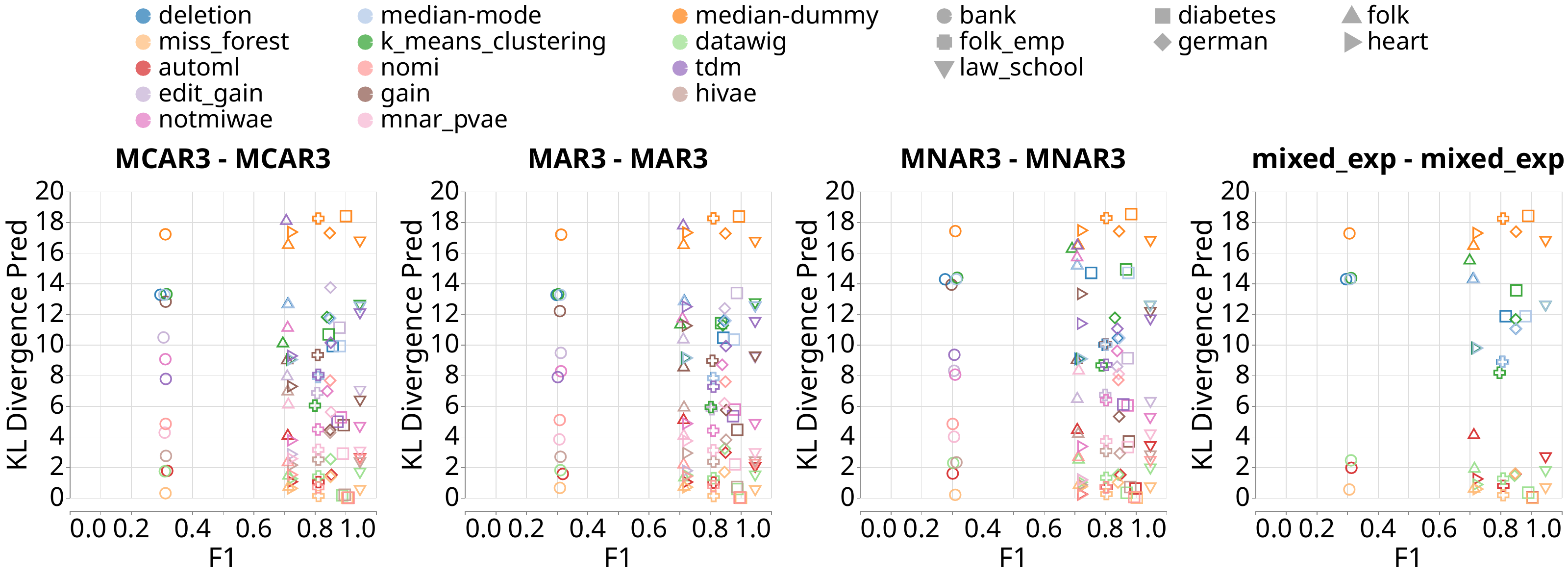}
    \caption{KL divergence (imputation) vs F1 (model)}
    \label{fig:imputation-kld-extended}
\end{subfigure}
\vspace{-0.4cm}
\caption{Imputation quality vs. model performance: imputation correctness (F1, RMSE and KL divergence) may not be indicative of model correctness (F1).}
\label{fig:imputation-quality-no-shift-extended}
\end{figure}

\begin{figure}[H]
\begin{subfigure}[h]{\linewidth}
    \includegraphics[width=\linewidth]{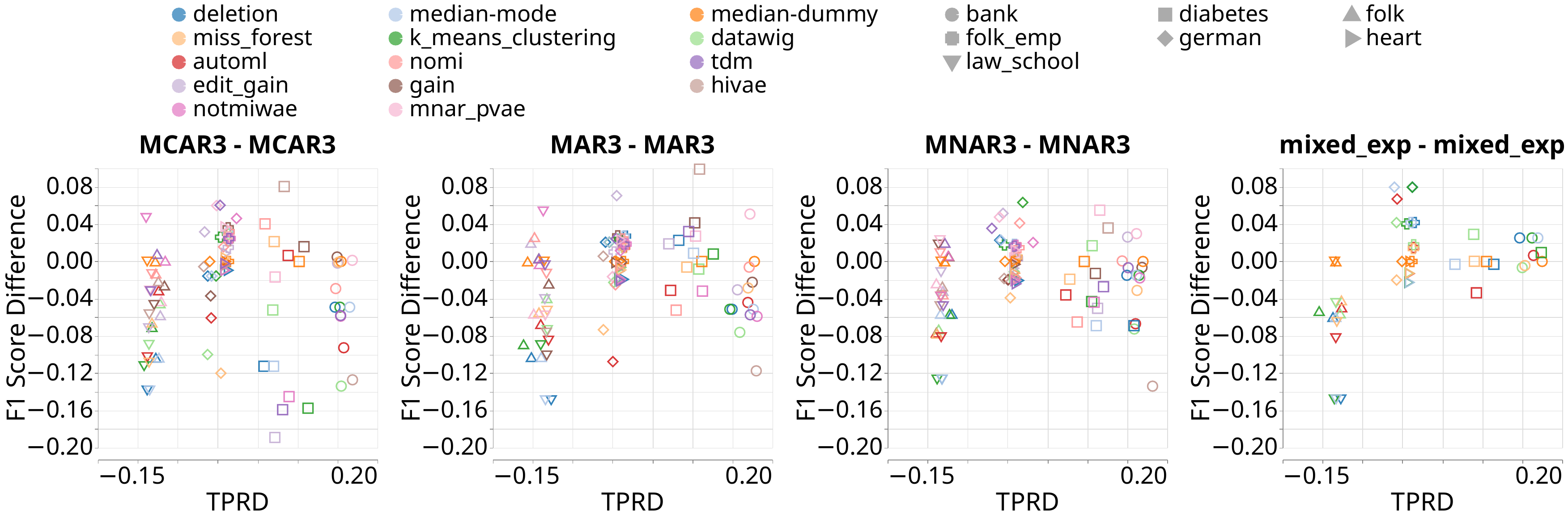}
    \caption{F1 score difference (imputation) vs TPRD (model)}
    \label{fig:imputation-f1-diff-extended}
\end{subfigure}

\begin{subfigure}[h]{\linewidth}
    \includegraphics[width=\linewidth]{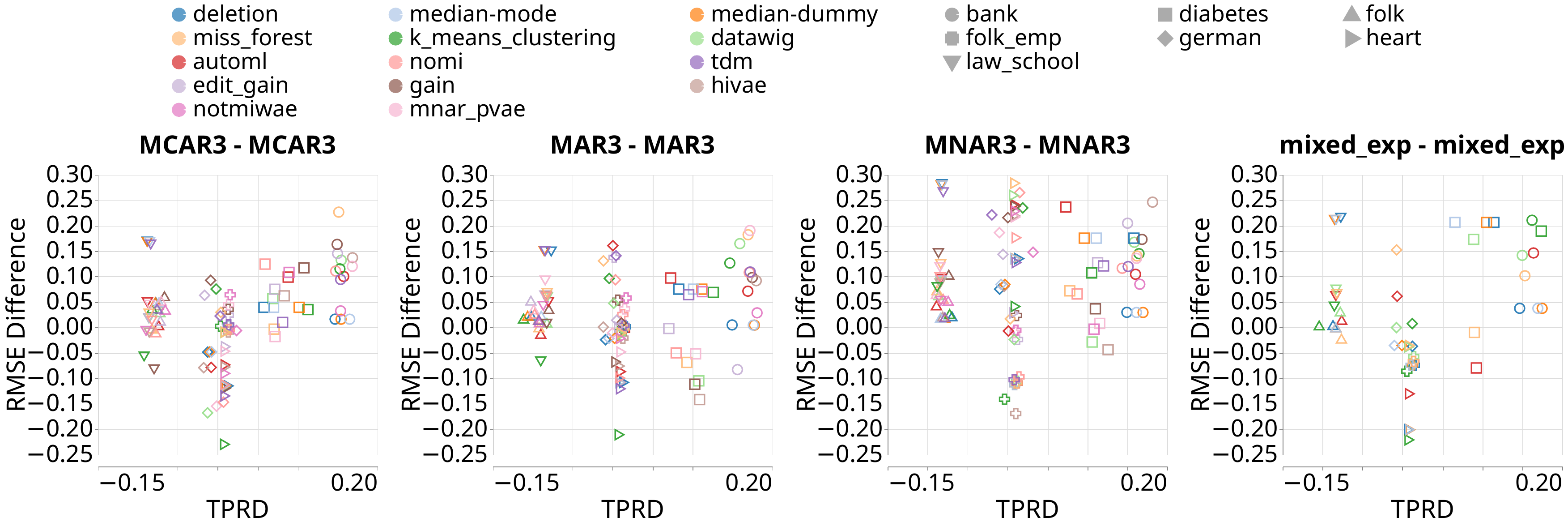}
    \caption{RMSE difference (imputation) vs TPRD (model)}
    \label{fig:imputation-rmse-diff-extended}
\end{subfigure}

\begin{subfigure}[h]{\linewidth}
    \includegraphics[width=\linewidth]{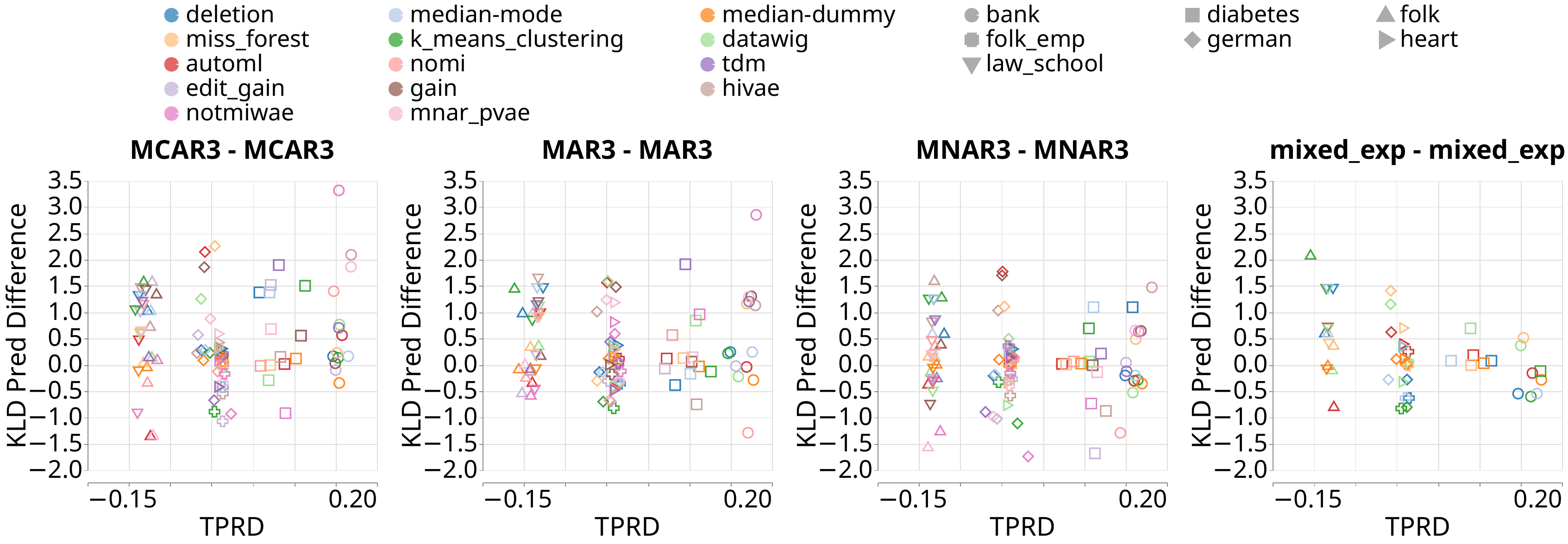}
    \caption{KL divergence difference (imputation) vs TPRD (model)}
    \label{fig:imputation-kld-diff-extended}
\end{subfigure}
\vspace{-0.4cm}
\caption{Imputation fairness vs model fairness: imputation fairness (F1 difference, RMSE difference and KL divergence difference) may not be indicative of model fairness (TPRD).}
\label{fig:imputation-fairness-no-shift-extended}
\end{figure}

\begin{figure*}[h!]
\begin{subfigure}[h]{0.9\linewidth}
    \includegraphics[width=\linewidth]{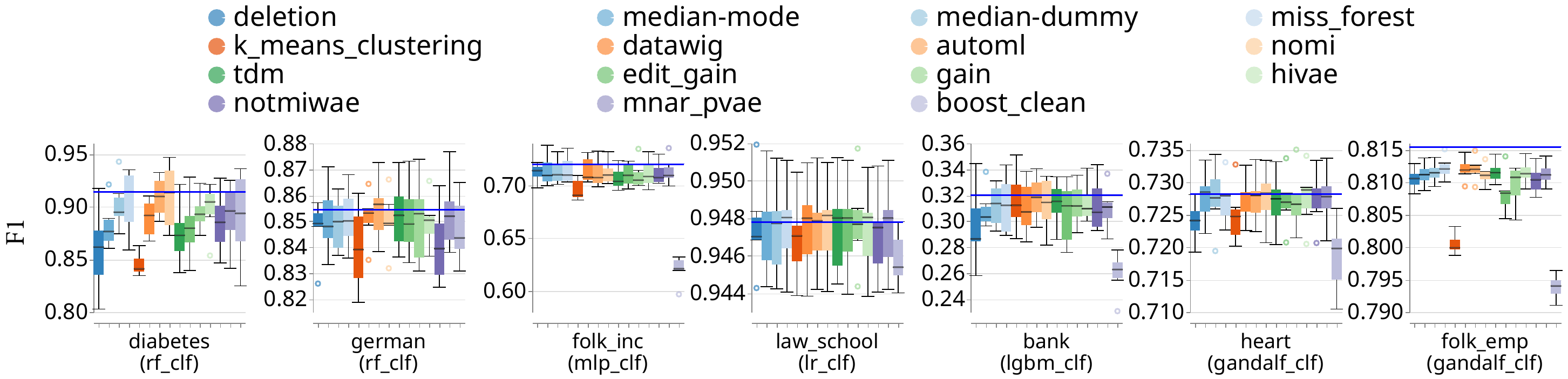}
    \caption{Missing Completely At Random (\mcar)}
\end{subfigure}

\begin{subfigure}[h]{0.9\linewidth}
    \includegraphics[width=\linewidth]{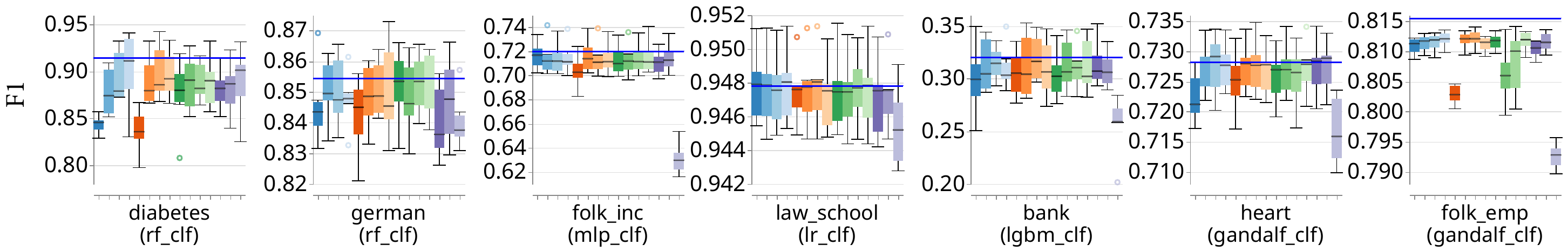}
    \caption{Missing At Random (\mar)}
\end{subfigure}

\begin{subfigure}[h]{0.9\linewidth}
    \includegraphics[width=\linewidth]{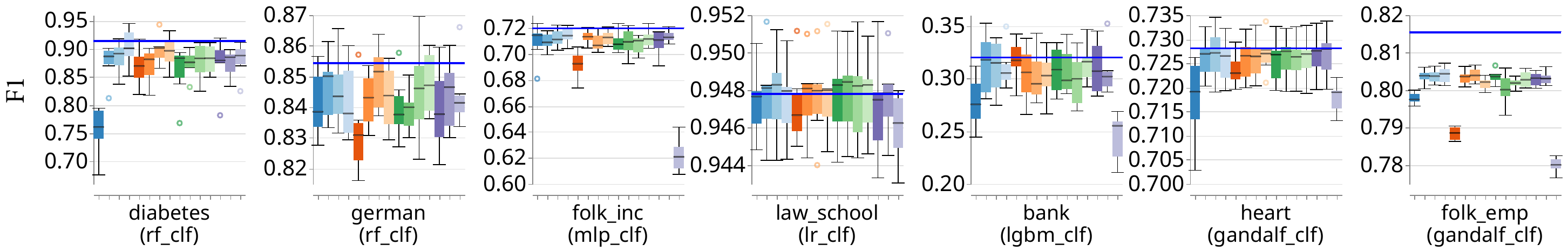}
    \caption{Missing Not At Random (\mnar)}
\end{subfigure}

\begin{subfigure}[h]{0.9\linewidth}
    \centering
    \includegraphics[width=\linewidth]{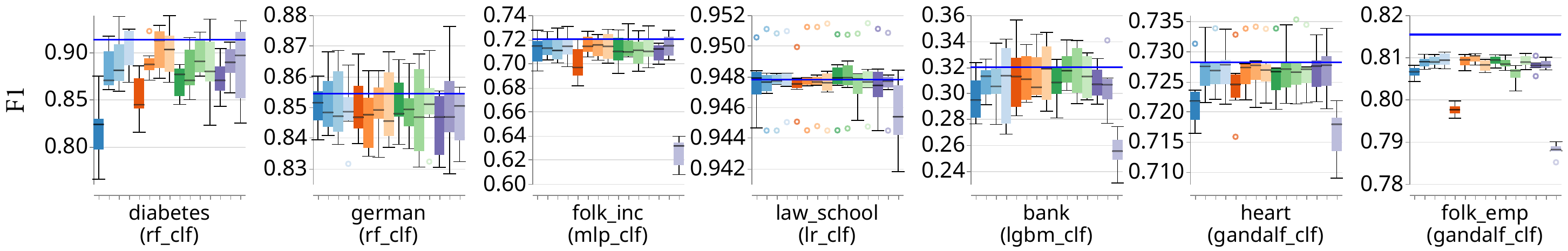}
    \caption{Mixed missingness (\mcar \& \mar \& \mnar)}
\end{subfigure}

\caption{F1 of best performing models (shown in figure) for different imputation strategies (colors in the legend), datasets (x-axis), and missingness mechanisms (subplots). Datasets are in increasing order by size. Blue line shows median performance of the model trained on clean data.}
\label{fig:exp1-F1}
\end{figure*}

\begin{figure*}[h!]
\begin{subfigure}[h]{0.9\linewidth}
    \includegraphics[width=\linewidth]{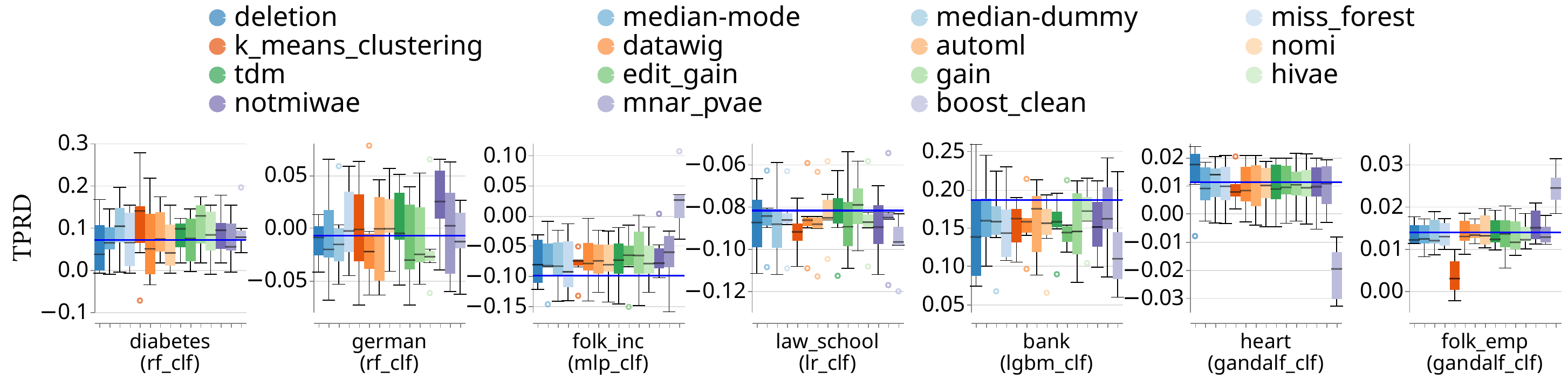}
    \caption{Missing Completely At Random (\mcar)}
\end{subfigure}

\begin{subfigure}[h]{0.9\linewidth}
    \includegraphics[width=\linewidth]{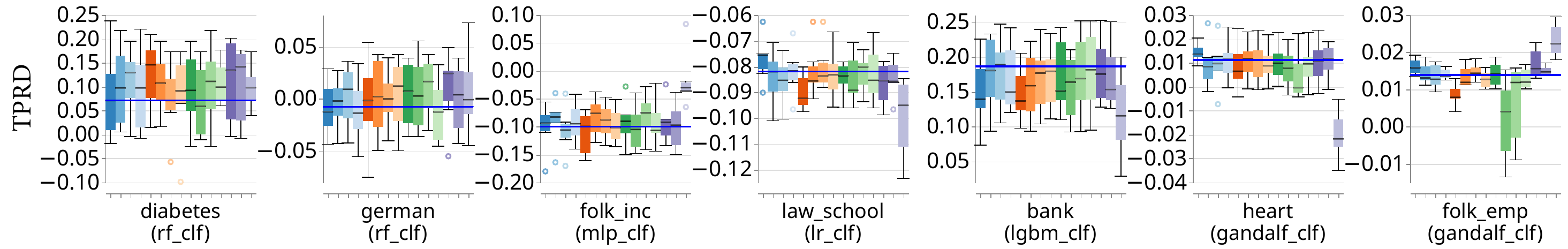}
    \caption{Missing At Random (\mar)}
\end{subfigure}

\begin{subfigure}[h]{0.9\linewidth}
    \includegraphics[width=\linewidth]{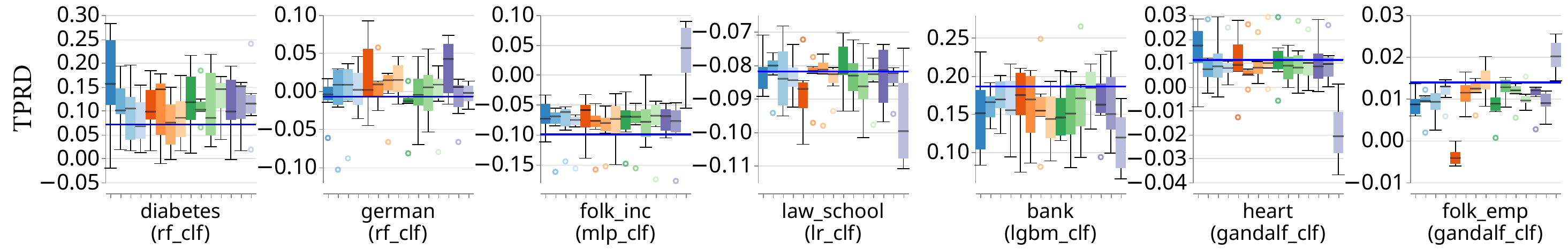}
    \caption{Missing Not At Random (\mnar)}
\end{subfigure}

\begin{subfigure}[h]{0.9\linewidth}
    \includegraphics[width=\linewidth]{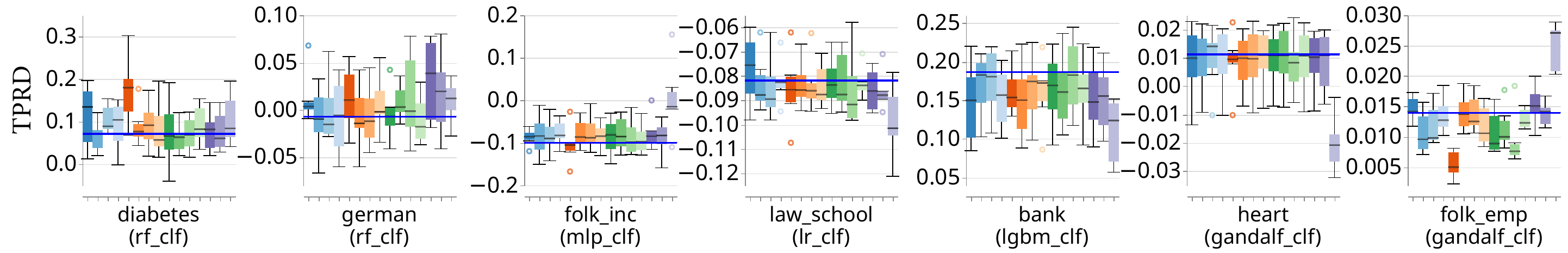}
    \caption{Mixed missingness (\mcar \& \mar \& \mnar)}
\end{subfigure}

\caption{True Positive Rate Difference (unfairness) of best performing models (shown in figure) for different imputation strategies (colors in the legend), datasets ($x$-axis), and missingness mechanisms (subplots). Values close to 0 are desirable. Datasets are in increasing order by size. Blue line shows median TPRD of the model trained on clean data.}
\label{fig:exp1-TPR}
\end{figure*}

\begin{figure*}[h!]
\begin{subfigure}[h]{0.9\linewidth}
    \centering
    \includegraphics[width=\linewidth]{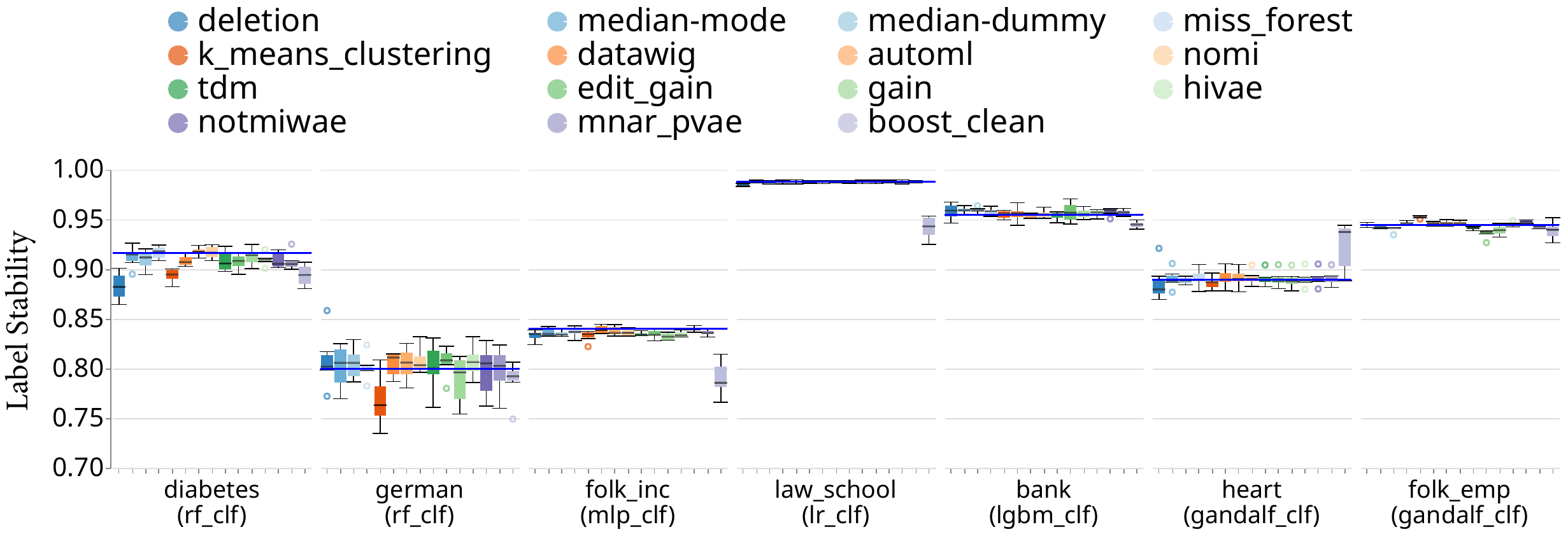}
    \vspace{-0.5cm}
    \caption{Missing Completely At Random (\mcar)}
\end{subfigure}

\begin{subfigure}[h]{0.9\linewidth}
    \centering
    \includegraphics[width=\linewidth]{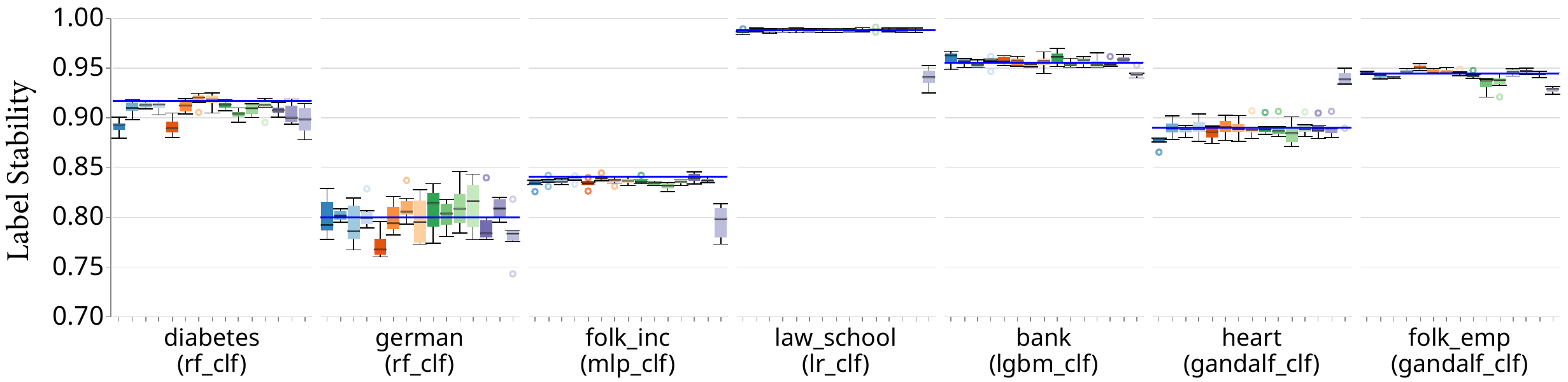}
    \caption{Missing At Random (\mar)}
\end{subfigure}

\begin{subfigure}[h]{0.9\linewidth}
    \centering
    \includegraphics[width=\linewidth]{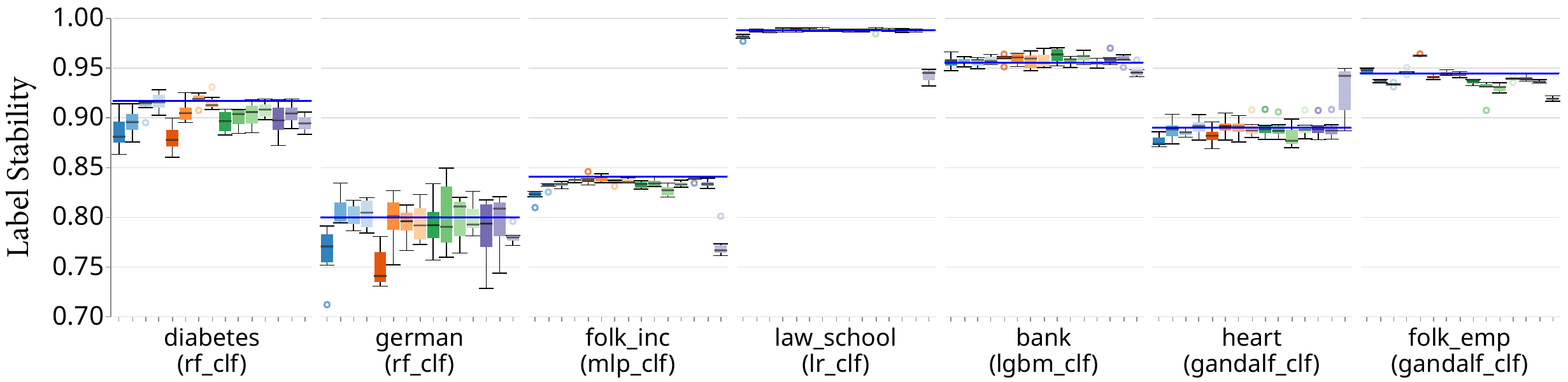}
    \caption{Missing Not At Random (\mnar)}
\end{subfigure}

\begin{subfigure}[h]{0.9\linewidth}
    \centering
    \includegraphics[width=\linewidth]{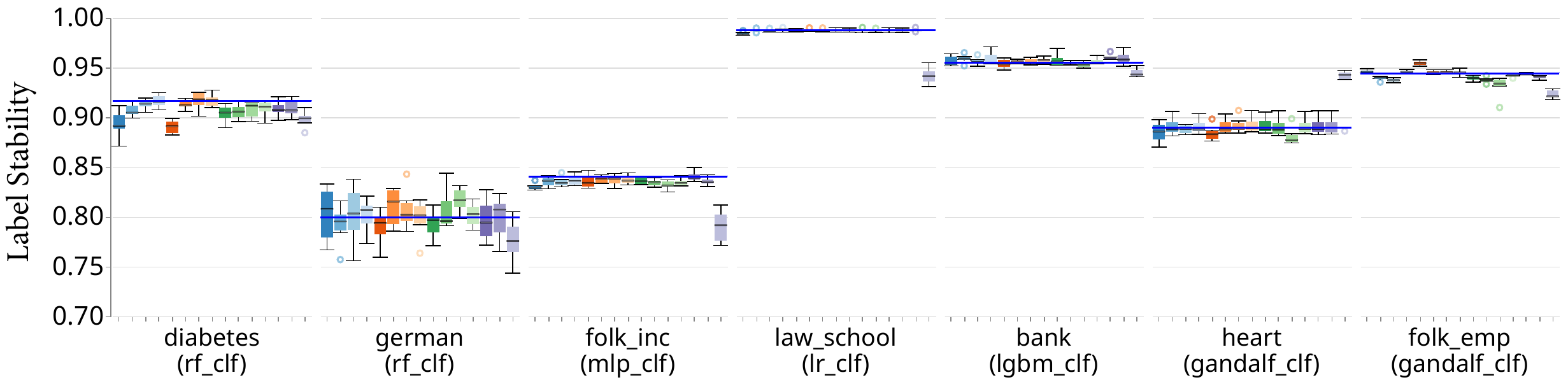}
    \caption{Mixed missingness (\mcar \& \mar \& \mnar)}
\end{subfigure}

\caption{Label Stability of best performing models (shown in figure) for different imputation strategies (colors in the legend), datasets (x-axis), and missingness mechanisms (subplots). Values close to 1 are desirable. Datasets are ordered in increasing order by size. The blue line indicates median performance of the model trained on clean data.}
\label{fig:exp1-label-stability}
\vspace{-0.5cm}
\end{figure*}

\begin{figure*}[h!]
\begin{subfigure}[h]{0.9\linewidth}
    \includegraphics[width=\linewidth]{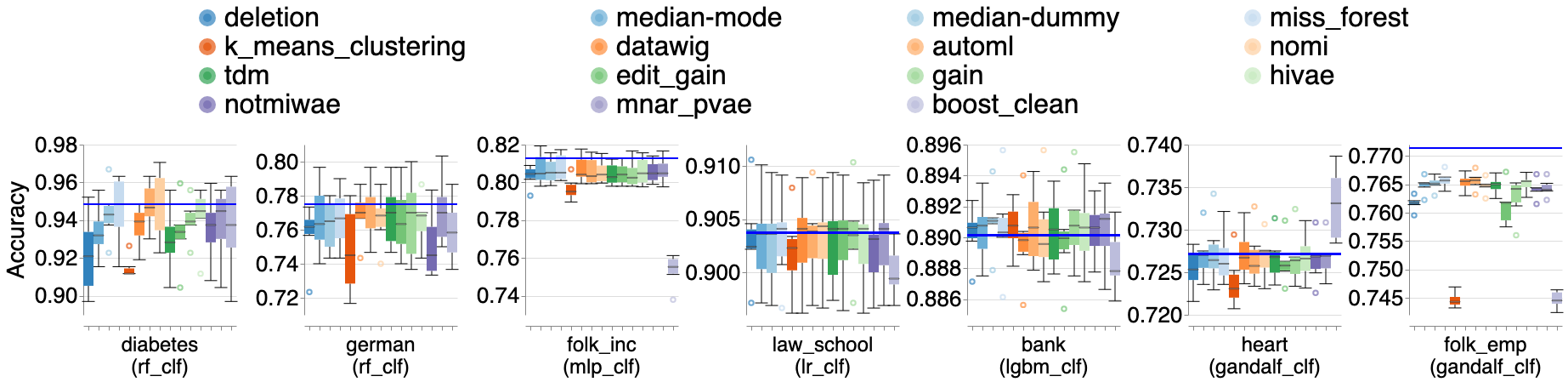}
    \caption{Missing Completely At Random (\mcar)}
\end{subfigure}

\begin{subfigure}[h]{0.9\linewidth}
    \includegraphics[width=\linewidth]{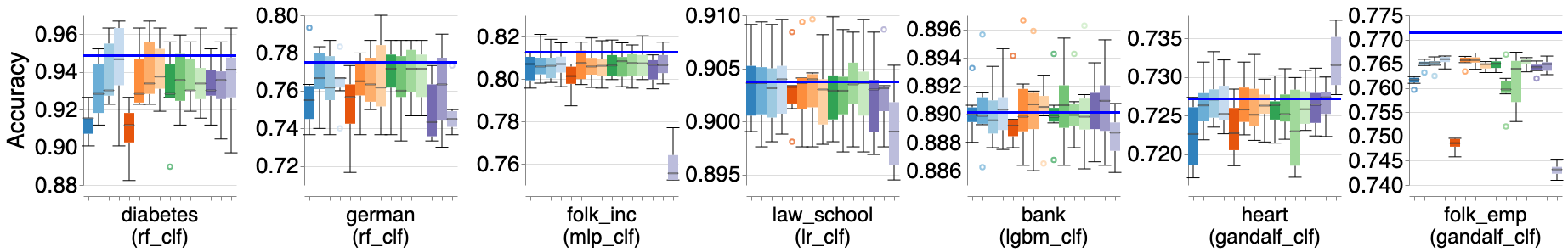}
    \caption{Missing At Random (\mar)}
\end{subfigure}

\begin{subfigure}[h]{0.9\linewidth}
    \includegraphics[width=\linewidth]{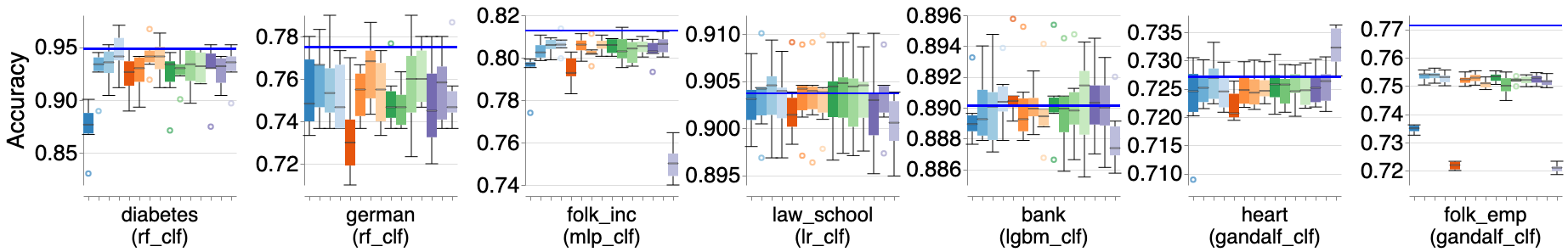}
    \caption{Missing Not At Random (\mnar)}
\end{subfigure}

\begin{subfigure}[h]{0.9\linewidth}
    \includegraphics[width=\linewidth]{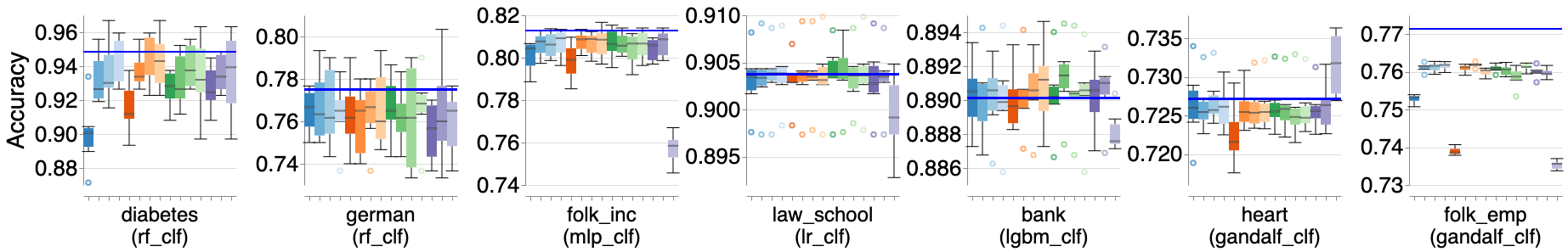}
    \caption{Mixed missingness (\mcar \& \mar \& \mnar)}
\end{subfigure}

\caption{Accuracy of best performing models (shown in figure) for different imputation strategies (colors in the legend), datasets (x-axis), and missingness mechanisms (subplots). Datasets are ordered in increasing order by size. The blue line shows median performance of a model trained on clean data.}
\label{fig:exp1-accuracy}
\end{figure*}

\begin{figure*}[h!]
\begin{subfigure}[h]{0.9\linewidth}
    \includegraphics[width=\linewidth]{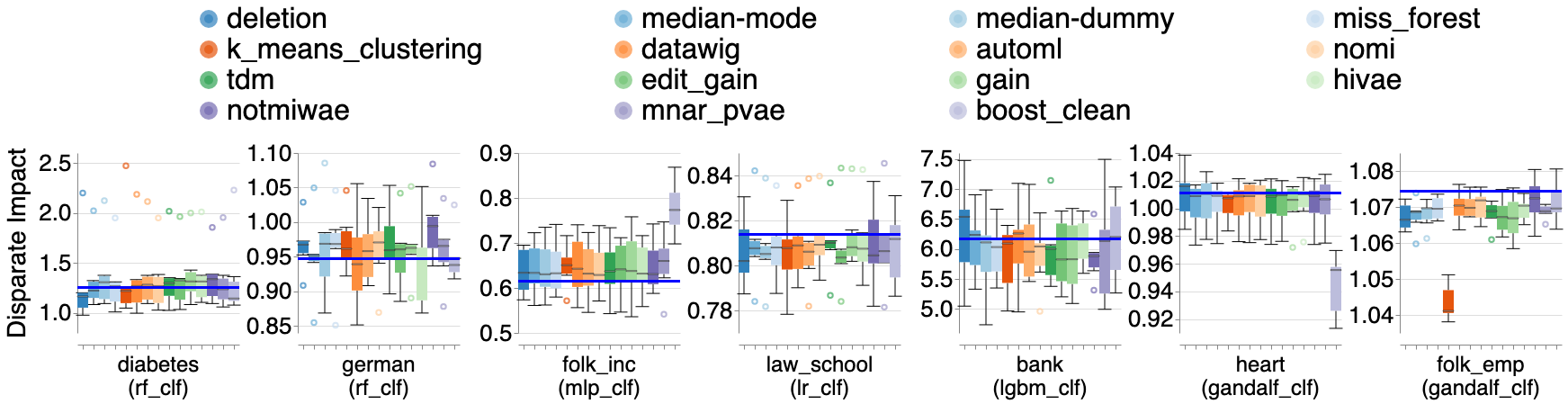}
    \caption{Missing Completely At Random (MCAR)}
\end{subfigure}

\begin{subfigure}[h]{0.9\linewidth}
    \includegraphics[width=\linewidth]{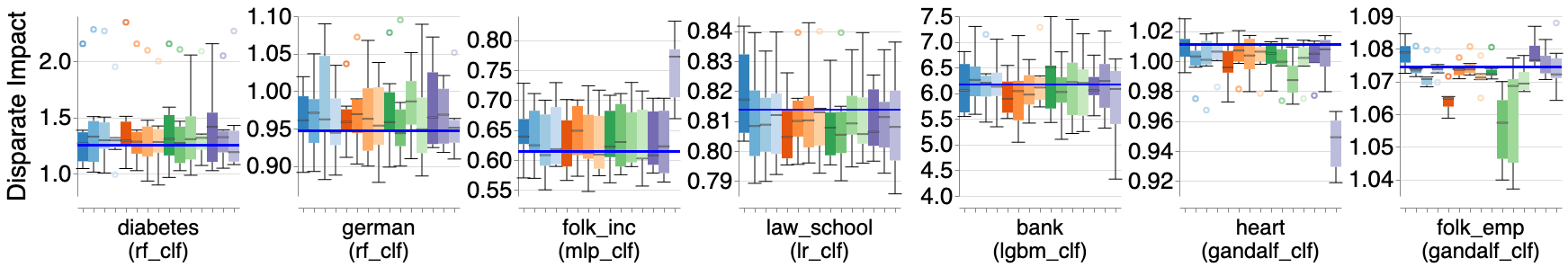}
    \caption{Missing At Random (MAR)}
\end{subfigure}

\begin{subfigure}[h]{0.9\linewidth}
    \includegraphics[width=\linewidth]{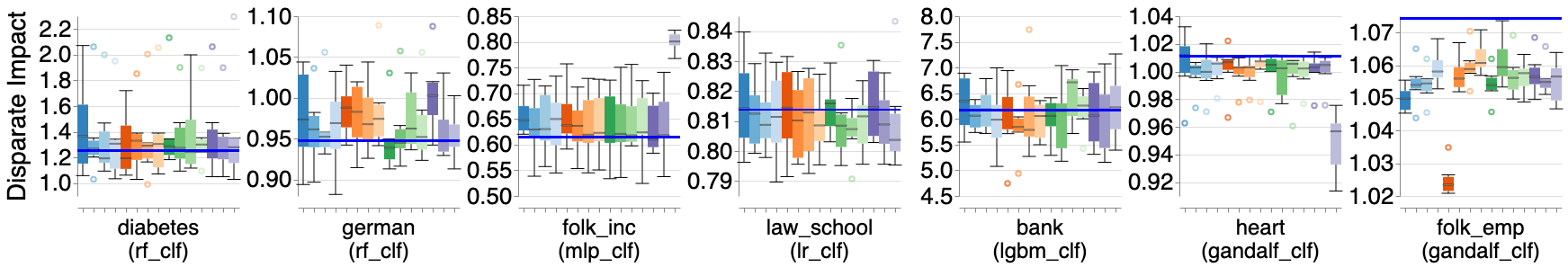}
    \caption{Missing Not At Random (MNAR)}
\end{subfigure}

\begin{subfigure}[h]{0.9\linewidth}
    \includegraphics[width=\linewidth]{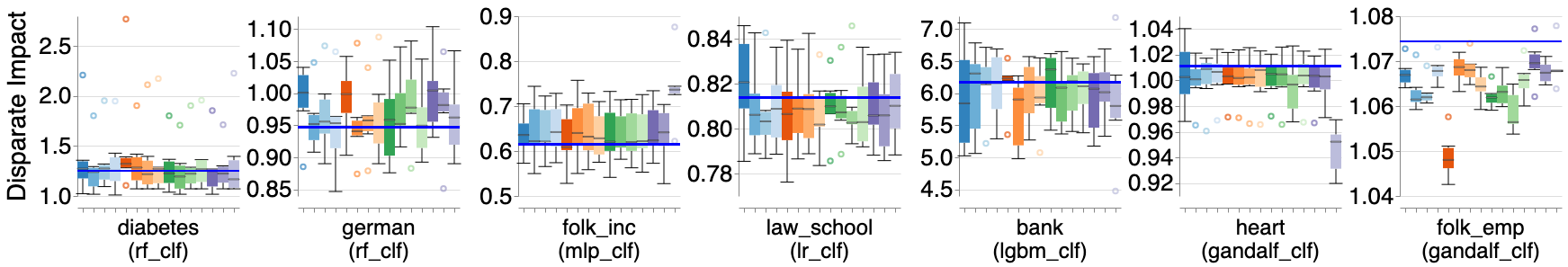}
    \caption{Mixed missingness (\mcar \& \mar \& \mnar)}
\end{subfigure}

\caption{Disparate Impact (unfairness) of best performing models for different imputation strategies (colors in the legend), datasets (x-axis), and missingness mechanisms (subplots). Values close to 1 are ideal/fair. Datasets are ordered in increasing order by size. The blue line indicates the median performance of a model trained on clean data.}
\label{fig:exp1-DI}
\end{figure*}

\begin{figure*}[h!]
\begin{subfigure}[h]{0.9\linewidth}
    \includegraphics[width=\linewidth]{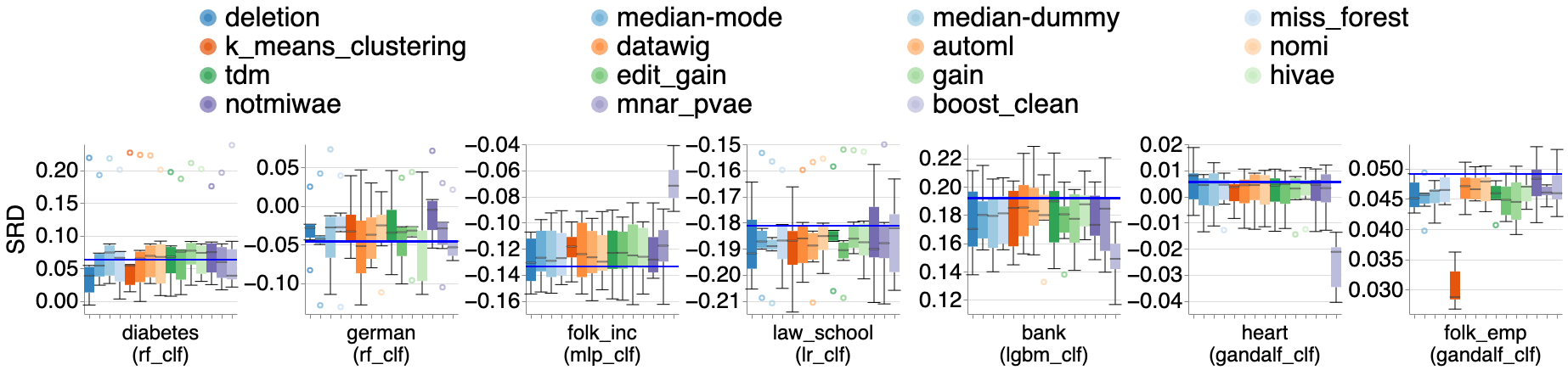}
    \caption{Missing Completely At Random (MCAR)}
\end{subfigure}

\begin{subfigure}[h]{0.9\linewidth}
    \includegraphics[width=\linewidth]{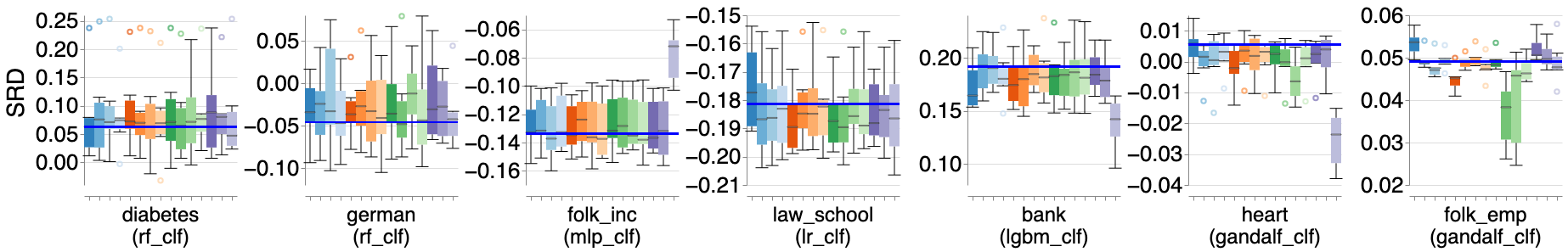}
    \caption{Missing At Random (MAR)}
\end{subfigure}

\begin{subfigure}[h]{0.9\linewidth}
    \includegraphics[width=\linewidth]{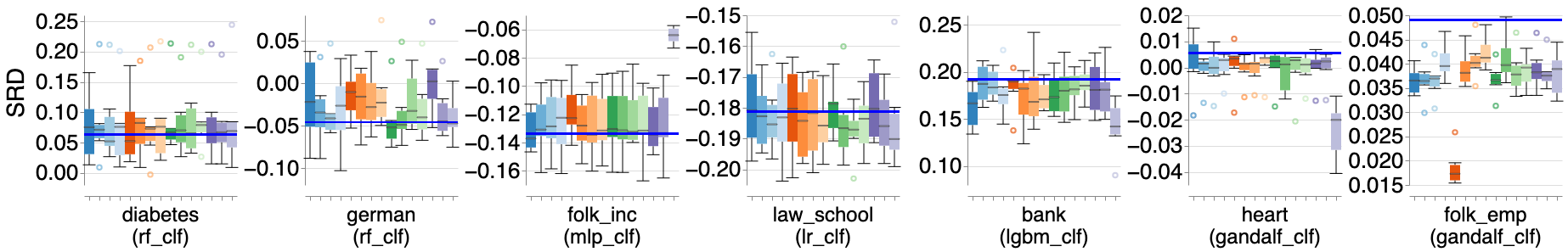}
    \caption{Missing Not At Random (MNAR)}
\end{subfigure}

\begin{subfigure}[h]{0.9\linewidth}
    \includegraphics[width=\linewidth]{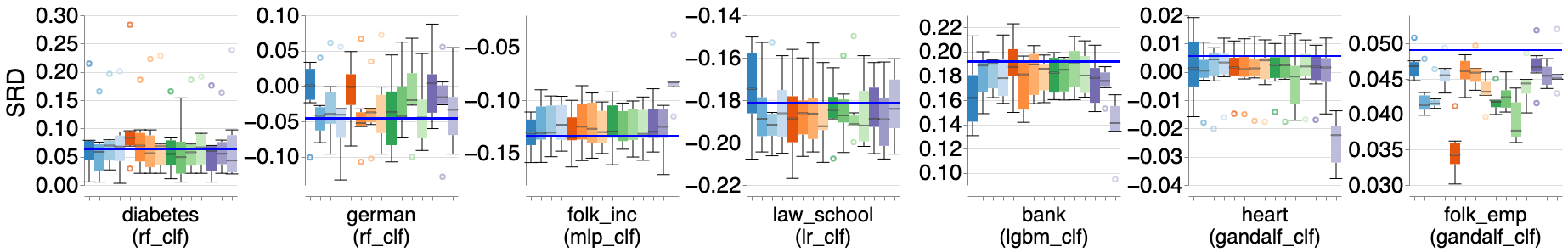}
    \caption{Mixed missingness (\mcar \& \mar \& \mnar)}
\end{subfigure}

\caption{Selection Rate Difference (unfairness) of best performing models for different imputation strategies (colors in the legend), datasets (x-axis), and missingness mechanisms (subplots). Values close to 0 are ideal/fair. Datasets are ordered in increasing order by size. The blue line indicates the median performance of a model trained on clean data.}
\label{fig:exp1-SPD}
\end{figure*}

\begin{figure*}[h!]
\begin{subfigure}[h]{0.9\linewidth}
    \includegraphics[width=\linewidth]{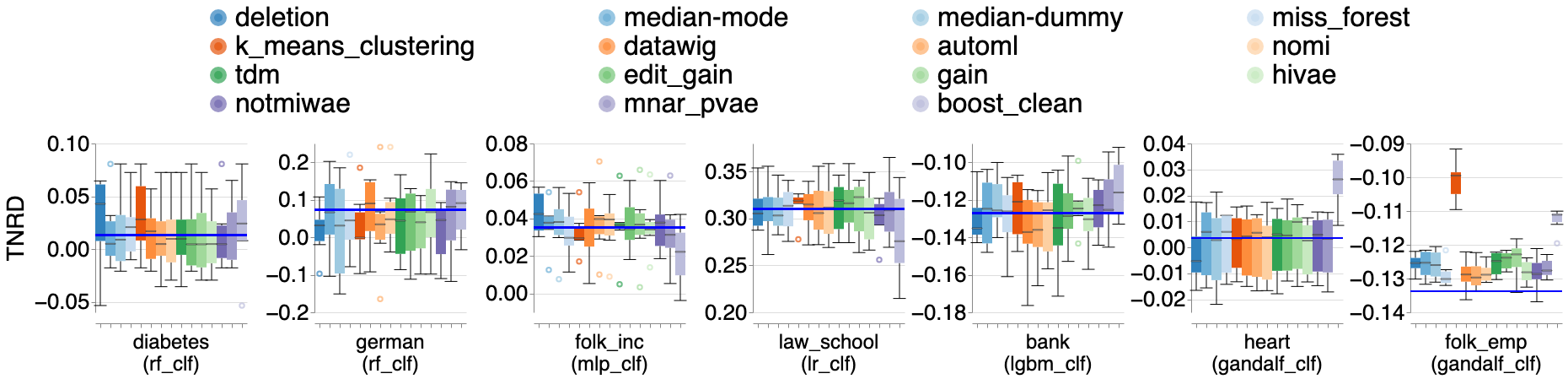}
    \caption{Missing Completely At Random (MCAR)}
\end{subfigure}

\begin{subfigure}[h]{0.9\linewidth}
    \includegraphics[width=\linewidth]{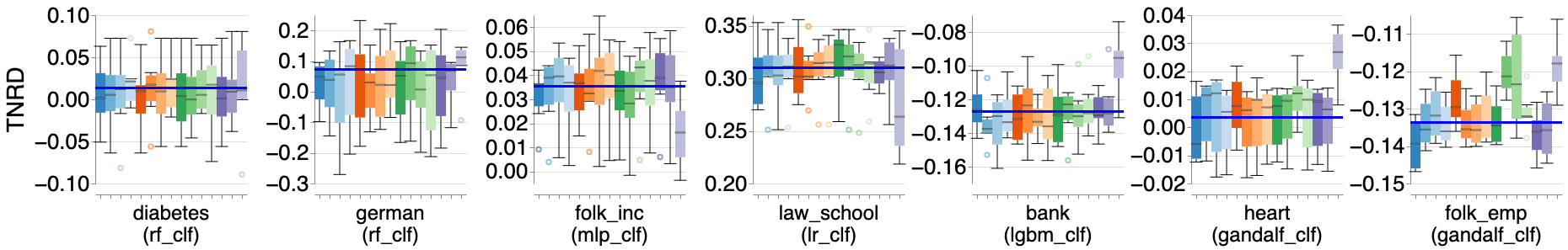}
    \caption{Missing At Random (MAR)}
\end{subfigure}

\begin{subfigure}[h]{0.9\linewidth}
    \includegraphics[width=\linewidth]{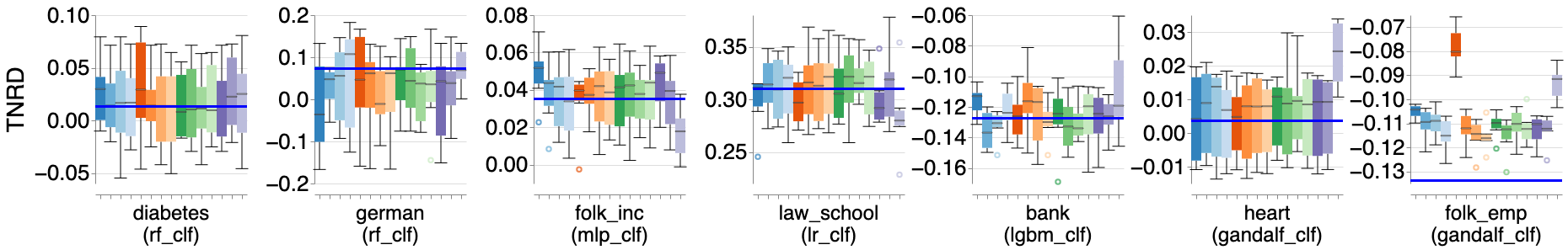}
    \caption{Missing Not At Random (MNAR)}
\end{subfigure}

\begin{subfigure}[h]{0.9\linewidth}
    \includegraphics[width=\linewidth]{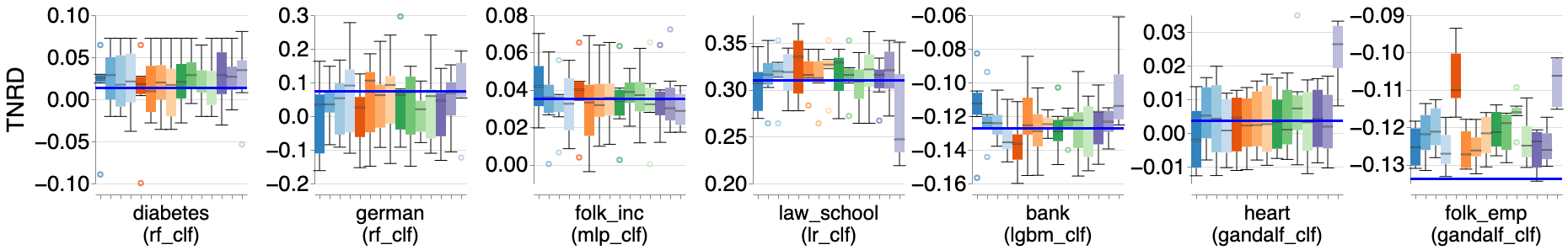}
    \caption{Mixed missingness (\mcar \& \mar \& \mnar)}
\end{subfigure}

\caption{True Negative Rate Difference (unfairness) of best performing models for different imputation strategies (colors in the legend), datasets (x-axis), and missingness mechanisms (subplots).}
\label{fig:exp1-TNR}
\end{figure*}

In this section, we extend the results discussed in Section~\ref{sec:exp1-single} by presenting plots for additional metrics of downstream model performance. Figures \ref{fig:exp1-F1}, \ref{fig:exp1-TPR}, \ref{fig:exp1-label-stability}, \ref{fig:exp1-accuracy}, \ref{fig:exp1-DI}, \ref{fig:exp1-SPD}, and \ref{fig:exp1-TNR} illustrate F1, True Positive Rate Difference, Label Stability, Accuracy, Disparate Impact, Selection Rate Difference, and True Negative Rate Difference of best performing models, respectively, for different imputation techniques, datasets, and missingness mechanisms. Figures~\ref{fig:imputation-quality-no-shift-extended} and \ref{fig:imputation-fairness-no-shift-extended} present scatter plots for all imputers that extend scatter plots from Section~\ref{sec:imputation-quality}. Fairness metrics are calculated for an intersectional group when multiple sensitive attributes are present in the dataset, and for a single group when there is only one sensitive attribute (see Table~\ref{tab:dataset-info} for reference). 

Overall, the plots support the assertion made in Section~\ref{sec:exp1-single-model-fairness} that the impact of \mvi on fairness (error-disparity) is most strongly influenced by the baseline disparity of the model trained on clean (complete, no missingness) data. This implies that it is hard to determine in advance whether data cleaning will worsen fairness.

%% file: tables/imputation_runtime_mcar.tex
\begin{tabular}{llllllll}
\toprule
 \textbf{Imputer}         & \makecell[tl]{\textbf{diabetes}\\\textbf{(633, 17)}}   & \makecell[tl]{\textbf{german}\\\textbf{(700, 21)}}   & \makecell[tl]{\textbf{folk$\_$inc}\\\textbf{(12000, 10)}}   & \makecell[tl]{\textbf{law$\_$school}\\\textbf{(16638, 11)}}   & \makecell[tl]{\textbf{bank}\\\textbf{(32003, 13)}}   & \makecell[tl]{\textbf{heart}\\\textbf{(56000, 11)}}   & \makecell[tl]{\textbf{folk$\_$emp}\\\textbf{(242112, 16)}}   \\
\midrule
 median-dummy             & $0.015 \pm 0.001$                                      & $0.016 \pm 0.001$                                    & $0.021 \pm 0.000$                                           & $0.025 \pm 0.001$                                             & $0.029 \pm 0.000$                                    & $0.057 \pm 0.002$                                     & $0.846 \pm 0.042$                                            \\
 median-mode              & $0.013 \pm 0.001$                                      & $0.016 \pm 0.001$                                    & $0.023 \pm 0.000$                                           & $0.029 \pm 0.001$                                             & $0.034 \pm 0.000$                                    & $0.072 \pm 0.002$                                     & $1 \pm 0.037$                                                \\
 deletion                 & $0.014 \pm 0.001$                                      & $0.015 \pm 0.000$                                    & $0.027 \pm 0.001$                                           & $0.028 \pm 0.001$                                             & $0.049 \pm 0.001$                                    & $0.100 \pm 0.006$                                     & $1 \pm 0.036$                                                \\
 nomi                     & $11 \pm 4$                                             & $16 \pm 9$                                           & $25 \pm 2$                                                  & $23 \pm 2$                                                    & $31 \pm 1$                                           & $36 \pm 2$                                            & $206 \pm 5$                                                  \\
 edit$\_$gain             & $2 \pm 0.133$                                          & $2 \pm 0.155$                                        & $14 \pm 0.188$                                              & $22 \pm 2$                                                    & $31 \pm 0.792$                                       & $62 \pm 7$                                            & $216 \pm 4$                                                  \\
 mnar$\_$pvae             & $8 \pm 1$                                              & $10 \pm 1$                                           & $13 \pm 1$                                                  & $44 \pm 43$                                                   & $129 \pm 121$                                        & $43 \pm 0.986$                                        & $228 \pm 6$                                                  \\
 tdm                      & $897 \pm 11$                                           & $1034 \pm 6$                                         & $1289 \pm 9$                                                & $1095 \pm 19$                                                 & $1720 \pm 68$                                        & $1195 \pm 22$                                         & $1448 \pm 25$                                                \\
 notmiwae                 & $157 \pm 84$                                           & $249 \pm 149$                                        & $600 \pm 8$                                                 & $751 \pm 3$                                                   & $749 \pm 349$                                        & $1472 \pm 637$                                        & $2978 \pm 578$                                               \\
 gain                     & $257 \pm 26$                                           & $261 \pm 5$                                          & $1112 \pm 72$                                               & $1421 \pm 39$                                                 & $1902 \pm 33$                                        & $2748 \pm 48$                                         & $5668 \pm 146$                                               \\
 hivae                    & $70 \pm 1$                                             & $90 \pm 0.228$                                       & $701 \pm 8$                                                 & $1017 \pm 6$                                                  & $2531 \pm 320$                                       & $3874 \pm 86$                                         & $7161 \pm 415$                                               \\
 k$\_$means$\_$clustering & $0.921 \pm 0.318$                                      & $1 \pm 0.326$                                        & $1 \pm 0.117$                                               & $18 \pm 1$                                                    & $26 \pm 1$                                           & $114 \pm 24$                                          & $7499 \pm 720$                                               \\
 miss$\_$forest           & $94 \pm 14$                                            & $240 \pm 92$                                         & $1837 \pm 577$                                              & $2150 \pm 989$                                                & $4488 \pm 1404$                                      & $6269 \pm 1610$                                       & $15154 \pm 5008$                                             \\
 datawig                  & $238 \pm 69$                                           & $279 \pm 66$                                         & $601 \pm 70$                                                & $2104 \pm 336$                                                & $4654 \pm 755$                                       & $7791 \pm 921$                                        & $33102 \pm 1695$                                             \\
 automl                   & $2219 \pm 367$                                         & $1969 \pm 131$                                       & $5751 \pm 604$                                              & $6936 \pm 252$                                                & $14203 \pm 1105$                                     & $19315 \pm 3505$                                      & $104290 \pm 19094$                                           \\
\bottomrule
\end{tabular}

%% file: tables/imputation_runtime_mar.tex
\begin{tabular}{llllllll}
\toprule
 \textbf{Imputer}         & \makecell[tl]{\textbf{diabetes}\\\textbf{(633, 17)}}   & \makecell[tl]{\textbf{german}\\\textbf{(700, 21)}}   & \makecell[tl]{\textbf{folk$\_$inc}\\\textbf{(12000, 10)}}   & \makecell[tl]{\textbf{law$\_$school}\\\textbf{(16638, 11)}}   & \makecell[tl]{\textbf{bank}\\\textbf{(32003, 13)}}   & \makecell[tl]{\textbf{heart}\\\textbf{(56000, 11)}}   & \makecell[tl]{\textbf{folk$\_$emp}\\\textbf{(242112, 16)}}   \\
\midrule
 median-dummy             & $0.015 \pm 0.001$                                      & $0.016 \pm 0.000$                                    & $0.021 \pm 0.000$                                           & $0.025 \pm 0.001$                                             & $0.029 \pm 0.000$                                    & $0.057 \pm 0.001$                                     & $0.883 \pm 0.044$                                            \\
 median-mode              & $0.013 \pm 0.001$                                      & $0.016 \pm 0.001$                                    & $0.023 \pm 0.000$                                           & $0.027 \pm 0.000$                                             & $0.033 \pm 0.000$                                    & $0.070 \pm 0.003$                                     & $1 \pm 0.018$                                                \\
 deletion                 & $0.014 \pm 0.001$                                      & $0.015 \pm 0.001$                                    & $0.026 \pm 0.000$                                           & $0.028 \pm 0.000$                                             & $0.048 \pm 0.000$                                    & $0.095 \pm 0.004$                                     & $1 \pm 0.031$                                                \\
 nomi                     & $10 \pm 3$                                             & $14 \pm 8$                                           & $23 \pm 1$                                                  & $24 \pm 3$                                                    & $29 \pm 1$                                           & $38 \pm 3$                                            & $191 \pm 3$                                                  \\
 edit$\_$gain             & $2 \pm 0.088$                                          & $2 \pm 0.120$                                        & $13 \pm 0.193$                                              & $20 \pm 2$                                                    & $30 \pm 1$                                           & $63 \pm 7$                                            & $209 \pm 8$                                                  \\
 mnar$\_$pvae             & $8 \pm 0.480$                                          & $10 \pm 1$                                           & $11 \pm 0.646$                                              & $16 \pm 0.633$                                                & $35 \pm 1$                                           & $41 \pm 1$                                            & $213 \pm 10$                                                 \\
 tdm                      & $907 \pm 19$                                           & $984 \pm 11$                                         & $1303 \pm 8$                                                & $1182 \pm 43$                                                 & $1335 \pm 12$                                        & $1556 \pm 57$                                         & $1477 \pm 8$                                                 \\
 notmiwae                 & $235 \pm 140$                                          & $273 \pm 0.540$                                      & $586 \pm 1$                                                 & $797 \pm 10$                                                  & $803 \pm 395$                                        & $1498 \pm 553$                                        & $2894 \pm 487$                                               \\
 gain                     & $299 \pm 11$                                           & $288 \pm 4$                                          & $1140 \pm 25$                                               & $1590 \pm 34$                                                 & $1971 \pm 34$                                        & $2880 \pm 44$                                         & $6659 \pm 171$                                               \\
 k$\_$means$\_$clustering & $0.886 \pm 0.341$                                      & $1 \pm 0.319$                                        & $1 \pm 0.122$                                               & $19 \pm 2$                                                    & $26 \pm 1$                                           & $107 \pm 22$                                          & $7395 \pm 715$                                               \\
 hivae                    & $71 \pm 0.788$                                         & $99 \pm 1$                                           & $734 \pm 13$                                                & $1134 \pm 11$                                                 & $2337 \pm 28$                                        & $3585 \pm 84$                                         & $7557 \pm 468$                                               \\
 miss$\_$forest           & $112 \pm 14$                                           & $251 \pm 87$                                         & $1874 \pm 606$                                              & $2976 \pm 914$                                                & $4493 \pm 1218$                                      & $7443 \pm 2031$                                       & $19323 \pm 5650$                                             \\
 datawig                  & $271 \pm 83$                                           & $325 \pm 85$                                         & $642 \pm 18$                                                & $2398 \pm 735$                                                & $4858 \pm 695$                                       & $7820 \pm 908$                                        & $29815 \pm 5383$                                             \\
 automl                   & $2172 \pm 187$                                         & $1822 \pm 236$                                       & $5495 \pm 797$                                              & $7036 \pm 507$                                                & $15469 \pm 2313$                                     & $19357 \pm 3697$                                      & $96490 \pm 8667$                                             \\
\bottomrule
\end{tabular}

%% file: tables/imputation_runtime_mnar.tex
\begin{tabular}{llllllll}
\toprule
 \textbf{Imputer}         & \makecell[tl]{\textbf{diabetes}\\\textbf{(633, 17)}}   & \makecell[tl]{\textbf{german}\\\textbf{(700, 21)}}   & \makecell[tl]{\textbf{folk$\_$inc}\\\textbf{(12000, 10)}}   & \makecell[tl]{\textbf{law$\_$school}\\\textbf{(16638, 11)}}   & \makecell[tl]{\textbf{bank}\\\textbf{(32003, 13)}}   & \makecell[tl]{\textbf{heart}\\\textbf{(56000, 11)}}   & \makecell[tl]{\textbf{folk$\_$emp}\\\textbf{(242112, 16)}}   \\
\midrule
 median-dummy             & $0.015 \pm 0.001$                                      & $0.015 \pm 0.000$                                    & $0.021 \pm 0.001$                                           & $0.025 \pm 0.001$                                             & $0.029 \pm 0.001$                                    & $0.057 \pm 0.002$                                     & $0.773 \pm 0.005$                                            \\
 median-mode              & $0.015 \pm 0.000$                                      & $0.016 \pm 0.001$                                    & $0.023 \pm 0.000$                                           & $0.027 \pm 0.000$                                             & $0.034 \pm 0.001$                                    & $0.067 \pm 0.003$                                     & $0.867 \pm 0.006$                                            \\
 deletion                 & $0.014 \pm 0.000$                                      & $0.013 \pm 0.001$                                    & $0.025 \pm 0.000$                                           & $0.027 \pm 0.000$                                             & $0.046 \pm 0.000$                                    & $0.086 \pm 0.004$                                     & $1 \pm 0.118$                                                \\
 mnar$\_$pvae             & $8 \pm 0.560$                                          & $26 \pm 44$                                          & $14 \pm 0.825$                                              & $16 \pm 0.755$                                                & $34 \pm 0.941$                                       & $51 \pm 23$                                           & $215 \pm 8$                                                  \\
 edit$\_$gain             & $2 \pm 0.129$                                          & $2 \pm 0.144$                                        & $13 \pm 0.250$                                              & $21 \pm 2$                                                    & $30 \pm 0.386$                                       & $62 \pm 7$                                            & $219 \pm 2$                                                  \\
 nomi                     & $12 \pm 5$                                             & $14 \pm 5$                                           & $25 \pm 2$                                                  & $24 \pm 3$                                                    & $38 \pm 1$                                           & $51 \pm 1$                                            & $722 \pm 32$                                                 \\
 tdm                      & $877 \pm 4$                                            & $1050 \pm 12$                                        & $1361 \pm 51$                                               & $1295 \pm 287$                                                & $1312 \pm 17$                                        & $1276 \pm 73$                                         & $1409 \pm 66$                                                \\
 notmiwae                 & $132 \pm 100$                                          & $226 \pm 96$                                         & $527 \pm 1$                                                 & $1049 \pm 20$                                                 & $608 \pm 267$                                        & $1518 \pm 543$                                        & $3414 \pm 814$                                               \\
 gain                     & $248 \pm 4$                                            & $349 \pm 4$                                          & $1084 \pm 29$                                               & $1531 \pm 29$                                                 & $1875 \pm 41$                                        & $2980 \pm 56$                                         & $5869 \pm 153$                                               \\
 hivae                    & $64 \pm 0.228$                                         & $100 \pm 1$                                          & $761 \pm 7$                                                 & $1023 \pm 11$                                                 & $2529 \pm 61$                                        & $3887 \pm 82$                                         & $7374 \pm 267$                                               \\
 k$\_$means$\_$clustering & $0.887 \pm 0.328$                                      & $1 \pm 0.323$                                        & $1 \pm 0.120$                                               & $18 \pm 2$                                                    & $26 \pm 1$                                           & $106 \pm 23$                                          & $7382 \pm 805$                                               \\
 miss$\_$forest           & $123 \pm 28$                                           & $275 \pm 90$                                         & $1442 \pm 365$                                              & $2558 \pm 818$                                                & $4281 \pm 1601$                                      & $4384 \pm 972$                                        & $19227 \pm 4033$                                             \\
 datawig                  & $247 \pm 46$                                           & $232 \pm 30$                                         & $581 \pm 48$                                                & $2142 \pm 380$                                                & $4952 \pm 674$                                       & $7048 \pm 1119$                                       & $30238 \pm 3335$                                             \\
 automl                   & $1704 \pm 78$                                          & $1677 \pm 309$                                       & $5147 \pm 463$                                              & $6707 \pm 992$                                                & $13011 \pm 2231$                                     & $19299 \pm 2315$                                      & $106635 \pm 17043$                                           \\
\bottomrule
\end{tabular}

%% file: tables/imputation_runtime_mixed.tex
\begin{tabular}{llllllll}
\toprule
 \textbf{Imputer}         & \makecell[tl]{\textbf{diabetes}\\\textbf{(633, 17)}}   & \makecell[tl]{\textbf{german}\\\textbf{(700, 21)}}   & \makecell[tl]{\textbf{folk$\_$inc}\\\textbf{(12000, 10)}}   & \makecell[tl]{\textbf{law$\_$school}\\\textbf{(16638, 11)}}   & \makecell[tl]{\textbf{bank}\\\textbf{(32003, 13)}}   & \makecell[tl]{\textbf{heart}\\\textbf{(56000, 11)}}   & \makecell[tl]{\textbf{folk$\_$emp}\\\textbf{(242112, 16)}}   \\
\midrule
 median-dummy             & $0.009 \pm 0.000$                                      & $0.010 \pm 0.001$                                    & $0.019 \pm 0.002$                                           & $0.021 \pm 0.000$                                             & $0.021 \pm 0.001$                                    & $0.040 \pm 0.001$                                     & $0.590 \pm 0.007$                                            \\
 median-mode              & $0.009 \pm 0.000$                                      & $0.010 \pm 0.000$                                    & $0.023 \pm 0.002$                                           & $0.019 \pm 0.000$                                             & $0.024 \pm 0.000$                                    & $0.059 \pm 0.003$                                     & $0.748 \pm 0.015$                                            \\
 deletion                 & $0.008 \pm 0.000$                                      & $0.009 \pm 0.001$                                    & $0.017 \pm 0.001$                                           & $0.018 \pm 0.000$                                             & $0.039 \pm 0.001$                                    & $0.067 \pm 0.002$                                     & $0.927 \pm 0.013$                                            \\
 mnar$\_$pvae             & $8 \pm 0.717$                                          & $9 \pm 0.997$                                        & $18 \pm 4$                                                  & $12 \pm 0.505$                                                & $22 \pm 0.776$                                       & $33 \pm 0.939$                                        & $169 \pm 3$                                                  \\
 edit$\_$gain             & $2 \pm 0.126$                                          & $2 \pm 0.145$                                        & $14 \pm 0.252$                                              & $21 \pm 2$                                                    & $29 \pm 1$                                           & $62 \pm 9$                                            & $215 \pm 2$                                                  \\
 nomi                     & $10 \pm 2$                                             & $13 \pm 5$                                           & $17 \pm 1$                                                  & $16 \pm 0.530$                                                & $19 \pm 0.962$                                       & $26 \pm 1$                                            & $306 \pm 38$                                                 \\
 tdm                      & $1048 \pm 260$                                         & $1024 \pm 16$                                        & $1236 \pm 21$                                               & $1114 \pm 20$                                                 & $1281 \pm 40$                                        & $1212 \pm 16$                                         & $1463 \pm 24$                                                \\
 notmiwae                 & $122 \pm 81$                                           & $123 \pm 83$                                         & $507 \pm 0.743$                                             & $618 \pm 1$                                                   & $502 \pm 128$                                        & $1082 \pm 406$                                        & $2492 \pm 1023$                                              \\
 gain                     & $241 \pm 7$                                            & $294 \pm 7$                                          & $1126 \pm 26$                                               & $1392 \pm 19$                                                 & $2106 \pm 38$                                        & $2960 \pm 43$                                         & $6396 \pm 241$                                               \\
 hivae                    & $68 \pm 0.624$                                         & $93 \pm 1$                                           & $785 \pm 31$                                                & $1119 \pm 16$                                                 & $2405 \pm 111$                                       & $3828 \pm 264$                                        & $6559 \pm 119$                                               \\
 k$\_$means$\_$clustering & $39 \pm 0.618$                                         & $50 \pm 0.800$                                       & $104 \pm 2$                                                 & $1239 \pm 23$                                                 & $3914 \pm 194$                                       & $3820 \pm 185$                                        & $7432 \pm 873$                                               \\
 miss$\_$forest           & $114 \pm 13$                                           & $207 \pm 76$                                         & $1878 \pm 553$                                              & $2434 \pm 685$                                                & $3966 \pm 1017$                                      & $7253 \pm 1789$                                       & $27730 \pm 5046$                                             \\
 datawig                  & $1627 \pm 541$                                         & $272 \pm 55$                                         & $591 \pm 46$                                                & $2802 \pm 518$                                                & $5892 \pm 480$                                       & $7708 \pm 468$                                        & $31086 \pm 4558$                                             \\
 automl                   & $1715 \pm 149$                                         & $1750 \pm 172$                                       & $5843 \pm 461$                                              & $6531 \pm 509$                                                & $12888 \pm 1099$                                     & $18248 \pm 1324$                                      & $110488 \pm 14166$                                           \\
\bottomrule
\end{tabular}

%% file: appendix/shift-additional.tex
\section{Additional Experimental Results for Missingness Shift}
\label{apdx:shift-additional}

\subsection{Fixed Train and Test Missingness Fractions}
\label{apdx:fixed-shift-additional}

\begin{figure*}[h!]
\begin{subfigure}[h]{0.49\linewidth}
    \centering
    \includegraphics[width=\linewidth]{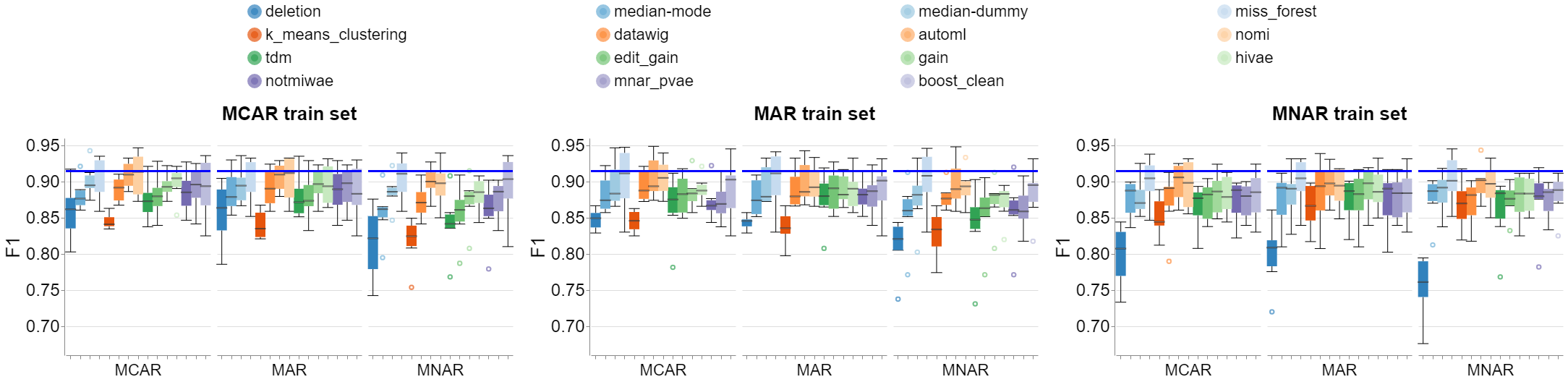}
    \caption{Diabetes (905 samples)}
\end{subfigure}
\hfill
\begin{subfigure}[h]{0.49\linewidth}
    \includegraphics[width=\linewidth]{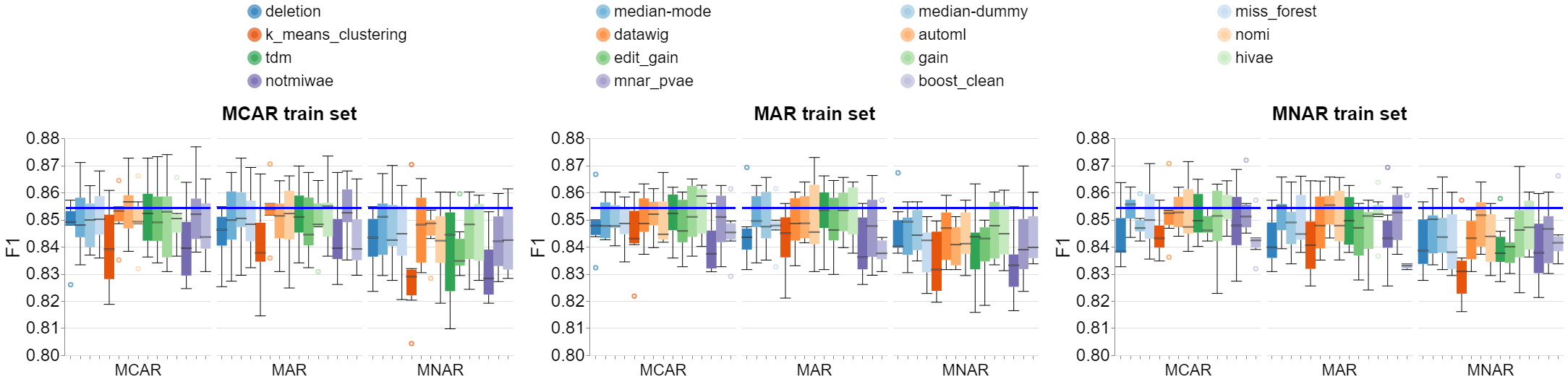}
    \caption{German (1000 samples)}
\end{subfigure}
\begin{subfigure}[h]{0.49\linewidth}
    \includegraphics[width=\linewidth]{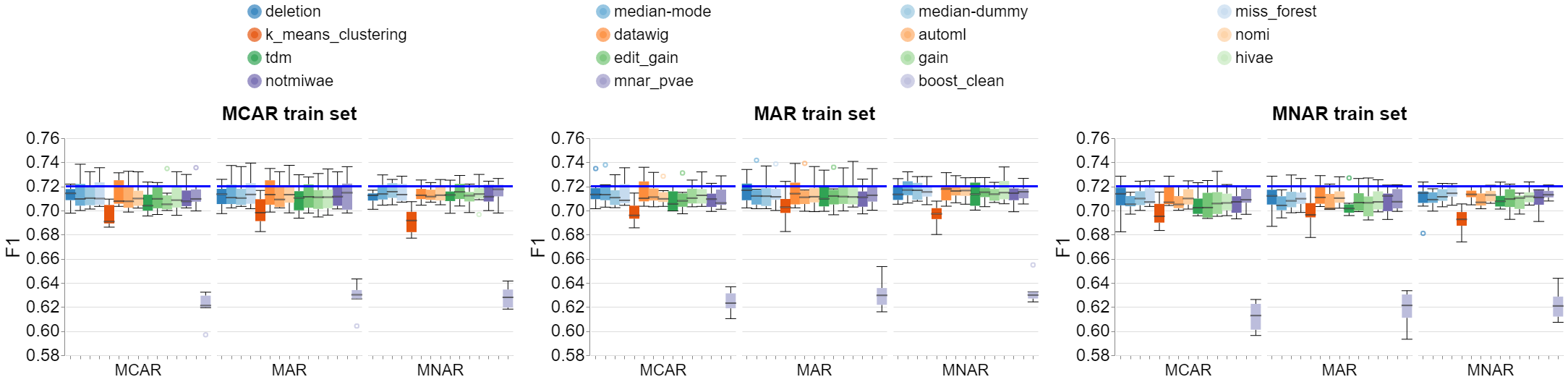}
    \caption{Folk Income (15,000 samples)}
\end{subfigure}
\hfill
\begin{subfigure}[h]{0.49\linewidth}
    \includegraphics[width=\linewidth]{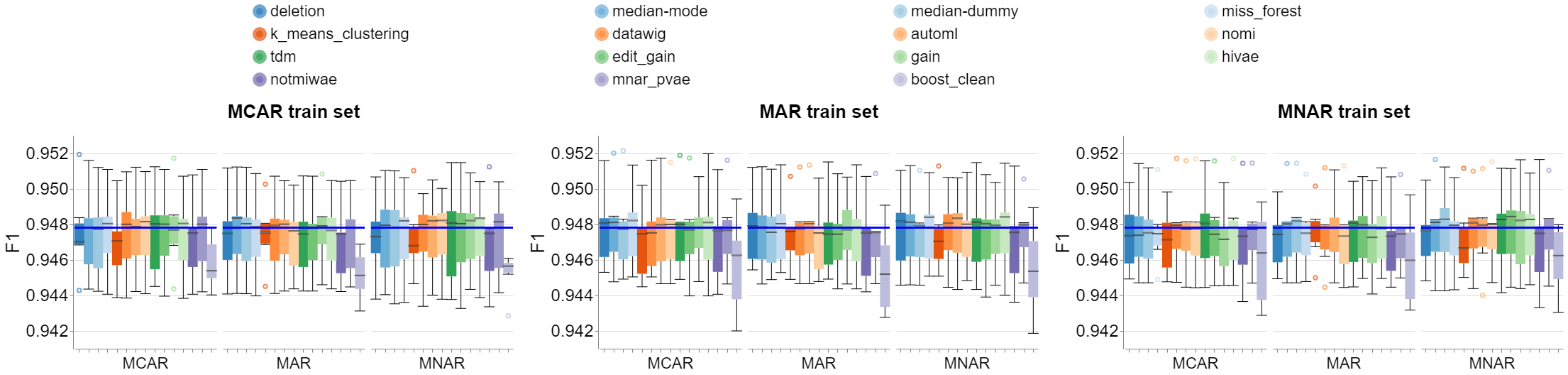}
    \caption{Law School (20,798 samples)}
\end{subfigure}
\begin{subfigure}[h]{0.49\linewidth}
    \includegraphics[width=\linewidth]{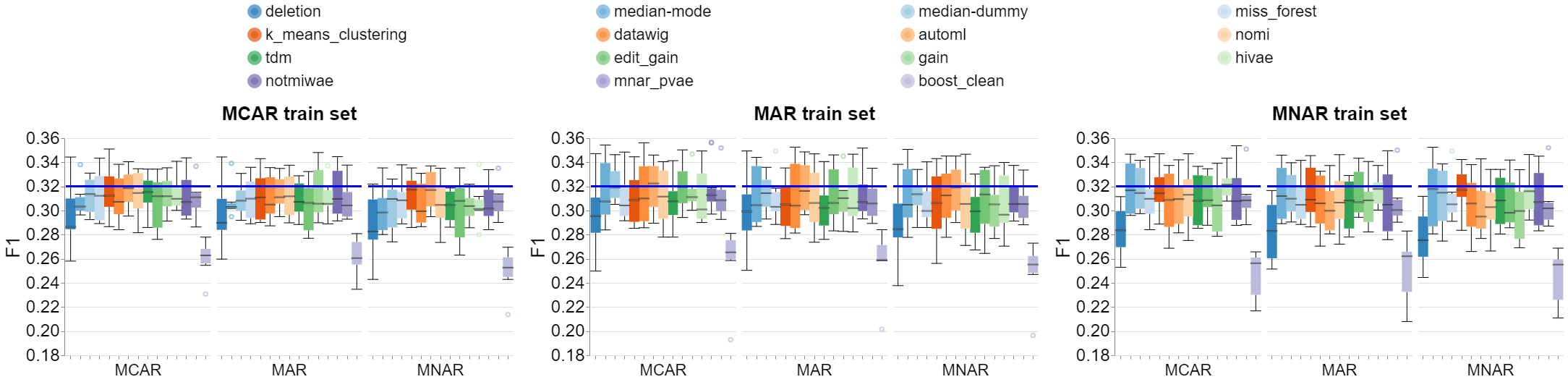}
    \caption{Bank (40,004 samples)}
\end{subfigure}
\hfill
\begin{subfigure}[h]{0.49\linewidth}
    \includegraphics[width=\linewidth]{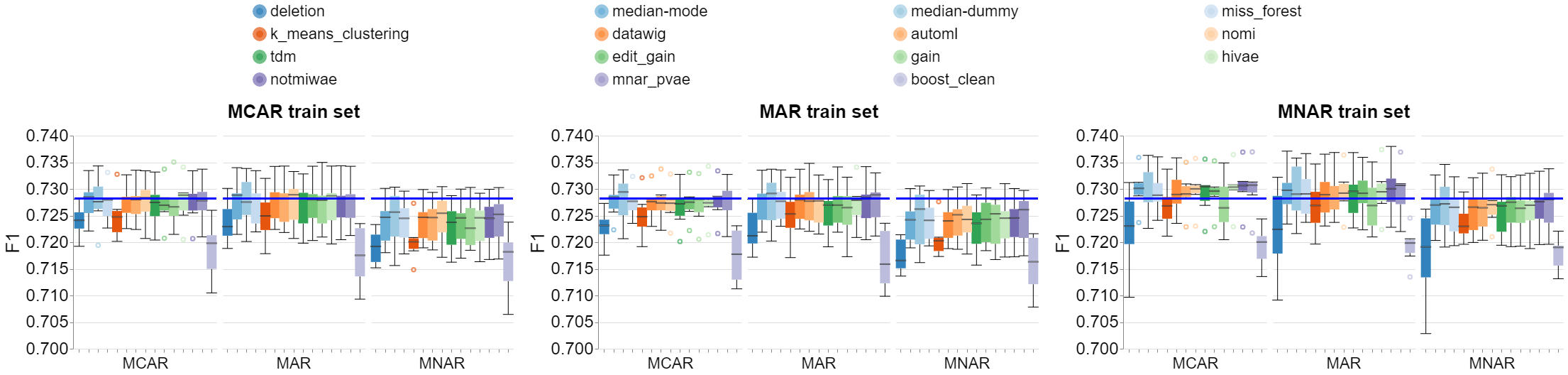}
    \caption{Heart (70,000 samples)}
\end{subfigure}
\hfill
\begin{subfigure}[h]{0.6\linewidth}
    \includegraphics[width=\linewidth]{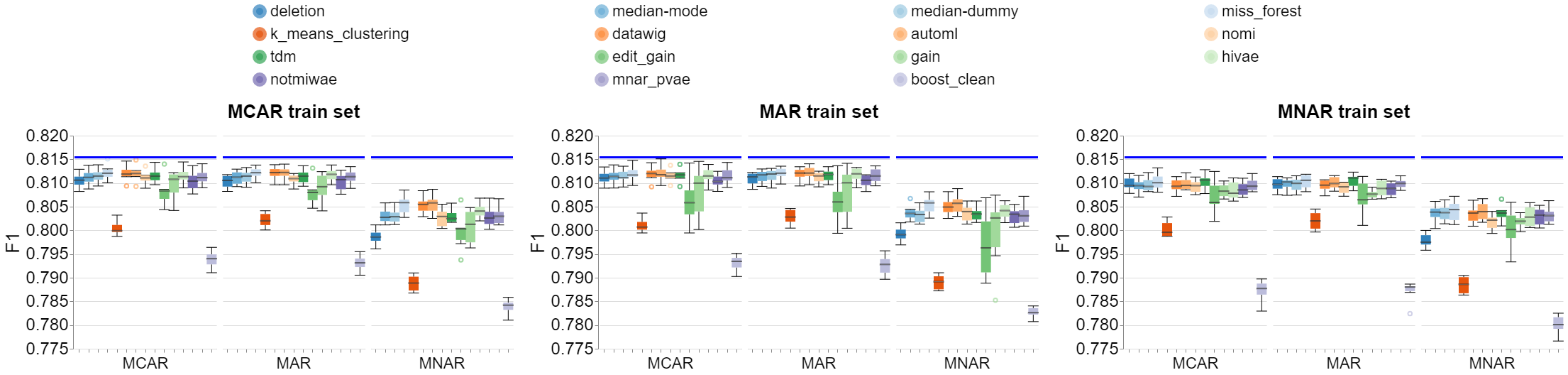}
    \caption{Folk Employment (302,640 samples)}
\end{subfigure}

\vspace{-0.3cm}
\caption{F1 of best performing models for different imputation strategies (colors in legend) on different datasets (subplots) under missingness shift (Scenarios S1-S9). The blue line indicates the performance of a model trained on clean data.}
\label{fig:app-all-F1}

\vspace{9cm}
\end{figure*}

\textbf{F1 of predictive model.}
F1 scores of predictive models trained with various \mvi techniques under different train and test missingness conditions, including missingness shift (scenarios S1-9 from Table~\ref{tab:scenarios}), are presented in Figure~\ref{fig:app-all-F1}. The plot demonstrates that F1 scores of predictive models are sensitive to missingness shifts across different datasets, as also discussed in Section~\ref{sec:shift-f1}. The impact of missingness shift is most significant for MCAR train \& MNAR test and MAR train \& MNAR test scenarios on the \diabetes, \german, and \heart datasets. This observation aligns with the statement in~\cite{Shadbahr_nature_medicine} that MNAR is more complex to model than MCAR and MAR. Additionally, \missforest emerges as the most robust technique in handling missingness shifts across all datasets and settings.

\begin{figure}[H]
\begin{subfigure}[h]{\linewidth}
    \includegraphics[width=\linewidth]{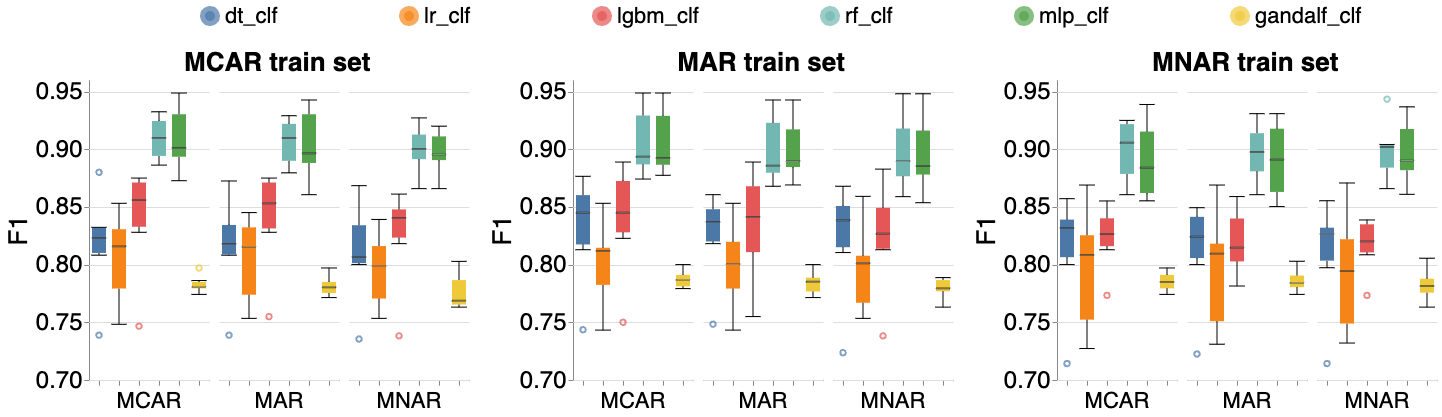}
    \caption{AutoML}
\end{subfigure}
\begin{subfigure}[h]{\linewidth}
    \includegraphics[width=\linewidth]{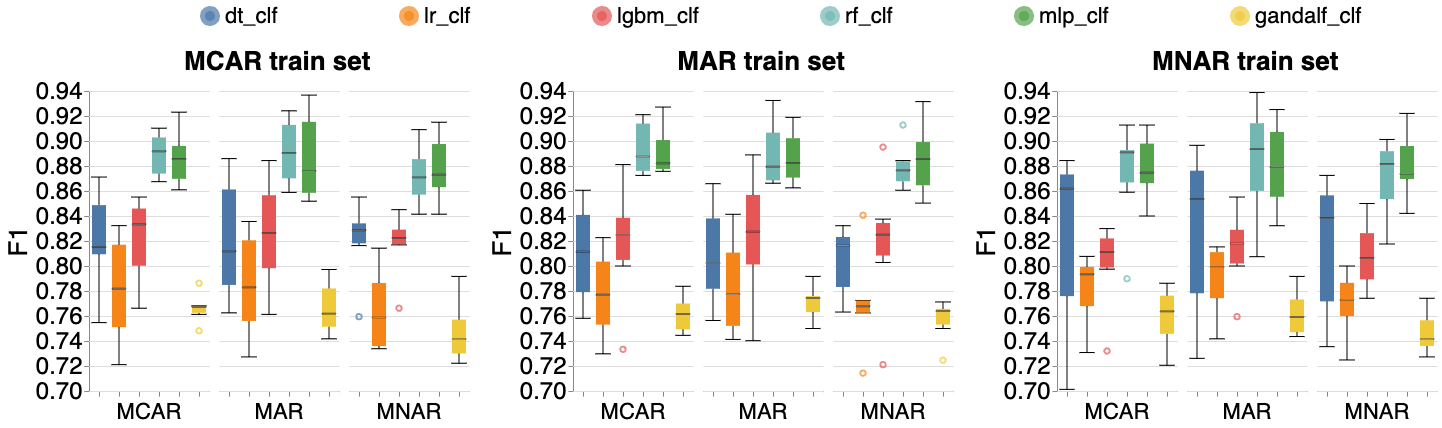}
    \caption{Datawig}
\end{subfigure}
\begin{subfigure}[h]{\linewidth}
    \includegraphics[width=\linewidth]{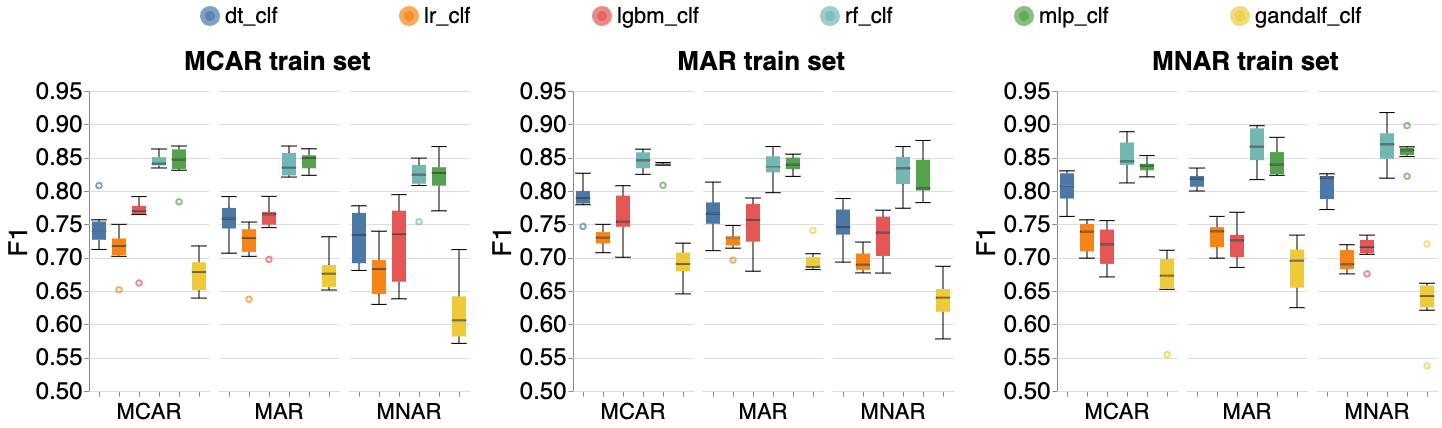}
    \caption{K-Means Clustering}
\end{subfigure}
\begin{subfigure}[h]{\linewidth}
    \centering
    \includegraphics[width=\linewidth]{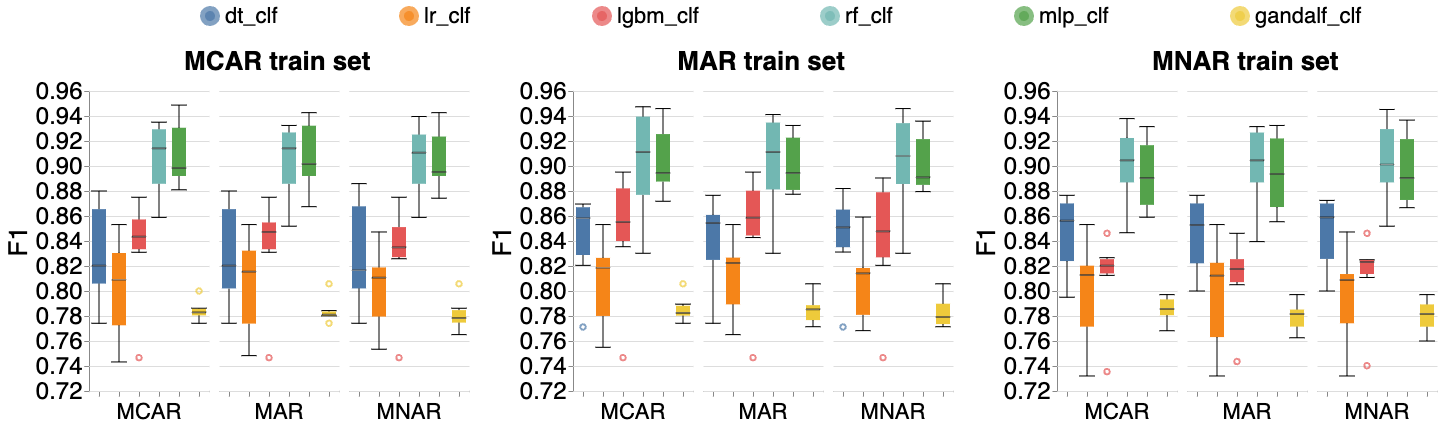}
    \caption{MissForest}
\end{subfigure}

\caption{F1 of different models (colors in legend) on the \diabetes dataset for ML and DL-based \mvi techniques (subplots) under missingness shift (Scenarios S1-S9).}
\label{fig:app-F1-diabetes-models}
\end{figure}

Figure~\ref{fig:app-F1-diabetes-models} further supports these findings by showing the performance of different model types in terms of F1 scores on \diabetes after applying various ML and DL-based null imputers.

\begin{figure*}[h!]
\begin{subfigure}[h]{0.49\linewidth}
    \centering
    \includegraphics[width=\linewidth]{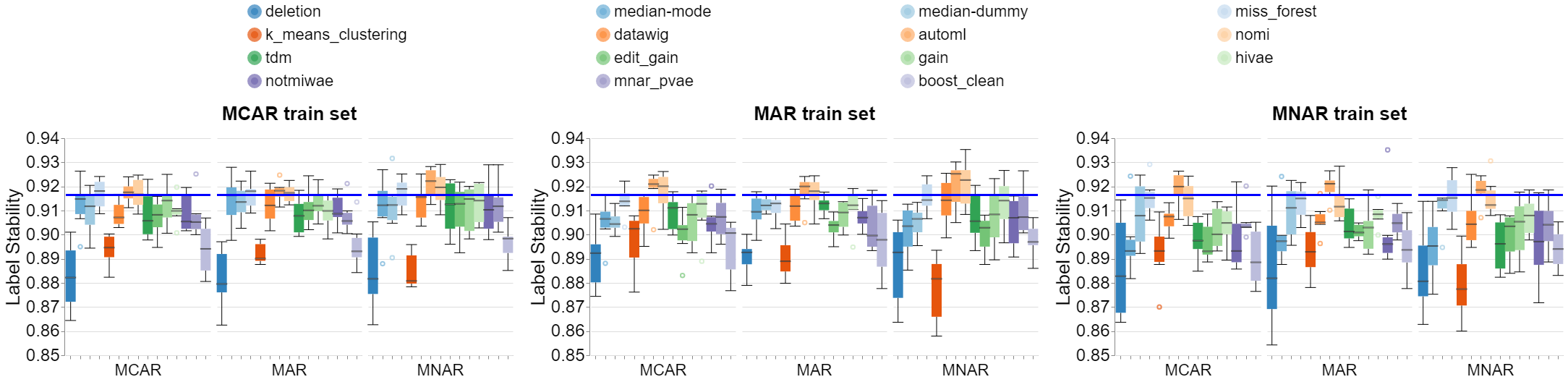}
    \caption{Diabetes (905 samples)}
\end{subfigure}
\hfill
\begin{subfigure}[h]{0.49\linewidth}
    \includegraphics[width=\linewidth]{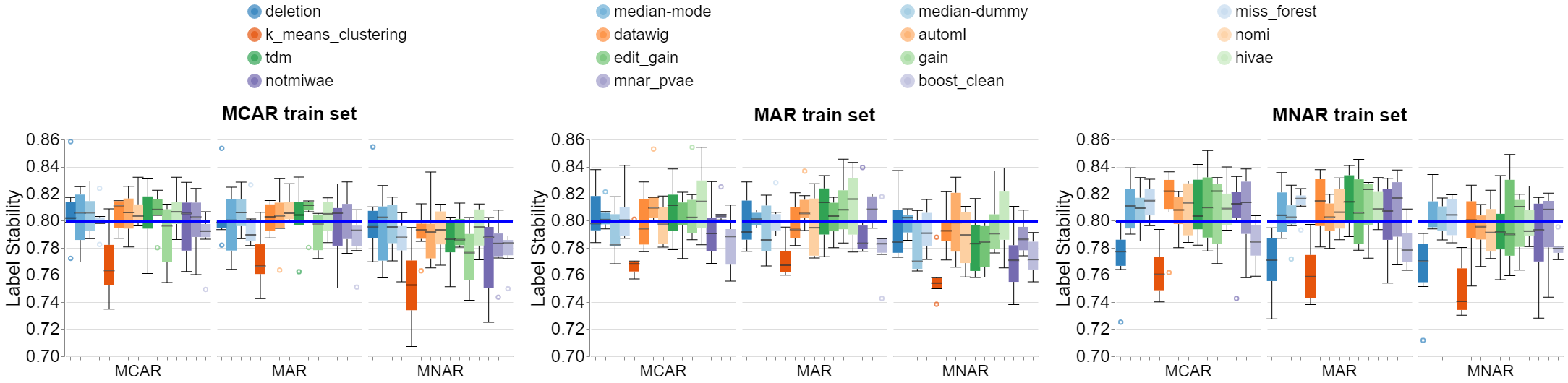}
    \caption{German (1000 samples)}
\end{subfigure}
\begin{subfigure}[h]{0.49\linewidth}
    \includegraphics[width=\linewidth]{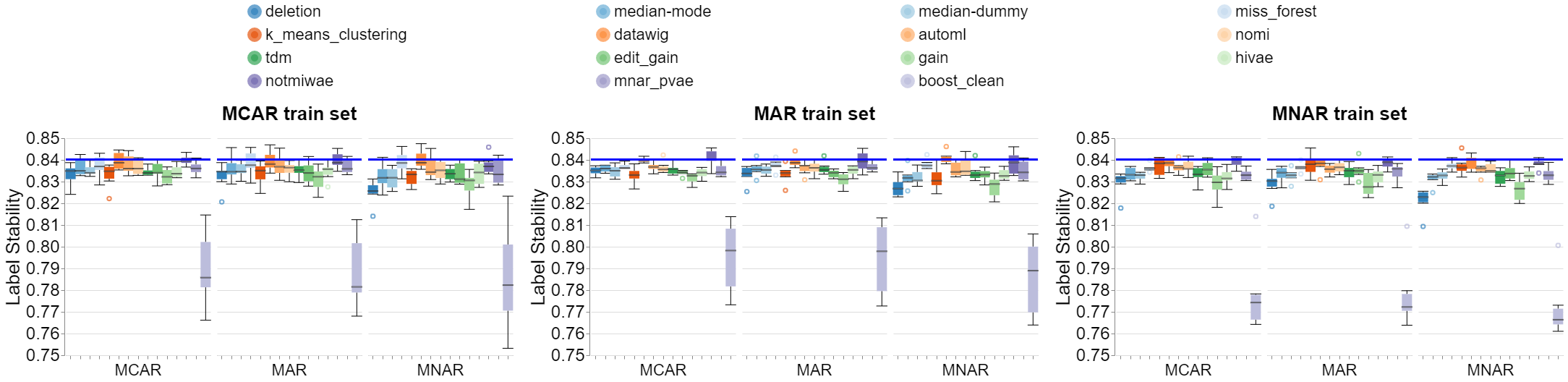}
    \caption{Folk Income (15,000 samples)}
\end{subfigure}
\hfill
\begin{subfigure}[h]{0.49\linewidth}
    \includegraphics[width=\linewidth]{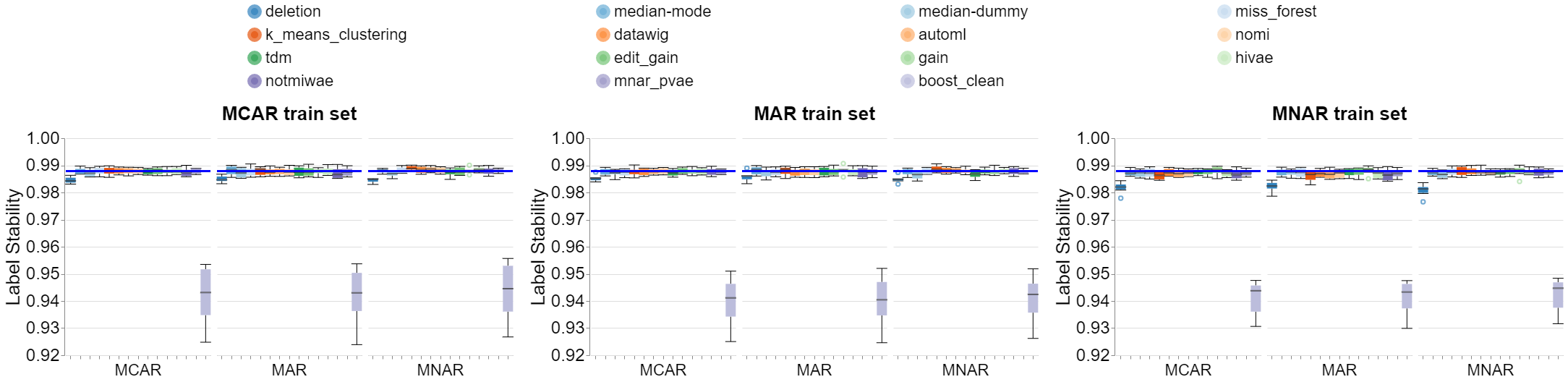}
    \caption{Law School (20,798 samples)}
\end{subfigure}
\begin{subfigure}[h]{0.49\linewidth}
    \includegraphics[width=\linewidth]{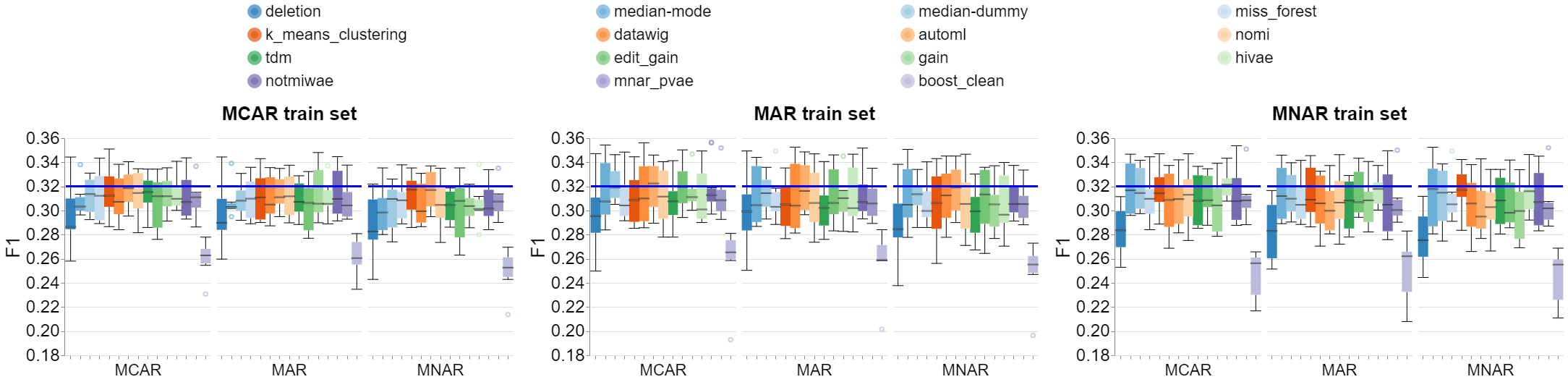}
    \caption{Bank (40,004 samples)}
\end{subfigure}
\hfill
\begin{subfigure}[h]{0.49\linewidth}
    \includegraphics[width=\linewidth]{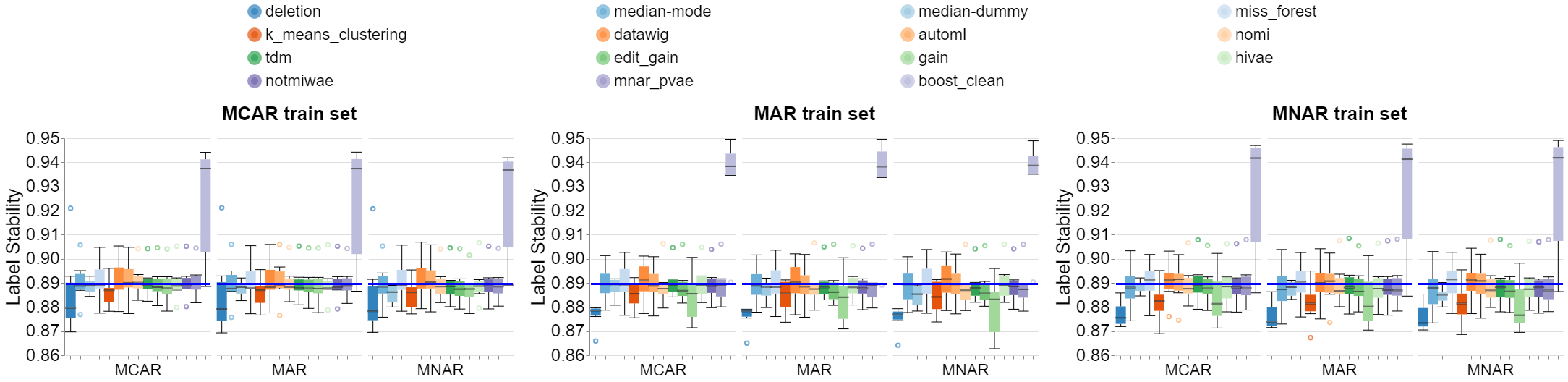}
    \caption{Heart (70,000 samples)}
\end{subfigure}
\hfill
\begin{subfigure}[h]{0.6\linewidth}
    \includegraphics[width=\linewidth]{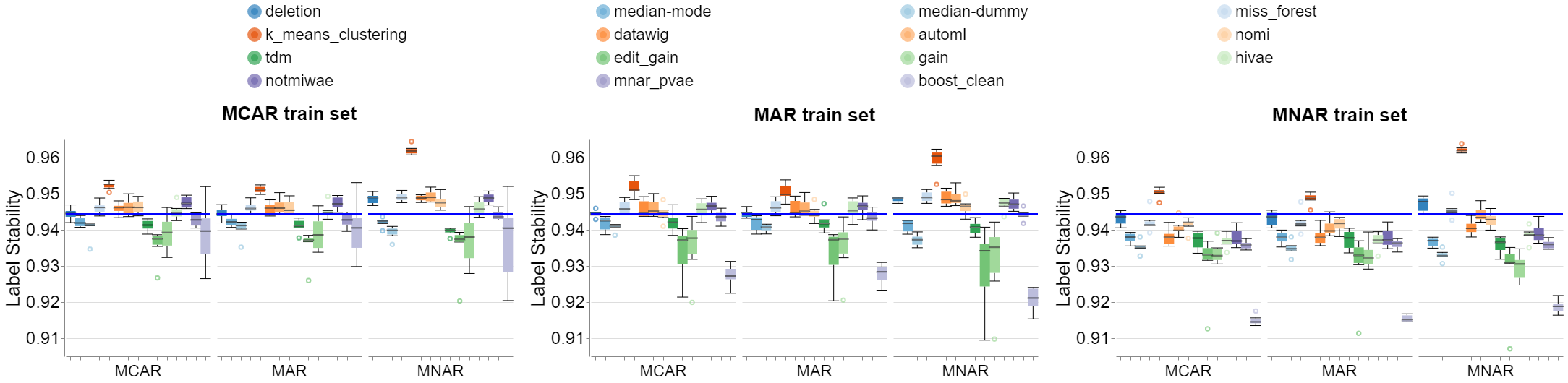}
    \caption{Folk Employment (302,640 samples)}
\end{subfigure}

\vspace{-0.3cm}
\caption{Label stability of best performing models for different imputation strategies (colors in legend) on different datasets (subplots) under missingness shift (Scenarios S1-S9). The blue line indicates the performance of a model trained on clean data.}
\label{fig:exp2-LS}

\vspace{9cm}
\end{figure*}

\textbf{Stability of predictive model.}
The stability of various top-performing models using different \mvi techniques under missingness shift is illustrated in Figure~\ref{fig:exp2-LS}. The plot indicates that missingness shift does not significantly impact model stability for most \mvi techniques, compared to the impact observed in F1 scores. However, the plot shows that the choice of \mvi technique can lead to notable differences in model stability for small-sized datasets, whereas changes in stability are less significant for datasets with more than 15K rows. This suggests that ML models exhibit better stability with larger datasets. Additionally, the plot supports our observation in Section~\ref{sec:exp1-single-model-stability} that the state-of-the-art multiple imputation technique \boostclean shows lower model stability across different datasets compared to other \mvi techniques, even if it holds good model accuracy. 

\begin{figure}[H]
\begin{subfigure}[h]{\linewidth}
    \includegraphics[width=\linewidth]{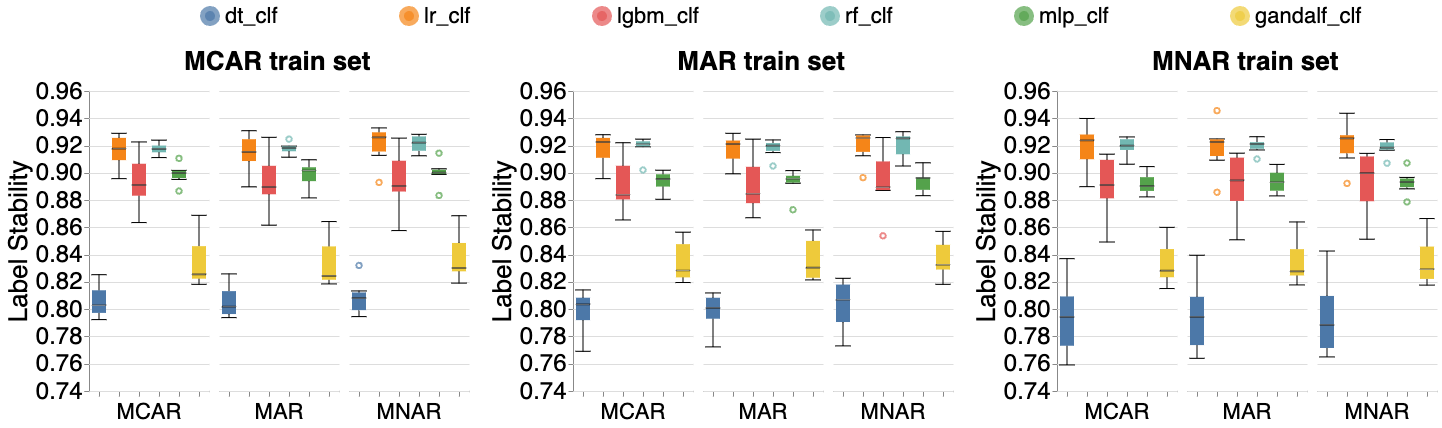}
    \caption{AutoML}
\end{subfigure}
\begin{subfigure}[h]{\linewidth}
    \includegraphics[width=\linewidth]{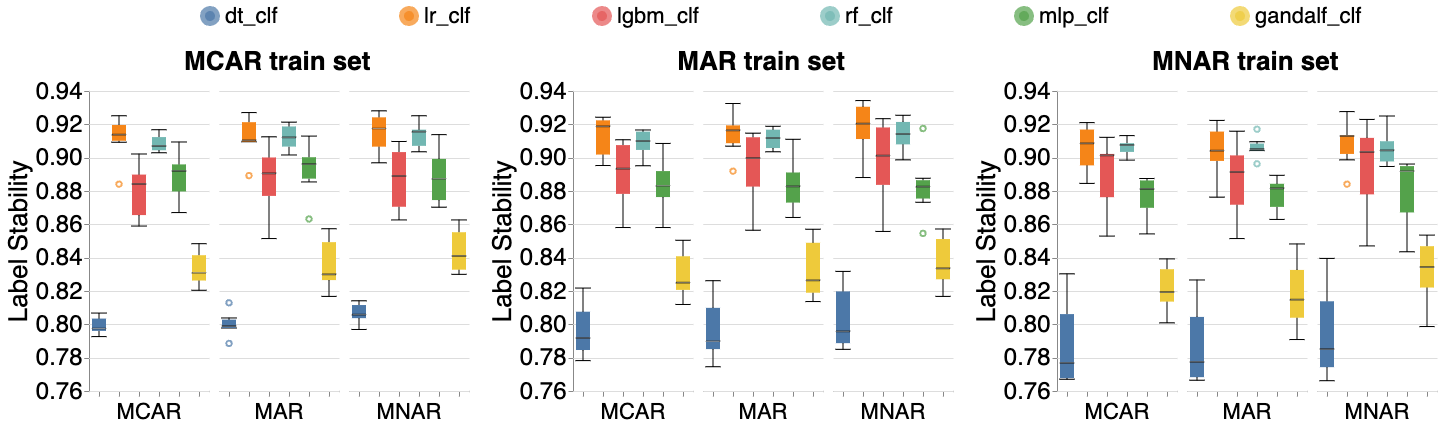}
    \caption{Datawig}
\end{subfigure}
\begin{subfigure}[h]{\linewidth}
    \includegraphics[width=\linewidth]{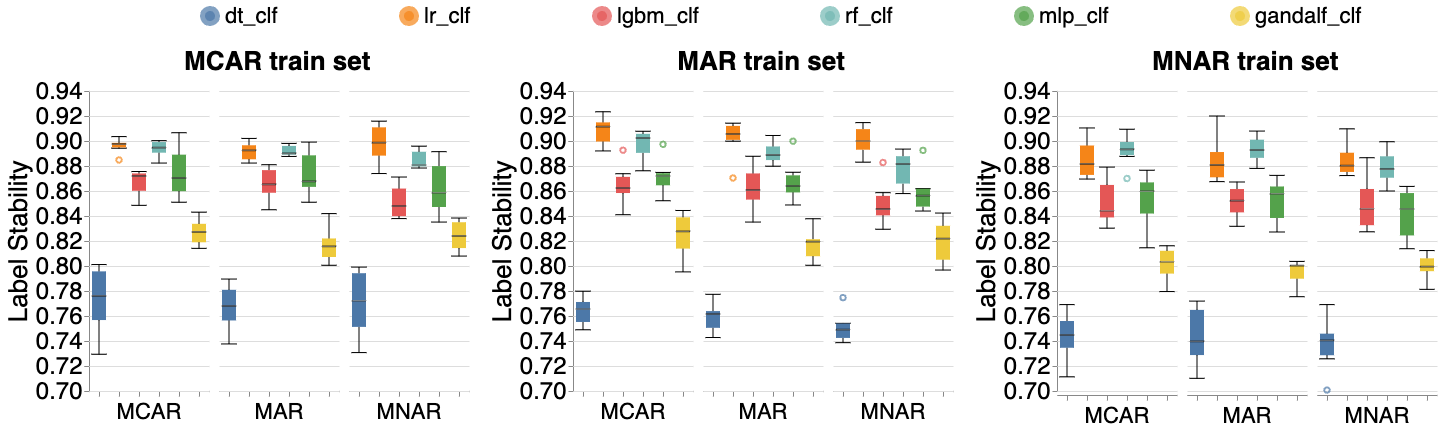}
    \caption{K-Means Clustering}
\end{subfigure}
\begin{subfigure}[h]{\linewidth}
    \centering
    \includegraphics[width=\linewidth]{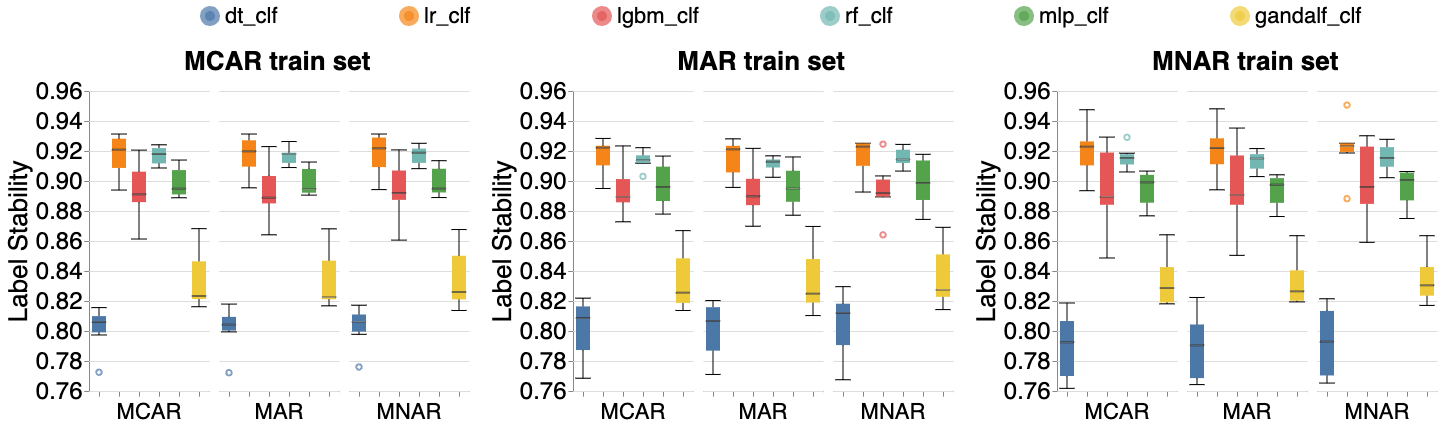}
    \caption{MissForest}
\end{subfigure}
\caption{Stability of different models (colors in legend) on the \diabetes dataset for ML and DL-based \mvi techniques (subplots) under missingness shift (Scenarios S1-S9).}
\label{fig:exp2-diabetes-models-stability}
\end{figure}

Figure~\ref{fig:exp2-diabetes-models-stability} further demonstrates the effect of model type on stability for the \diabetes dataset.

\vspace{1cm}

\begin{figure*}[h!]
    \begin{minipage}{\linewidth}
        \begin{multicols}{2}
            \textbf{Fairness of predictive model.} 
            The fairness of predictive models according to true positive rate difference is illustrated in Figure~\ref{fig:app-all-tprd}. The plot reveals that model fairness is highly sensitive to missingness shift. For instance, for the \diabetes dataset, a predictive model trained on an MCAR train set can exhibit worse TPRD on an MNAR test set compared to an MCAR test set. Conversely, for the \bank dataset, a predictive model trained on an MCAR train set shows better TPRD on an MNAR test set than on an MCAR test set. We further present the impact of missingness shifts on other fairness metrics, decomposed by different model types, in Figures~\ref{fig:missingness_shift_di_diabetes}-\ref{fig:missingness_shift_spd_heart}. These plots illustrate the effects on Disparate Impact, Selection Rate Difference, and True Negative Rate Difference across ML- and DL-based imputers and all our datasets.
        \end{multicols}
    \end{minipage}

    \vspace{5em}

    \centering
    \begin{minipage}{\linewidth}
        \centering
        \begin{subfigure}[h]{0.85\linewidth}
            \centering
            \includegraphics[width=\linewidth]{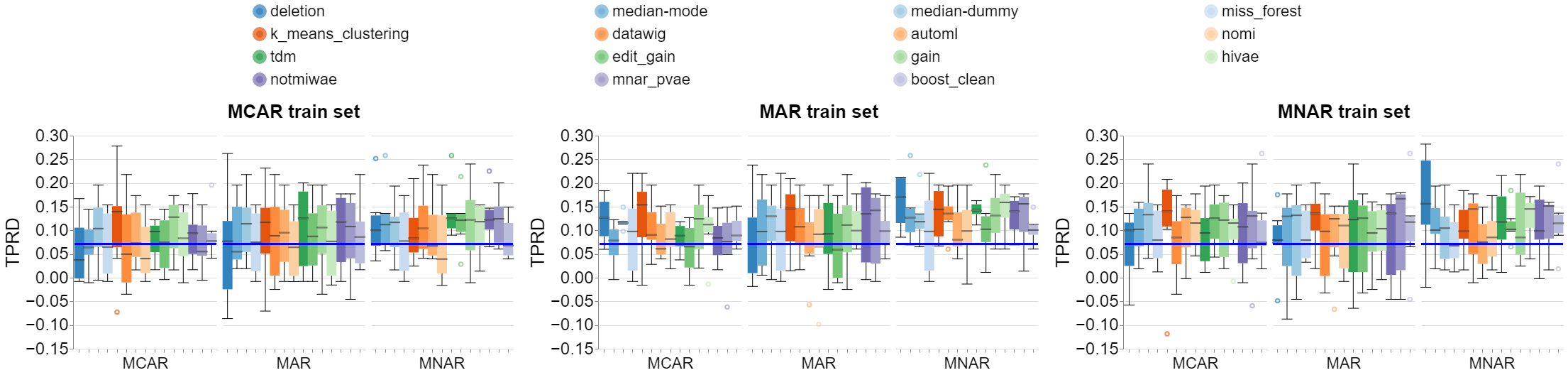}
            \caption{Diabetes and Random Forest}
        \end{subfigure}

        \centering
        \begin{subfigure}[h]{0.85\linewidth}
            \centering
            \includegraphics[width=\linewidth]{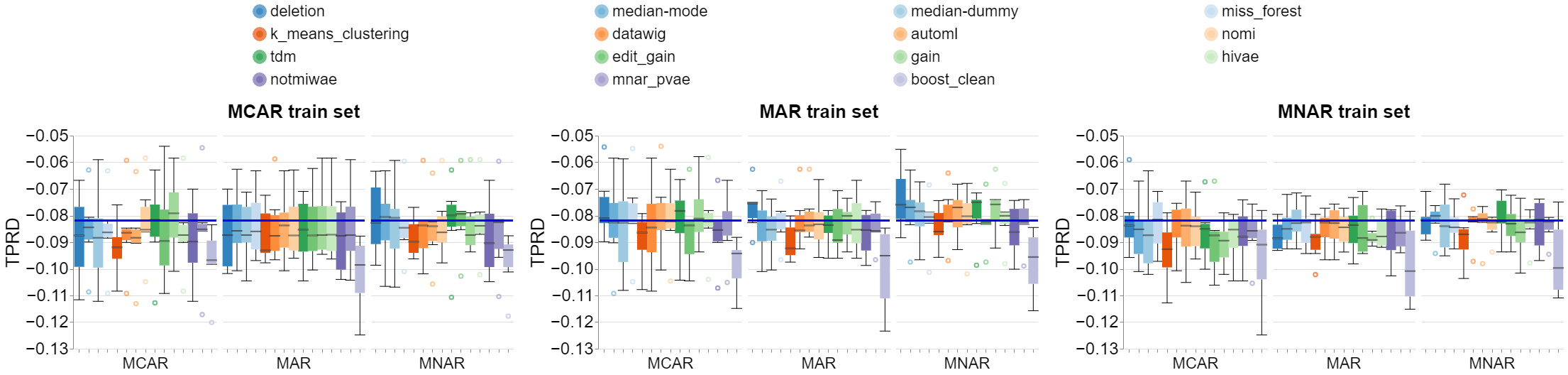}
            \caption{Law School and Logistic Regression}
        \end{subfigure}

        \centering
        \begin{subfigure}[h]{0.85\linewidth}
            \centering
            \includegraphics[width=\linewidth]{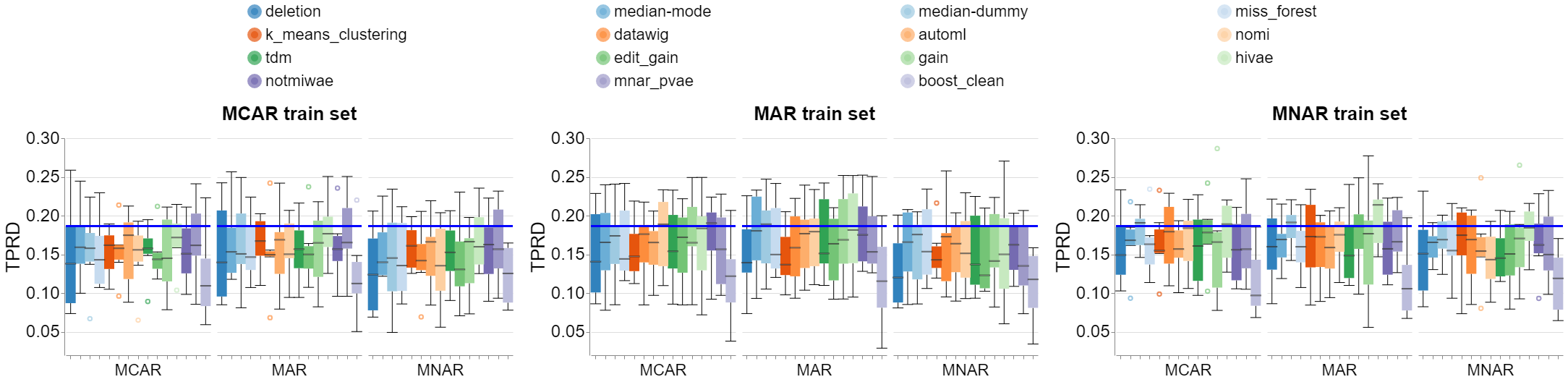}
            \caption{Bank and Gradient Boosted Trees}
        \end{subfigure}
        
        \caption{Fairness, measured by True Positive Rate Difference, of best performing models for different imputation strategies (colors in legend) on different datasets (subplots) under missingness shift (Scenarios S1-9)}
        \label{fig:app-all-tprd}
    \end{minipage}
    \vspace{2cm}
\end{figure*}


\begin{figure*}[h!]
\begin{subfigure}[h]{0.475\linewidth}
    \centering
    \includegraphics[width=\linewidth]{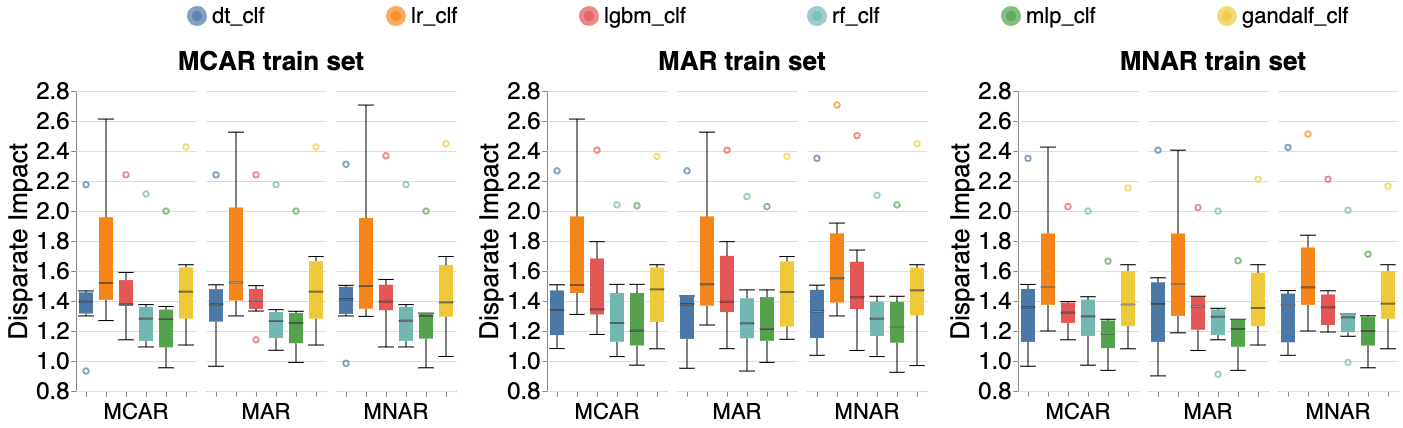}
    \caption{AutoML}
\end{subfigure}
\hfill
\begin{subfigure}[h]{0.475\linewidth}
    \includegraphics[width=\linewidth]{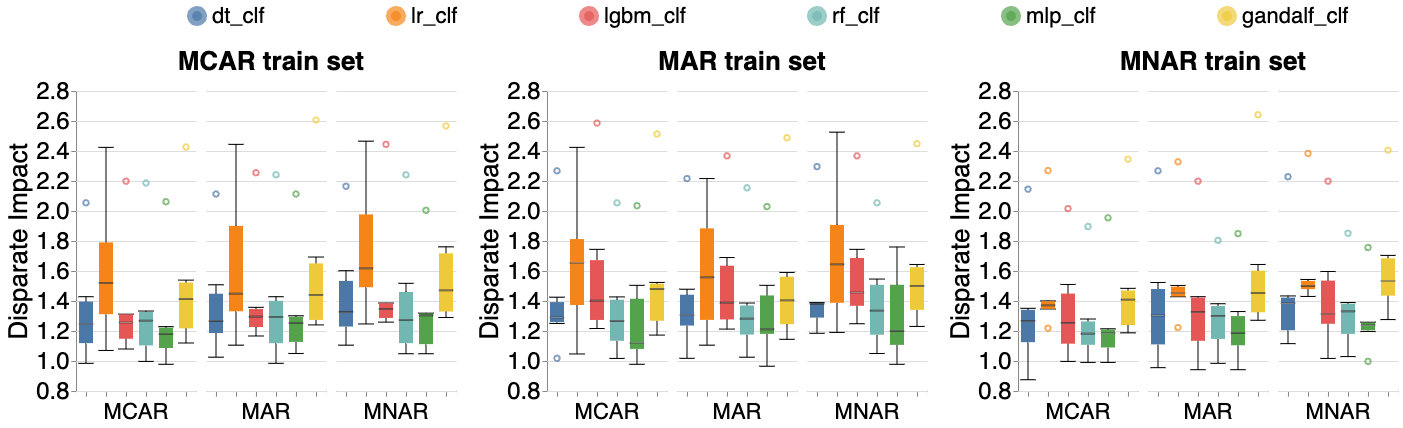}
    \caption{Datawig}
\end{subfigure}
\begin{subfigure}[h]{0.475\linewidth}
    \includegraphics[width=\linewidth]{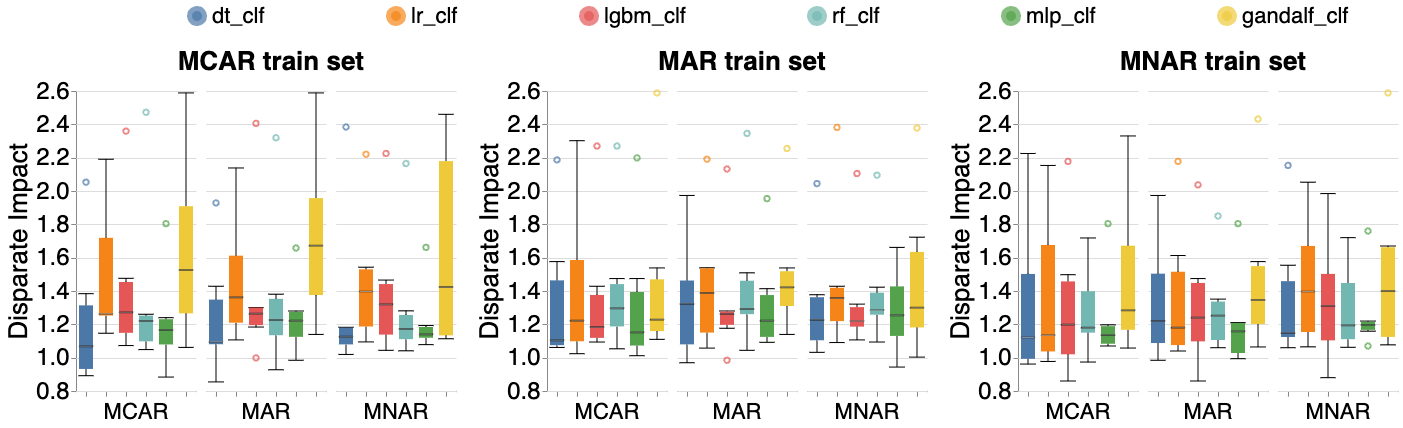}
    \caption{K-Means Clustering}
\end{subfigure}
\hfill
\begin{subfigure}[h]{0.475\linewidth}
    \includegraphics[width=\linewidth]{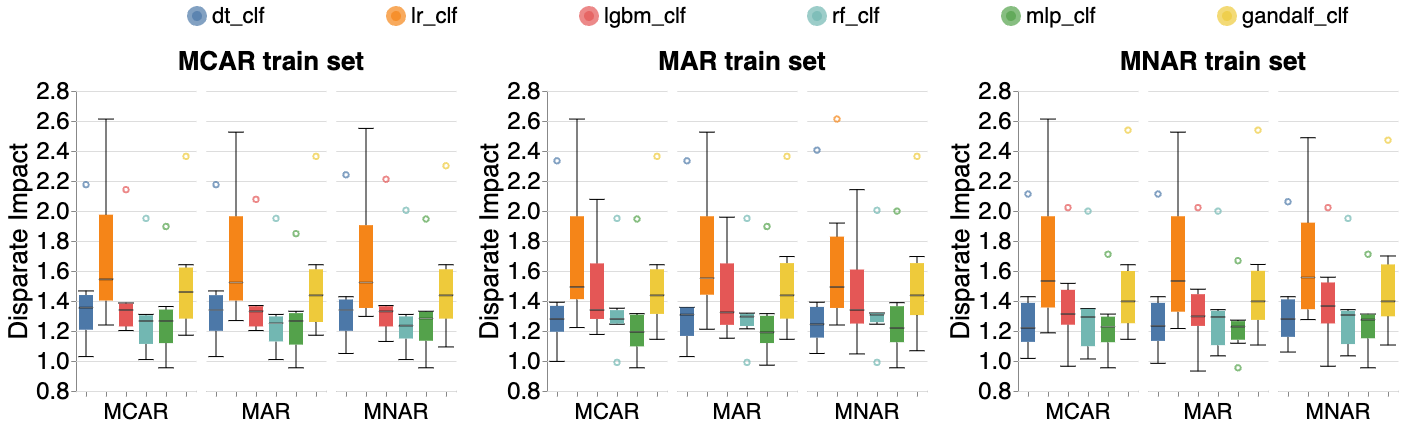}
    \caption{MissForest}
\end{subfigure}

\vspace{-0.3cm}
\caption{Disparate Impact of different models (colors in legend) on the \diabetes dataset for ML and DL-based \mvi techniques (subplots) under missingness shift.}
\label{fig:missingness_shift_di_diabetes}
\end{figure*}

\begin{figure}[h!]
\begin{subfigure}[h]{\linewidth}
    \includegraphics[width=\linewidth]{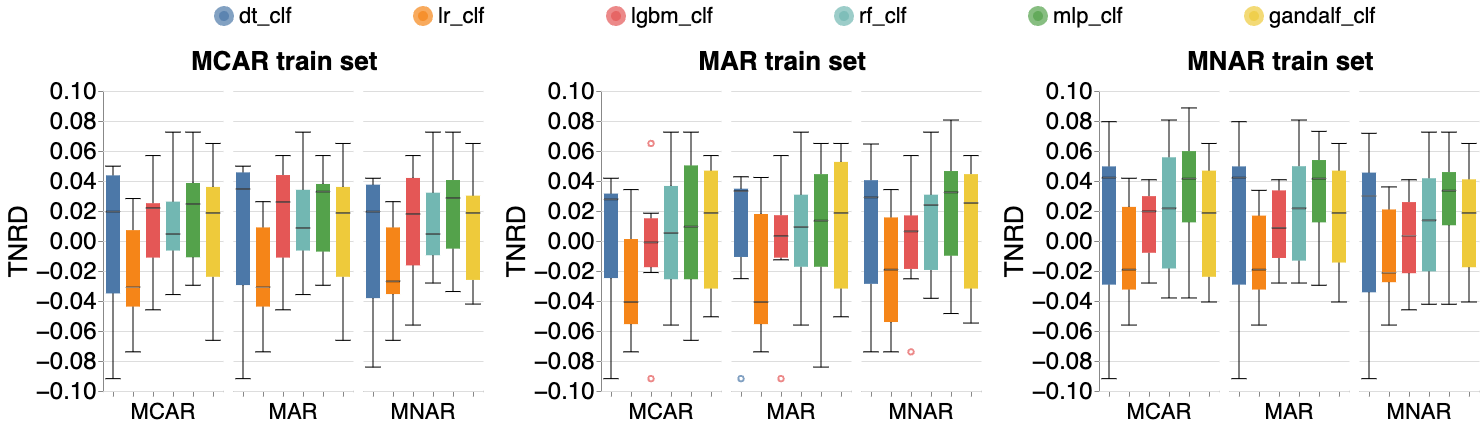}
    \caption{AutoML}
\end{subfigure}

\begin{subfigure}[h]{\linewidth}
    \includegraphics[width=\linewidth]{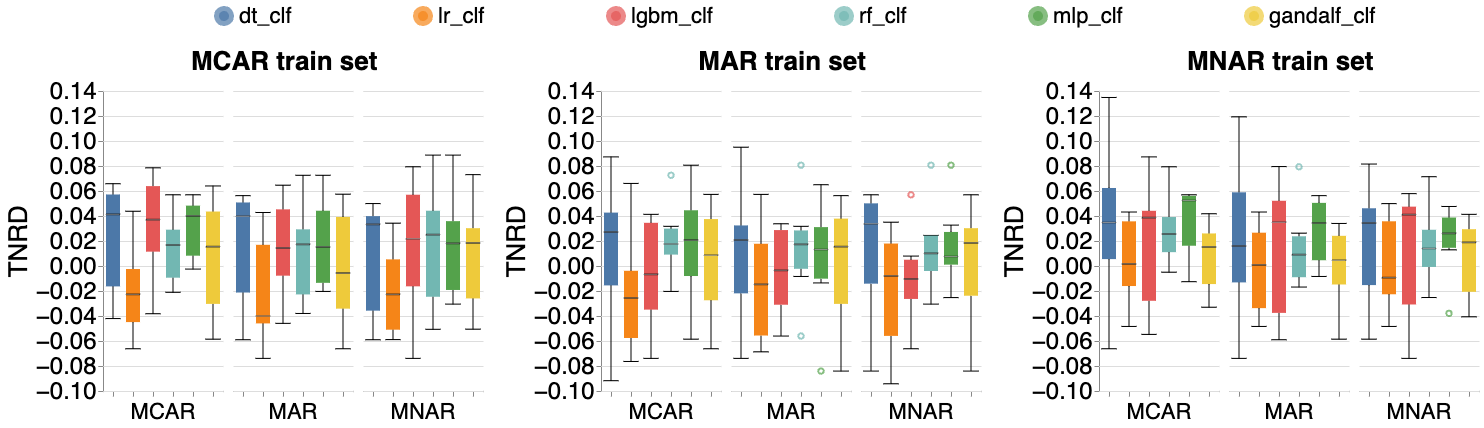}
    \caption{Datawig}
\end{subfigure}

\begin{subfigure}[h]{\linewidth}
    \includegraphics[width=\linewidth]{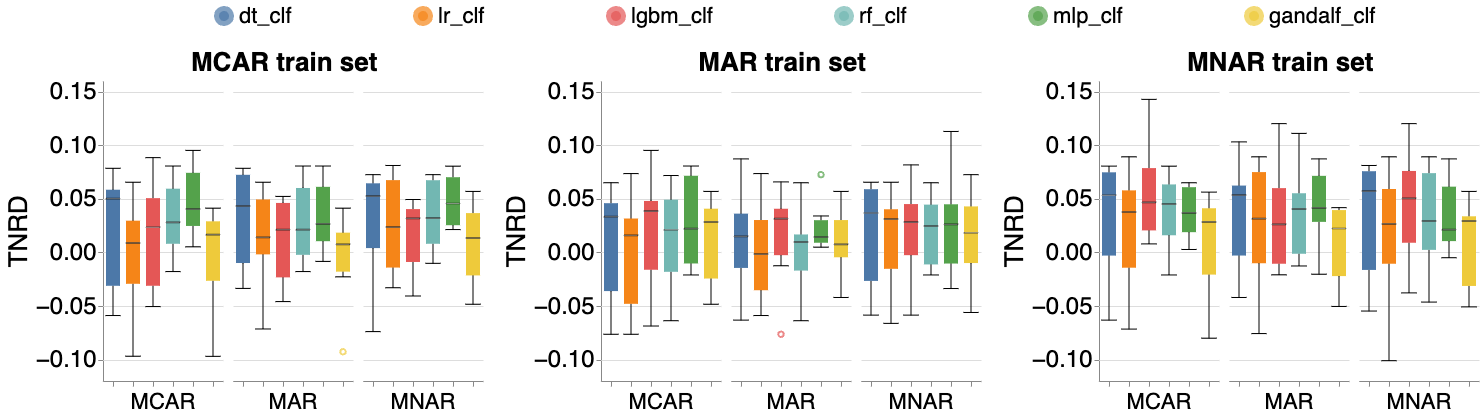}
    \caption{K-Means Clustering}
\end{subfigure}

\begin{subfigure}[h]{\linewidth}
    \includegraphics[width=\linewidth]{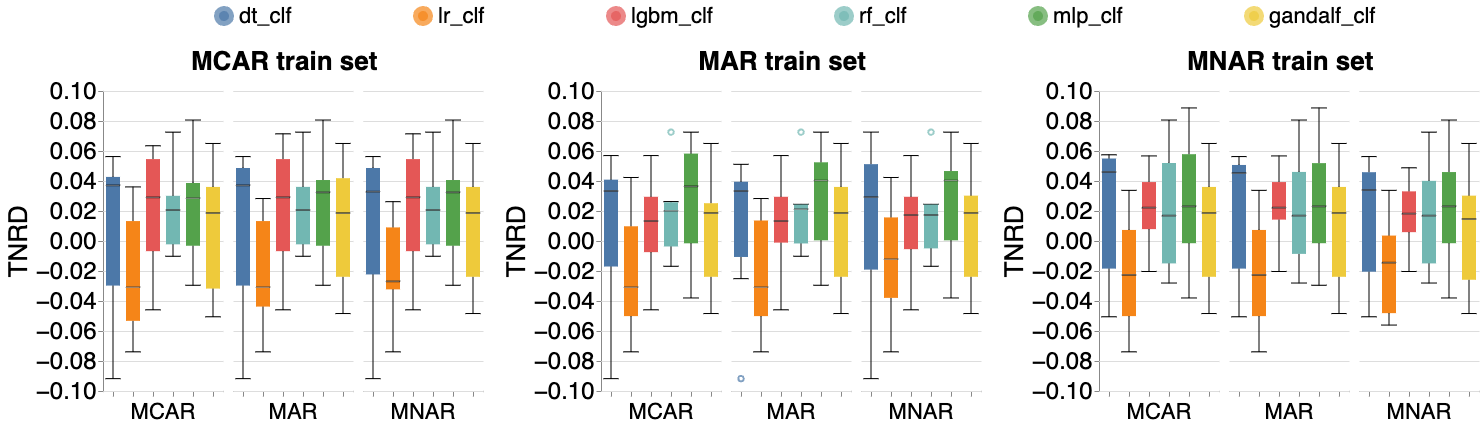}
    \caption{MissForest}
\end{subfigure}

\vspace{-0.3cm}
\caption{True Negative Rate Difference of different models (colors in legend) on the \diabetes dataset for ML and DL-based \mvi techniques (subplots) under missingness shift.}
\label{fig:missingness_shift_eq_odds_tnr_diabetes}
\end{figure}

\begin{figure}[h!]
\begin{subfigure}[h]{\linewidth}
    \includegraphics[width=\linewidth]{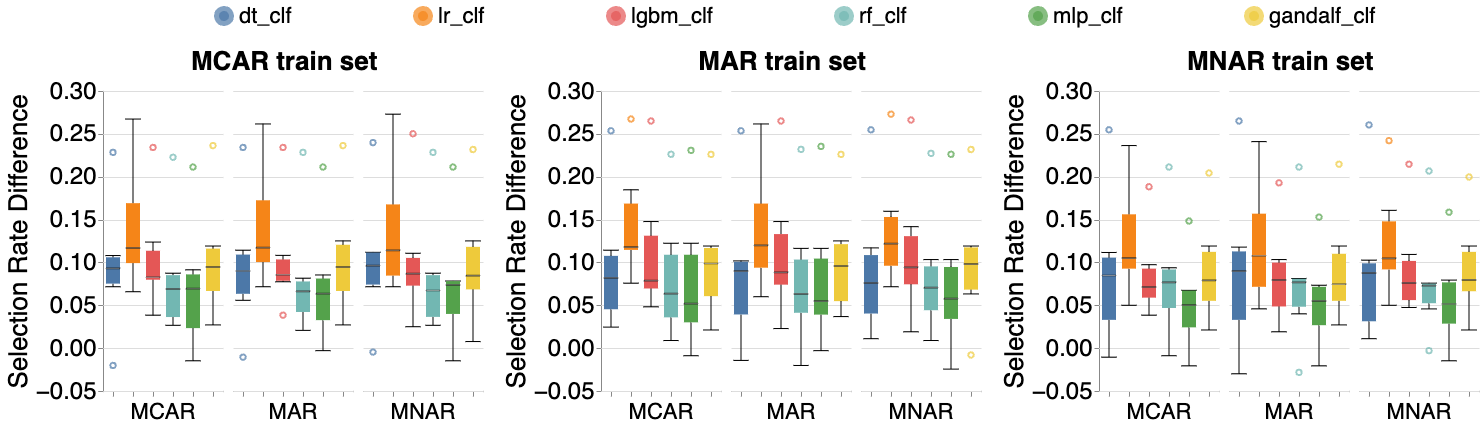}
    \caption{AutoML}
\end{subfigure}

\begin{subfigure}[h]{\linewidth}
    \includegraphics[width=\linewidth]{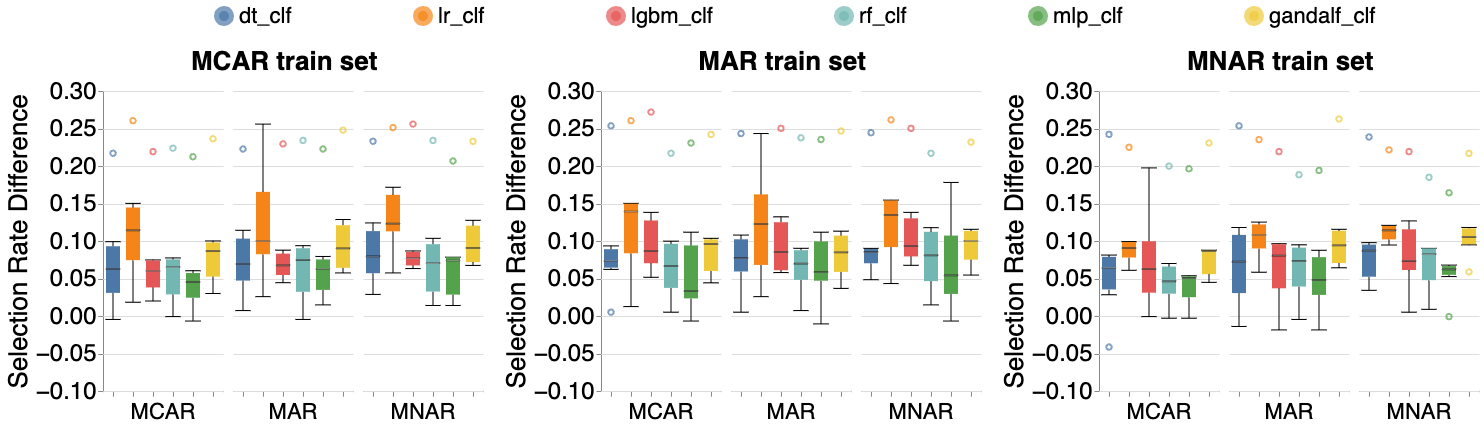}
    \caption{Datawig}
\end{subfigure}

\begin{subfigure}[h]{\linewidth}
    \includegraphics[width=\linewidth]{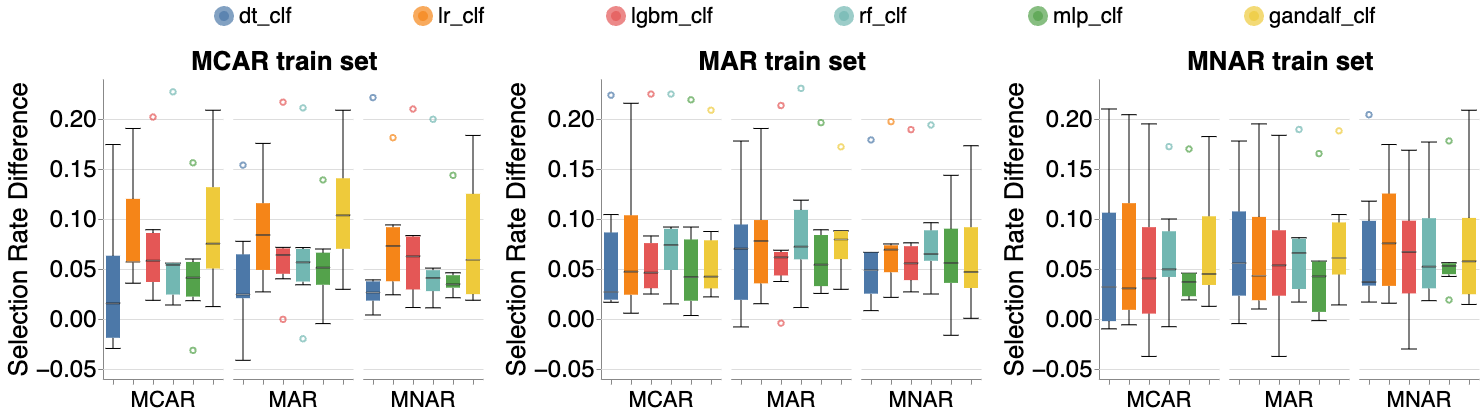}
    \caption{K-Means Clustering}
\end{subfigure}

\begin{subfigure}[h]{\linewidth}
    \includegraphics[width=\linewidth]{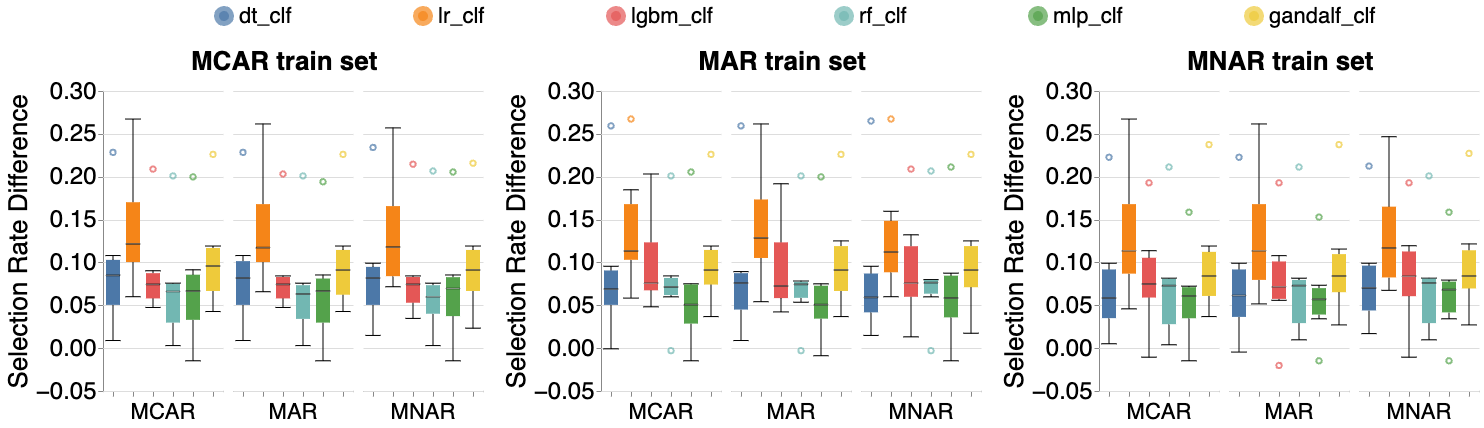}
    \caption{MissForest}
\end{subfigure}

\vspace{-0.3cm}
\caption{Selection Rate Difference of different models (colors in legend) on the \diabetes dataset for ML and DL-based \mvi techniques (subplots) under missingness shift.}
\label{fig:missingness_shift_spd_diabetes}
\end{figure}


\begin{figure*}[h!]
\begin{subfigure}[h]{0.475\linewidth}
    \centering
    \includegraphics[width=\linewidth]{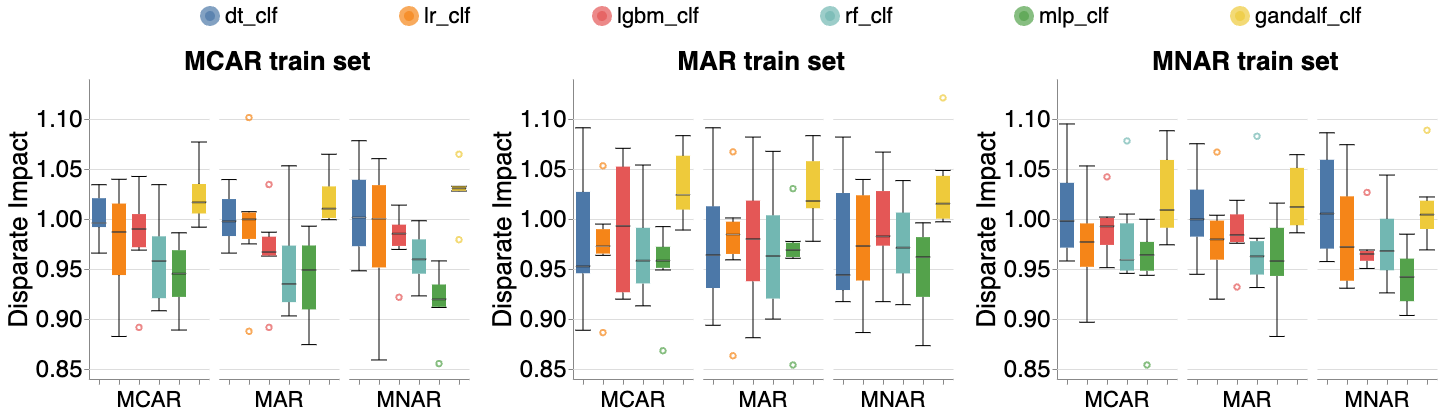}
    \caption{AutoML}
\end{subfigure}
\hfill
\begin{subfigure}[h]{0.475\linewidth}
    \includegraphics[width=\linewidth]{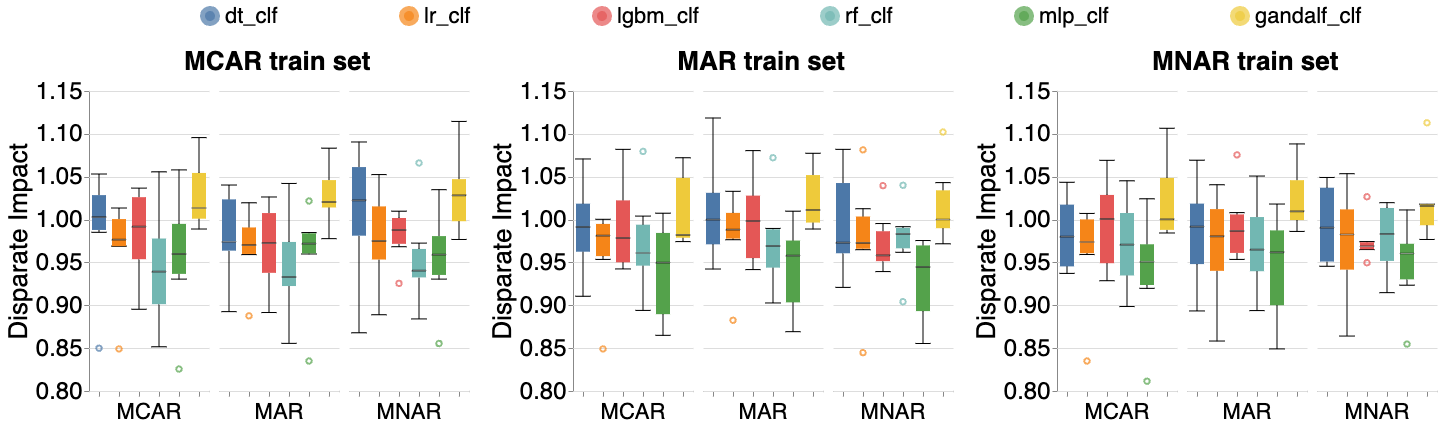}
    \caption{Datawig}
\end{subfigure}
\begin{subfigure}[h]{0.475\linewidth}
    \includegraphics[width=\linewidth]{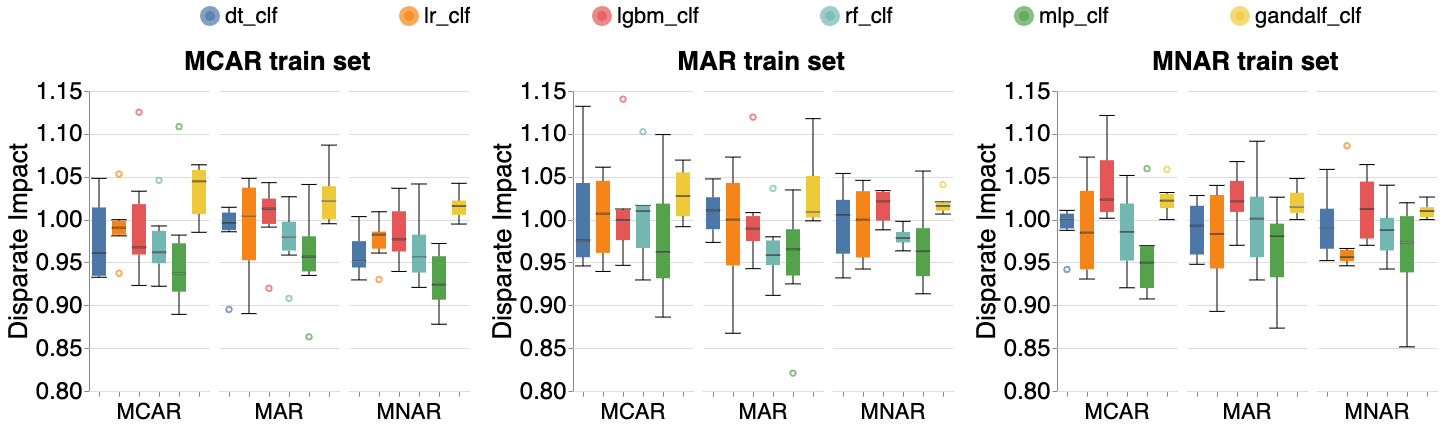}
    \caption{K-Means Clustering}
\end{subfigure}
\hfill
\begin{subfigure}[h]{0.475\linewidth}
    \includegraphics[width=\linewidth]{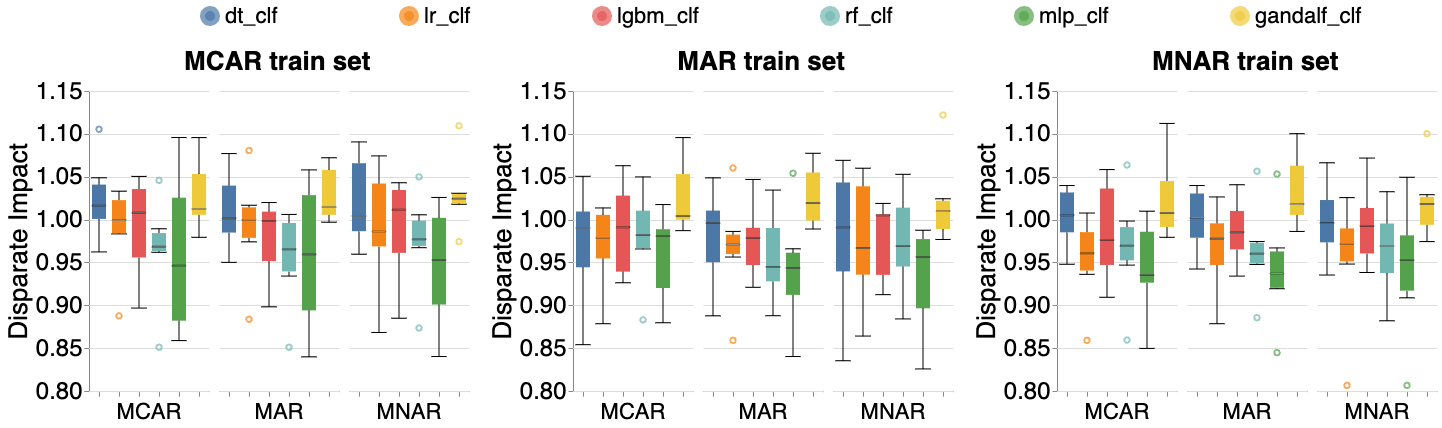}
    \caption{MissForest}
\end{subfigure}

\vspace{-0.3cm}
\caption{Disparate Impact of different models (colors in legend) on the \german dataset for ML and DL-based \mvi techniques (subplots) under missingness shift.}
\label{fig:missingness_shift_di_german}
\end{figure*}

\begin{figure}[h!]
\begin{subfigure}[h]{\linewidth}
    \includegraphics[width=\linewidth]{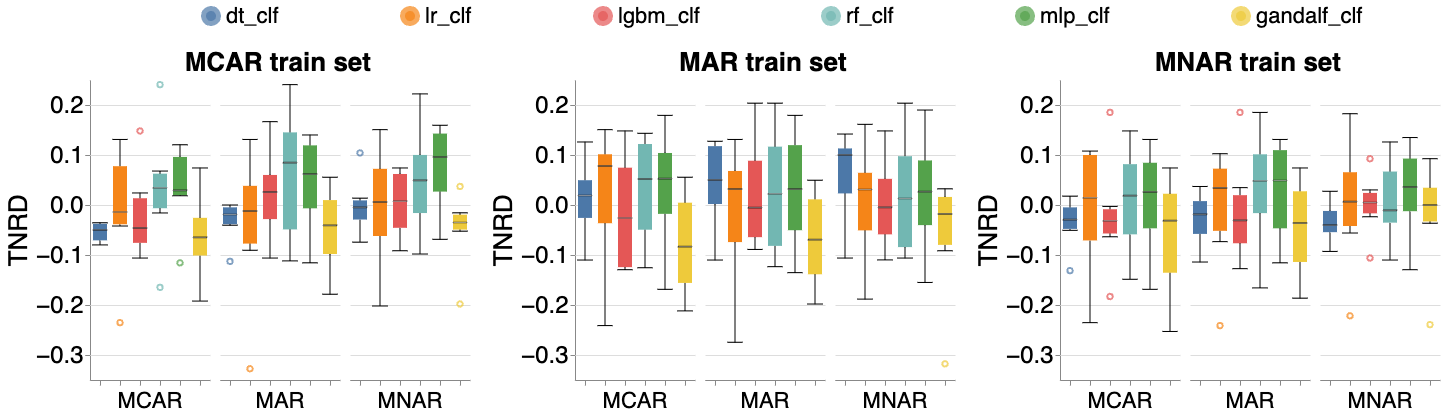}
    \caption{AutoML}
\end{subfigure}

\begin{subfigure}[h]{\linewidth}
    \includegraphics[width=\linewidth]{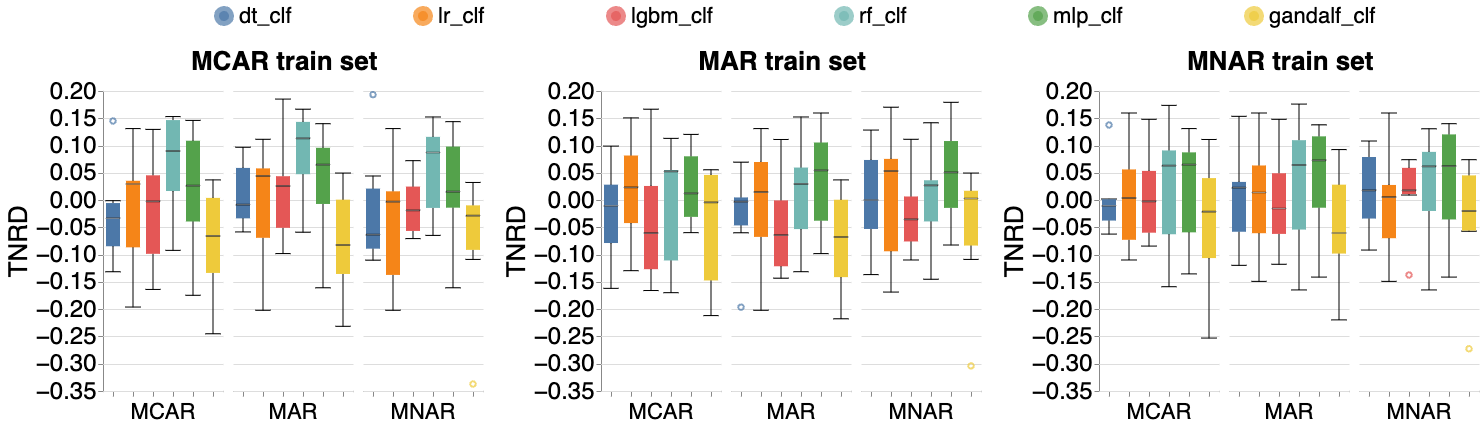}
    \caption{Datawig}
\end{subfigure}

\begin{subfigure}[h]{\linewidth}
    \includegraphics[width=\linewidth]{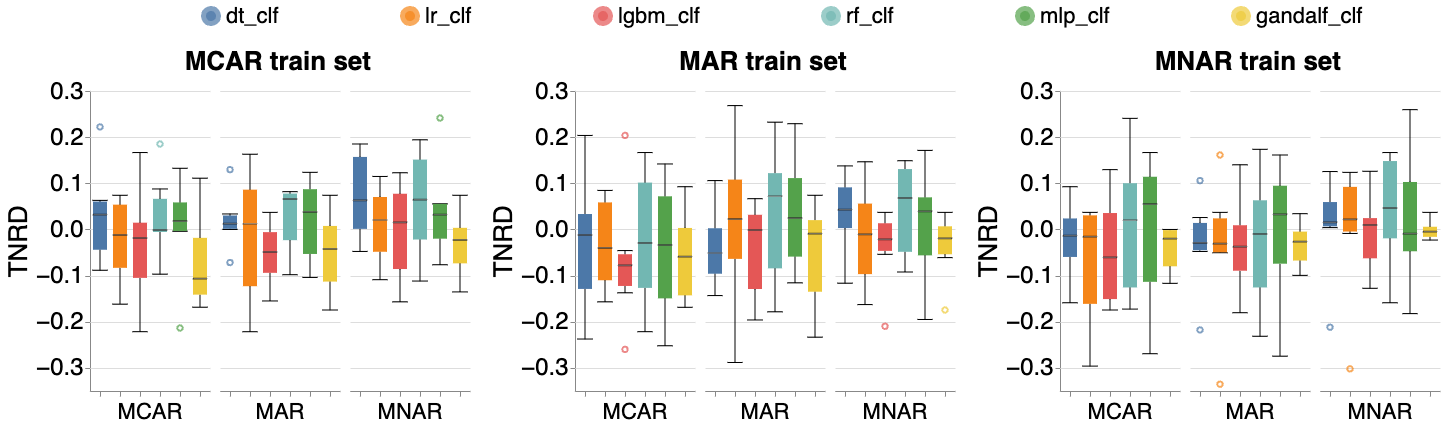}
    \caption{K-Means Clustering}
\end{subfigure}

\begin{subfigure}[h]{\linewidth}
    \includegraphics[width=\linewidth]{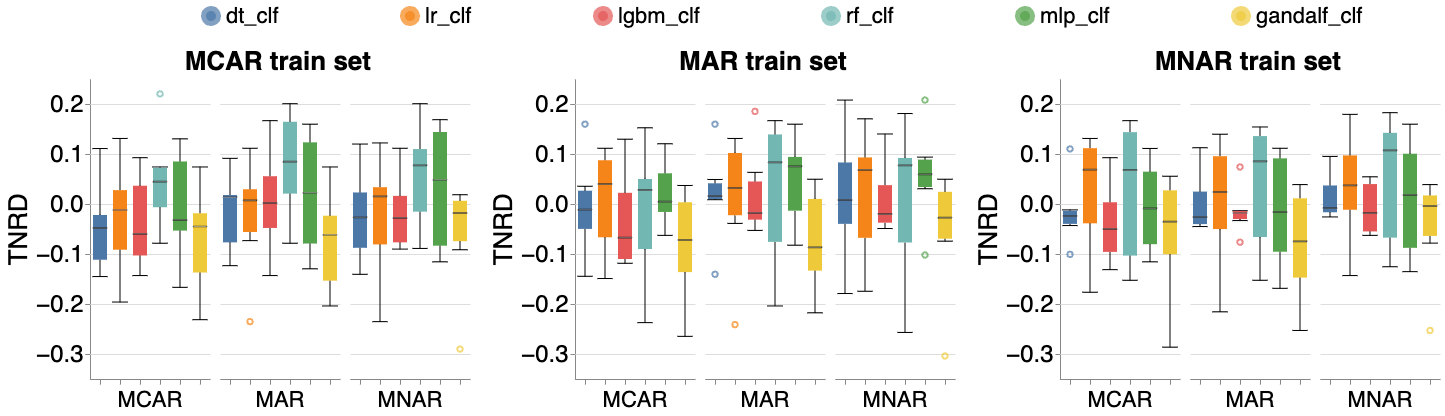}
    \caption{MissForest}
\end{subfigure}

\vspace{-0.3cm}
\caption{True Negative Rate Difference of different models (colors in legend) on the \german dataset for ML and DL-based \mvi techniques (subplots) under missingness shift.}
\label{fig:missingness_shift_eq_odds_tnr_german}
\end{figure}

\begin{figure}[h!]
\begin{subfigure}[h]{\linewidth}
    \includegraphics[width=\linewidth]{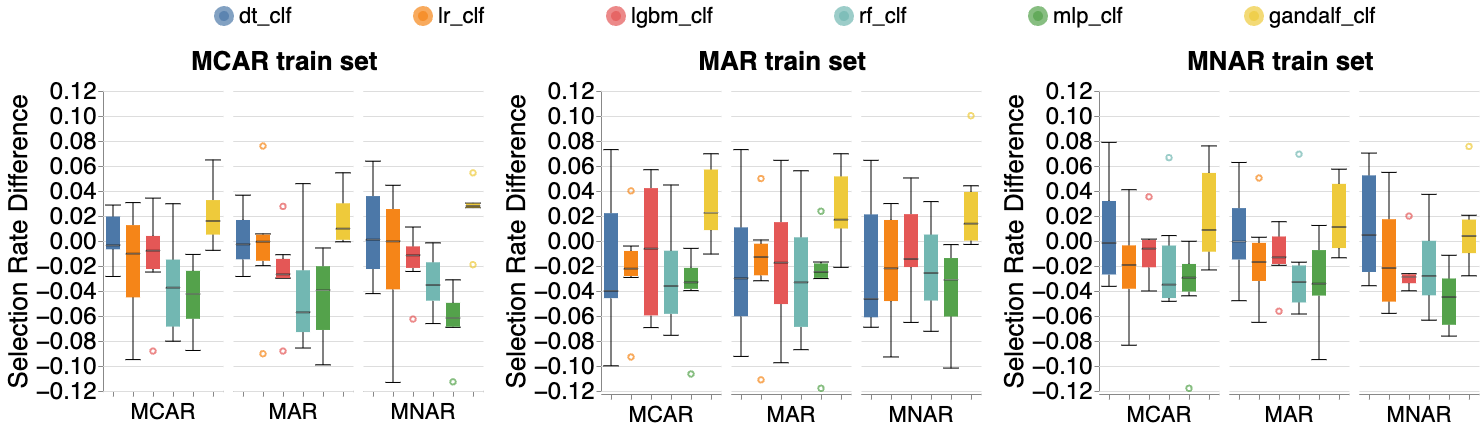}
    \caption{AutoML}
\end{subfigure}

\begin{subfigure}[h]{\linewidth}
    \includegraphics[width=\linewidth]{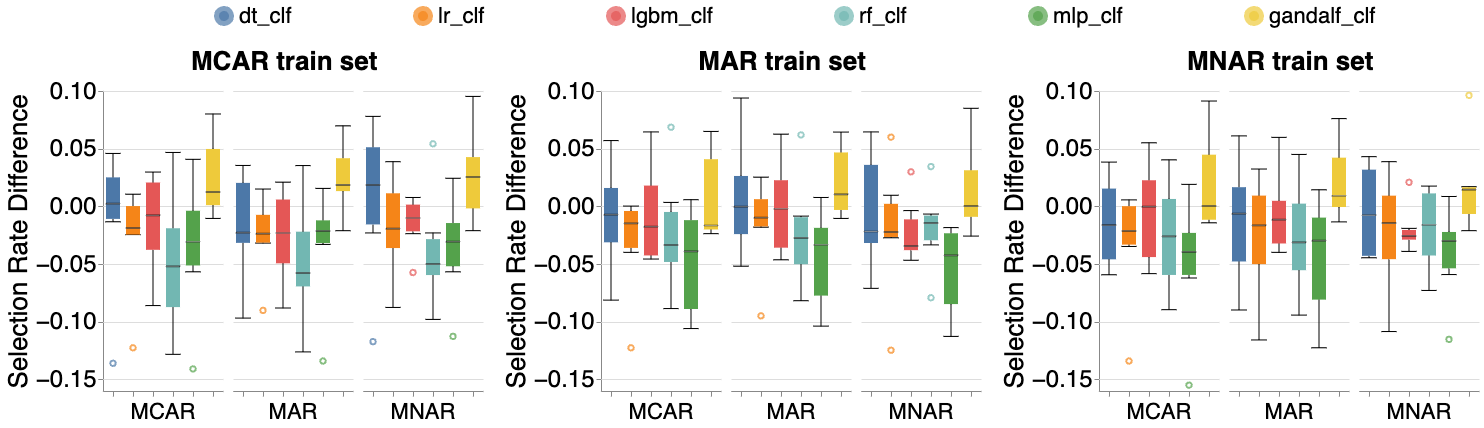}
    \caption{Datawig}
\end{subfigure}

\begin{subfigure}[h]{\linewidth}
    \includegraphics[width=\linewidth]{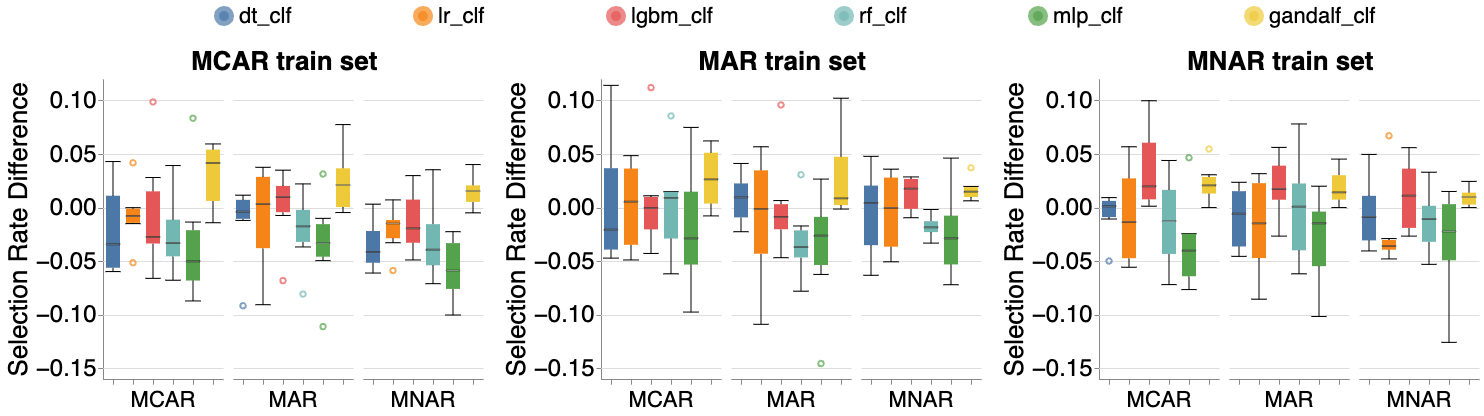}
    \caption{K-Means Clustering}
\end{subfigure}

\begin{subfigure}[h]{\linewidth}
    \includegraphics[width=\linewidth]{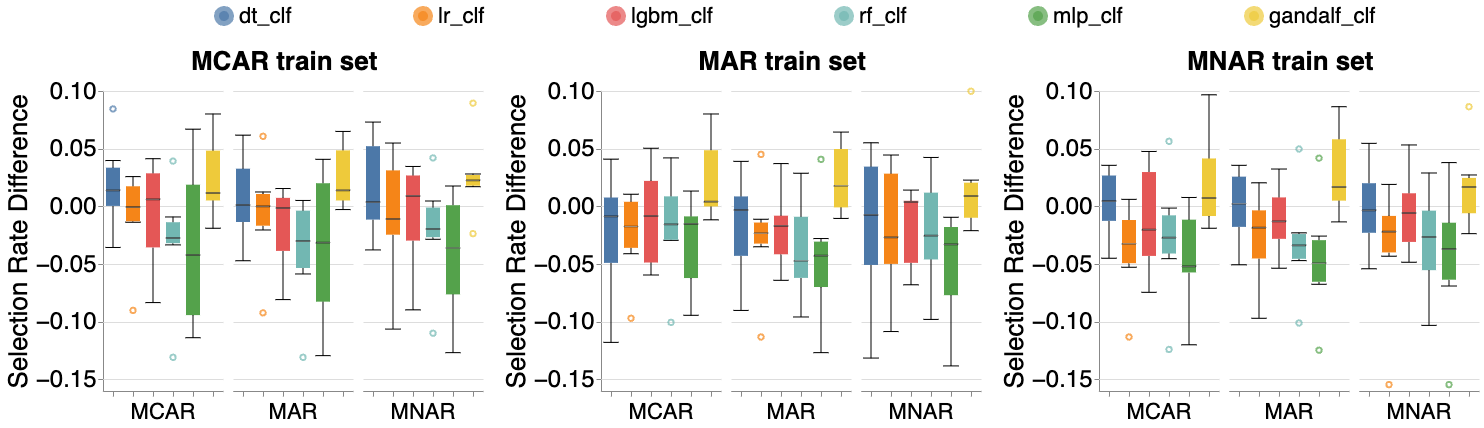}
    \caption{MissForest}
\end{subfigure}

\vspace{-0.3cm}
\caption{Selection Rate Difference of different models (colors in legend) on the \german dataset for ML and DL-based \mvi techniques (subplots) under missingness shift.}
\label{fig:missingness_shift_spd_german}
\end{figure}


\begin{figure*}[h!]
\begin{subfigure}[h]{0.475\linewidth}
    \centering
    \includegraphics[width=\linewidth]{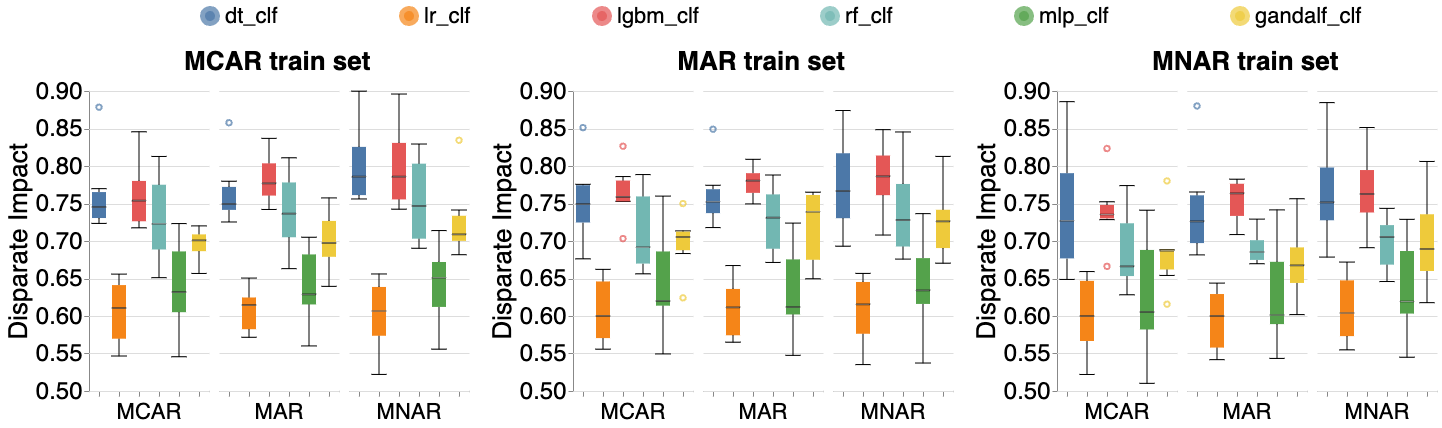}
    \caption{AutoML}
\end{subfigure}
\hfill
\begin{subfigure}[h]{0.475\linewidth}
    \includegraphics[width=\linewidth]{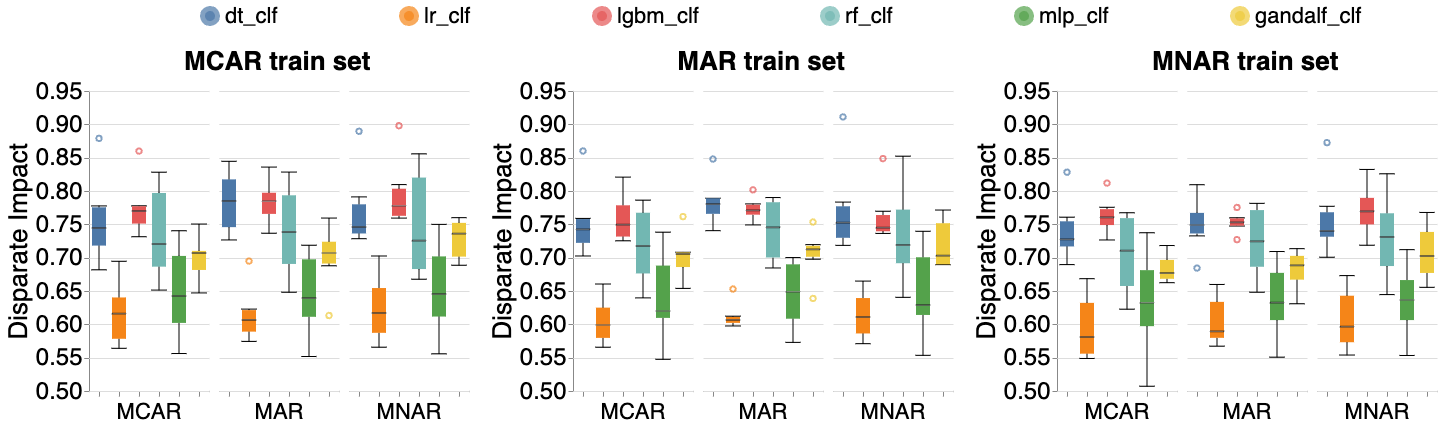}
    \caption{Datawig}
\end{subfigure}
\begin{subfigure}[h]{0.475\linewidth}
    \includegraphics[width=\linewidth]{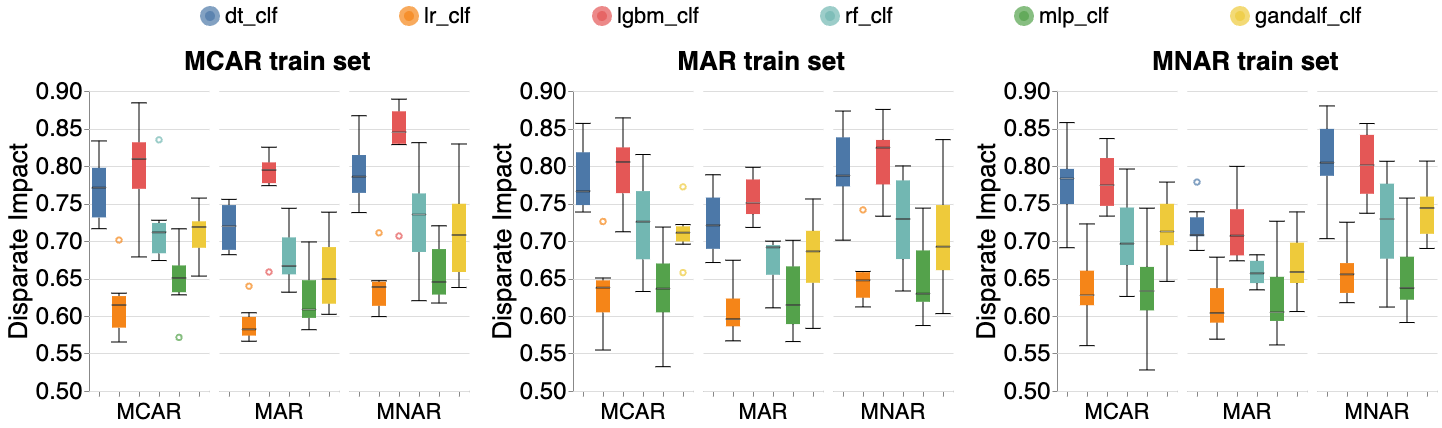}
    \caption{K-Means Clustering}
\end{subfigure}
\hfill
\begin{subfigure}[h]{0.475\linewidth}
    \includegraphics[width=\linewidth]{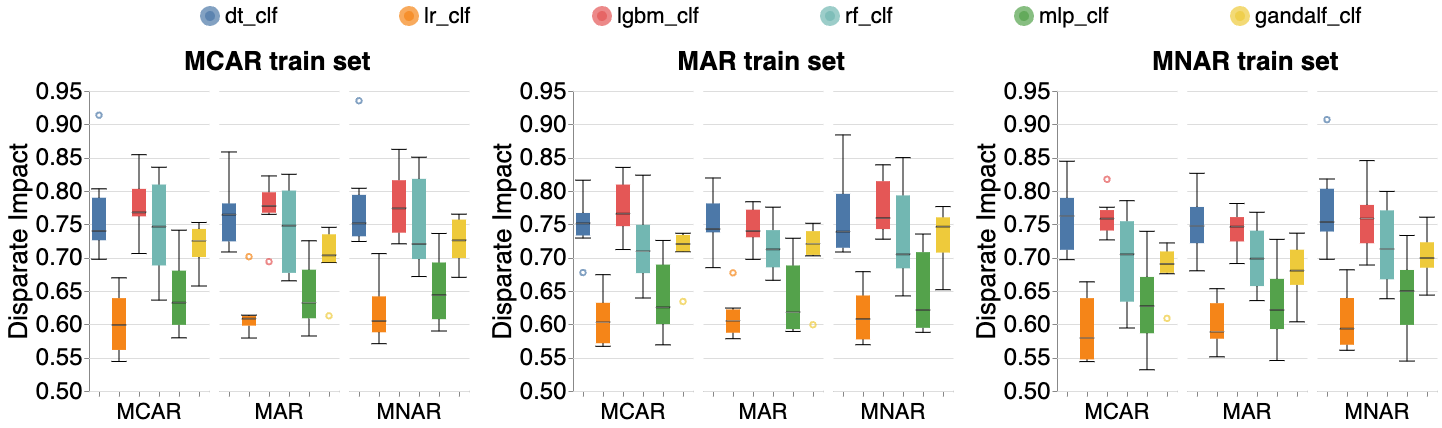}
    \caption{MissForest}
\end{subfigure}
 
\vspace{-0.3cm}
\caption{Disparate Impact of different models (colors in legend) on the \folk dataset for ML and DL-based \mvi techniques (subplots) under missingness shift.}
\label{fig:missingness_shift_di_folk}
\end{figure*}

\begin{figure}[h!]
\begin{subfigure}[h]{\linewidth}
    \includegraphics[width=\linewidth]{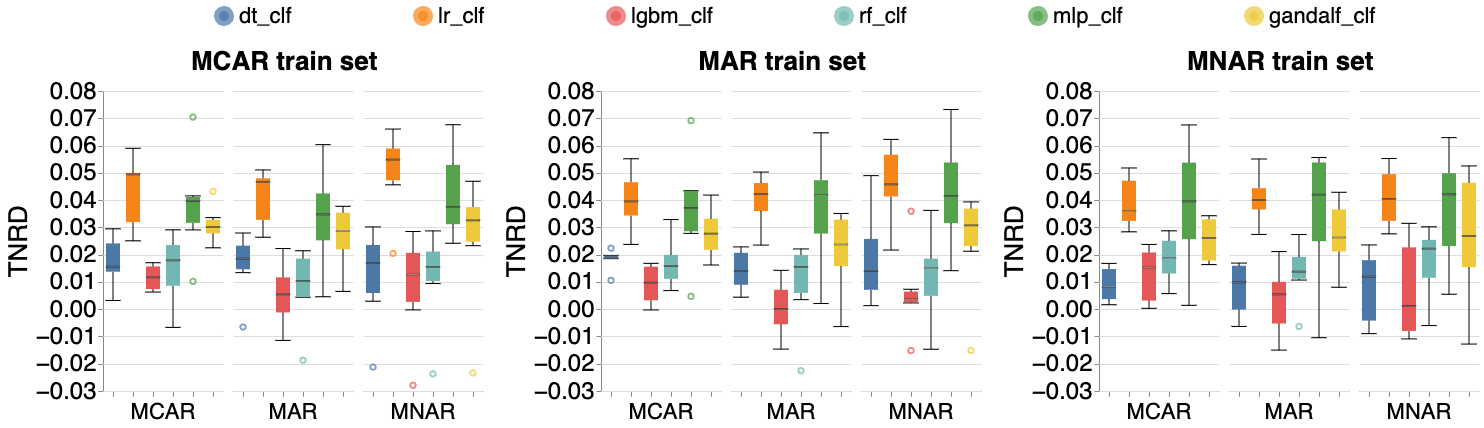}
    \caption{AutoML}
\end{subfigure}

\begin{subfigure}[h]{\linewidth}
    \includegraphics[width=\linewidth]{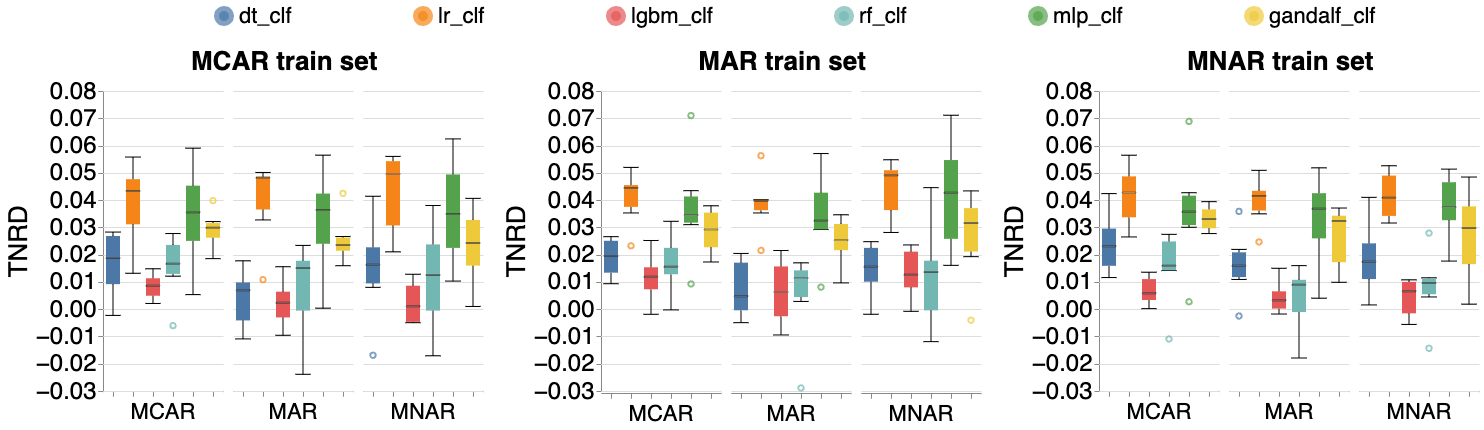}
    \caption{Datawig}
\end{subfigure}

\begin{subfigure}[h]{\linewidth}
    \includegraphics[width=\linewidth]{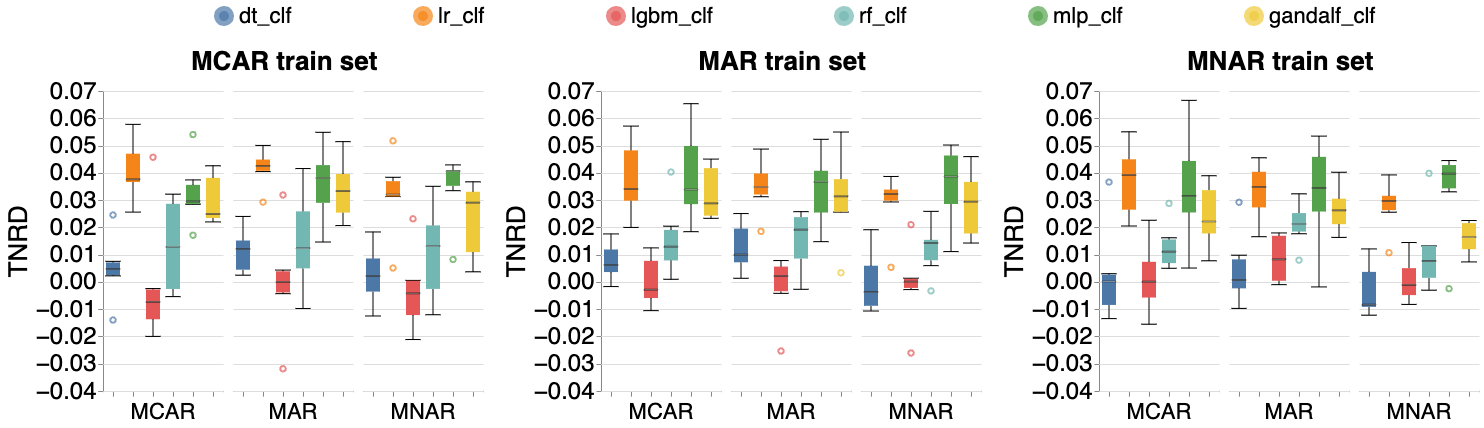}
    \caption{K-Means Clustering}
\end{subfigure}

\begin{subfigure}[h]{\linewidth}
    \includegraphics[width=\linewidth]{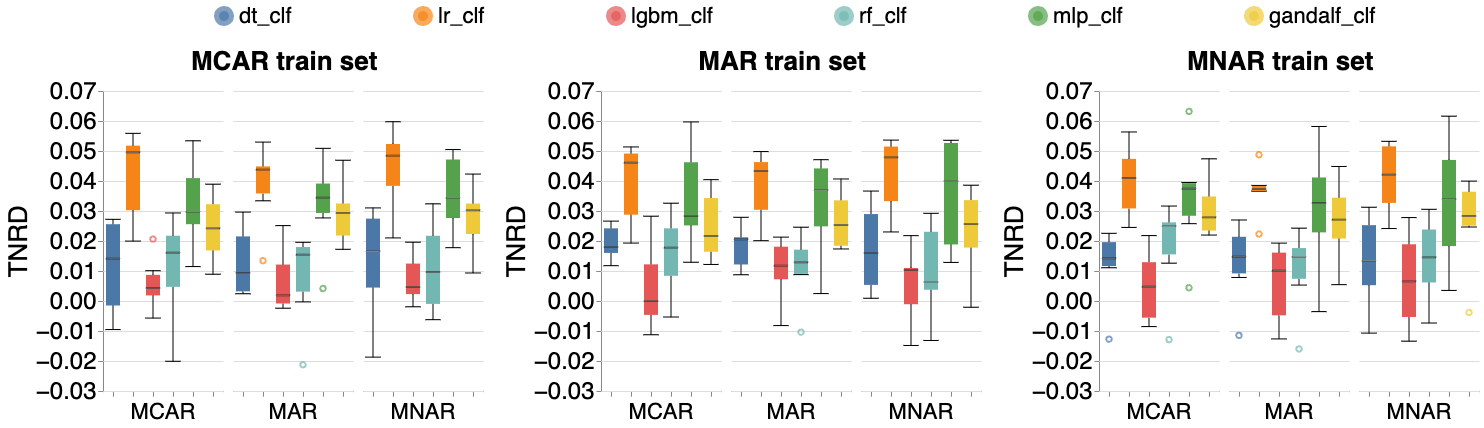}
    \caption{MissForest}
\end{subfigure}

\vspace{-0.3cm}
\caption{True Negative Rate Difference of different models (colors in legend) on the \folk dataset for ML and DL-based \mvi techniques (subplots) under missingness shift.}
\label{fig:missingness_shift_eq_odds_tnr_folk}
\end{figure}

\begin{figure}[h!]
\begin{subfigure}[h]{\linewidth}
    \includegraphics[width=\linewidth]{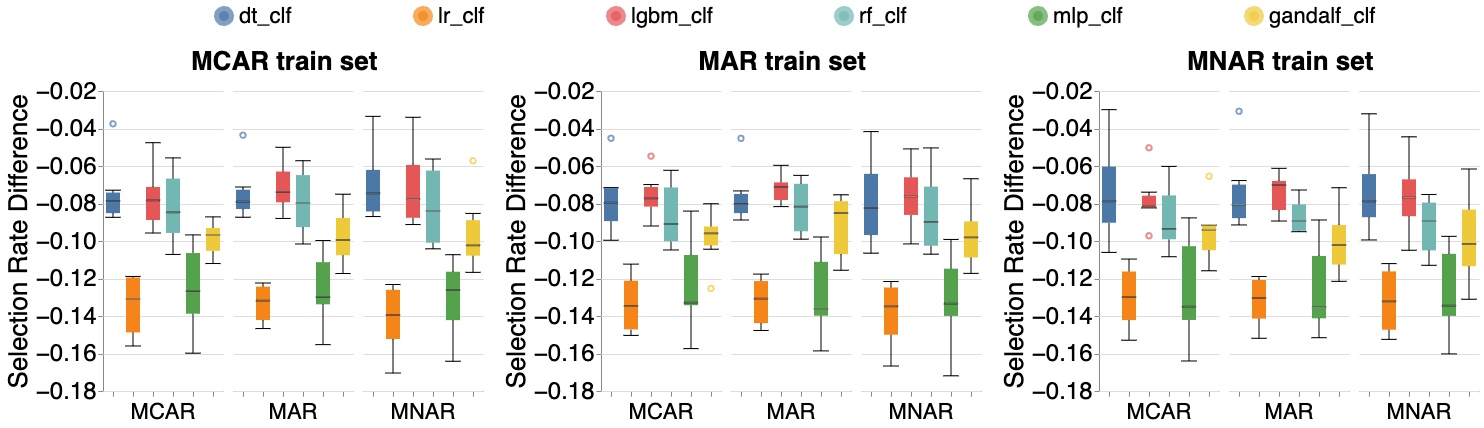}
    \caption{AutoML}
\end{subfigure}

\begin{subfigure}[h]{\linewidth}
    \includegraphics[width=\linewidth]{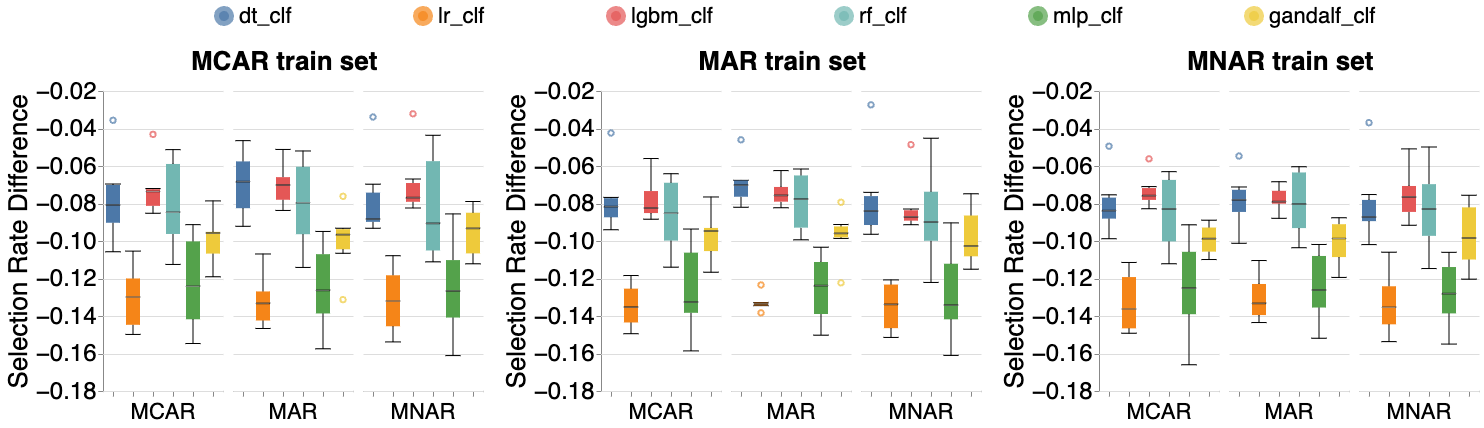}
    \caption{Datawig}
\end{subfigure}

\begin{subfigure}[h]{\linewidth}
    \includegraphics[width=\linewidth]{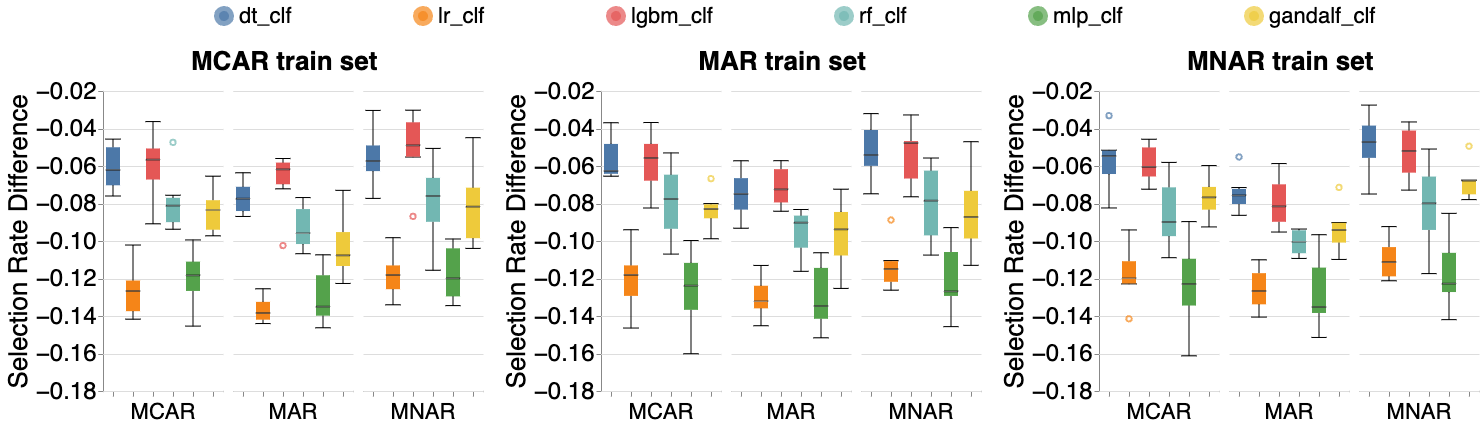}
    \caption{K-Means Clustering}
\end{subfigure}

\begin{subfigure}[h]{\linewidth}
    \includegraphics[width=\linewidth]{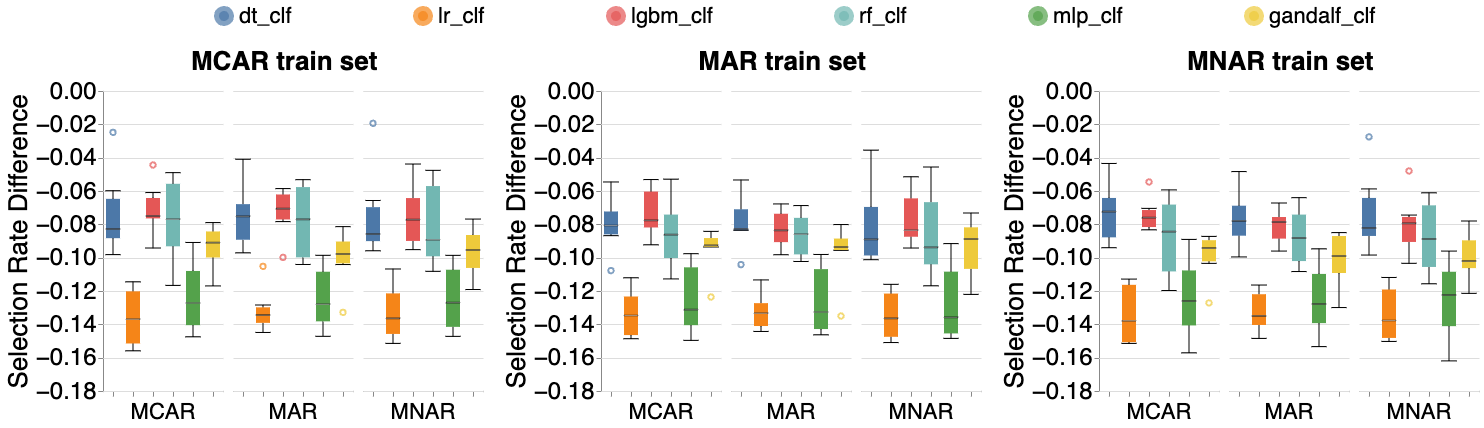}
    \caption{MissForest}
\end{subfigure}

\vspace{-0.3cm}
\caption{Selection Rate Difference of different models (colors in legend) on the \folk dataset for ML and DL-based \mvi techniques (subplots) under missingness shift.}
\label{fig:missingness_shift_spd_folk}
\end{figure}


\begin{figure*}[h!]
\begin{subfigure}[h]{0.475\linewidth}
    \centering
    \includegraphics[width=\linewidth]{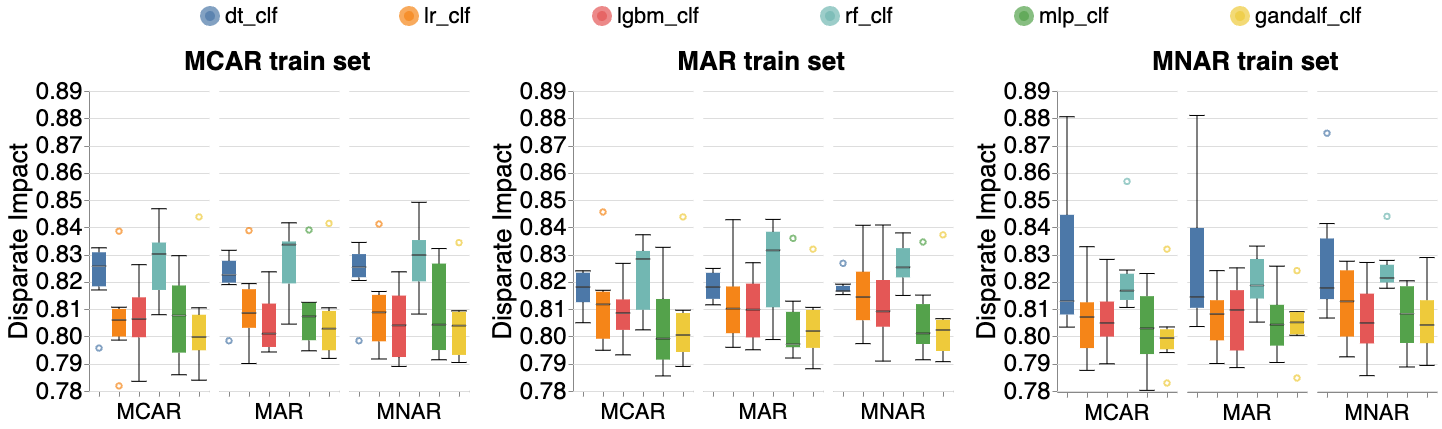}
    \caption{AutoML}
\end{subfigure}
\hfill
\begin{subfigure}[h]{0.475\linewidth}
    \includegraphics[width=\linewidth]{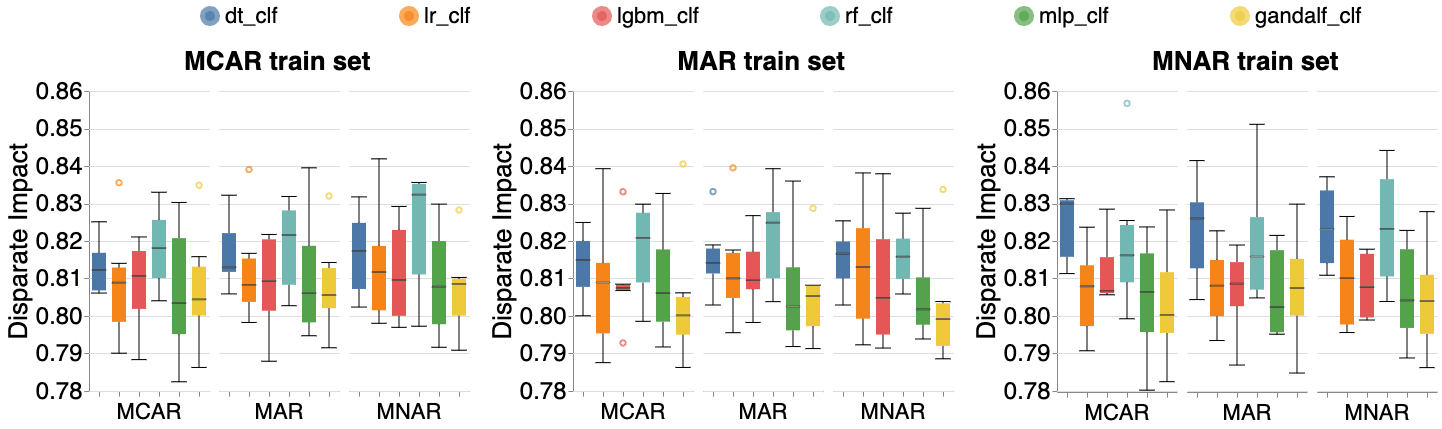}
    \caption{Datawig}
\end{subfigure}
\begin{subfigure}[h]{0.475\linewidth}
    \includegraphics[width=\linewidth]{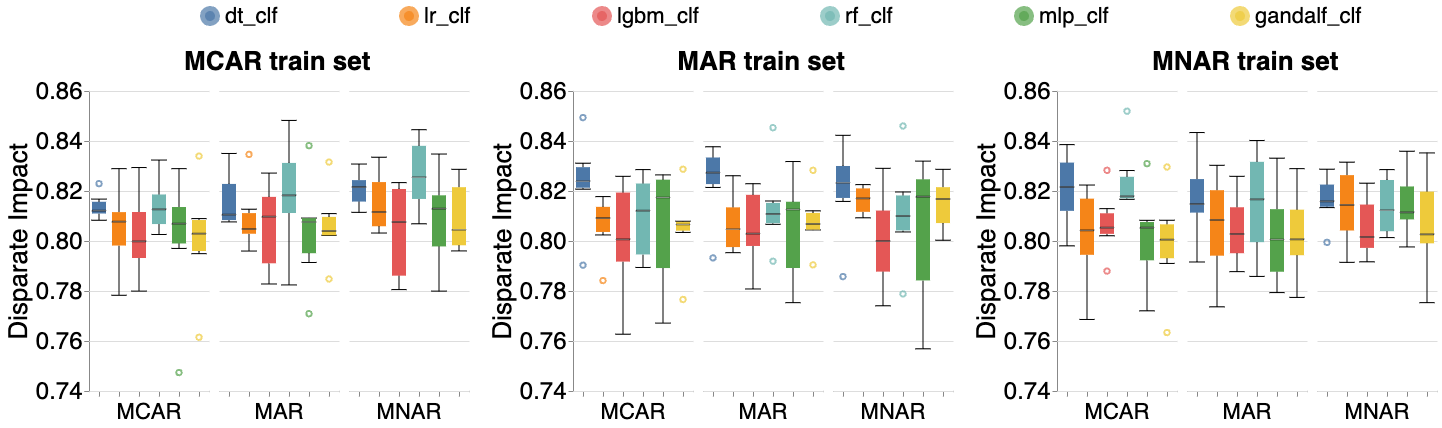}
    \caption{K-Means Clustering}
\end{subfigure}
\hfill
\begin{subfigure}[h]{0.475\linewidth}
    \includegraphics[width=\linewidth]{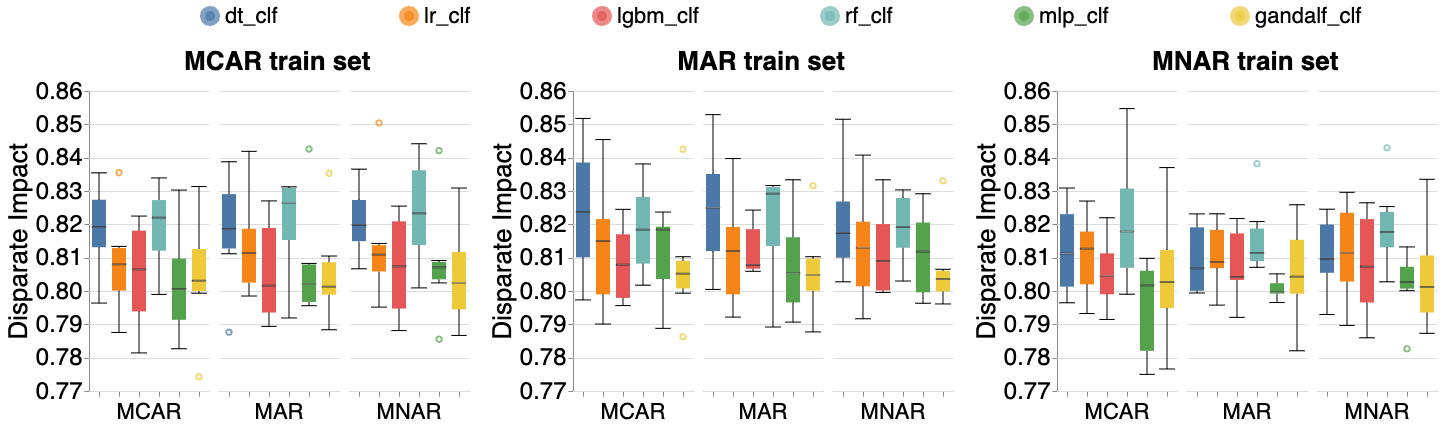}
    \caption{MissForest}
\end{subfigure}

\vspace{-0.3cm}
\caption{Disparate Impact of different models (colors in legend) on the \law dataset for ML and DL-based \mvi techniques (subplots) under missingness shift.}
\label{fig:missingness_shift_di_law}
\end{figure*}

\begin{figure}[h!]
\begin{subfigure}[h]{\linewidth}
    \includegraphics[width=\linewidth]{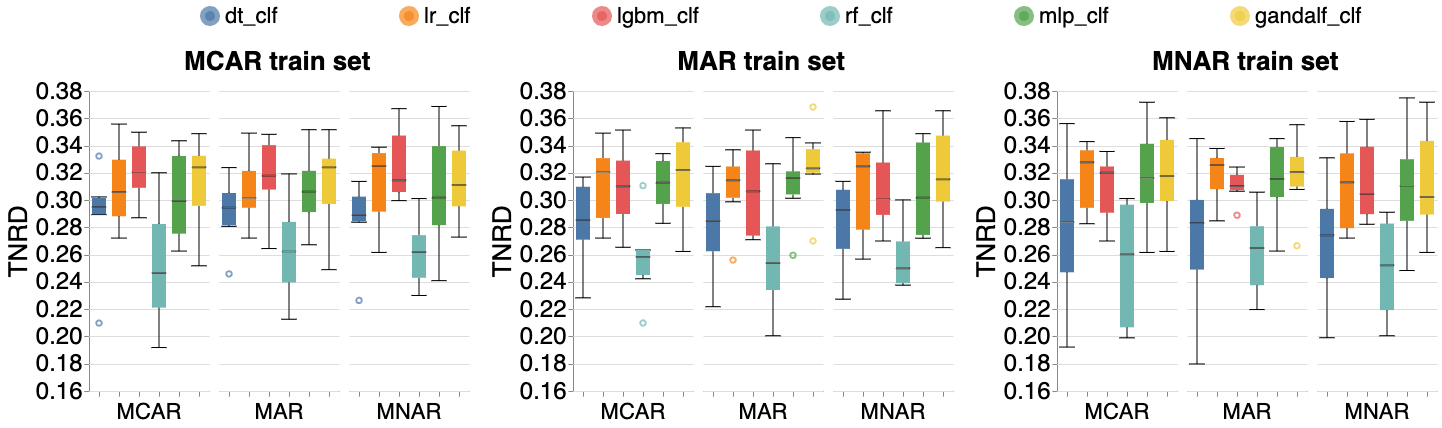}
    \caption{AutoML}
\end{subfigure}

\begin{subfigure}[h]{\linewidth}
    \includegraphics[width=\linewidth]{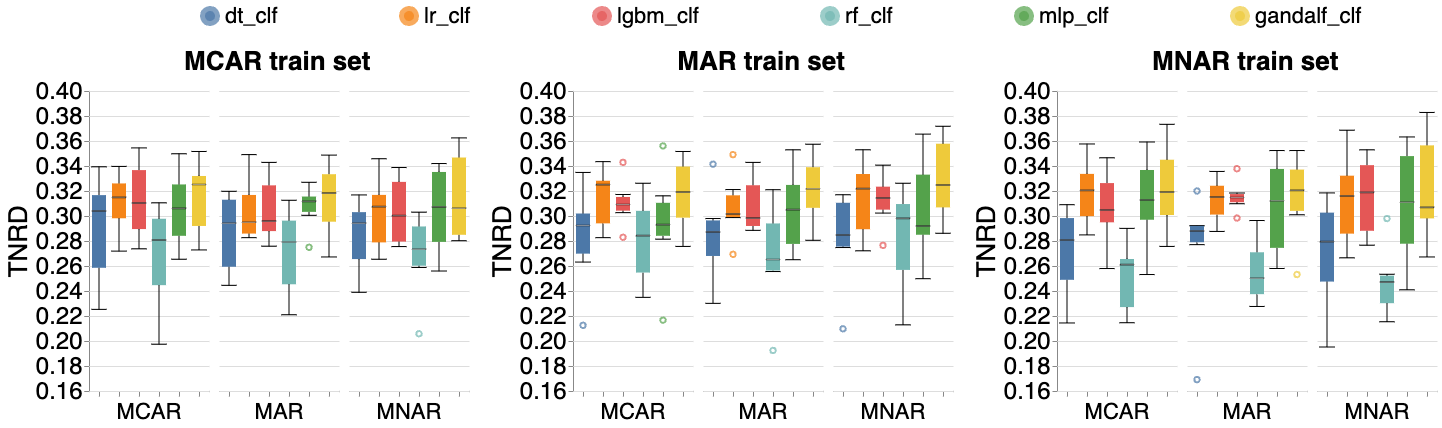}
    \caption{Datawig}
\end{subfigure}

\begin{subfigure}[h]{\linewidth}
    \includegraphics[width=\linewidth]{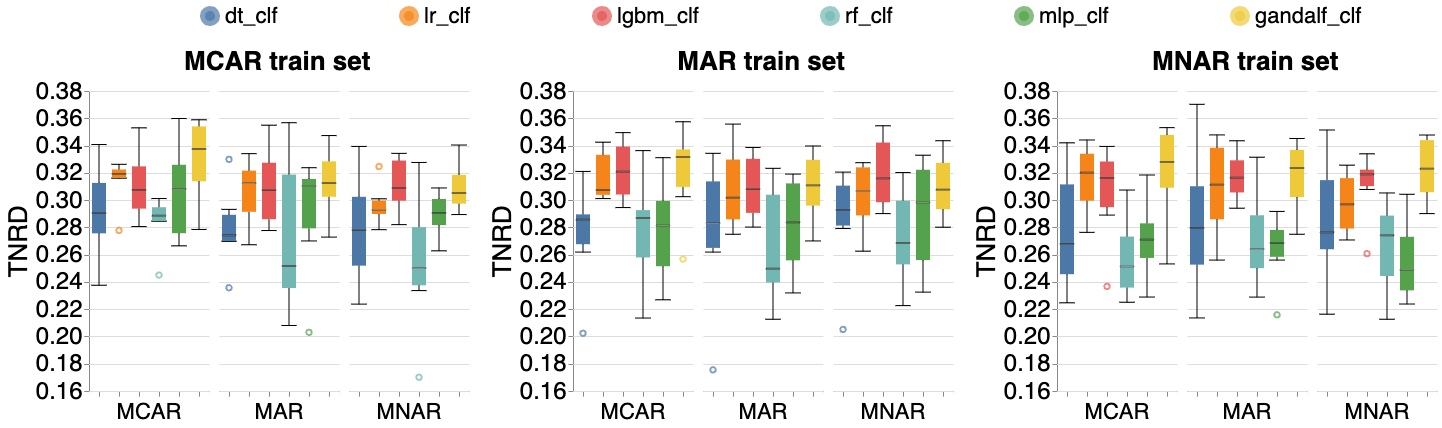}
    \caption{K-Means Clustering}
\end{subfigure}

\begin{subfigure}[h]{\linewidth}
    \includegraphics[width=\linewidth]{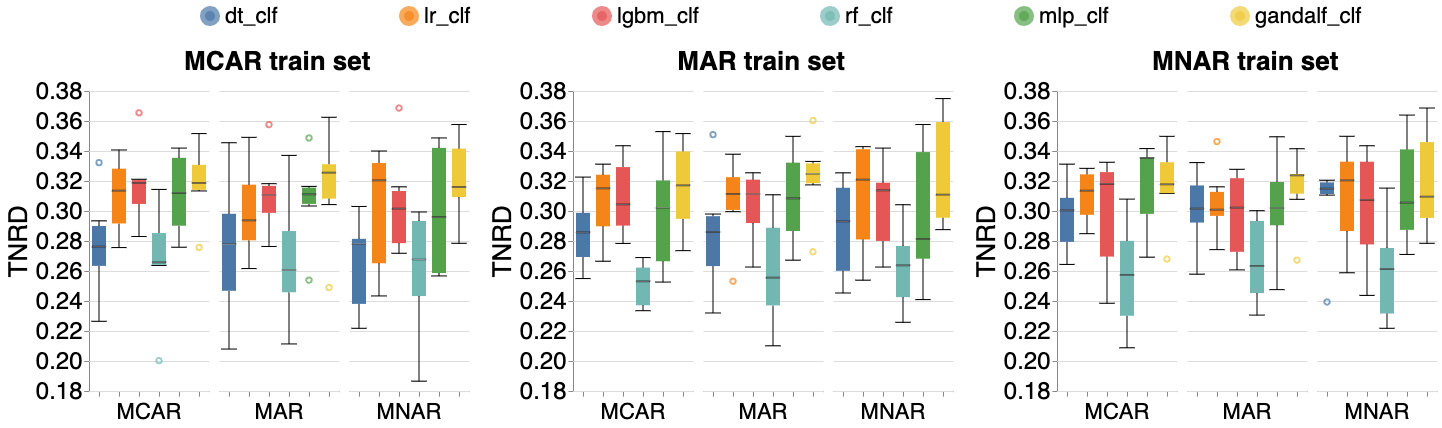}
    \caption{MissForest}
\end{subfigure}

\vspace{-0.3cm}
\caption{True Negative Rate Difference of different models (colors in legend) on the \law dataset for ML and DL-based \mvi techniques (subplots) under missingness shift.}
\label{fig:missingness_shift_eq_odds_tnr_law}
\end{figure}

\begin{figure}[h!]
\begin{subfigure}[h]{\linewidth}
    \includegraphics[width=\linewidth]{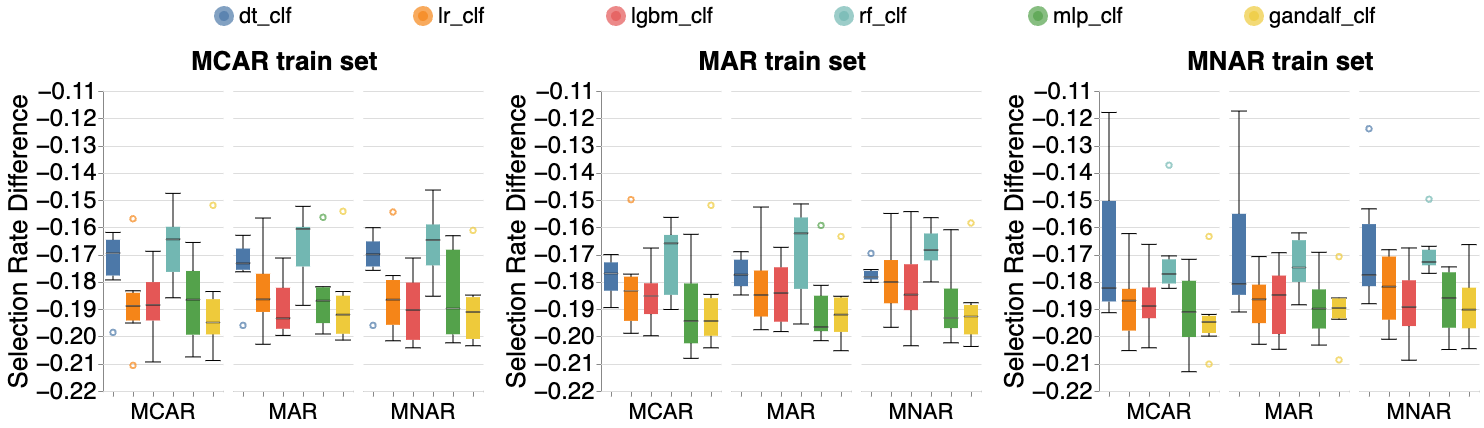}
    \caption{AutoML}
\end{subfigure}

\begin{subfigure}[h]{\linewidth}
    \includegraphics[width=\linewidth]{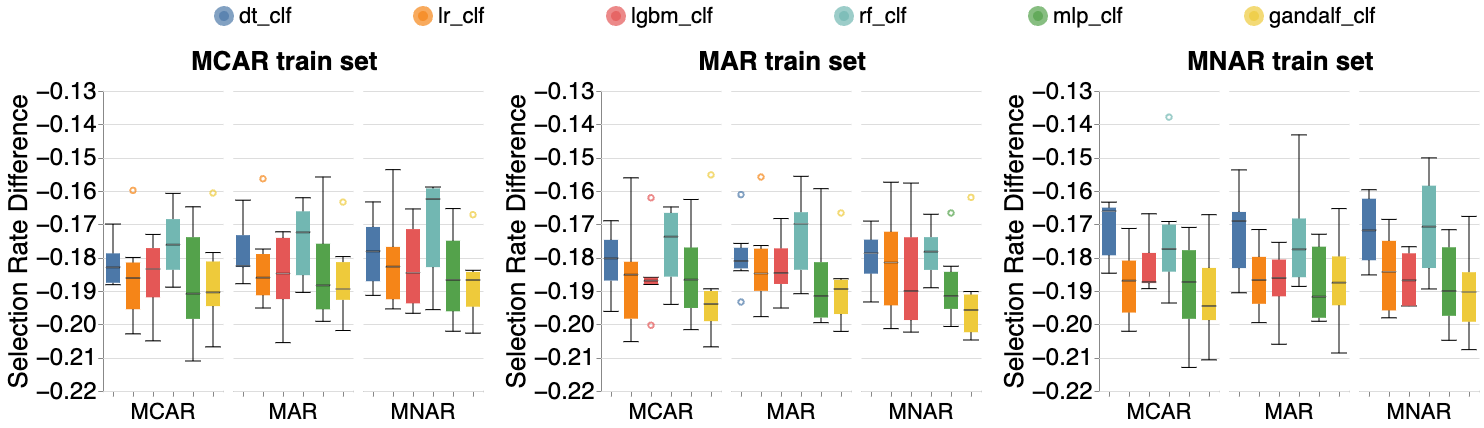}
    \caption{Datawig}
\end{subfigure}

\begin{subfigure}[h]{\linewidth}
    \includegraphics[width=\linewidth]{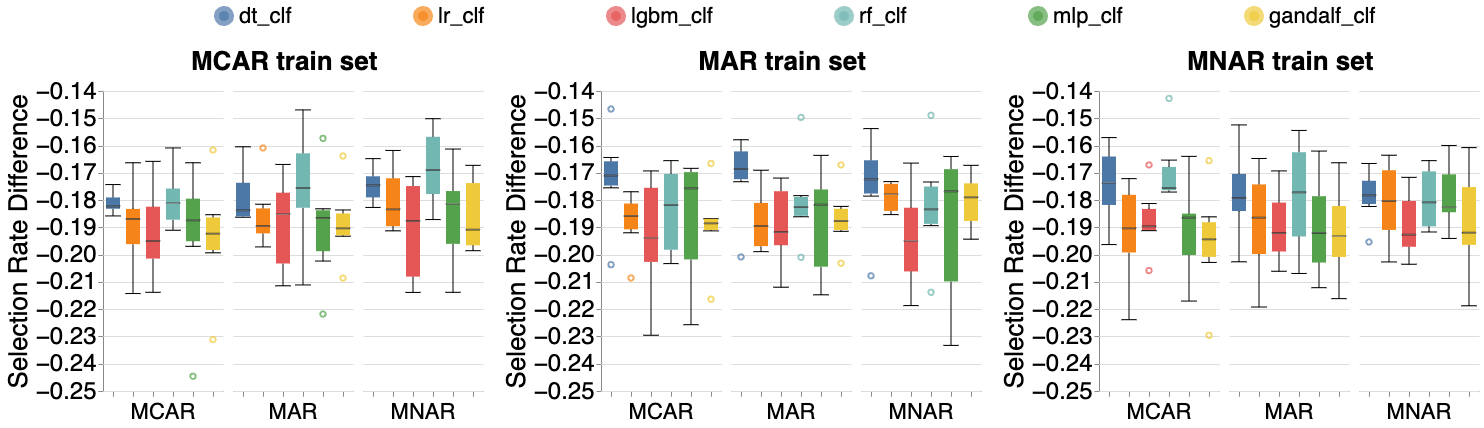}
    \caption{K-Means Clustering}
\end{subfigure}

\begin{subfigure}[h]{\linewidth}
    \includegraphics[width=\linewidth]{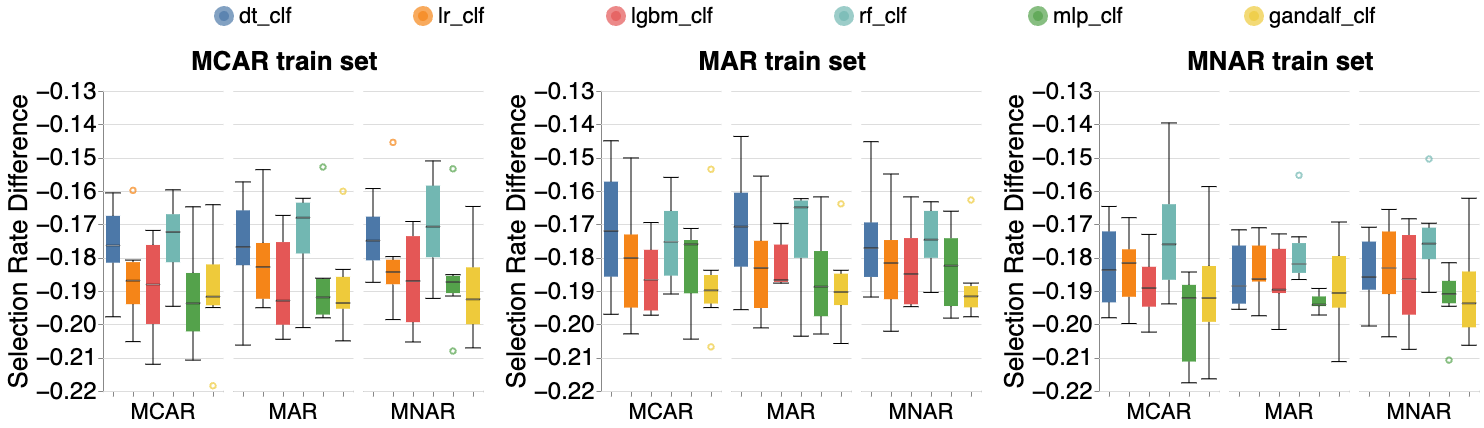}
    \caption{MissForest}
\end{subfigure}

\vspace{-0.3cm}
\caption{Selection Rate Difference of different models (colors in legend) on the \law dataset for ML and DL-based \mvi techniques (subplots) under missingness shift.}
\label{fig:missingness_shift_spd_law}
\end{figure}


\begin{figure*}[h!]
\begin{subfigure}[h]{0.475\linewidth}
    \centering
    \includegraphics[width=\linewidth]{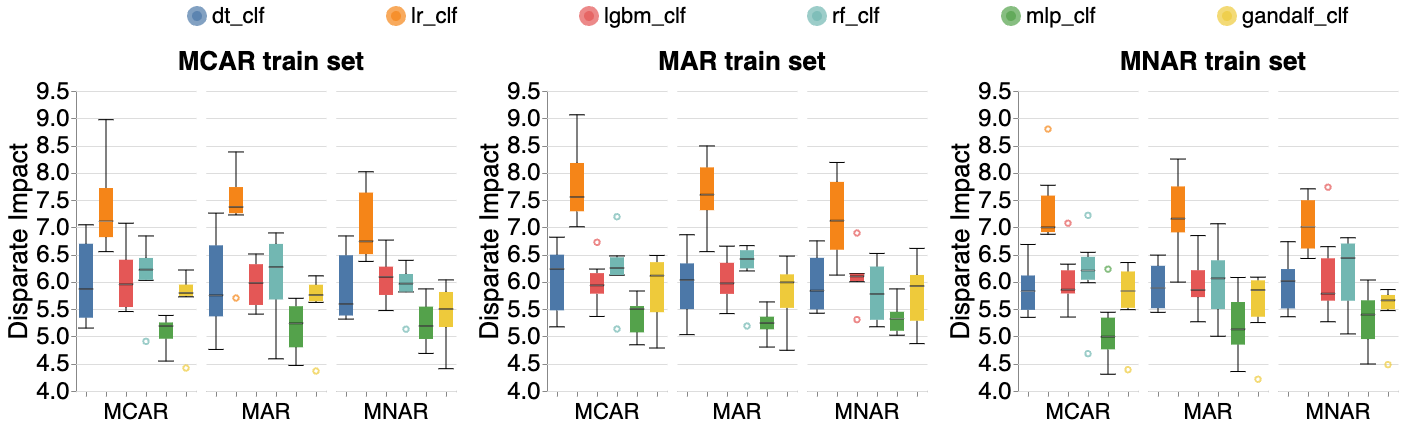}
    \caption{AutoML}
\end{subfigure}
\hfill
\begin{subfigure}[h]{0.475\linewidth}
    \includegraphics[width=\linewidth]{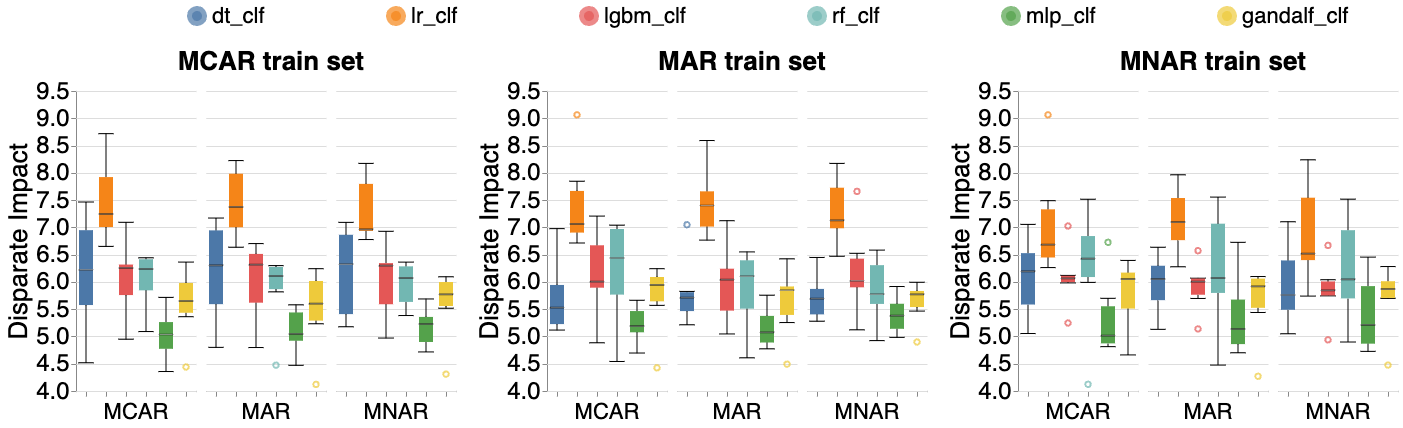}
    \caption{Datawig}
\end{subfigure}
\begin{subfigure}[h]{0.475\linewidth}
    \includegraphics[width=\linewidth]{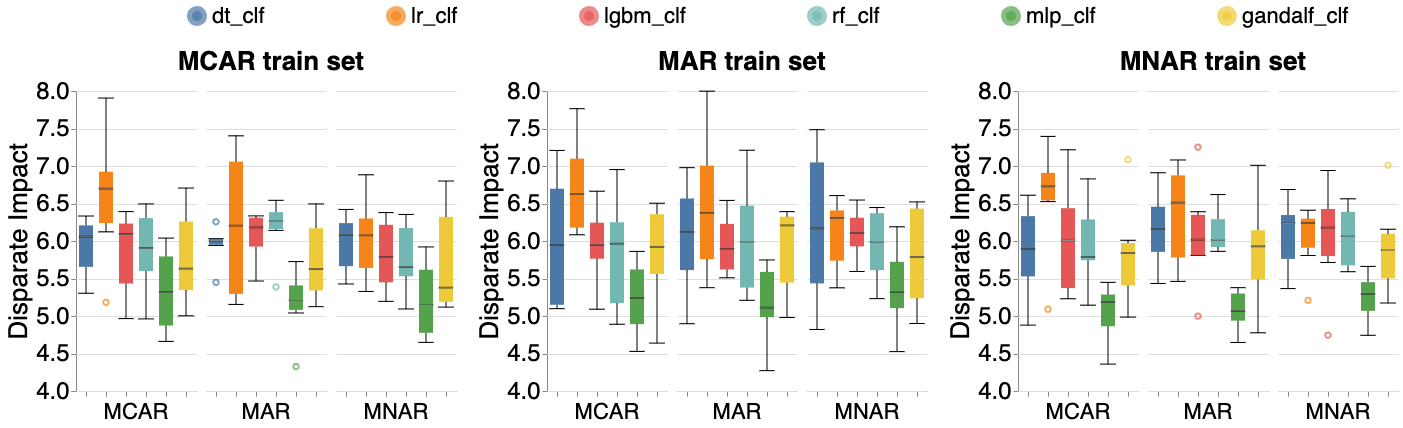}
    \caption{K-Means Clustering}
\end{subfigure}
\hfill
\begin{subfigure}[h]{0.475\linewidth}
    \includegraphics[width=\linewidth]{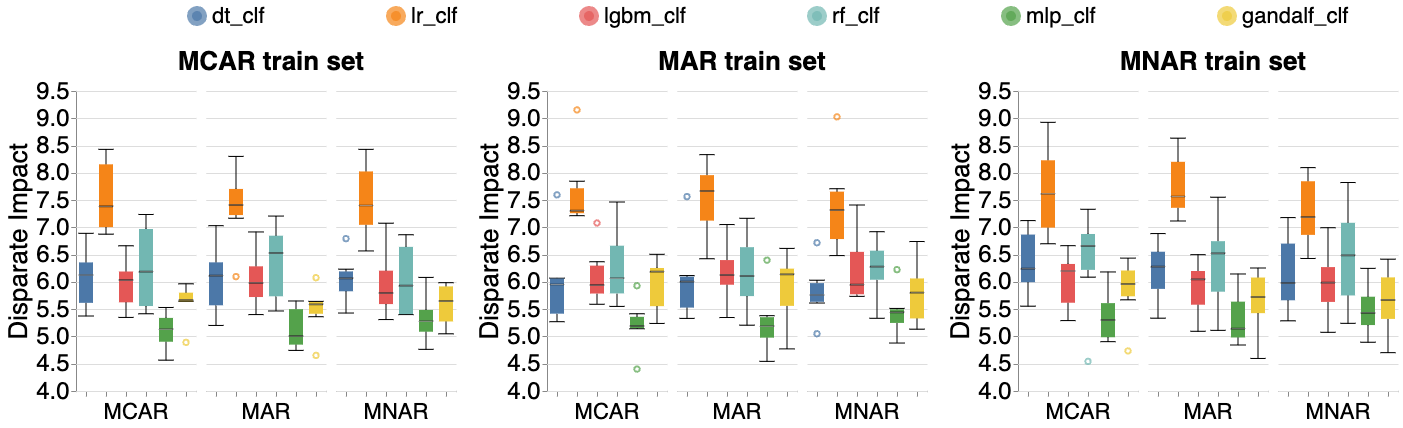}
    \caption{MissForest}
\end{subfigure}

\vspace{-0.3cm}
\caption{Disparate Impact of different models (colors in legend) on the \bank dataset for ML and DL-based \mvi techniques (subplots) under missingness shift.}
\label{fig:missingness_shift_di_bank}
\end{figure*}

\begin{figure}[h!]
\begin{subfigure}[h]{\linewidth}
    \includegraphics[width=\linewidth]{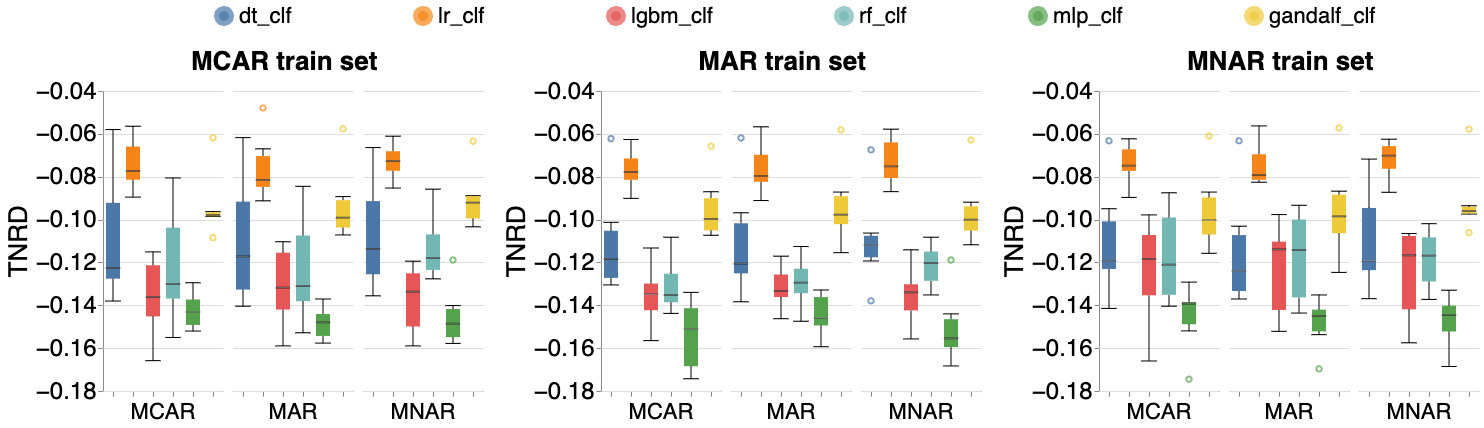}
    \caption{AutoML}
\end{subfigure}

\begin{subfigure}[h]{\linewidth}
    \includegraphics[width=\linewidth]{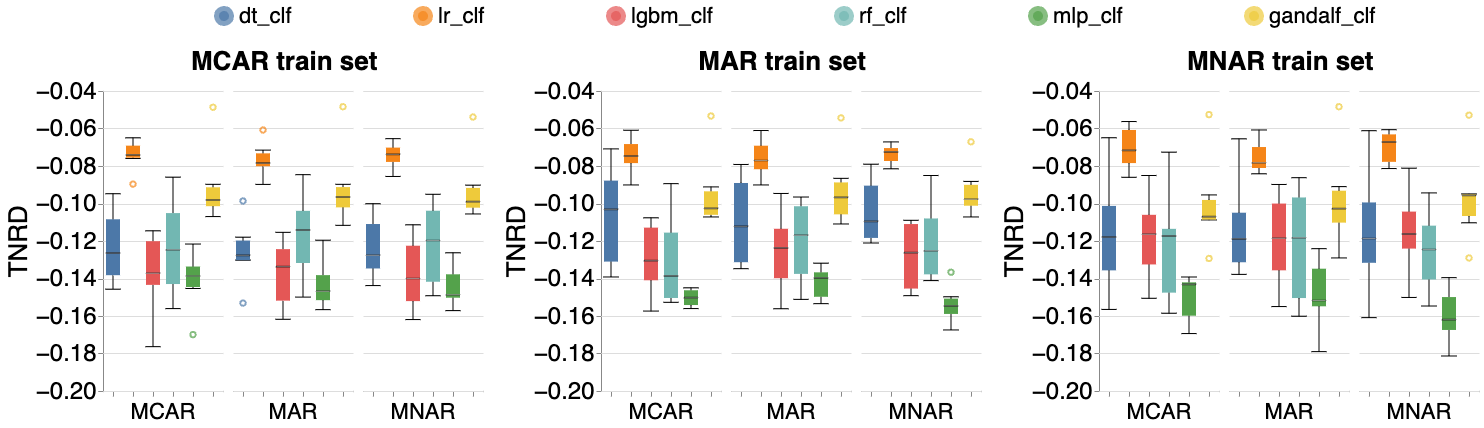}
    \caption{Datawig}
\end{subfigure}

\begin{subfigure}[h]{\linewidth}
    \includegraphics[width=\linewidth]{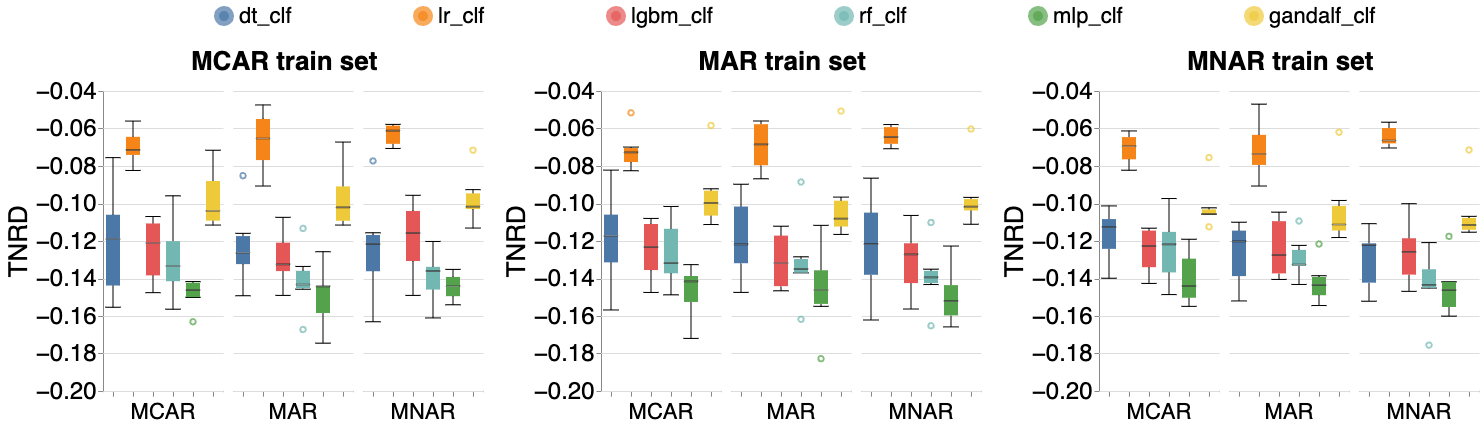}
    \caption{K-Means Clustering}
\end{subfigure}

\begin{subfigure}[h]{\linewidth}
    \includegraphics[width=\linewidth]{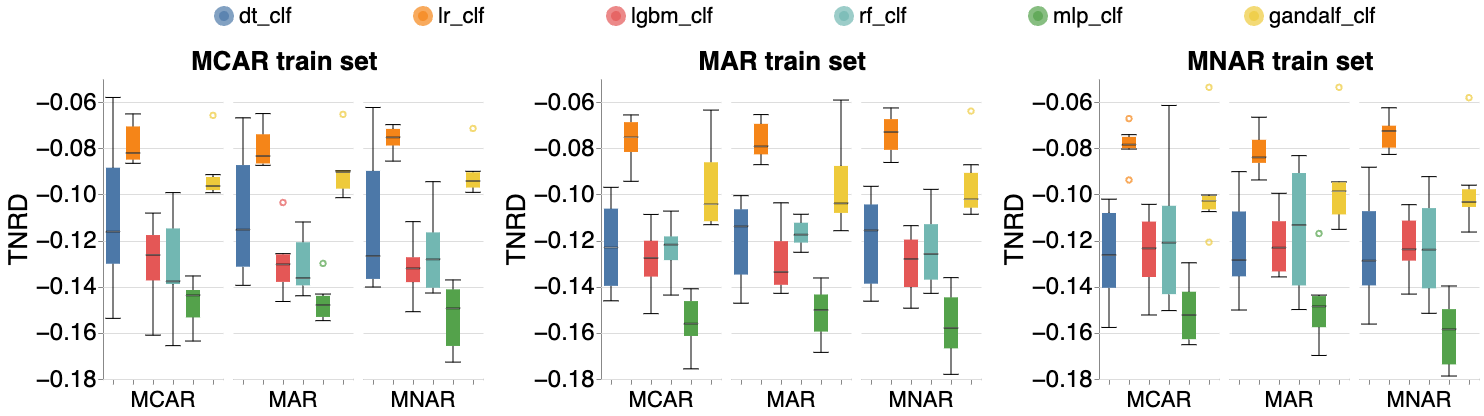}
    \caption{MissForest}
\end{subfigure}

\vspace{-0.3cm}
\caption{True Negative Rate Difference of different models (colors in legend) on the \bank dataset for ML and DL-based \mvi techniques (subplots) under missingness shift.}
\label{fig:missingness_shift_eq_odds_tnr_bank}
\end{figure}

\begin{figure}[h!]
\begin{subfigure}[h]{\linewidth}
    \includegraphics[width=\linewidth]{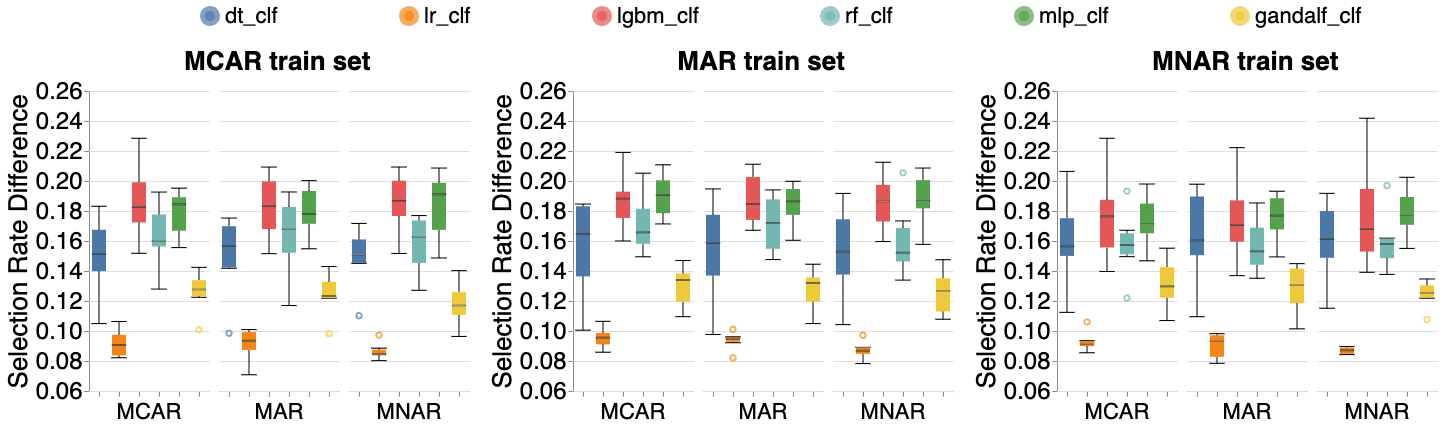}
    \caption{AutoML}
\end{subfigure}

\begin{subfigure}[h]{\linewidth}
    \includegraphics[width=\linewidth]{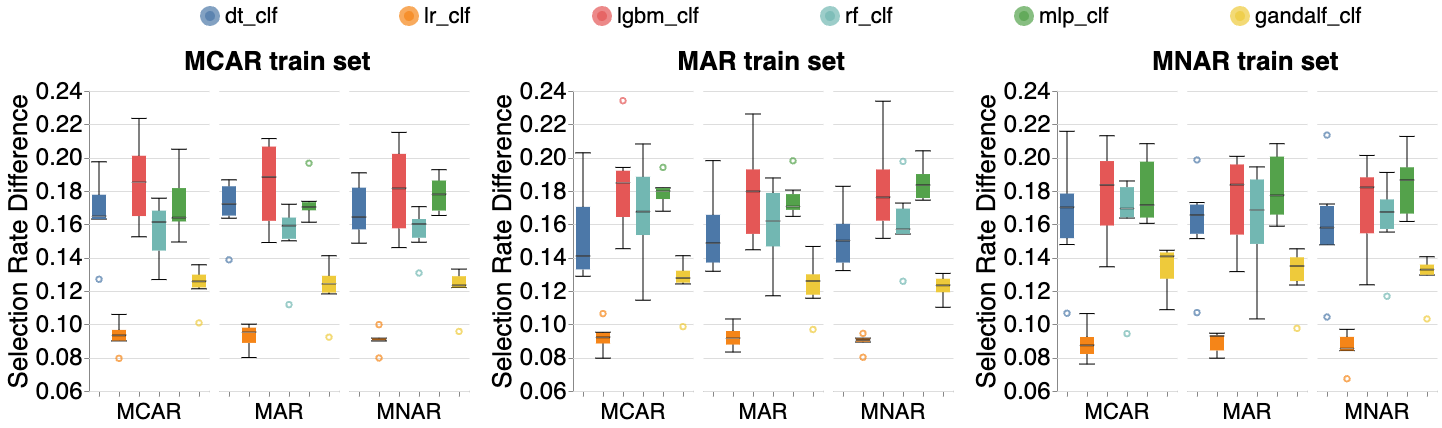}
    \caption{Datawig}
\end{subfigure}

\begin{subfigure}[h]{\linewidth}
    \includegraphics[width=\linewidth]{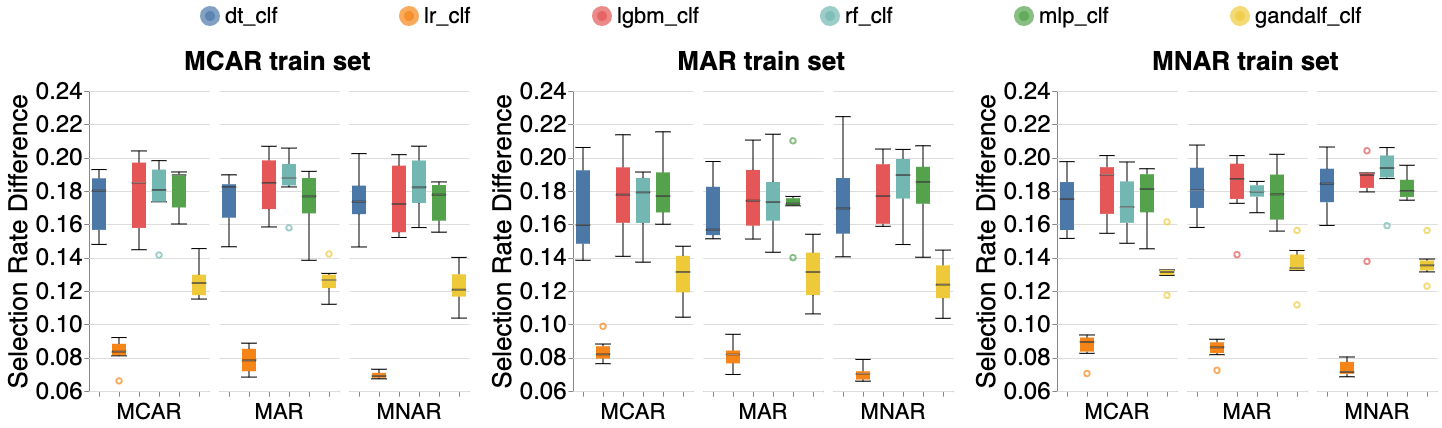}
    \caption{K-Means Clustering}
\end{subfigure}

\begin{subfigure}[h]{\linewidth}
    \includegraphics[width=\linewidth]{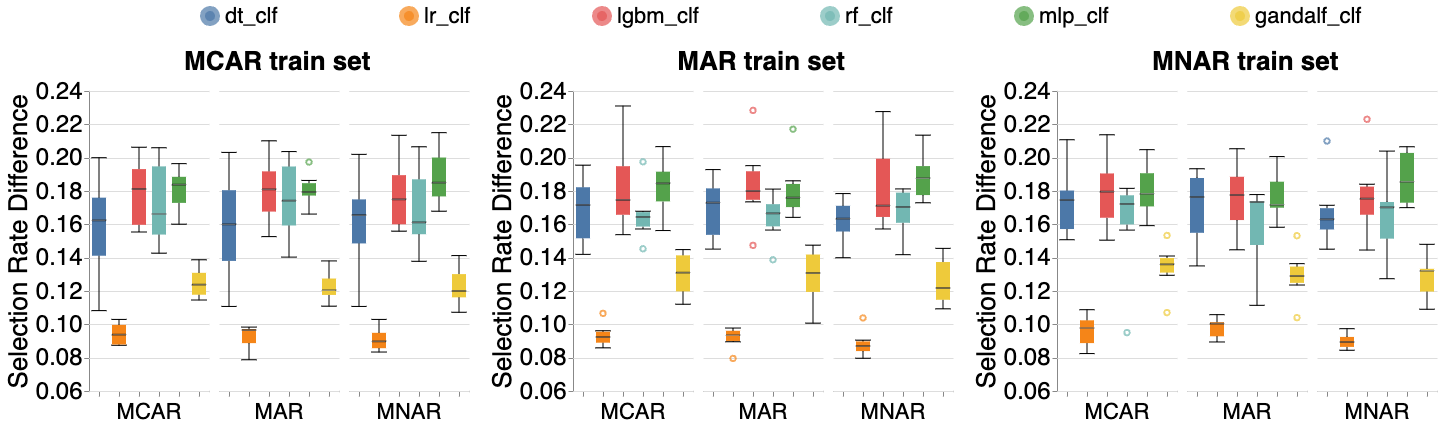}
    \caption{MissForest}
\end{subfigure}

\vspace{-0.3cm}
\caption{Selection Rate Difference of different models (colors in legend) on the \bank dataset for ML and DL-based \mvi techniques (subplots) under missingness shift.}
\label{fig:missingness_shift_spd_bank}
\end{figure}


\begin{figure*}[h!]
\begin{subfigure}[h]{0.475\linewidth}
    \centering
    \includegraphics[width=\linewidth]{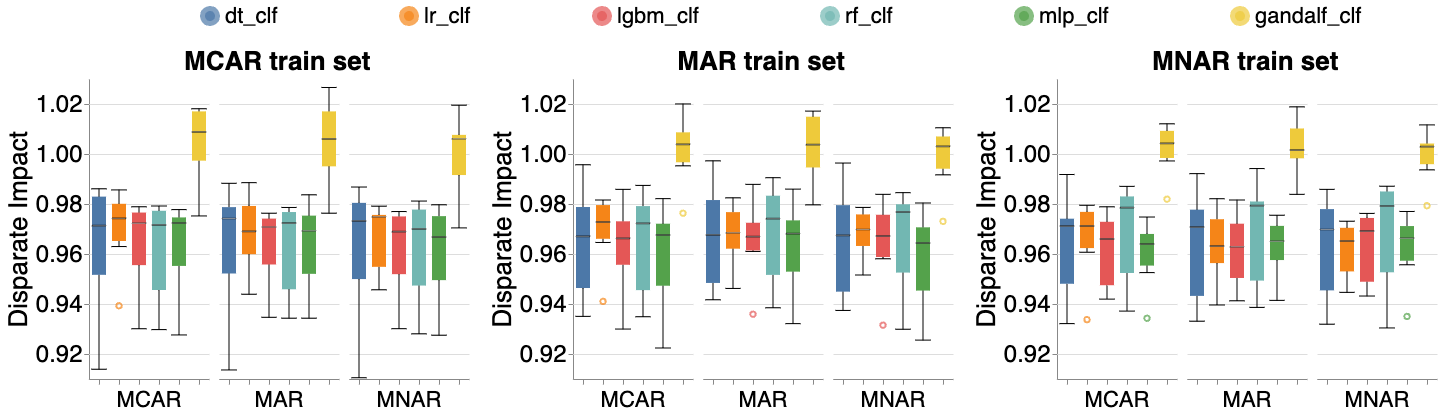}
    \caption{AutoML}
\end{subfigure}
\hfill
\begin{subfigure}[h]{0.475\linewidth}
    \includegraphics[width=\linewidth]{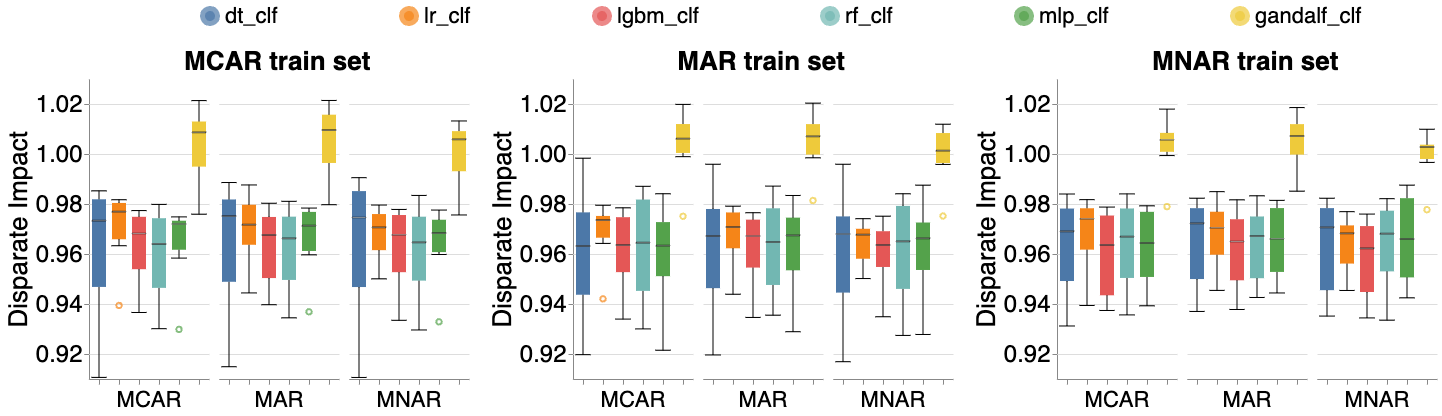}
    \caption{Datawig}
\end{subfigure}
\begin{subfigure}[h]{0.475\linewidth}
    \includegraphics[width=\linewidth]{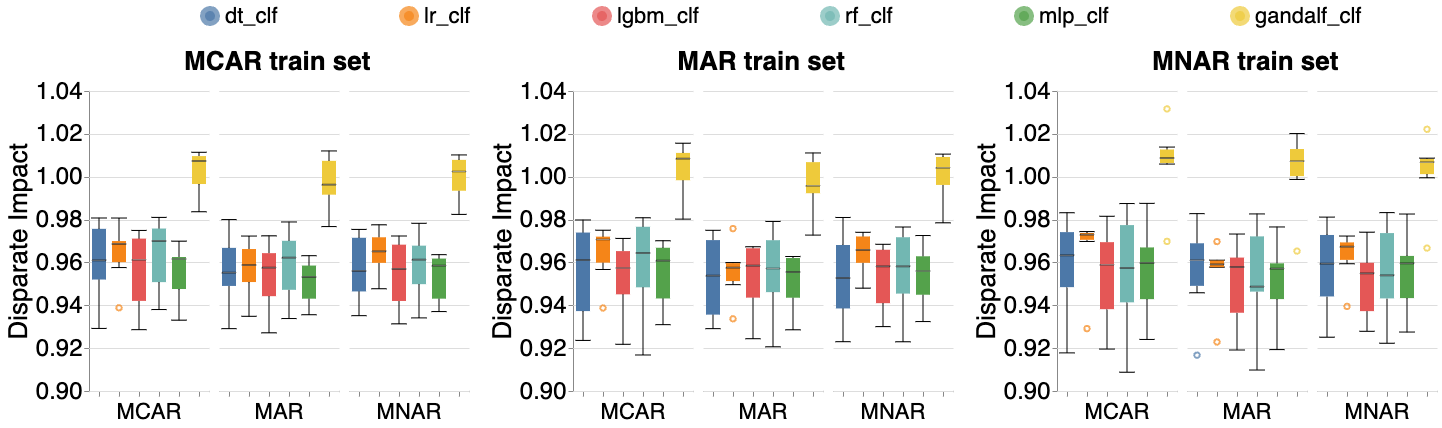}
    \caption{K-Means Clustering}
\end{subfigure}
\hfill
\begin{subfigure}[h]{0.475\linewidth}
    \includegraphics[width=\linewidth]{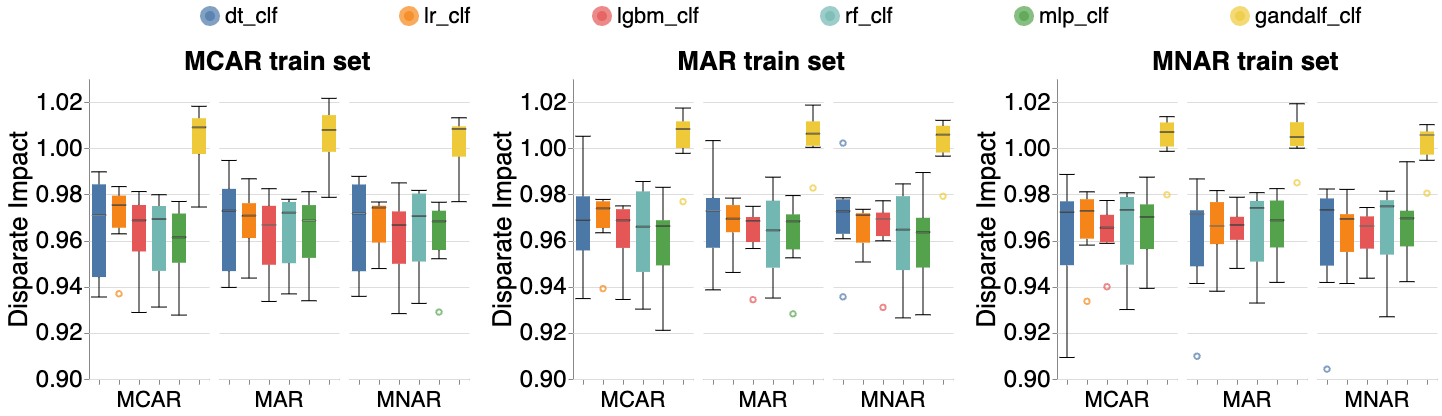}
    \caption{MissForest}
\end{subfigure}

\vspace{-0.3cm}
\caption{Disparate Impact of different models (colors in legend) on the \heart dataset for ML and DL-based \mvi techniques (subplots) under missingness shift.}
\label{fig:missingness_shift_di_heart}
\end{figure*}

\begin{figure}[h!]
\begin{subfigure}[h]{\linewidth}
    \includegraphics[width=\linewidth]{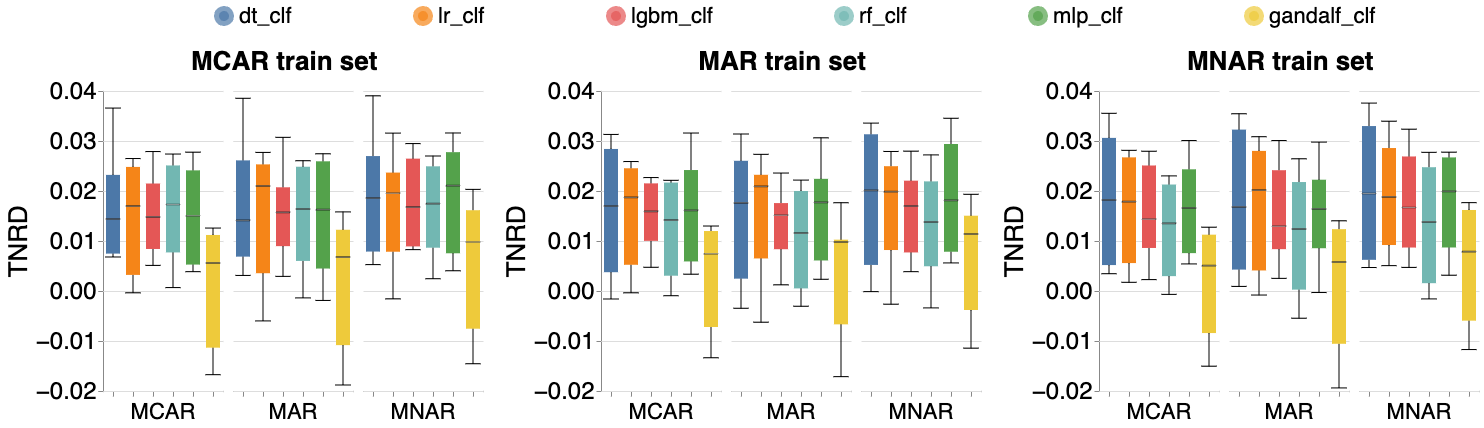}
    \caption{AutoML}
\end{subfigure}

\begin{subfigure}[h]{\linewidth}
    \includegraphics[width=\linewidth]{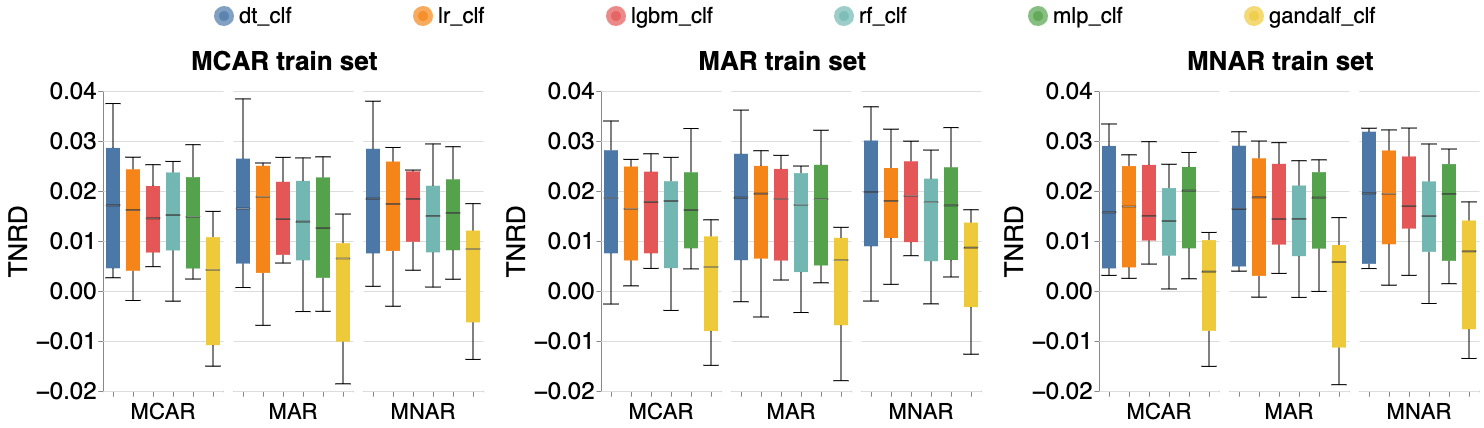}
    \caption{Datawig}
\end{subfigure}

\begin{subfigure}[h]{\linewidth}
    \includegraphics[width=\linewidth]{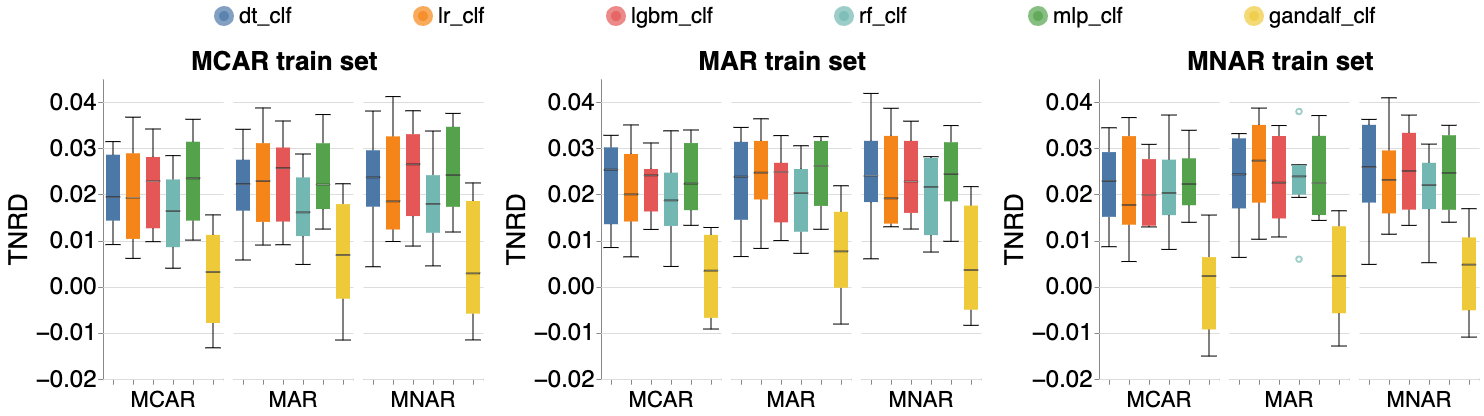}
    \caption{K-Means Clustering}
\end{subfigure}

\begin{subfigure}[h]{\linewidth}
    \includegraphics[width=\linewidth]{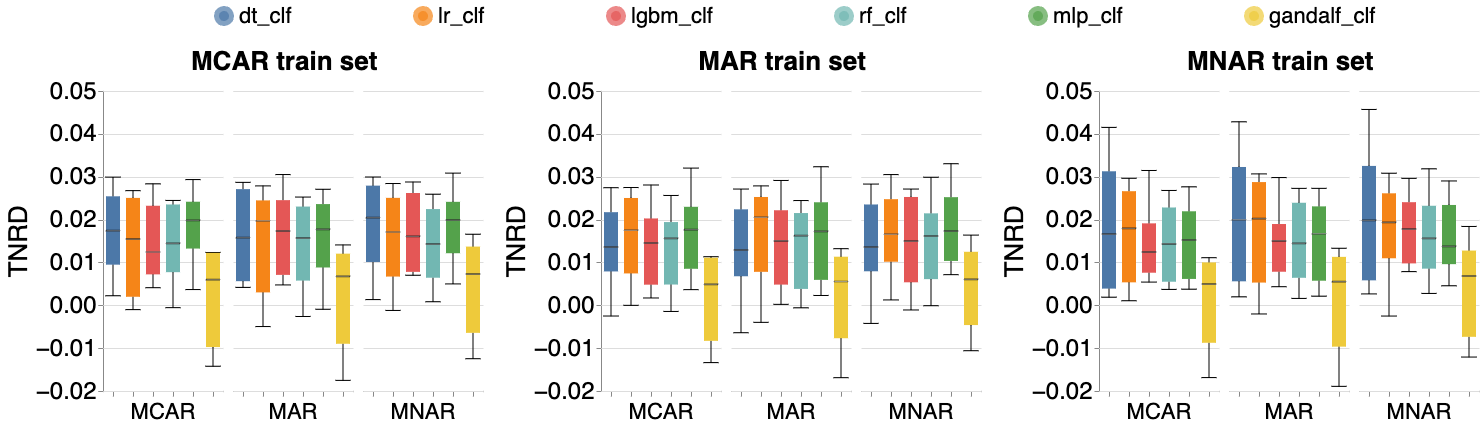}
    \caption{MissForest}
\end{subfigure}

\vspace{-0.3cm}
\caption{True Negative Rate Difference of different models (colors in legend) on the \heart dataset for ML and DL-based \mvi techniques (subplots) under missingness shift.}
\label{fig:missingness_shift_eq_odds_tnr_heart}
\end{figure}

\begin{figure}[h!]
\begin{subfigure}[h]{\linewidth}
    \includegraphics[width=\linewidth]{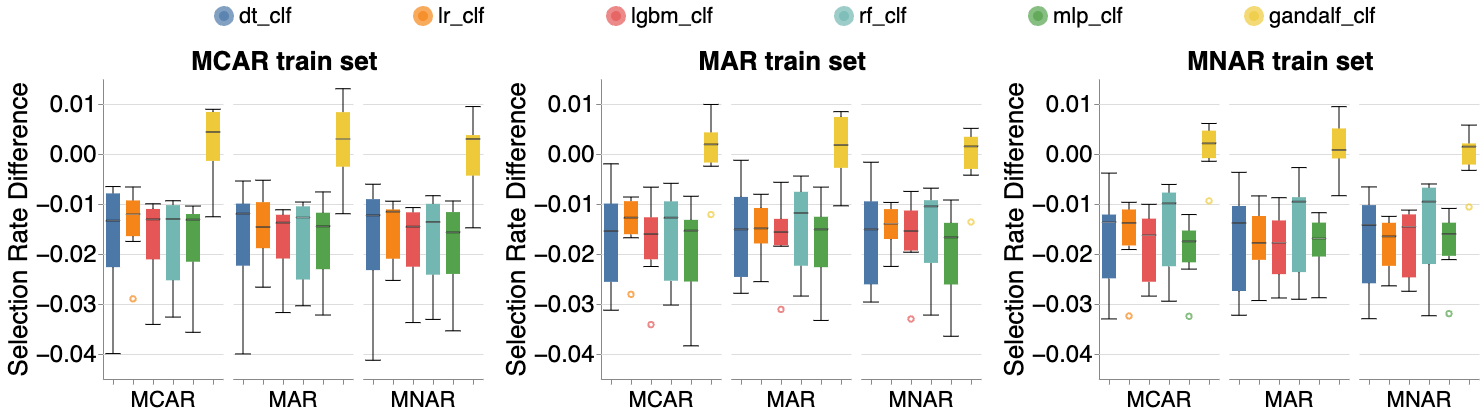}
    \caption{AutoML}
\end{subfigure}

\begin{subfigure}[h]{\linewidth}
    \includegraphics[width=\linewidth]{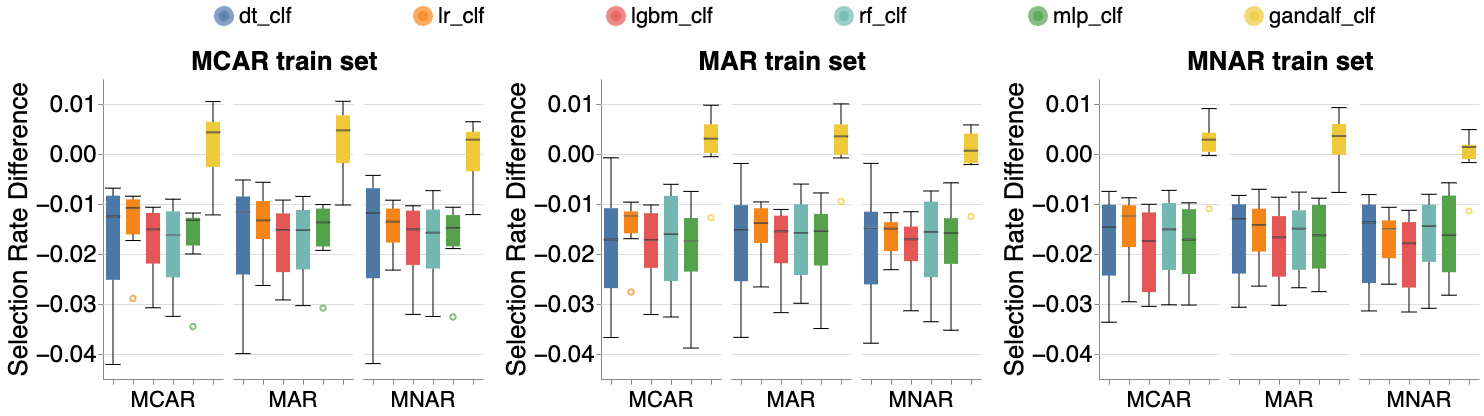}
    \caption{Datawig}
\end{subfigure}

\begin{subfigure}[h]{\linewidth}
    \includegraphics[width=\linewidth]{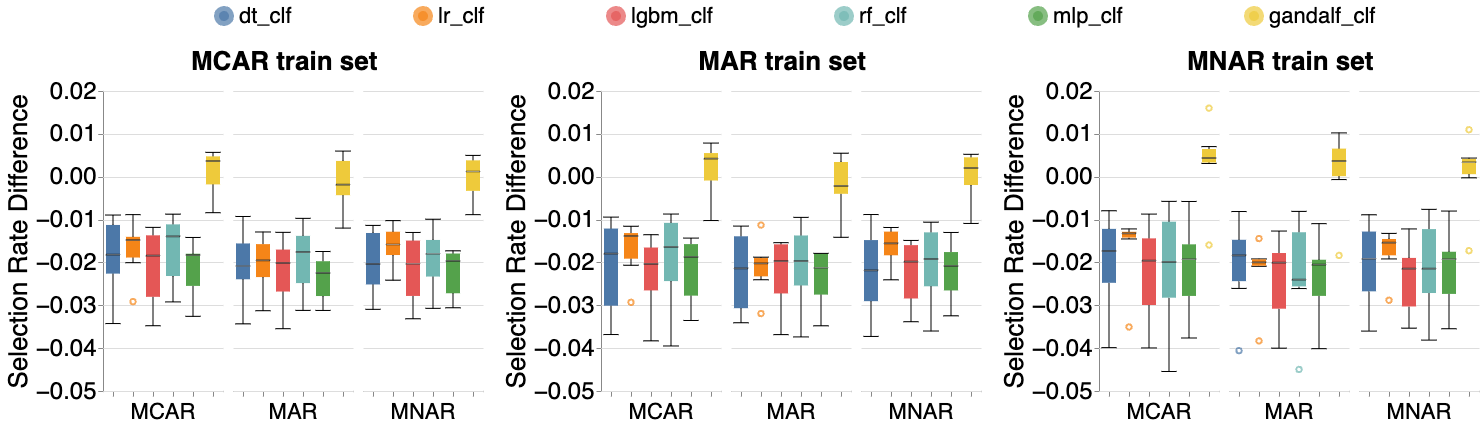}
    \caption{K-Means Clustering}
\end{subfigure}

\begin{subfigure}[h]{\linewidth}
    \includegraphics[width=\linewidth]{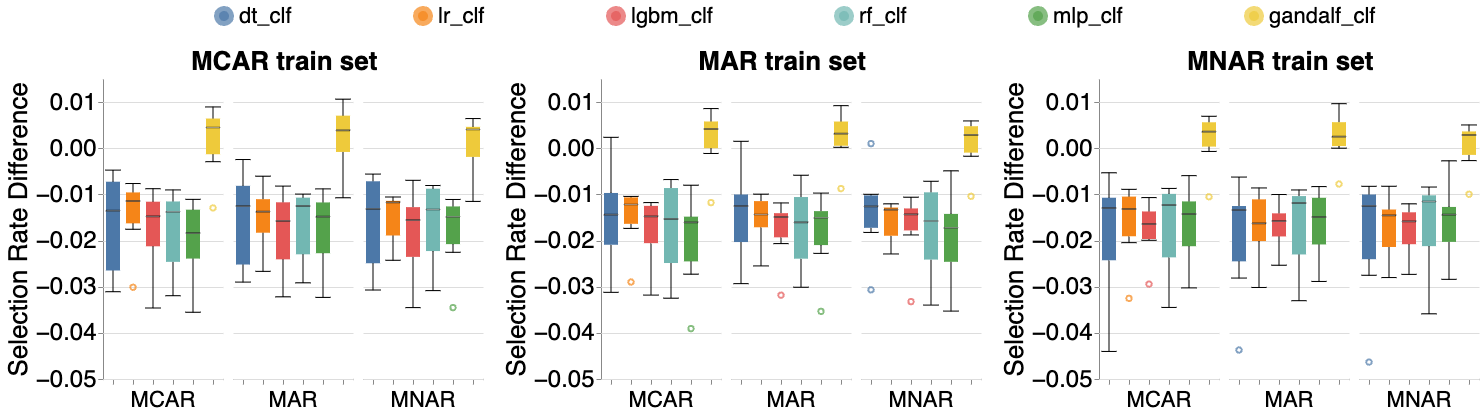}
    \caption{MissForest}
\end{subfigure}

\vspace{-0.3cm}
\caption{Selection Rate Difference of different models (colors in legend) on the \heart dataset for ML and DL-based \mvi techniques (subplots) under missingness shift.}
\label{fig:missingness_shift_spd_heart}
\vspace{1cm}
\end{figure}



\subsection{Variable Train and Test Missingness Rates}
\label{apdx:variable-shift-additional}

\begin{figure}[t!]
    \centering
    \includegraphics[width=0.9\linewidth]{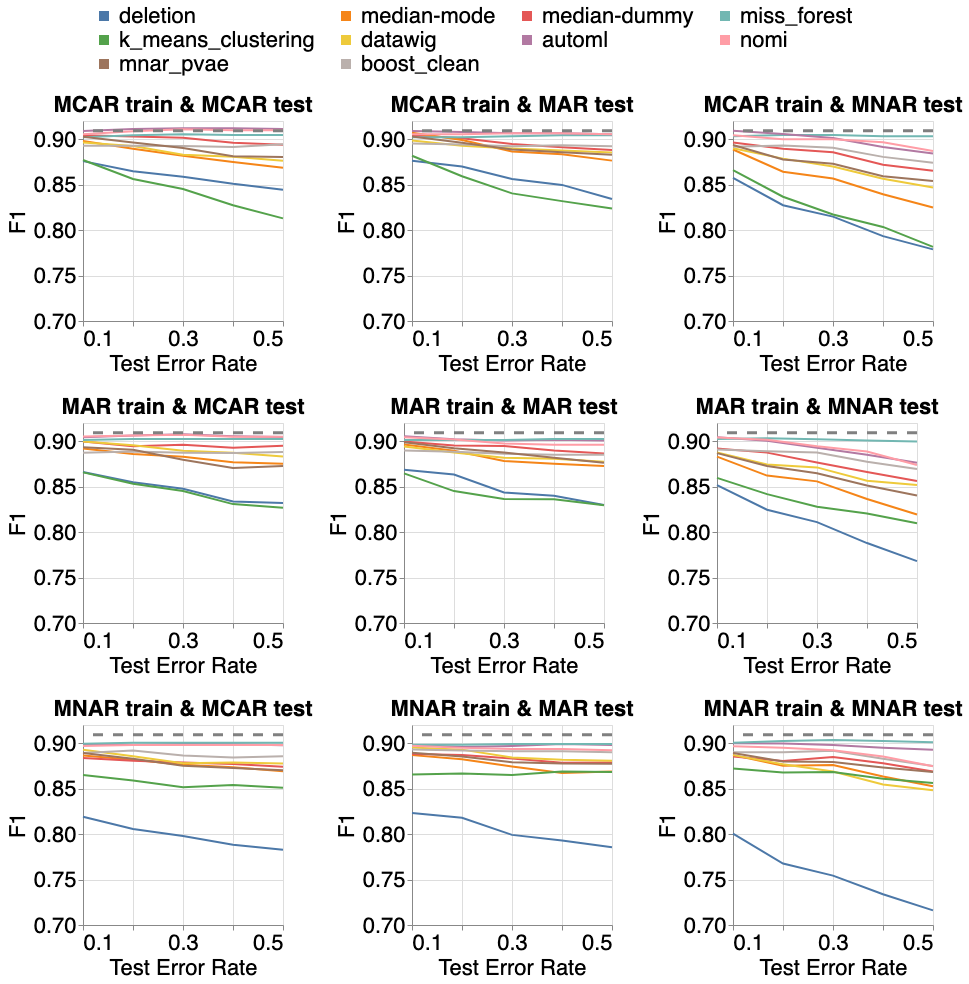}
    \caption{F1 of the Random Forest model on \diabetes, as a function of test missingness rate, under different missingness scenarios. The dashed line indicates performance of the model trained on clean data.}
    \label{fig:diabetes-test-error-f1}
\end{figure}

\textbf{Correctness under variable test set missingness.}
\label{apdx:f1-shift-additional}
This subsection supplements the discussion of Section~\ref{sec:shift-f1}. Figure \ref{fig:diabetes-test-error-f1} shows effect of varying test missingness rates on the F1 of the Random Forest model on \diabetes, under missingness shifts.

\begin{figure}[t!]
    \centering
    \includegraphics[width=0.9\linewidth]{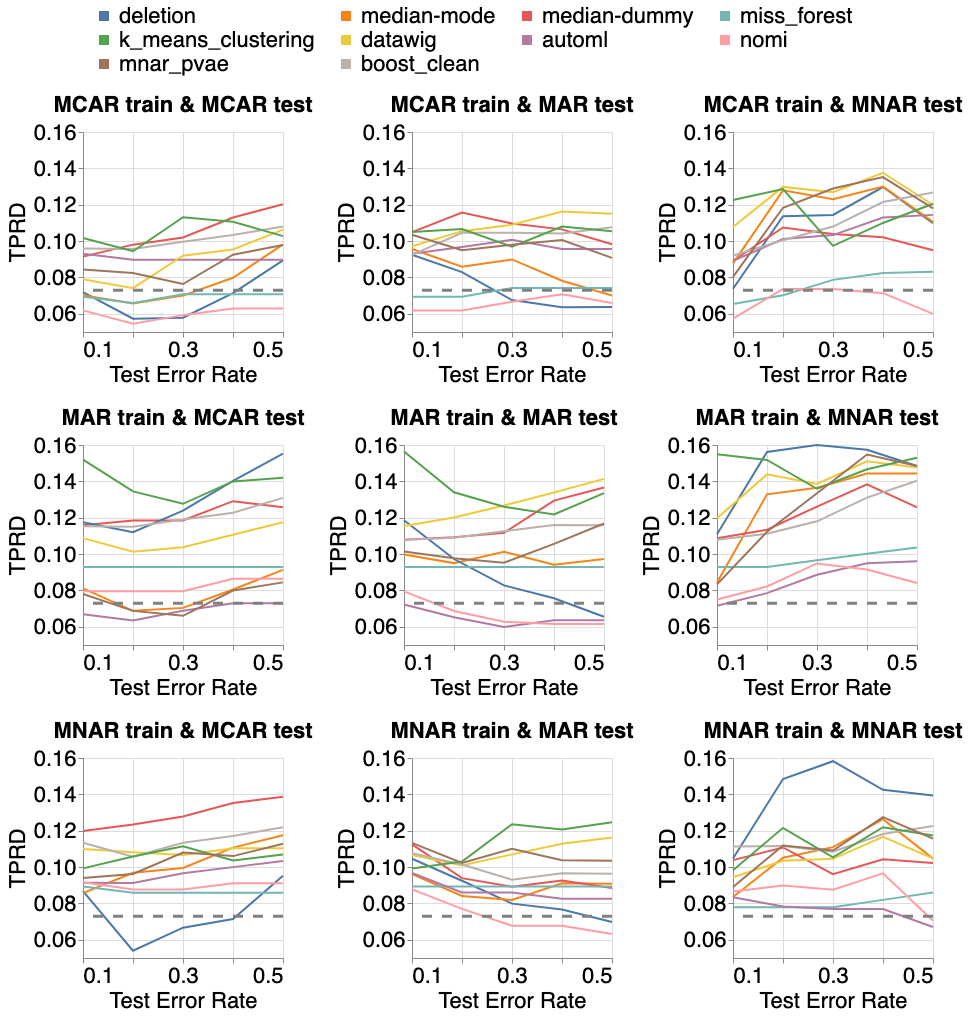}
    \caption{True Positive Rate Difference of the Random Forest model on \diabetes, as a function of test set error rate, under different missingness scenarios. The dashed line indicates the  performance of the model trained on clean data.}
    \label{fig:diabetes-test-error-tprd}
\end{figure}

\textbf{Fairness (TPRD) under variable test set missingness.}
\label{apdx:tprd-shift-additional}
This subsection supplements the discussion of Section~\ref{sec:shift-fairness}. Figure \ref{fig:diabetes-test-error-tprd} shows effect of varying test missingness rates on the True Positive Rate Difference (TPRD) of the Random Forest model on \diabetes, under missingness shifts.

\begin{figure}[t!]
    \centering
    \includegraphics[width=0.9\linewidth]{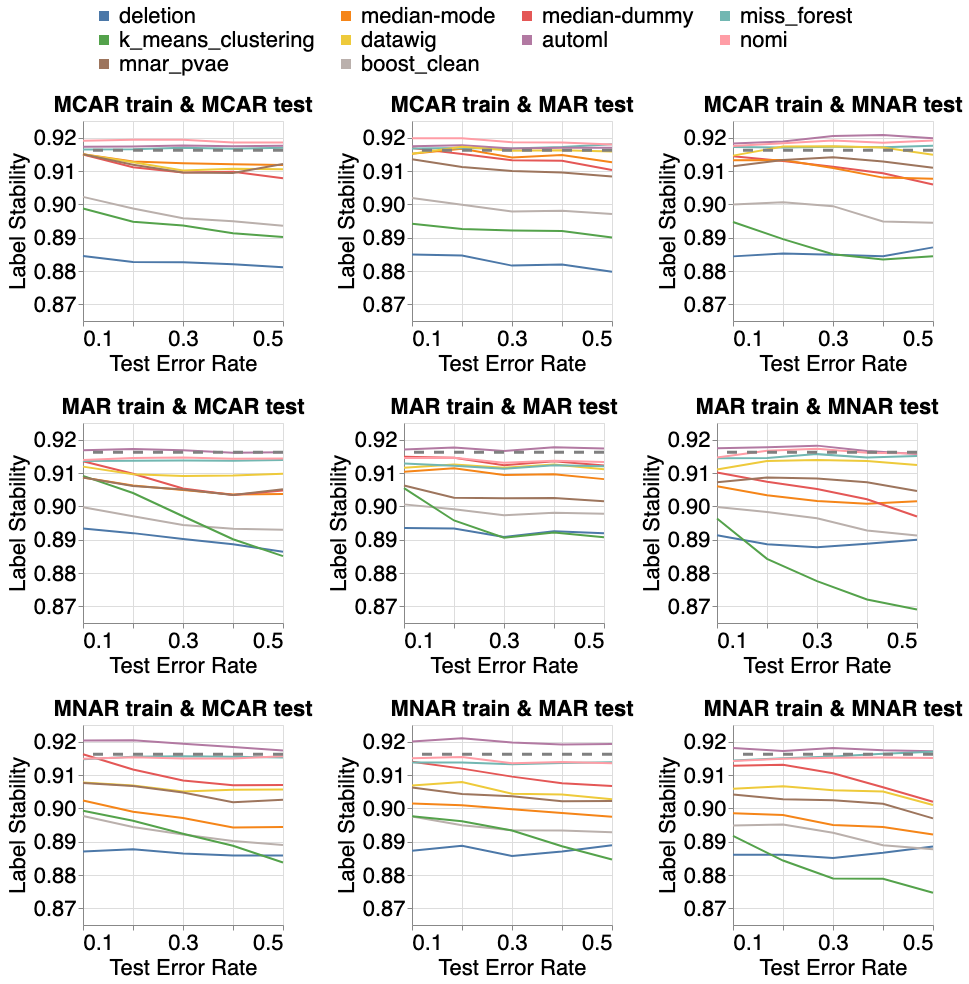}
    \caption{Label Stability of the Random Forest model on \diabetes, as a function of test missingness rate, under different missingness scenarios. The dashed line indicates the performance of the model trained on clean data.}
    \label{fig:diabetes-test-error-stability}
\end{figure}

\textbf{Stability under variable test set missingness.} 
\label{apdx:stability-shift-additional}
This subsection supplements the discussion of Section~\ref{sec:shift:stability}. It includes the label stability plots that were omitted in the main body of the paper due to space constraints. Figure \ref{fig:diabetes-test-error-stability} shows effect of varying test missingness rates on the Label Stability of the Random Forest model on \diabetes, under missingness shifts.

\begin{figure}[H]
    \centering
    \includegraphics[width=0.98\linewidth]{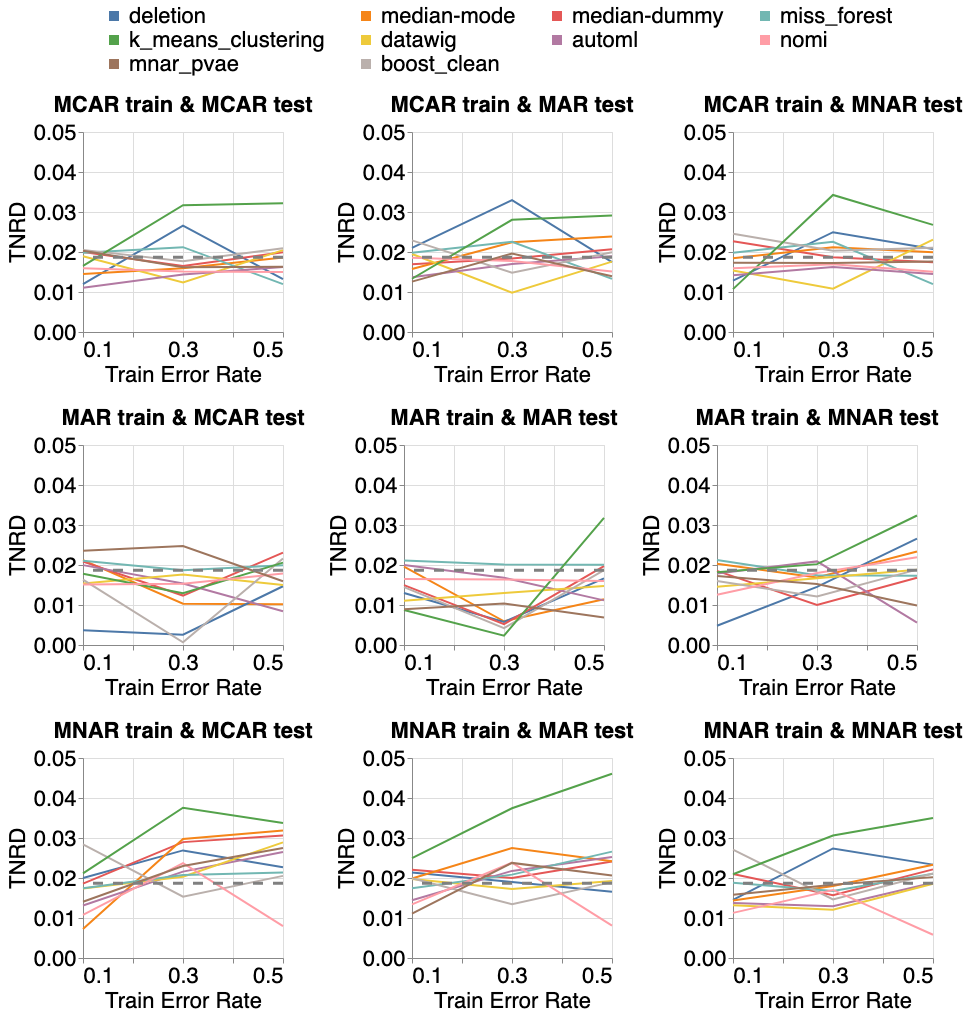}
    \caption{True Negative Rate Difference of the Random Forest model on \diabetes, as a function of train set error rate, under different missingness scenarios. The dashed line indicates the performance of the model trained on clean data.}
    \label{fig:diabetes-train-error-tnrd}
\end{figure}

\begin{figure}[H]
    \centering
    \includegraphics[width=0.98\linewidth]{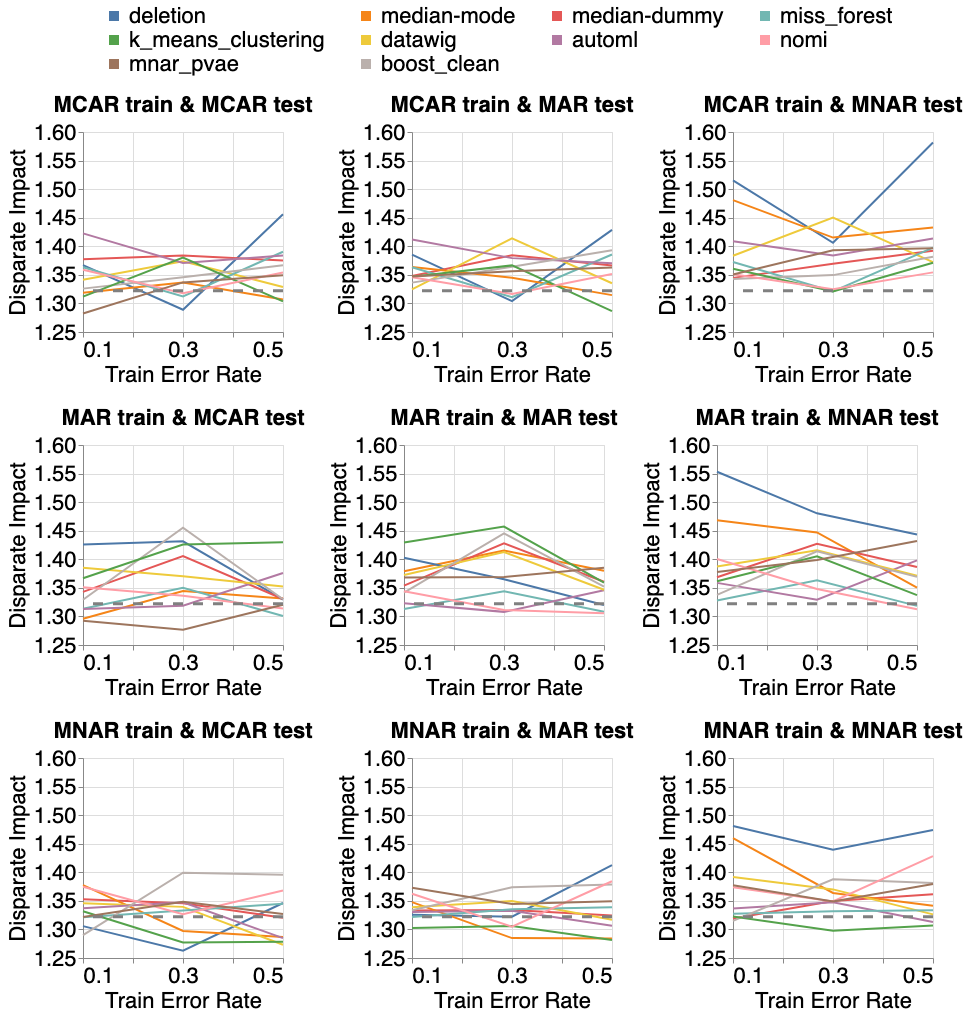}
    \caption{Disparate Impact of the Random Forest model on \diabetes, as a function of train set error rate, under different missingness scenarios. The dashed line indicates the performance of the model trained on clean data.}
    \label{fig:diabetes-train-error-di}
\end{figure}

\begin{figure}[H]
    \centering
    \includegraphics[width=0.93\linewidth]{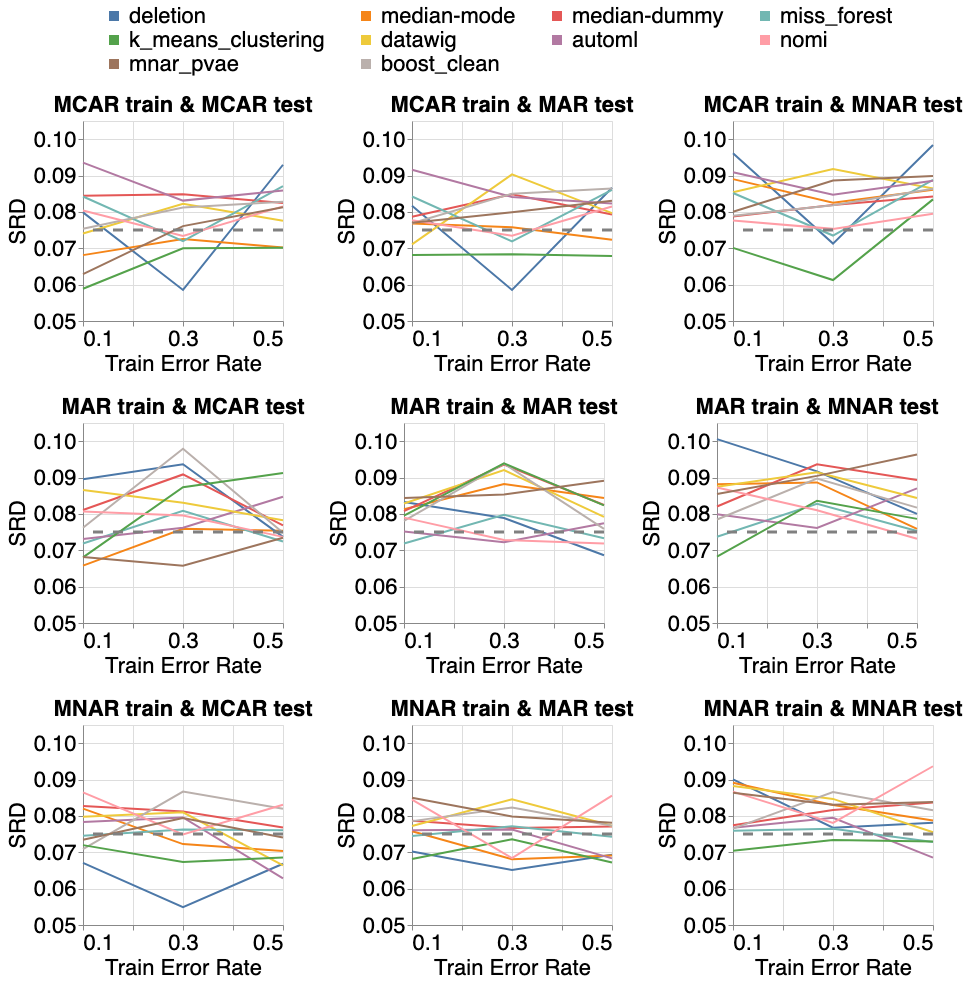}
    \caption{Selection Rate Difference of the Random Forest model on \diabetes, as a function of train set error rate, under different missingness scenarios. The dashed line indicates the performance of the model trained on clean data.}
    \label{fig:diabetes-train-error-spd}
\end{figure}

\textbf{Other fairness metrics under variable train set missingness.} Figures \ref{fig:diabetes-train-error-tnrd}, \ref{fig:diabetes-train-error-di}, and \ref{fig:diabetes-train-error-spd} for the \diabetes dataset and the Random Forest model illustrate the impact of train set missingness on True Negative Rate Difference, Disparate Impact, and Selection Rate Difference, respectively. Missingness shifts were modeled for all combinations of MCAR, MAR, and MNAR missingness patterns in the train and test sets, with the train set having 10\%, 30\%, or 50\% error rates, while the test set maintained a constant 30\% error rate across all settings.

The plots for additional fairness metrics support the assertion in Section~\ref{sec:shift-fairness} that null imputation methods are sensitive to the train missingness rate in terms of fairness. However, the impact varies across different fairness metrics. The highest deviation from the baseline value of each fairness metric is observed in the subfigures for MAR train sets at a 30\% error rate, matching the test error rate. For instance, the effect on TNRD is similar to that on TPRD in Section~\ref{sec:shift-fairness}, with most imputers showing an improvement in TNRD, whereas Disparate Impact and Selection Rate Difference worsen as the train error rate increases.

\begin{figure}[H]
    \centering
    \includegraphics[width=0.95\linewidth]{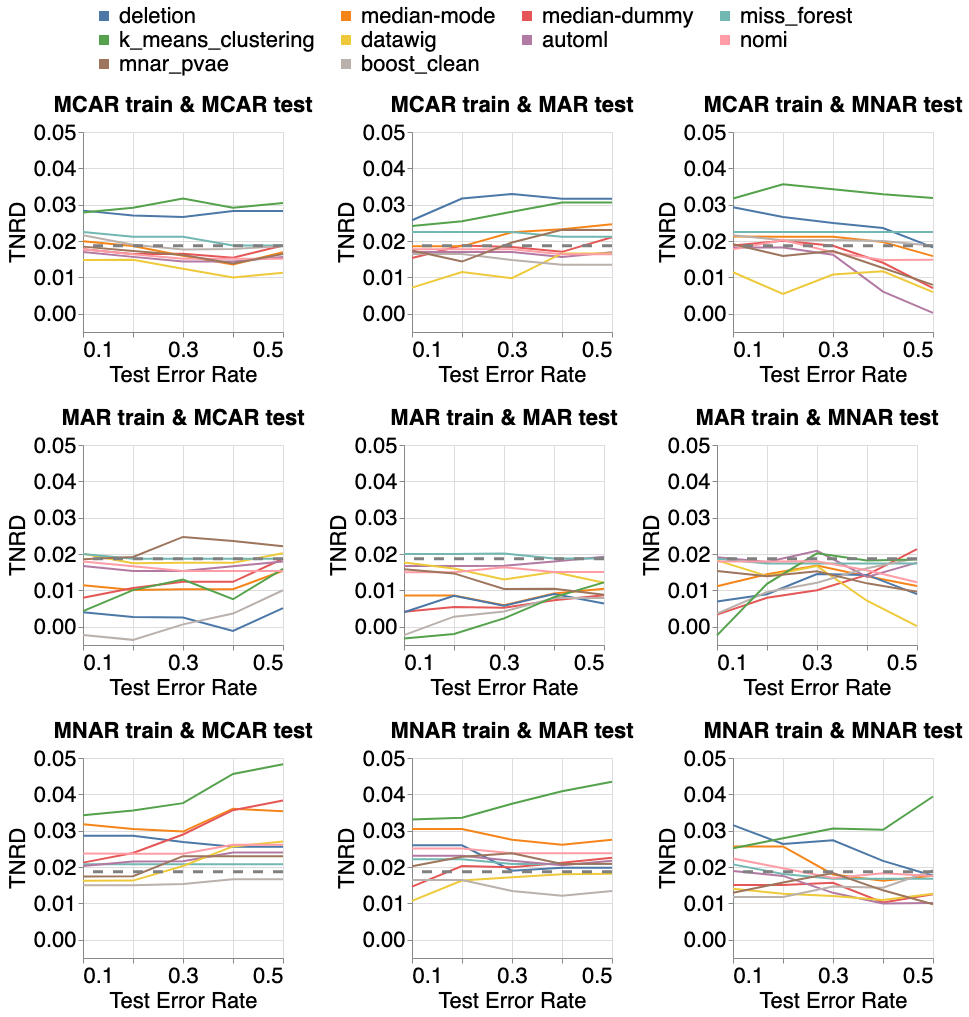}
    \caption{True Negative Rate Difference of the Random Forest model on \diabetes, as a function of test set error rate, under different missingness scenarios. The dashed line indicates the performance of the model trained on clean data.}
    \label{fig:diabetes-test-error-tnrd}
\end{figure}

\begin{figure}[H]
    \centering
    \includegraphics[width=0.95\linewidth]{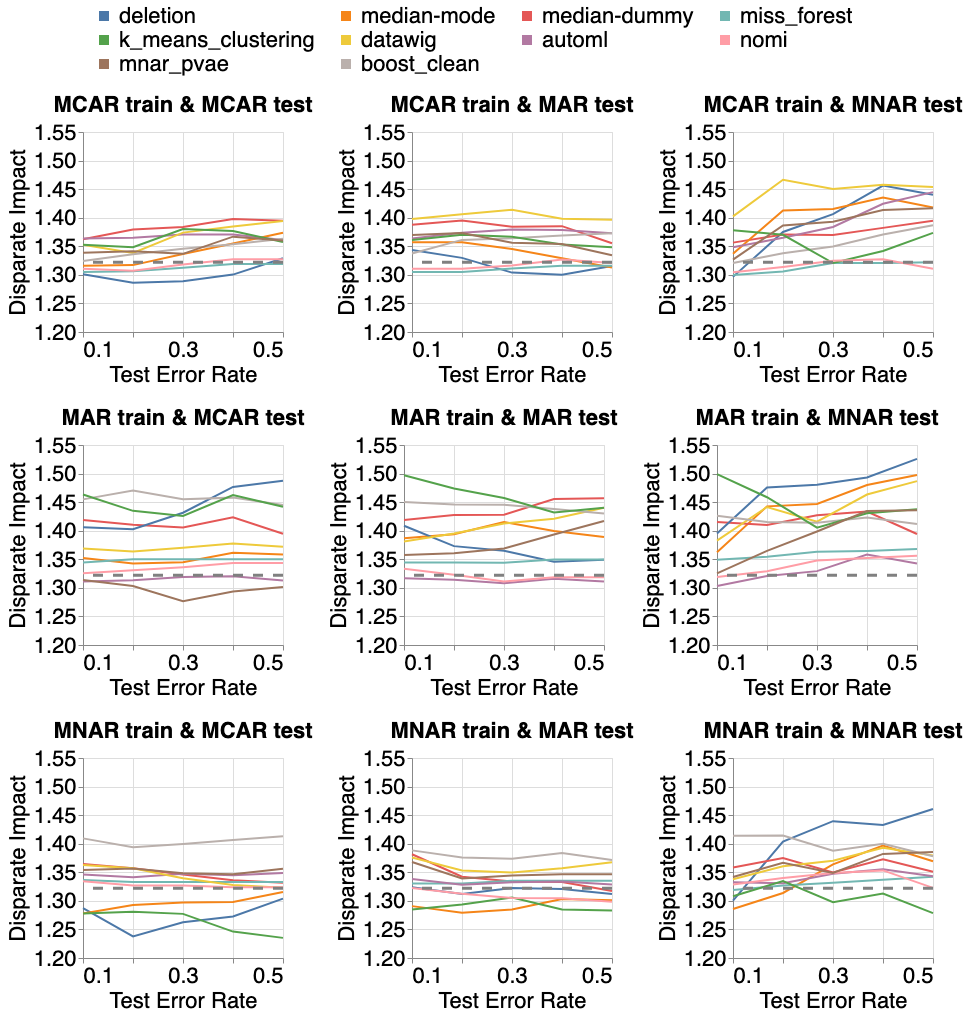}
    \caption{Disparate Impact of the Random Forest model on \diabetes, as a function of test set error rate, under different missingness scenarios. The dashed line indicates the performance of the model trained on clean data.}
    \label{fig:diabetes-test-error-di}
\end{figure}

\begin{figure}[H]
    \centering
    \includegraphics[width=0.96\linewidth]{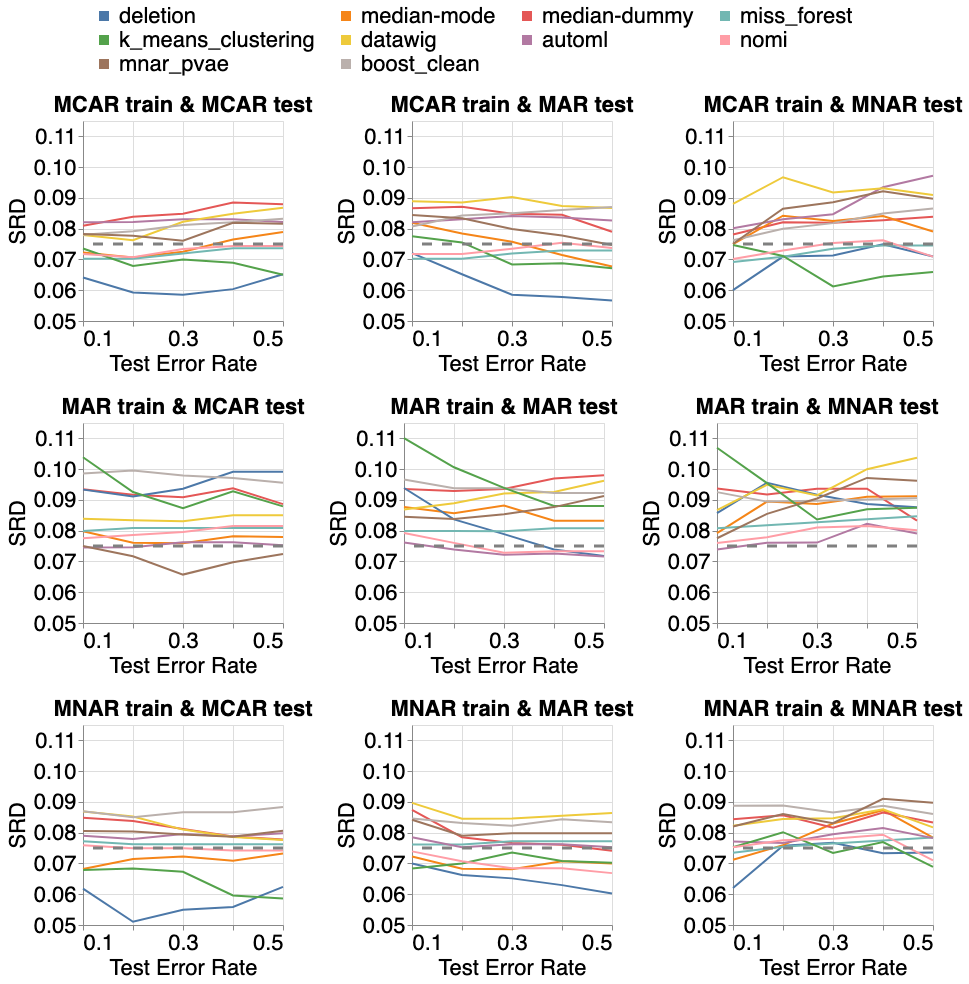}
    \caption{Selection Rate Difference of the Random Forest model on \diabetes, as a function of test set error rate, under different missingness scenarios. The dashed line indicates the performance of the model trained on clean data.}
    \label{fig:diabetes-test-error-spd}
\end{figure}

\textbf{Other fairness metrics under variable test set missingness.} Figures \ref{fig:diabetes-test-error-tnrd}, \ref{fig:diabetes-test-error-di}, and \ref{fig:diabetes-test-error-spd} for the \diabetes dataset and the Random Forest model show the impact of test set missingness on True Negative Rate Difference, Disparate Impact, and Selection Rate Difference, respectively. Missingness shifts were modeled for all combinations of MCAR, MAR, and MNAR missingness patterns in the train and test sets, with the test set having 10\%, 20\%, 30\%, 40\%, and 50\% error rates, while the train set maintained a constant 30\% error rate across all settings.

The plots reinforce the assertion in Section~\ref{sec:shift-fairness} that model fairness is highly sensitive to missingness shift. For example, the subfigures for MCAR train \& MNAR test and MAR train \& MNAR test show the largest deviations with increasing test error rates for all three fairness metrics, while plots for other settings remain relatively flat. Interestingly, when null imputers and models are fitted on the MNAR train set, model performance on the MNAR test set across different test error rates do not change significantly compared to MCAR train \& MNAR test and MAR train \& MNAR test settings. This indicates that the missingness mechanism in the train set can significantly affect model fairness on the MNAR test set. Additionally, \missforest, \automl, and \nomi proved to be the most robust \mvi techniques against test set missingness shifts under all conditions.

%% file: appendix/correlation.tex
\section{Spearman Correlation by Train and Test Missingness}
\label{sec:apdx-correlations}

In Figure~\ref{fig:correlations_train} in Section~\ref{sec:conclusion}, we reported correlations between \mvi technique, model type, test missingness and performance metrics (F1, fairness and stability), based on train missingness. In this section, we supplement this analysis by reporting correlations between \mvi technique, model type and performance metrics for each combination of train and test missingness. 

\begin{figure}[b]
\begin{subfigure}[h]{0.65\linewidth}
    \centering
    \includegraphics[width=\linewidth]{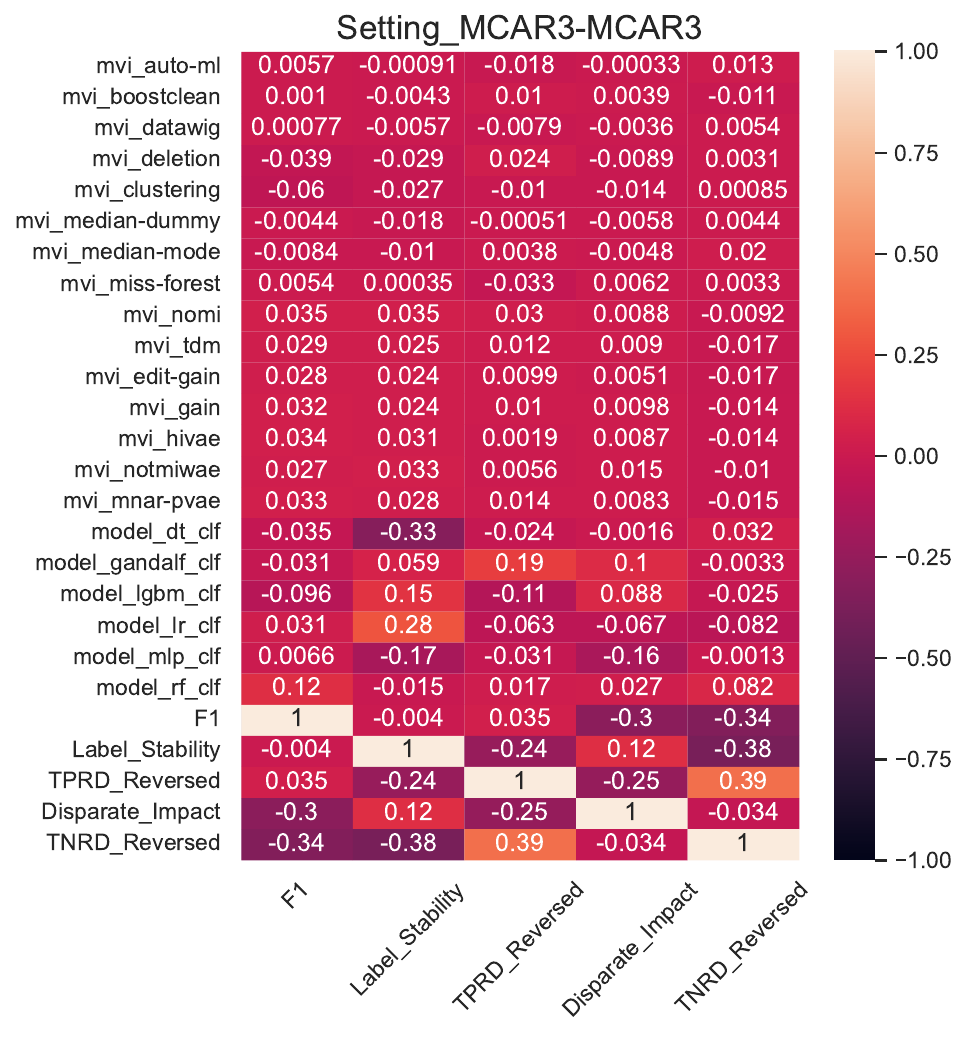}
    \caption{MCAR train, MCAR test}
\end{subfigure}
\begin{subfigure}[h]{0.65\linewidth}
    \centering
    \includegraphics[width=\linewidth]{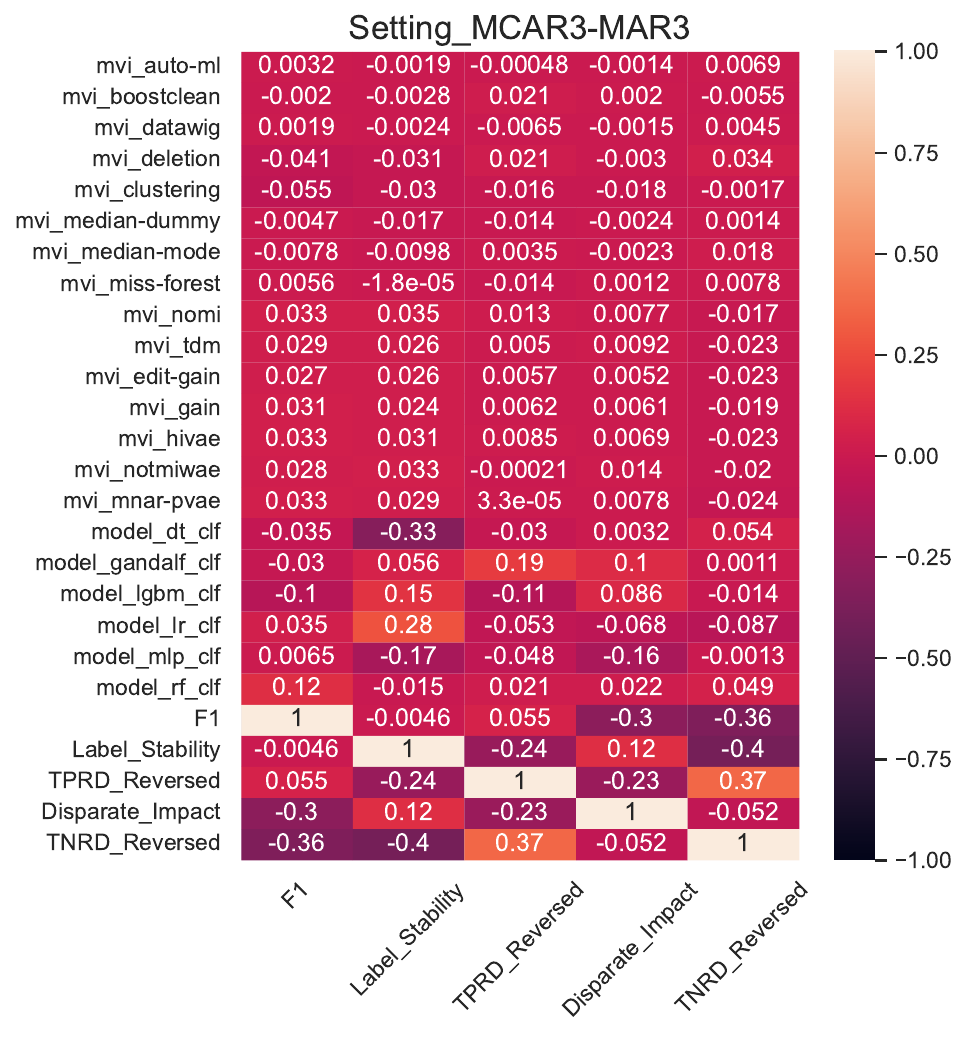}
    \caption{MCAR train, MAR test}
\end{subfigure}
\begin{subfigure}[h]{\linewidth}
    \centering
    \includegraphics[width=0.65\linewidth]{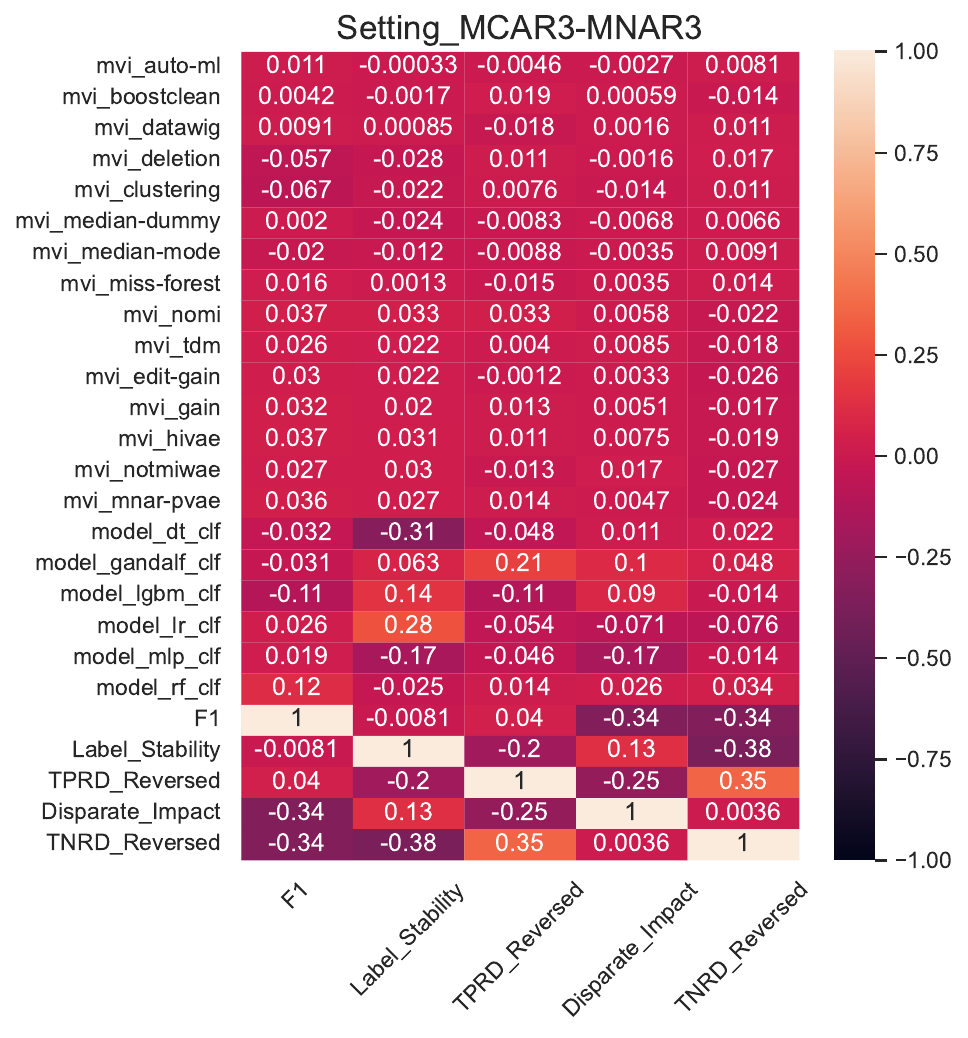}
    \caption{MCAR train, MNAR test}
\end{subfigure}
\caption{Spearman correlation ($\rho$) between \mvi technique, model type, test missingness and performance metrics (F1, fairness and stability) under MCAR train and different test missingnesses. TPRD and TNRD values close to 0 are ideal (fair), so we compute correlations using $TPRD\_Reversed = 1 - |TPRD|$ and $TNRD\_Reversed = 1 - |TNRD|$}
\label{fig:apdx-mcar-correlations}
\vspace{-1.0cm}
\end{figure}

\begin{figure}[h!]
\begin{subfigure}[h]{0.65\linewidth}
    \centering
    \includegraphics[width=\linewidth]{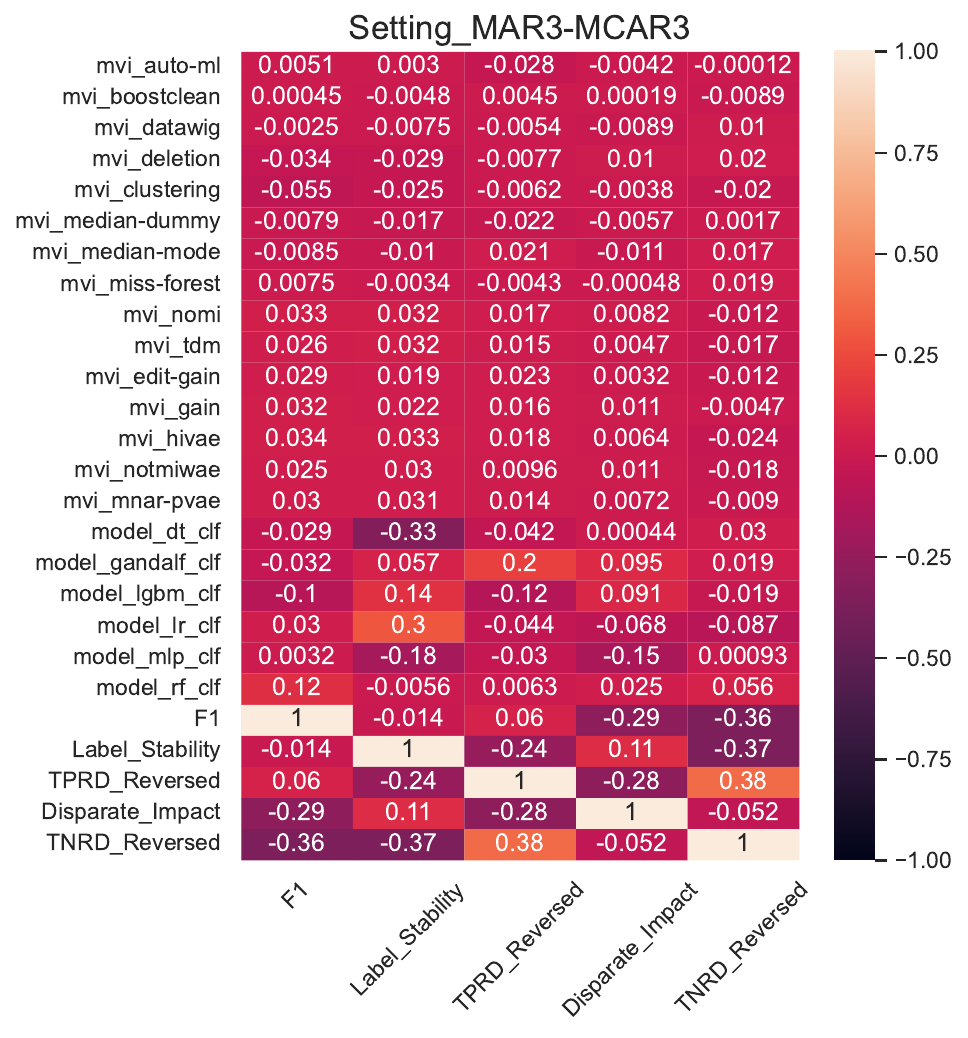}
    \caption{MAR train, MCAR test}
\end{subfigure}
\begin{subfigure}[h]{0.65\linewidth}
    \centering
    \includegraphics[width=\linewidth]{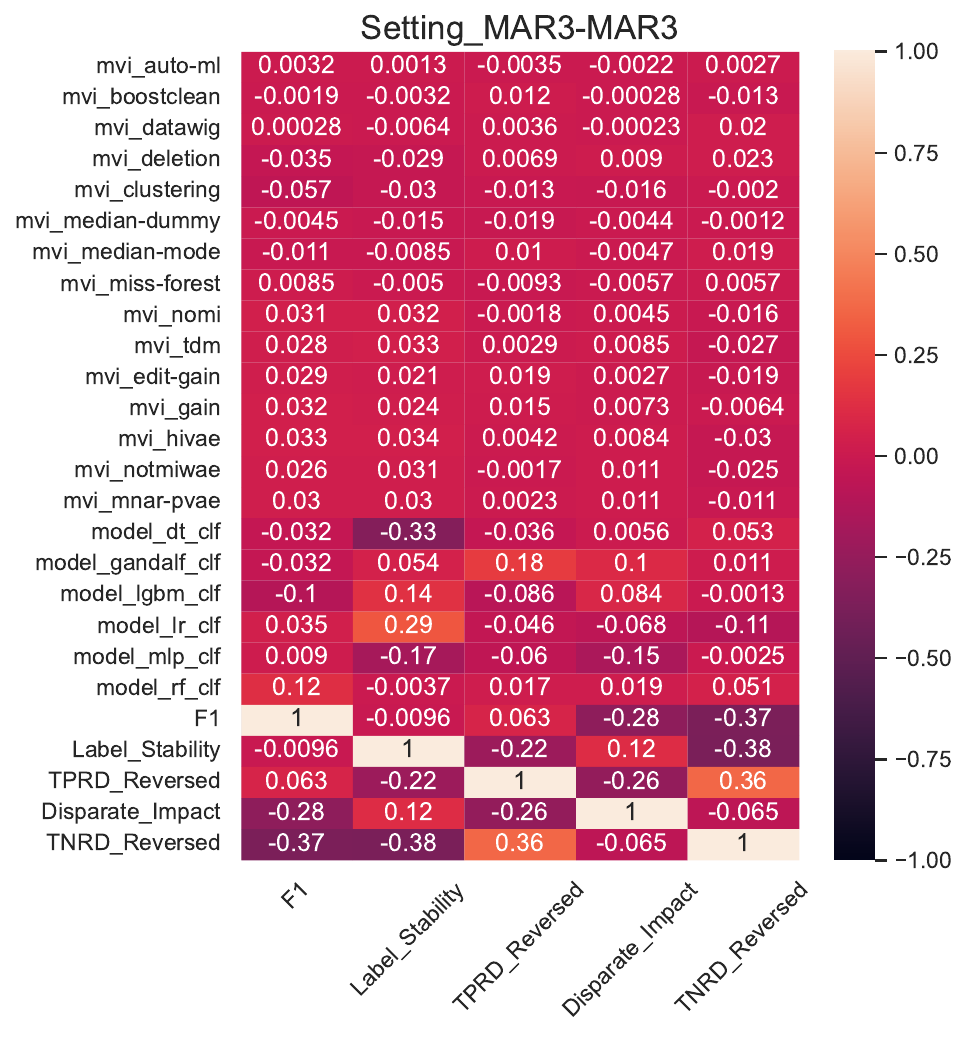}
    \caption{MAR train, MAR test}
\end{subfigure}
\begin{subfigure}[h]{\linewidth}
    \centering
    \includegraphics[width=0.65\linewidth]{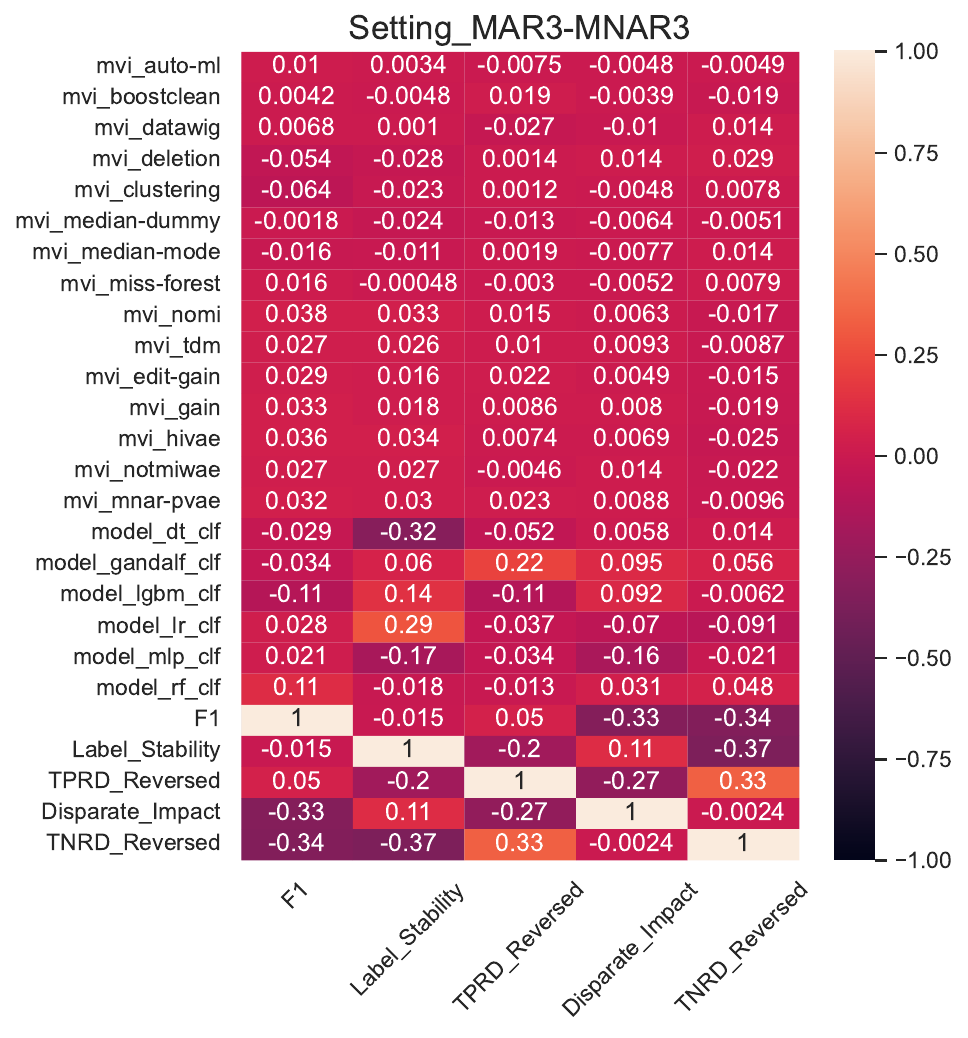}
    \caption{MAR train, MNAR test}
\end{subfigure}
\caption{Spearman correlation ($\rho$) between \mvi technique, model type, test missingness and performance metrics (F1, fairness and stability) under MAR train and different test missingnesses. TPRD and TNRD values close to 0 are ideal (fair), so we compute correlations using $TPRD\_Reversed = 1 - |TPRD|$ and $TNRD\_Reversed = 1 - |TNRD|$}
\label{fig:apdx-mar-correlations}
\vspace{-0.7cm}
\end{figure}

\begin{figure}[h!]
\begin{subfigure}[h]{0.65\linewidth}
    \centering
    \includegraphics[width=\linewidth]{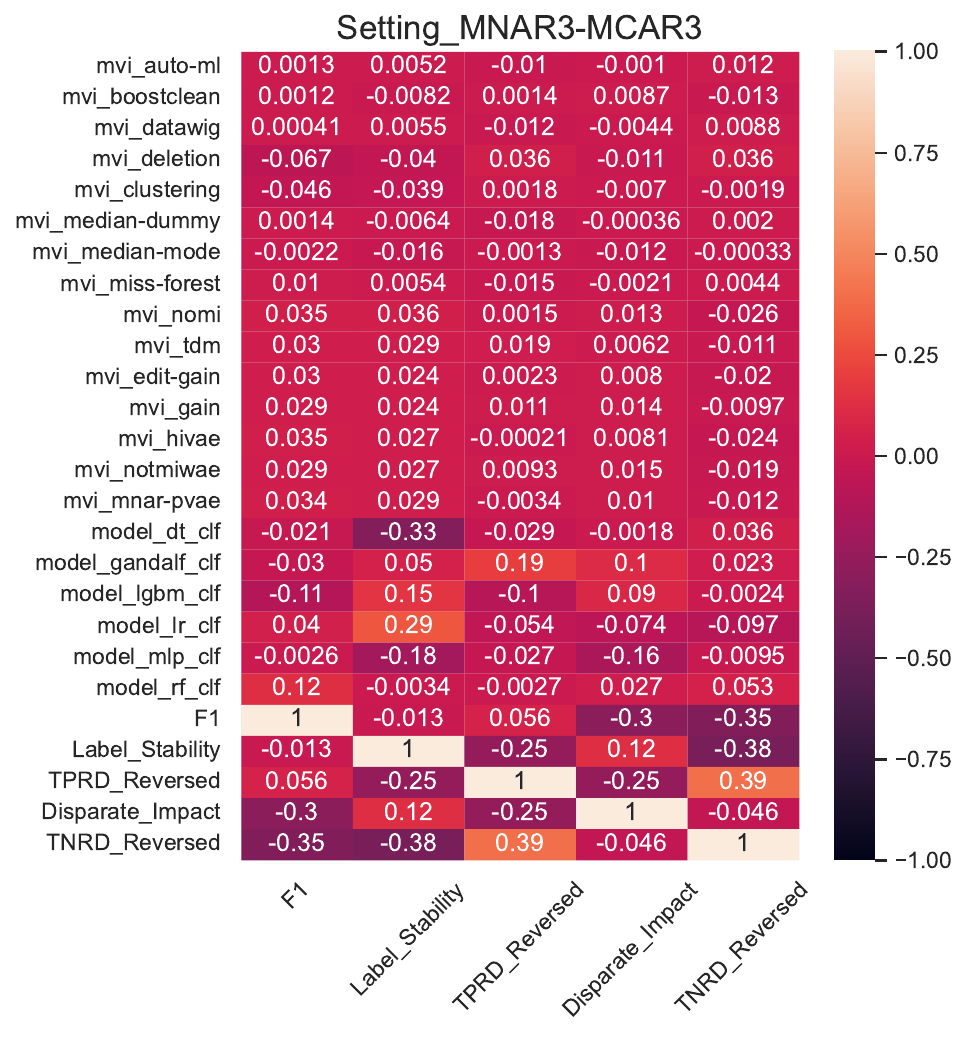}
    \caption{MNAR train, MCAR test}
\end{subfigure}
\begin{subfigure}[h]{0.65\linewidth}
    \centering
    \includegraphics[width=\linewidth]{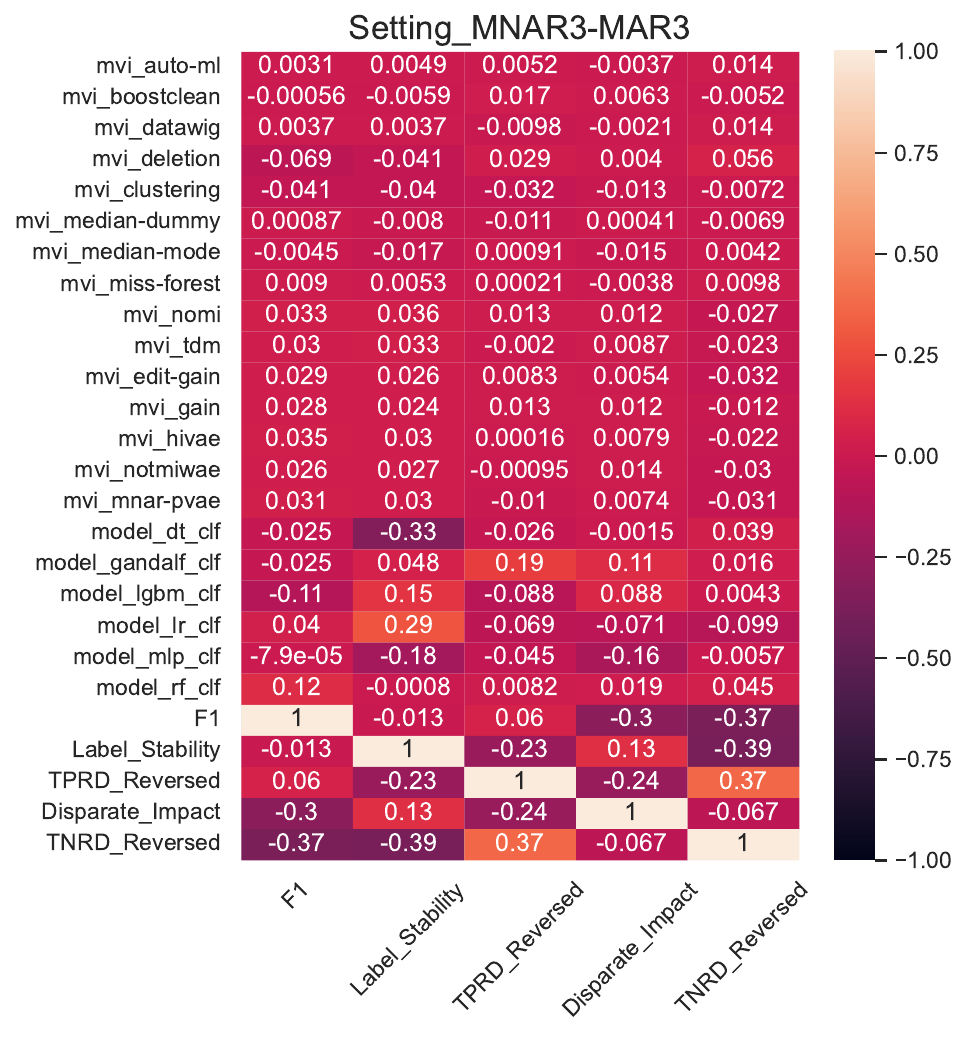}
    \caption{MNAR train, MAR test}
\end{subfigure}
\begin{subfigure}[h]{\linewidth}
    \centering
    \includegraphics[width=0.65\linewidth]{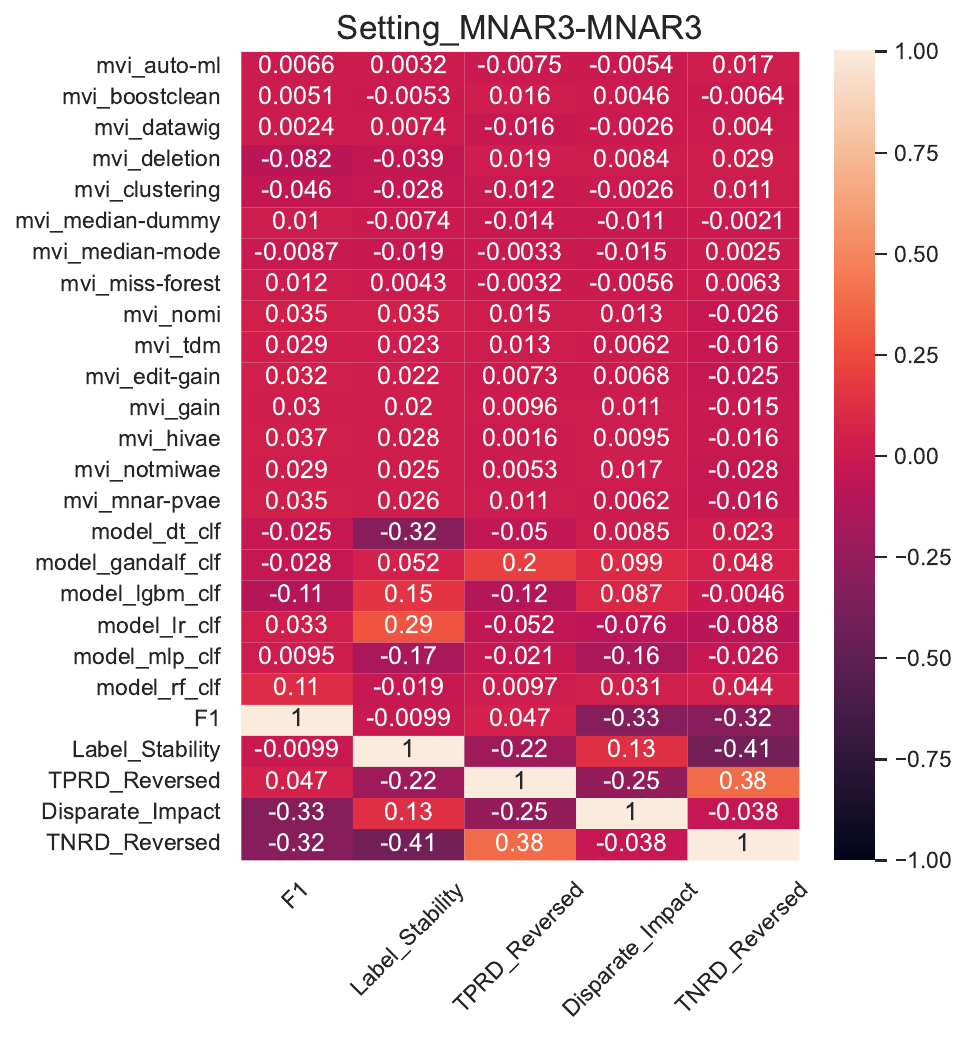}
    \caption{MNAR train, MNAR test}
\end{subfigure}
\caption{Spearman correlation ($\rho$) between \mvi technique, model type, test missingness and performance metrics (F1, fairness and stability) under MNAR train and different test missingnesses. TPRD and TNRD values close to 0 are ideal (fair), so we compute correlations using $TPRD\_Reversed = 1 - |TPRD|$ and $TNRD\_Reversed = 1 - |TNRD|$}
\label{fig:apdx-mnar-correlations}
\vspace{-0.7cm}
\end{figure}

Figure~\ref{fig:apdx-mcar-correlations} is for \mcar train and different test missingness, Figure~\ref{fig:apdx-mar-correlations} is for \mar train and different test missingness, and Figure~\ref{fig:apdx-mnar-correlations} is for \mnar train and different test missingness.

\newpage

